\documentclass[10pt,twocolumn,letterpaper]{article}

\usepackage[pagenumbers]{cvpr}

\usepackage{booktabs} %

\usepackage{hhline}
\usepackage{amsfonts}
\usepackage{amsmath}
\usepackage[ruled]{algorithm2e} %
\usepackage{subcaption}
\usepackage{pifont}
\usepackage{arydshln}
\usepackage{multirow}
\usepackage{multicol}
\usepackage{xcolor}
\usepackage[export]{adjustbox}
\usepackage[nor malem]{ulem}
\usepackage{xspace}
\usepackage{rotating}
\usepackage{xhfill}
\usepackage{tikz,bm}

\usepackage{appendix}

\usepackage[pagebackref,breaklinks,colorlinks]{hyperref}
\usepackage[capitalize]{cleveref}

\makeatletter
\usetikzlibrary{arrows.meta}

\crefname{section}{Sec.}{Secs.}
\Crefname{section}{Section}{Sections}
\Crefname{table}{Table}{Tables}
\crefname{table}{Tab.}{Tabs.}

\definecolor{mydarkblue}{rgb}{0,0.08,1}
\definecolor{mydarkgreen}{rgb}{0.02,0.6,0.02}
\definecolor{mydarkorange}{rgb}{0.40,0.2,0.02}

\makeatletter
\DeclareRobustCommand\onedot{\futurelet\@let@token\@onedot}
\def\@onedot{\ifx\@let@token.\else.\null\fi\xspace}

\def\limitarrowmain#1{%
\begin{tikzpicture}
\draw[-{stealth[scale=4]}] (-5.5,0) to (-1,0);
\end{tikzpicture}}

\def\limitarrowsup#1{%
\begin{tikzpicture}
\draw[-{stealth[scale=4]}] (-7,0) to (1,0);
\end{tikzpicture}}

\def\limitarrowadditional#1{%
\begin{tikzpicture}
\draw[-{stealth[scale=4]}] (-5,0) to (0,0);
\end{tikzpicture}}

\def\etal{\emph{et al}\onedot}

\def\etal{\emph{et al}\onedot}

\def\naive{na\"{\i}ve\xspace}

\newcommand{\p}{$\mathcal{P}$\xspace}
\newcommand{\pplus}{$\mathcal{P}+$\xspace}
\newcommand{\pstar}{$\mathcal{P}^*$\xspace}

\definecolor{mygreen}{RGB}{112,173,71}
\definecolor{myblue}{RGB}{91,155,213}
\definecolor{myorange}{RGB}{237,125,49}
\definecolor{myred}{RGB}{255,22,67}
\definecolor{honey}{RGB}{201,89,95}
\definecolor{amber}{RGB}{180,148,41}
\definecolor{darkblue}{RGB}{84,96,126}
\definecolor{yelloworange}{HTML}{FAA21A}
\definecolor{darkgreen}{HTML}{008000}
\definecolor{mypurple}{RGB}{127,60,141}
\definecolor{hotpink}{RGB}{207,28,144}

\let\shortcite\cite

\title{
A Neural Space-Time Representation for Text-to-Image Personalization
\vspace*{-0.3cm}
}

\begin{document}

\author{
Yuval Alaluf$^*$ \hspace{0.65cm} 
Elad Richardson$^*$ \hspace{0.65cm} 
Gal Metzer \hspace{0.65cm} 
Daniel Cohen-Or \\ \\
Tel Aviv University \vspace{0.2cm} \\
\small\url{https://NeuralTextualInversion.github.io/NeTI/} \vspace{-0.8cm}
}

\twocolumn[{%
\vspace{-1em}
\maketitle
\renewcommand\twocolumn[1][]{#1}%
\vspace{-0.1in}
\begin{center}
    \centering
    \includegraphics[width=0.98\textwidth]{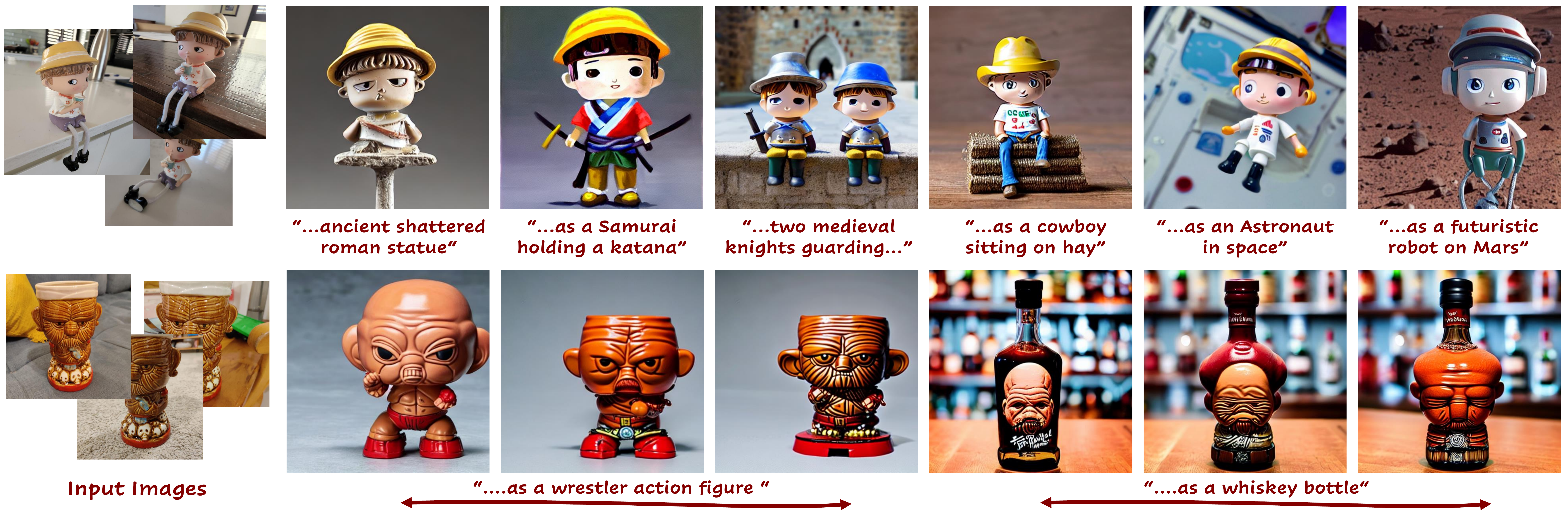}
    \vspace{-0.2cm}
    \captionof{figure}{
    Personalization results of our method under a variety of prompts. Our expressive representation enables one to generate novel compositions of personalized concepts that achieve high visual fidelity and editability without tuning the generative model. The bottom row shows our method's unique ability to control the reconstruction-editability tradeoff at inference time with a single trained model.
    }
    \label{fig:teaser}
\end{center}%
}]

\begin{abstract}
\vspace*{-0.2cm}
A key aspect of text-to-image personalization methods is the manner in which the target concept is represented within the generative process. This choice greatly affects the visual fidelity, downstream editability, and disk space needed to store the learned concept. In this paper, we explore a new text-conditioning space that is dependent on both the denoising process timestep (\textit{time}) and the denoising U-Net layers (\textit{space}) and showcase its compelling properties. A single concept in the space-time representation is composed of hundreds of vectors, one for each combination of \textit{time} and \textit{space}, making this space challenging to optimize directly. Instead, we propose to implicitly represent a concept in this space by optimizing a small neural mapper that receives the current \textit{time} and \textit{space} parameters and outputs the matching token embedding. In doing so, the entire personalized concept is represented by the parameters of the learned mapper, resulting in a compact, yet expressive representation. Similarly to other personalization methods, the output of our mapper resides in the input space of the text encoder. We observe that one can significantly improve the convergence and visual fidelity of the concept by introducing a \textit{textual bypass}, where the mapper additionally outputs a residual that is added to the \textit{output} of the text encoder. Finally, we show how one can impose an importance-based ordering over our implicit representation, providing control over the reconstruction and editability of the learned concept using a single trained model. We demonstrate the effectiveness of our approach over a range of concepts and prompts, showing our method's ability to generate high-quality and controllable compositions without fine-tuning any parameters of the generative model itself.

\end{abstract}

\def\thefootnote{*}\footnotetext{Denotes equal contribution}

\vspace*{-0.25cm}
\section{Introduction}
Recent large-scale text-to-image models~\cite{rombach2022high,saharia2022photorealistic,ramesh2022hierarchical} have quickly revolutionized the world of artistic creation, demonstrating an unprecedented ability to generate incredible and diverse visual content.
By learning to inject new concepts into these powerful models, numerous personalization techniques have further enabled users to create new artistic compositions of a unique subject or artistic style using a small set of images depicting the concept. 

\begin{figure*}
    \centering
    \includegraphics[width=\textwidth]{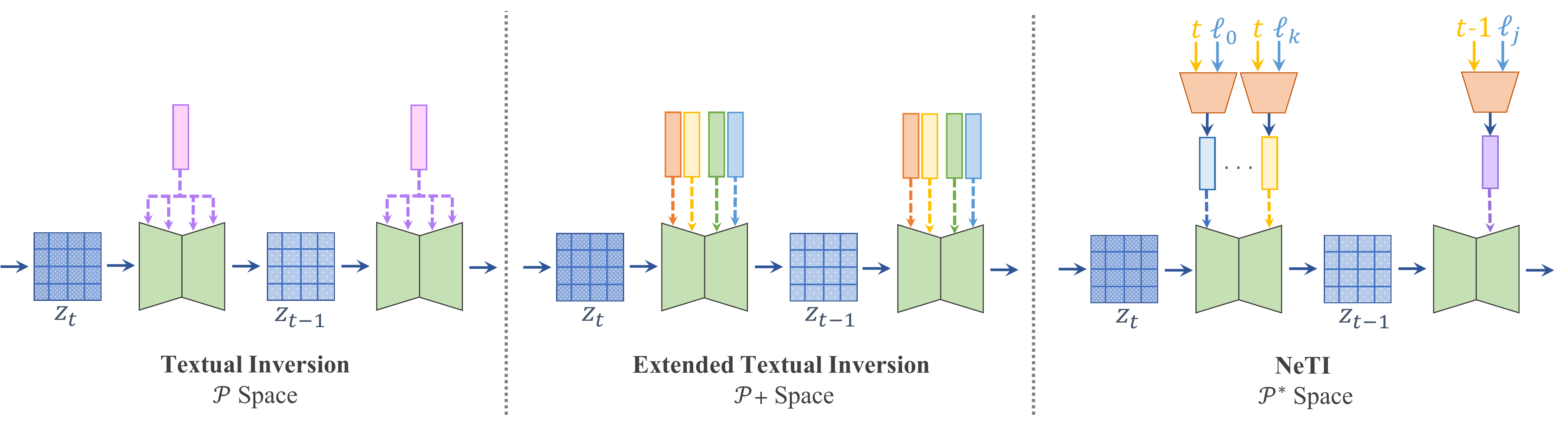}
    \\[-0.35cm]
    \caption{The personalization-by-inversion approaches and their text-conditioning spaces. Textual Inversion~\cite{gal2023image} (left) invert into the \p space where a single token embedding is learned for all timesteps and U-Net layers. Voynov~\etal~\shortcite{voynov2023p} (middle) introduce the \pplus space where different embeddings are optimized for each attention layer but are shared across all timesteps. Finally, (right) we introduce a NeTI, which utilizes a new \textit{space-time} representation learned implicitly via a small mapping layer that considers both the different U-Net layers and denoising timesteps.}
    \label{fig:overview}
    \vspace{-0.25cm}
\end{figure*}

Current personalization techniques can be categorized by how they treat the pretrained text-to-image model. The \textit{personalization-by-inversion} approach, first proposed in Gal~\etal~~\shortcite{gal2023image}, freezes the generative model and optimizes an input vector to represent the desired subject or artistic style. 
This vector resides in the recently dubbed \p space~\cite{voynov2023p} containing all possible input embeddings to the text encoder.
Alternatively, to better capture the target concept, Ruiz~\etal~\shortcite{ruiz2022dreambooth} proposed the \textit{personalization-by-fine-tuning} approach, where one directly fine-tunes the generative model to represent the user-specified concept. 
While this results in better reconstructions, it requires additional storage costs and is often more prone to overfitting unwanted details such as the image background.
Recently, Voynov~\etal~\shortcite{voynov2023p} demonstrated that one can improve inversion approaches by inverting into an extended input space, \pplus, where a different vector $p\in$~\p is learned for each attention layer in the denoising U-Net network. 
In doing so, they achieve improved reconstruction and editability of the target concept without tuning the generative model.

In this paper, we introduce a new text-conditioning space that is dependent on both the denoising process timestep and the U-Net layers, which we call \pstar. 
The \pstar space is composed of a set of vectors $p_t\in$~\pplus, one for each timestep $t$. This results in a richer representation space that considers the time-dependent nature of the denoising process.
A \naive approach for inverting a concept into the \pstar space would require optimizing hundreds of different vectors, one for each possible combination of timestep and U-Net layer. 
Instead, we propose to implicitly represent the \pstar space using a small neural mapper that receives the current timestep $t$ and the U-Net layer $\ell$ and outputs a vector $p\in$~\p, see~\Cref{fig:overview}. 
In a sense, the entire network represents a concept in \pstar defined by its learned parameters, resulting in a neural representation for Textual Inversion, which we dub \textit{NeTI}.
We show that a new concept can be learned by optimizing the parameters of our neural representation, similar to the standard optimization mechanism in Textual Inversion.

Next, we observe that while \pstar is more expressive compared to \p or \pplus, its potential to generate complex concepts is still dependent on the text encoder, as our learned embeddings are first passed through the encoder before being fed to the U-Net model. 
Unfortunately, completely skipping the text encoder and working directly within the U-Net's input space is not a viable option, as the text encoder is essential for maintaining editability through the mixing of our concept's learned representation with the other prompt tokens.
To overcome this issue, we propose a \textit{textual bypass} technique where we learn an additional residual vector for each space-time input and add it to the text encoder's \textit{output}. 
In this formulation, our neural mapper outputs two vectors, one which is fed into the text encoder and is mixed with the other prompt tokens, and a second bypass vector that incorporates additional information that was not captured by the text encoder.
Our experiments demonstrate that using our \textit{textual bypass} can significantly speed up the convergence and enhance visual fidelity, reaching fidelity that is competitive with fine-tuning methods, all without needing to alter the weights of the generative model, see~\Cref{fig:teaser}. 

A general problem with personalization methods, including NeTI, is their inherent tradeoff between reconstruction quality and editability~\cite{tov2021designing,zhu2020improved, gal2023image}.
We investigate this phenomenon in the context of NeTI and propose two techniques to mitigate and control this issue.
First, we observe that the norms of existing token embeddings have a specific distribution while the learned mapping network can output embeddings deviating greatly from this distribution. We show that setting the norm of the network output to a constant value, taken from an existing token in \p, significantly improves the editability of our learned concept.
Next, we propose an extension to our method that allows one to control the balance between the reconstruction and editability of the concept \textit{at inference time}. 
This is achieved by imposing an importance-based ordering over our implicit representation. After training, gradually removing elements from our ordered representation allows one to control the reconstruction-editability tradeoff and reduces NeTI's required storage footprint.

We demonstrate the effectiveness of our \pstar space and NeTI over a range of concepts and prompts when compared to existing personalization approaches.
We additionally analyze \pstar and the attributes learned at different denoising timesteps. Finally, we demonstrate the appealing properties of our ordered representations for controlling the reconstruction-editability tradeoff at inference time.

\section{Related Work}

\paragraph{\textbf{Text-Guided Synthesis.}} 
Recent advancements in large-scale autoregressive models~\cite{ramesh2021zero,yu2022scaling}
and diffusion models~\cite{ho2020denoising,nichol2021improved,dhariwal2021diffusion} have resulted in unprecedented diversity and fidelity in visual content creation guided by a free-form text prompt~\cite{ramesh2022hierarchical,nichol2021glide,rombach2022high,saharia2022photorealistic,balaji2023ediffi}. While extremely expressive, these methods do not directly support the usage of user-specified concepts, resulting in research on \textit{inversion} and \textit{personalization} for diffusion models.

\vspace{-0.3cm}
\paragraph{\textbf{Inversion.}}
Image inversion is the process of finding a latent code that can be passed to a generator to reconstruct a given image~\cite{zhu2016generative,xia2022gan}.
In the context of diffusion models, inversion often refers to the process of finding an initial noise latent that can be iteratively denoised into the target image~\cite{dhariwal2021diffusion,ramesh2022hierarchical,mokady2022nulltext}. 
This initial noise latent can then be used for editing the given input image using text prompts~\cite{kawar2023imagic,pnpDiffusion2022,couairon2022diffedit,liew2022magicmix}, but is less suited for representing new personalized concepts.

\vspace{-0.3cm}
\paragraph{\textbf{Personalization.}}
In the task of personalization, we are interested in adapting a given model to better capture a given subject or concept. In the context of text-to-image synthesis, the personalized model should enable synthesizing novel images of a specific target concept using a free-form text prompt.  In~\cite{gal2023image, cohen2022my} it was first observed that personalization can be approached as an inversion problem where text embeddings are optimized to describe the target concept. Alternatively, other methods have resorted to fine-tuning the diffusion model directly~\cite{ruiz2022dreambooth,kumari2022customdiffusion,tewel2023keylocked} where one key distinction between these methods is the subset of the network which they choose to optimize.
A new line of work has recently explored encoder-based approaches for mapping a given concept to its textual representation~\cite{gal2023encoderbased,shi2023instantbooth,wei2023elite}. 
The personalization of text-to-image models has given rise to various downstream applications such as image editing~\cite{kawar2023imagic,valevski2022unitune} and personalized 3D generation~\cite{raj2023dreambooth3d,lin2023magic3d,richardson2023texture, metzer2022latent}. 

\vspace{-0.3cm}
\paragraph{\textbf{Spaces for Inversion and Personalization.}}
Numerous works have already analyzed the latent spaces of pretrained text-to-image diffusion models~\cite{kwon2023diffusion,Haas2023DiscoveringID,Park2023UnsupervisedDO,zhu2023boundary}. 
Most relevant to our work is the text-conditioning space of the pretrained text-to-image model. In Textual Inversion, Gal~\etal~\shortcite{gal2023image} invert a given concept into a single vector representation residing in the input space of the text encoder. This space has been recently termed the \p space. 
Voynov~\etal~\shortcite{voynov2023p} propose an extended \pplus latent space composed of a set of vectors $p\in$~\p, one for each layer of the U-Net denoising network. They demonstrate that this \textit{space}-dependent latent space results in improved reconstructions and higher editability compared to the smaller \p space. In the context of \textit{time}-dependent representation, previous works have demonstrated that using different inputs for different timesteps of a diffusion model has intriguing properties~\cite{liew2022magicmix,patashnik2023localizing} with Gal~\etal~\shortcite{gal2023encoderbased} also using a timestep-conditioned encoder. However, to the best of our knowledge, the resulting latent space was not directly investigated. In this work, we introduce the \pstar latent space that is dependent on both \textit{time} and \textit{space} by considering the different attention layers and the time-dependent nature of the denoising process. We further extend the inversion space using the \textit{textual bypass} which resides outside the text encoder input space used in all previous work.

\vspace{-0.1cm}
\paragraph{\textbf{Ordered Representations.}}
Ordered representations, such as principal component analysis (PCA), in which different dimensions have different degrees of importance, are widely used in machine learning and statistics. 
However, in the context of inversion and personalization spaces, this property is not commonly used~\cite{gal2023image,voynov2023p,abdal2019image2stylegan,abdal2020image2stylegan++}.
Rippel~\etal~\shortcite{rippel2014learning} showed that one can encourage a neural network to learn an ordered representation by simply introducing a special form of dropout on the hidden units of a neural network, which is proved to be an exact equivalence to PCA for the linear case. Inspired by their work, we propose to bring back this property by applying a similar technique to our learned representation, resulting in an ordered representation that can be used to achieve inference-time control of the reconstruction-editability tradeoff.

\begin{figure*}
    \centering
    \includegraphics[width=0.95\textwidth]{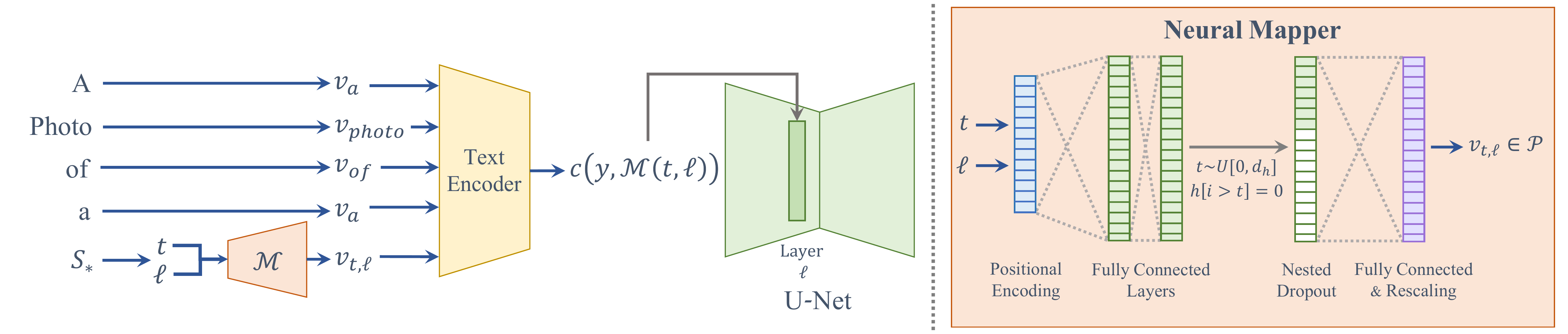}
    \\[-0.2cm]
    \caption{Method overview. Given a text prompt $p$ containing our concept $S_*$, we compute the token embedding $\mathcal{M}(t,\ell) = v_{t,\ell}\in$~\p representing the concept at the current timestep $t$ and U-Net layer $\ell$. This embedding, along with the embeddings of the other input tokens are passed to the pretrained text encoder to obtain the conditioning vector $c(y, \mathcal{M}(t,\ell)) \in \mathbb{R}^{N\times D}$. This conditioning is fed to the U-Net network at the $\ell$-th cross-attention layer. 
    During training, we optimize the parameters of the neural mapper $\mathcal{M}$, shown to the right, using a reconstruction objective (\Cref{eq:ours_objective}) while all other components remain fixed.
    }
    \vspace{-0.4cm}
    \label{fig:method}
\end{figure*}

\section{Preliminaries}

\paragraph{\textbf{Latent Diffusion Models.}}
We apply our inversion technique over the Stable Diffusion (SD) model~\cite{rombach2022high}.
In SD, an encoder $\mathcal{E}$ is trained to map an image $x\in\mathcal{X}$ into a spatial latent code $z = \mathcal{E}(x)$ while a decoder $\mathcal{D}$ is tasked with reconstructing the input image, i.e., $\mathcal{D}(\mathcal{E}(x)) \approx x$.
Given the trained autoencoder, a diffusion model is trained to produce latent codes within this learned latent space, and can be conditioned on an additional input vector $c(y)$ for some input prompt $y$. 
The training objective is given by: 
\begin{equation}~\label{eq:ldm}
    \mathcal{L} = \mathbb{E}_{z\sim\mathcal{E}(x),y,\varepsilon\sim\mathcal{N}(0,1),t} \left [ || \varepsilon - \varepsilon_\theta(z_t, t, c(y)) ||_2^2 \right ].
\end{equation}
Here, at each timestep $t$, the denoising network $\varepsilon_\theta$ is tasked with removing the noise added to the latent code given the noised latent $z_t$, the timestep $t$, and the conditioning vector $c(y)$.
To generate $c(y)$ the input prompt is first split into a series of $N$ pre-defined tokens, which are then mapped to $N$ corresponding embedding vectors, one for each token. These token embeddings are then passed to a pretrained CLIP text encoder~\cite{radford2021learning} which outputs a conditioning vector $c(y) \in \mathbb{R}^{N\times D}$ where $N=77$ is the number of input tokens and $D=768$ is the dimension of each output vector.

\vspace{-0.2cm}
\paragraph{\textbf{Textual Inversion.}}
In Textual Inversion, Gal~\etal~\shortcite{gal2023image} introduce a new token $S_*$ and a corresponding embedding vector $v_*\in$~\p representing the concept. Given a small set of images depicting the concept, they directly optimize $v_*$ to minimize the objective given in~\Cref{eq:ldm}. That is, their optimization objective is defined as: 
\begin{equation}
    v_* = \arg \min_v \mathbb{E}_{z,y,\varepsilon,t} \left [ || \varepsilon - \varepsilon_\theta(z_t, t, c(y, v)) ||_2^2 \right ],
\end{equation}
where the conditioning $c(y,v)$ is now obtained using the optimized embedding vector $v$ representing our concept. 
Notably, the entire LDM remains fixed and only $v$ is optimized. 

\setlength{\abovedisplayskip}{7.5pt}
\setlength{\belowdisplayskip}{7.5pt}

\section{Method}~\label{sec:method}
Similar to Textual Inversion, we are interested in finding a personalized representation for a user-provided concept depicted using a small set of images. 
Rather than encoding the concept into a single embedding vector $v\in$~\p we project our concept into a more expressive space, \pstar, dependent on both the denoising timestep $t$ and the U-Net layer $\ell$ to which the conditioning vector is injected. 
Instead of directly optimizing all possible $v_{t,\ell}$ vectors that compose our concept, we choose to implicitly represent them via a simple neural mapper $\mathcal{M}$. Our mapper receives the current timestep $t$ and U-Net layer $\ell$ and outputs the corresponding token embedding $v_{t,\ell} \in$~\p representing the concept at the current timestep and layer. This embedding is then passed to the matching U-Net layer $\ell$, see~\Cref{fig:method}.

To train the neural mapper, we follow a similar optimization scheme to that of Textual Inversion but directly optimize the \textit{parameters} of the mapper. Formally, the objective is defined as:
\begin{align}~\label{eq:ours_objective}
\begin{split}
    \arg \min_\mathcal{M} \mathbb{E}_{z,y,\varepsilon,t,\ell} & \left [ || \varepsilon - \varepsilon_\theta(z_t, t, c(y,\mathcal{M}(t, \ell))) ||_2^2 \right ].
\end{split}
\end{align}
During training, we randomly sample timesteps $t \in [0,1000)$ and compute the above objective for $16$ different cross-attention layers $\{\ell_1,\dots,\ell_{16}\}$ of the denoising U-Net network at various resolutions, following Voynov~\etal~\shortcite{voynov2023p}. 
Intuitively, in the above formulation, after training, the entire target concept is encoded into the learned parameters of the neural mapper. 

At inference time, novel compositions of the concept can be created by adding the token $S_*$ to any text prompt. The trained neural mapper is queried to obtain the concept's learned token embedding $v_{t,\ell}$ for each combination of timestep $t=50,\dots,1$ and U-Net layer $\{\ell_1,\dots,\ell_{16}\}$. These embeddings are then passed to the text encoder to obtain the conditioning vector passed to the U-Net.

\subsection{Network Architecture}
Our neural mapper is illustrated on the right of~\Cref{fig:method}. The mapper receives as input a pair of scalars $(t,\ell)$ denoting the current timestep and U-Net layer. 
First, this input is passed through a positional encoding function $f(\cdot)$, discussed below, to transform the input into a high-dimensional vector $f(t,\ell)$, followed by two fully-connected layers.
Next, we optionally apply a Nested Dropout~\cite{rippel2014learning} technique over the hidden representation to impose an importance-based ordering over the learned representation, see~\Cref{sec:nested_dropout}.
The resulting compressed vector is passed through a final fully-connected layer to obtain a $768$-dimensional vector $v_{t,\ell}\in$~\p representing the concept at the current timestep and layer. 
In total, the resulting architecture contains approximately $460,000$ trainable parameters, which amounts to 2MB of disk space. As a reference to fine-tuning methods, DreamBooth~\cite{ruiz2022dreambooth} requires several GBs of disk space per concept with CustomDiffusion~\cite{kumari2022customdiffusion} requiring $\sim$75MB of disk space.

\vspace{-0.25cm}
\paragraph{\textbf{Positional Encoding}}
To introduce an inductive bias with respect to the timestep and U-Net layer, we apply a positional encoding on the input $(t,\ell)$. Specifically, each input $(t, \ell)$ is encoded with Random Fourier Features~\cite{rahimi2007random, tancik2020fourfeat} into a $2048$-dimensional vector, $f(t,\ell)\in\mathbb{R}^{2048}$, modulated by $1024$ random frequencies.
We then define a set of $160$ uniformly spaced anchor pairs $(t,\ell)$, encoded using $f$. This set of vectors is then used to form an encoding matrix $E\in\mathbb{R}^{160\times2048}$. 
The output of our positional encoding is then defined as $e_{t,\ell} = E \times f(t,\ell) \in \mathbb{R}^{160}$.

We observe that biasing the encoding toward nearby layers produces less favorable results. Hence, we choose the random frequencies such that the encodings are smooth with respect to time and well separated with respect to the U-Net layer. Additional details, an ablation study, and a visualization of our positional encoding are provided in~\Cref{sec:additional_details,sec:ablation_study,sec:additional_analysis}.

\begin{figure}
    \centering
    \setlength{\tabcolsep}{0.5pt}
    \addtolength{\belowcaptionskip}{-12.5pt}
    {\small
    \begin{tabular}{c@{\hspace{0.25cm}} c c@{\hspace{0.25cm}} c c}

        & \multicolumn{1}{c}{\begin{tabular}{c} w/o rescale \end{tabular}} &
        \multicolumn{1}{c}{\begin{tabular}{c} w/ rescale \end{tabular}} &
        \multicolumn{1}{c}{\begin{tabular}{c}  w/o rescale  \end{tabular}} &
        \multicolumn{1}{c}{\begin{tabular}{c} w/ rescale \end{tabular}} \\

        \includegraphics[width=0.0825\textwidth]{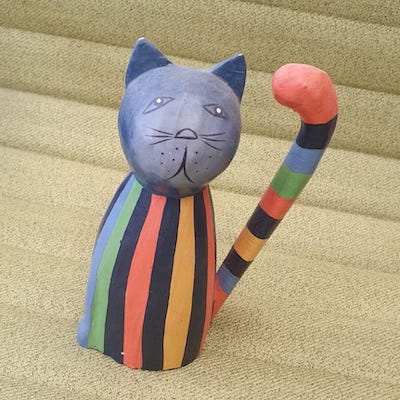} &
        \includegraphics[width=0.0825\textwidth]{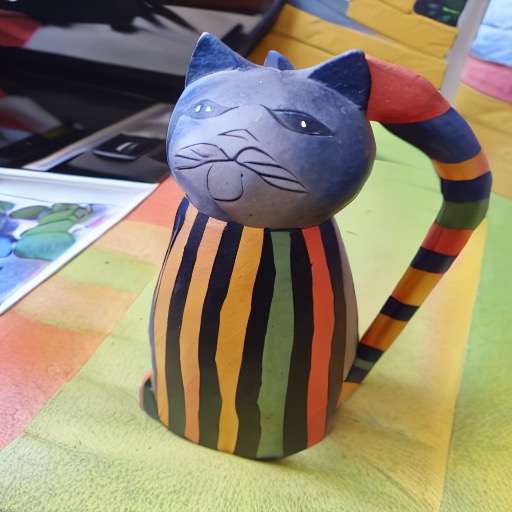} &
        \includegraphics[width=0.0825\textwidth]{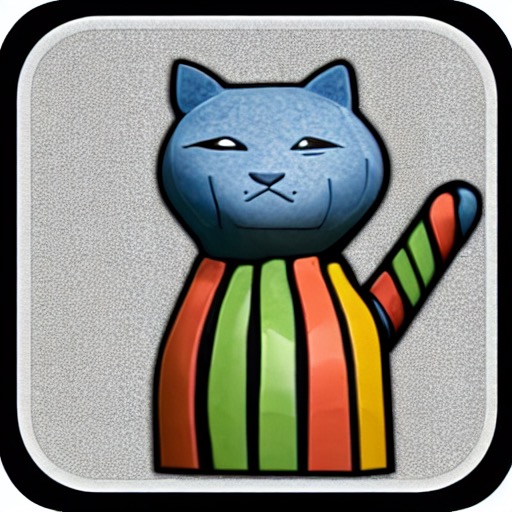} &
        \hspace{0.05cm}
        \includegraphics[width=0.0825\textwidth]{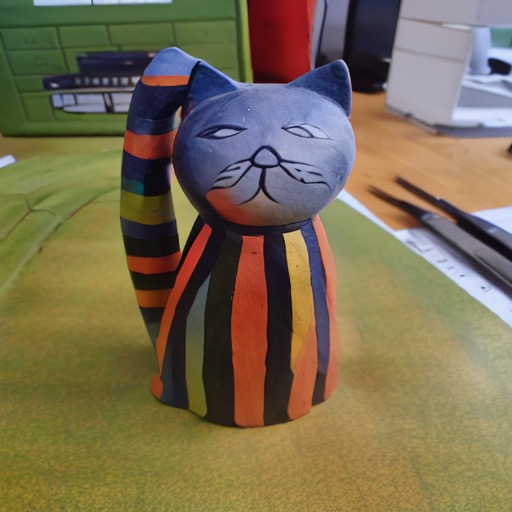} &
        \includegraphics[width=0.0825\textwidth]{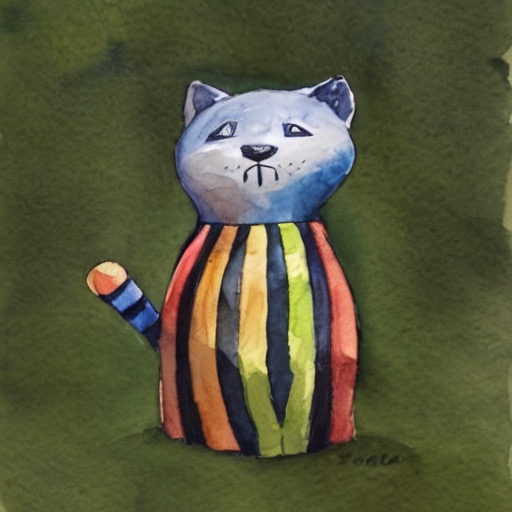} \\

        \begin{tabular}{c} Real \end{tabular} & \multicolumn{2}{c}{\begin{tabular}{c} ``An app icon...'' \end{tabular}} &
        \multicolumn{2}{c}{\begin{tabular}{c} ``A watercolor painting...'' \end{tabular}} \\

        \includegraphics[width=0.0825\textwidth]{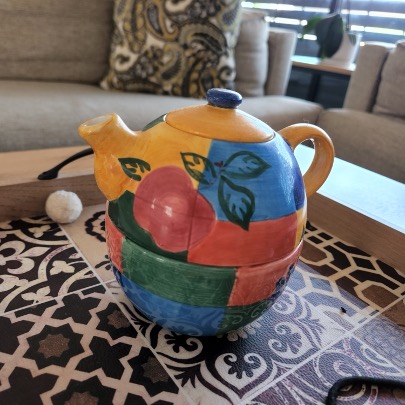} &
        \includegraphics[width=0.0825\textwidth]{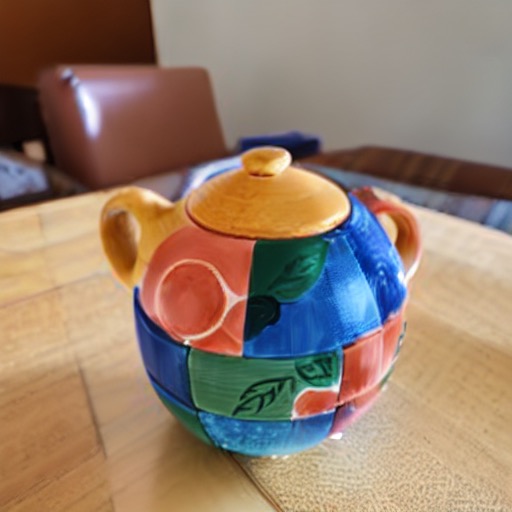} &
        \includegraphics[width=0.0825\textwidth]{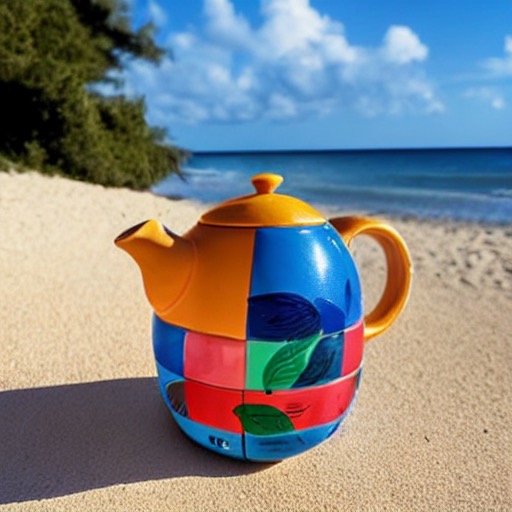} &
        \hspace{0.05cm}
        \includegraphics[width=0.0825\textwidth]{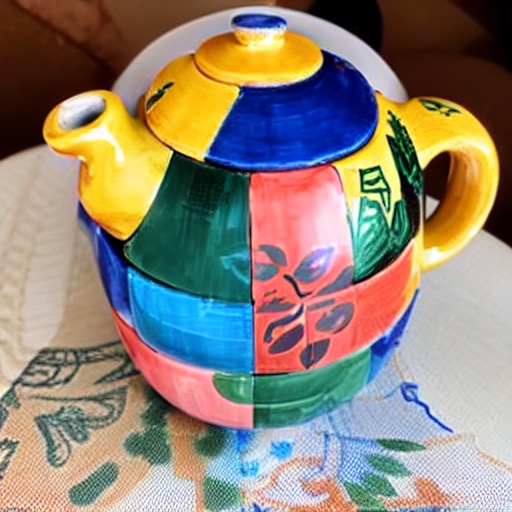} &
        \includegraphics[width=0.0825\textwidth]{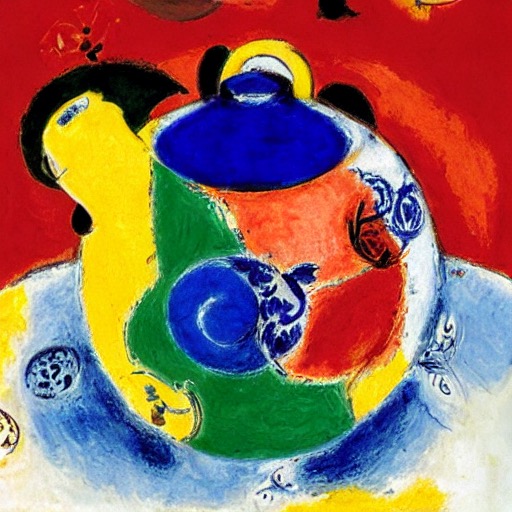} \\

        \begin{tabular}{c} Real \end{tabular} & \multicolumn{2}{c}{\begin{tabular}{c} ``$S_*$ on a beach'' \end{tabular}} &
        \multicolumn{2}{c}{\begin{tabular}{c} ``Marc Chagall painting...'' \end{tabular}}

    \\[-0.3cm]        
    \end{tabular}
    }
    \caption{Generation results when training with and without our rescaling technique. As can be seen, applying rescaling improves editability without harming visual fidelity.}
    \label{fig:output_rescaling}
\end{figure}

\vspace{-0.325cm}
\paragraph{\textbf{Output Rescaling}}
During optimization, the outputs of the neural mapper are unconstrained, resulting in representations that may reside far away from the true distribution of token embeddings typically passed to the text encoder. 
We find that such unnatural representations significantly harm the editability of the learned concept. 
To mitigate this issue, we find that it is enough to rescale the \textit{norm} of the network output to match the norm of real token embeddings. Specifically, we set the norm of the network output to be equal to the norm of the embedding of the concept's ``super-category'' token (e.g., for the second example in~\Cref{fig:output_rescaling}, we set the norm equal to the norm of ``teapot''). Formally, the normalized output of the network is given by: 
\begin{equation}
    \mathcal{M}'(t,\ell) = \frac{\mathcal{M}(t,\ell)}{||\mathcal{M}(t,\ell)||} ||v_{super}||
\end{equation}
where $v_{super}$ is the embedding of the ``super-category'' token.
As shown in~\Cref{fig:output_rescaling}, this simple normalization results in a large improvement in editability without harming reconstruction. 

\subsection{Imposing an Ordered Representation}~\label{sec:nested_dropout}
A common characteristic of inversion and personalization methods is the existence of a tradeoff between the reconstruction quality and the editability of the inverted concept, making it challenging to find an embedding that is both highly accurate and can still be edited with complex prompts. 
We observe that the dimensionality $d_h$ of the last hidden layer $h$ in our mapper greatly affects this tradeoff.
Notice that $d_h$ determines the number of $768$-dimensional vectors present in our last linear layer, where a larger $d_h$ allows the model to better ``overfit'' to the training images, resulting in more accurate reconstructions but at the possible cost of reduced editability.

Theoretically, one can train multiple neural mappers with different representation sizes and choose the one that bests balances reconstruction and editability for each concept. However, doing so is both cumbersome and impractical at scale. 
Instead, we introduce an importance-based ordering over our final hidden layer $h$ that allows for post-hoc control of the representation's dimensionality.
This is achieved by applying a variant of the Nested Dropout technique proposed in Rippel~\etal~\shortcite{rippel2014learning} over the output of $h$. 
Specifically, we uniformly sample a truncation value $t$ and zero out the outputs of $h$ above the truncation value,
\begin{align}~\label{eq:dropout}
    h[i > t] = 0 \quad \text{where}\,\, t \,\, \sim U(0, d_h],
\end{align}
see the right-hand side of ~\Cref{fig:method}. 
Zeroing out a subset of $h$ effectively means the corresponding set of $768$-dimensional vectors from the last linear layer will not be used for that specific truncation $t$. This has the same effect as dynamically changing $d_h$ without changing the model architecture itself. %
By randomly sampling the truncation values during training we encourage the network to be robust to different dimensionality sizes and encode more information into the first set of output vectors, which naturally have a lower truncation frequency. 

We note that our dropout mechanism is a simpler variant of Rippel~\etal~\shortcite{rippel2014learning} which focused on a retrieval setting and used a different sampling distribution with an additional sweeping mechanism. In~\Cref{sec:analysis}, we show that manually setting the truncation value $t$ during inference offers a new way to traverse the reconstruction-editability tradeoff for personalization. There, users are provided an added axis of control, enabling them to dynamically choose a truncation that best suits their concept and target prompt.
Importantly, as different concepts vary in their complexity, our method allows us to train all concepts with a single architecture and easily compress the learned representation by simply dropping the redundant vectors after training.

\begin{figure}
    \centering
    \includegraphics[width=0.475\textwidth]{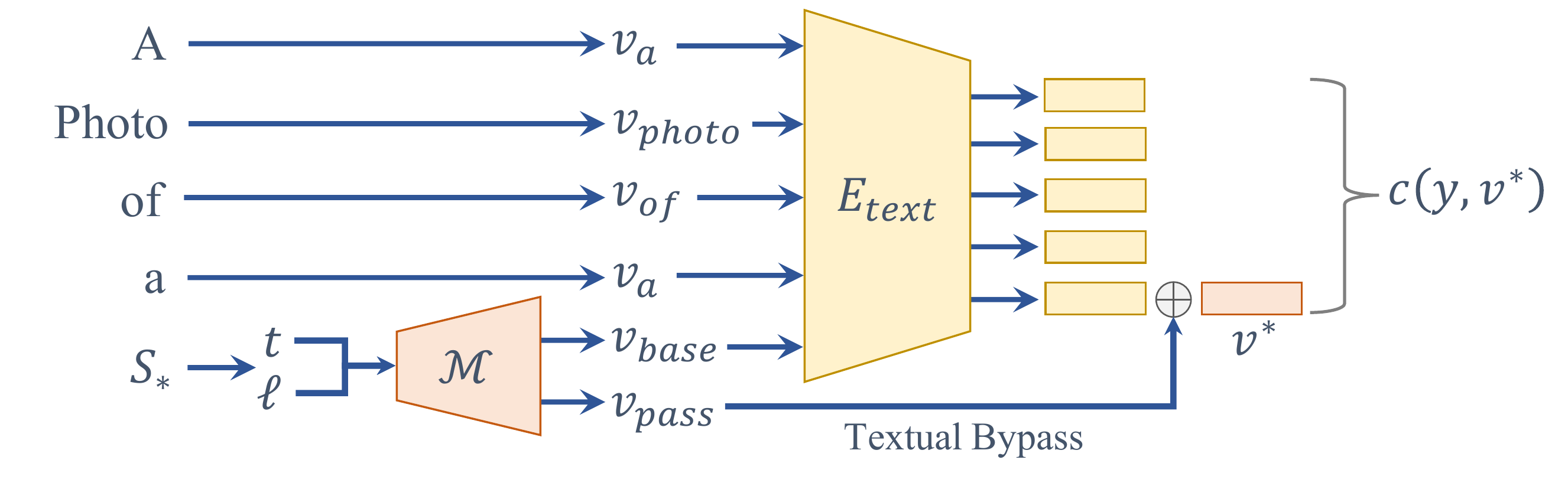}
    \\[-0.2cm]
    \caption{The textual bypass. Our mapper $\mathcal{M}$ outputs two vectors, $v_{base}$ and $v_{pass}$. As before, $v_{base}$ is passed to the text encoder $E_{text}$ along with the other token embeddings. Then, $v_{pass}$ is added to the \textit{output} of the text encoder corresponding to $S_*$. Doing so allows us to pass additional information directly to the U-Net that was not captured by the text encoder.}
    \label{fig:textual_bypass}
    \vspace{-0.2cm}
\end{figure}

\subsection{The Textual Bypass}~\label{sec:textual_bypass}
The outputs of our neural mapper reside in the \p space of the text encoder and every vector returned by our mapper is first passed through the text encoder alongside the rest of the prompt tokens. 
Inverting a concept directly into the U-Net's input space, without going through the text encoder, could potentially lead to much quicker convergence and more accurate reconstructions. 
However, doing so is not a practical solution by itself, as skipping the text encoder limits our ability to create new compositions since our concept is not seen by the text encoder alongside the other prompt tokens. 
Instead, we propose to learn two vectors using our neural mapper. The first vector, $v_{base}$ is fed into the text encoder, as presented above, and can therefore affect and be affected by the other tokens. 
We denote the corresponding output of the text encoder by $E_{text}(v_{base}) \in \mathbb{R}^{768}$. 

The second vector, $v_{pass}$ is a \textit{textual bypass} vector which is added as a residual to the output of the text encoder and can incorporate additional information directly into the U-Net. To avoid $v_{pass}$ from becoming too dominant we scale it to match the norm of $E_{text}(v_{base})$. Our final representation passed to the U-Net cross-attention layer is then defined as:
\begin{align}~\label{eq:dropout}
    v^* = E_{text}(v_{base}) + \alpha \frac{v_{pass}}{||v_{pass}||}  ||E_{text}(v_{base})||,
\end{align}
where we set $\alpha=0.2$ in all experiments. \Cref{fig:textual_bypass} presents our modified neural mapper with our textual bypass.

\begin{figure}
    \centering
    \renewcommand{\arraystretch}{0.3}
    \setlength{\tabcolsep}{0.5pt}

    {\small
    \begin{tabular}{c@{\hspace{0.15cm}} c c @{\hspace{0.1cm}} c c}

        \begin{tabular}{c} Real \end{tabular} &
        \multicolumn{2}{c}{Results for $v_{base}$} &
        \multicolumn{2}{c}{Results with $v_{pass}$} \\
        \\
        \includegraphics[width=0.085\textwidth]{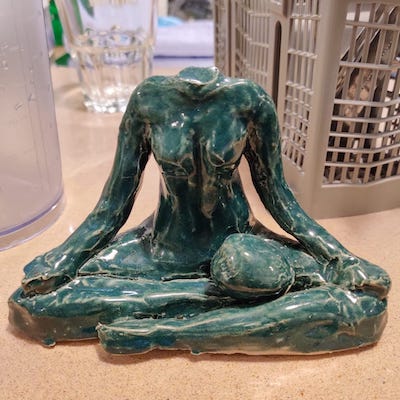} &
        \includegraphics[width=0.085\textwidth]{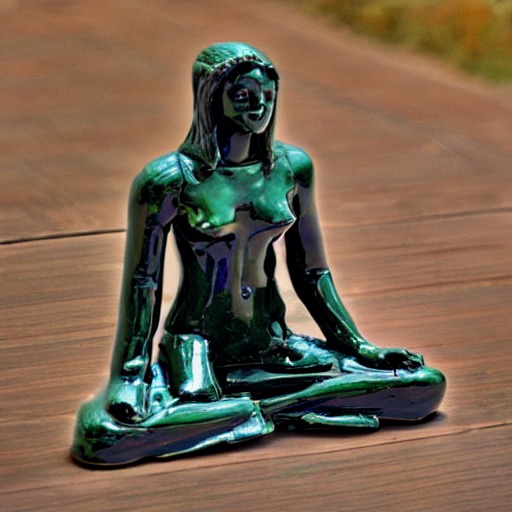} &
        \includegraphics[width=0.085\textwidth]{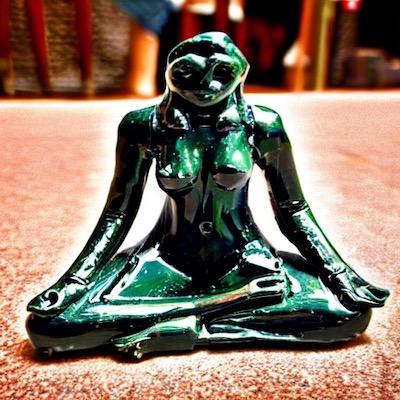} &
        \hspace{0.015cm}
        \includegraphics[width=0.085\textwidth]{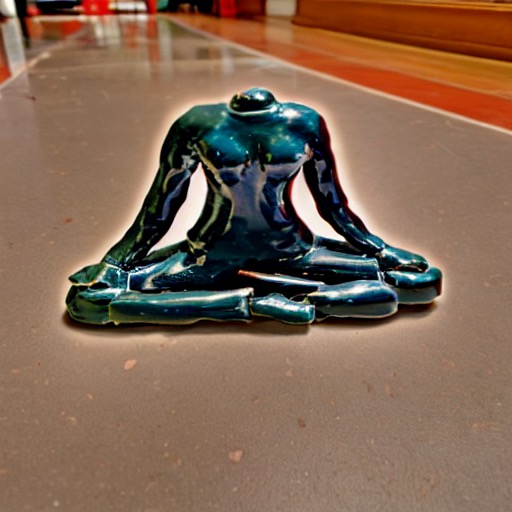} &
        \includegraphics[width=0.085\textwidth]{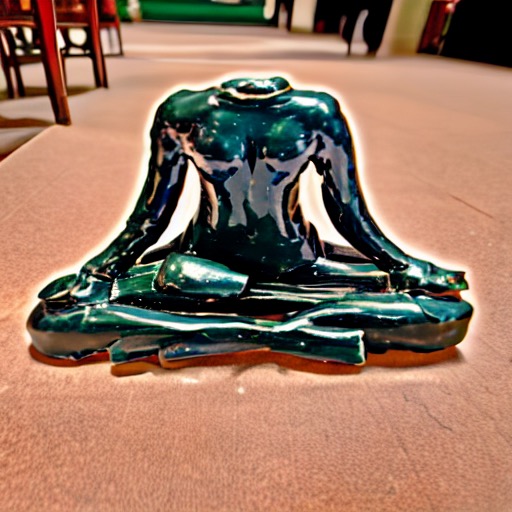} \\

        \includegraphics[width=0.085\textwidth]{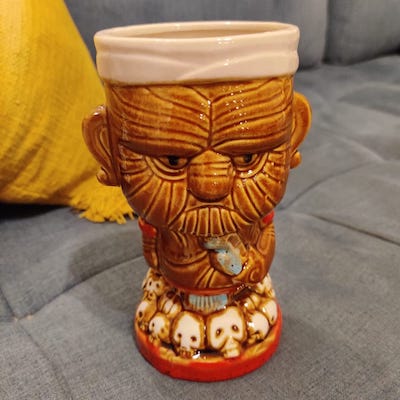} &
        \includegraphics[width=0.085\textwidth]{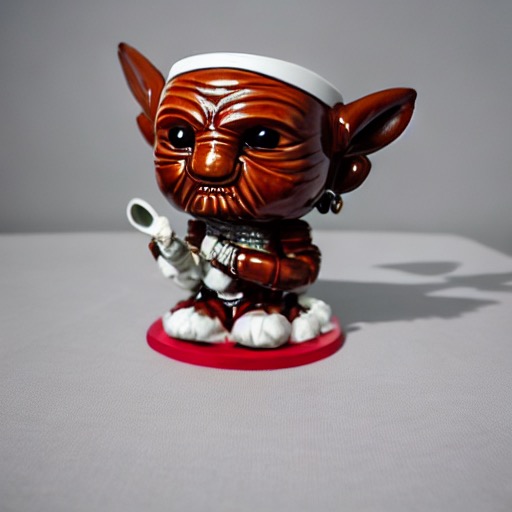} &
        \includegraphics[width=0.085\textwidth]{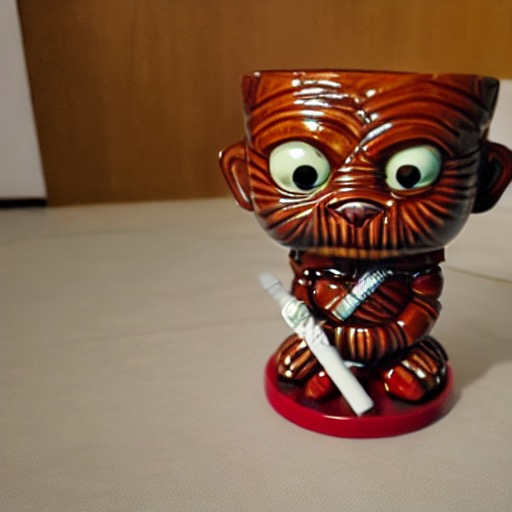} &
        \hspace{0.015cm}
        \includegraphics[width=0.085\textwidth]{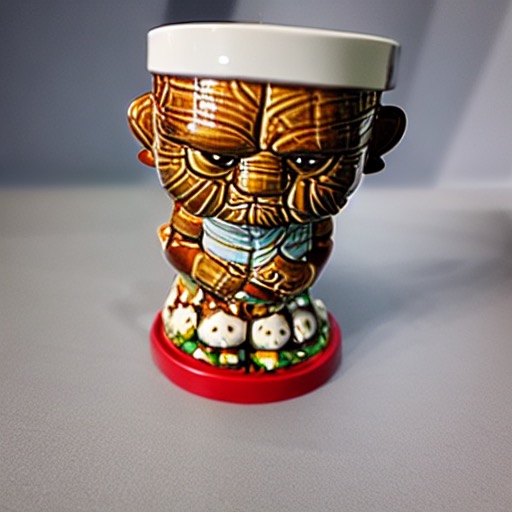} &
        \includegraphics[width=0.085\textwidth]{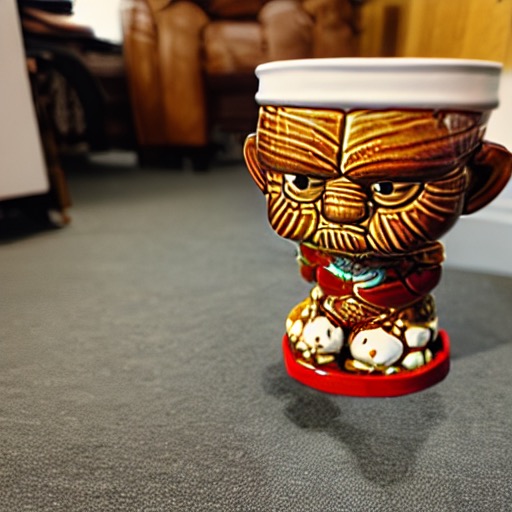} \\ \\

    \end{tabular}
    \\[-0.4cm]
    }
    \caption{When trained with \textit{textual bypass}, $v_{base}$ learns to reconstruct the coarse concept details with the bypass vector $v_{pass}$ further refining the results.}
    \vspace{-0.1cm}
    \label{fig:textual_bypass_ablation_main_paper}
\end{figure}

Intuitively our textual bypass should only affect the appearance of the learned concept and not its position or style which should be mostly governed by the text prompt. 
Thus, inspired by the recent key-locking mechanism of Tewel~\etal~\shortcite{tewel2023keylocked}, we use $v^*$ only as input for the values ($V$) of the cross-attention of the U-Net, while the inputs of the keys ($K$) are still set using the standard $v_{base}$ vector. 
Note that our approach differs from the mechanism of Tewel~\etal~\shortcite{tewel2023keylocked} where the keys are set using a fixed, manually determined word. Yet, our approach borrows from the same intuition regarding the roles of keys and values in cross-attention layers, as also discussed in~\cite{patashnik2023localizing}.
In ~\Cref{fig:textual_bypass_ablation_main_paper}, we illustrate the different aspects of the concept captured by $v_{base}$ and $v_{pass}$, showing how $v_{pass}$ is used by our mapper to refine the result of $v_{base}$.

\begin{figure*}
    \centering
    \setlength{\tabcolsep}{0.1pt}
    {\footnotesize
    \begin{tabular}{c@{\hspace{0.25cm}} c@{\hspace{0.25cm}} c@{\hspace{0.25cm}} c@{\hspace{0.25cm}} c@{\hspace{0.25cm}} c@{\hspace{0.25cm}} c@{\hspace{0.25cm}}}

        \includegraphics[width=0.13\textwidth]{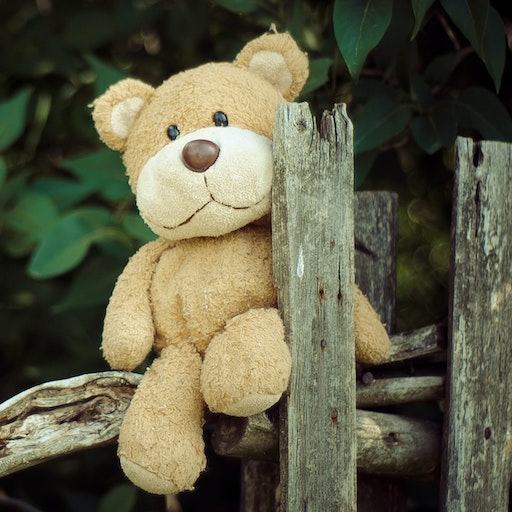} &
        \includegraphics[width=0.13\textwidth]{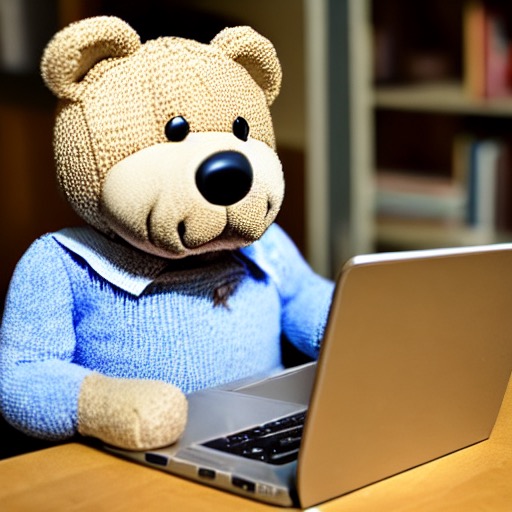} &
        \includegraphics[width=0.13\textwidth]{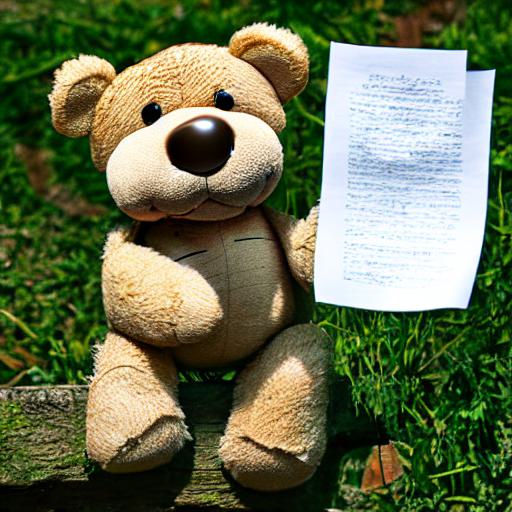} &
        \includegraphics[width=0.13\textwidth]{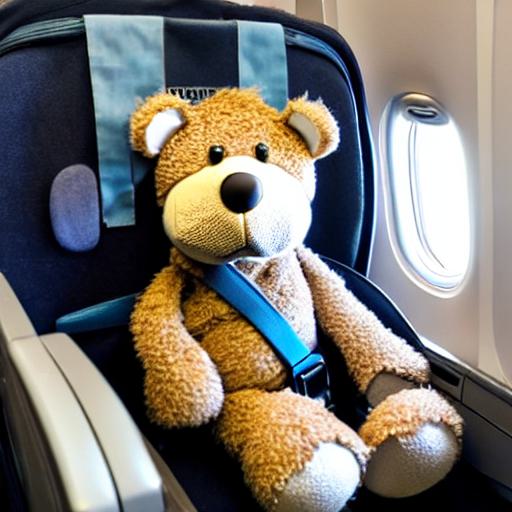} &
        \includegraphics[width=0.13\textwidth]{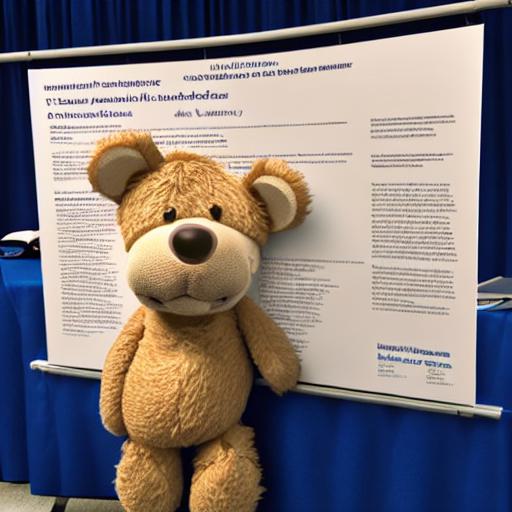} &
        \includegraphics[width=0.13\textwidth]{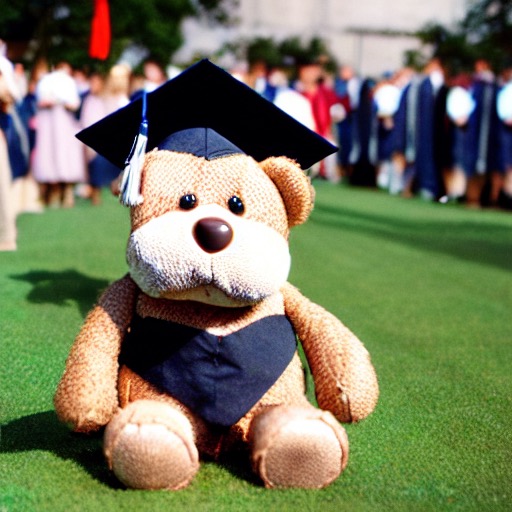} \\

        Real Sample &
        \begin{tabular}{c} ``A photo of $S_*$ writing a \\ paper on a laptop'' \end{tabular} &
        \begin{tabular}{c} ``$S_*$ holding up \\ his accepted paper'' \end{tabular} &
        \begin{tabular}{c} ``$S_*$ buckled in his \\ seat on a plane'' \end{tabular} &
        \begin{tabular}{c} ``$S_*$ presenting a \\ poster at a conference'' \end{tabular} &
        \begin{tabular}{c} ``A photo of $S_*$ graduating \\ after finishing his PhD'' \end{tabular} \\ \\[-0.185cm]

        \includegraphics[width=0.13\textwidth]{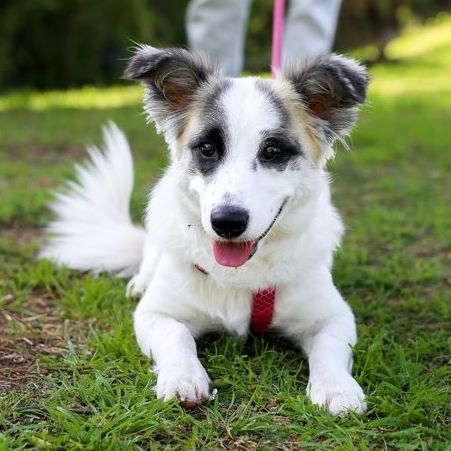} &
        \includegraphics[width=0.13\textwidth]{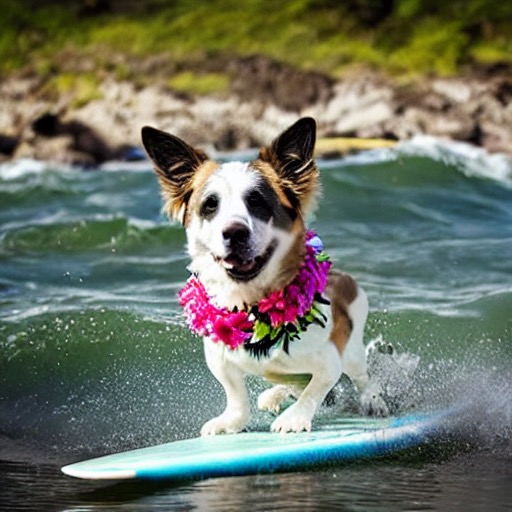} &
        \includegraphics[width=0.13\textwidth]{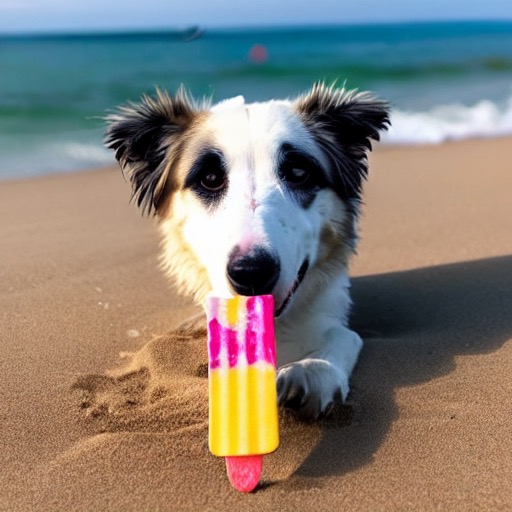} &
        \includegraphics[width=0.13\textwidth]{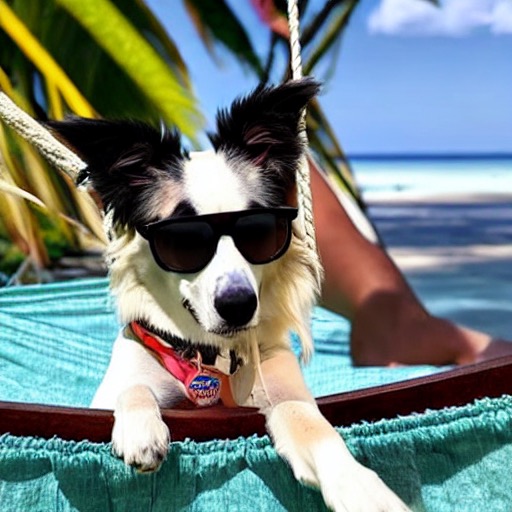} &
        \includegraphics[width=0.13\textwidth]{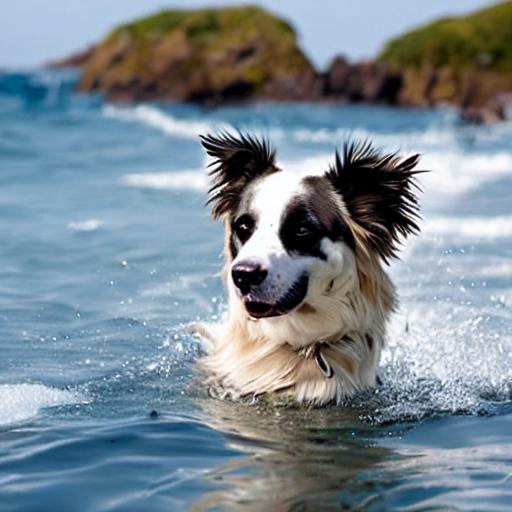} &
        \includegraphics[width=0.13\textwidth]{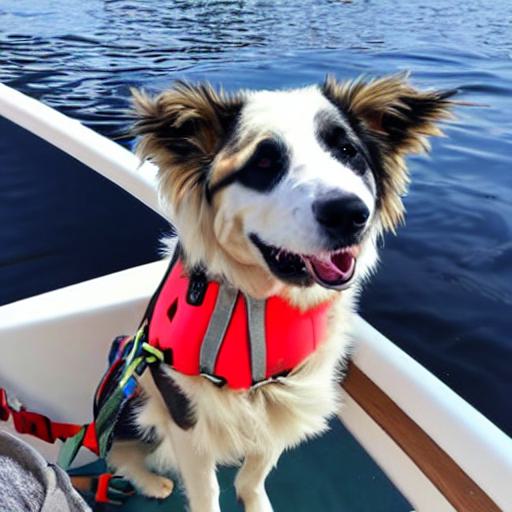} \\

        Real Sample &
        \begin{tabular}{c} ``A photo of $S_*$ \\ surfing on a wave, \\ wearing a floral lei'' \end{tabular} &
        \begin{tabular}{c} ``A photo of $S_*$ eating \\ a popsicle on the \\ beach'' \end{tabular} &
        \begin{tabular}{c} ``$S_*$ lounging in a \\ hammock on a tropical \\ beach, wearing \\ sunglasses'' \end{tabular} &
        \begin{tabular}{c} ``A photo of  $S_*$ \\ swimming in the \\ ocean'' \end{tabular} &
        \begin{tabular}{c} ``A photo of $S_*$ \\ wearing a life jacket \\ on a boat'' \end{tabular} \\ \\[-0.185cm]

        \includegraphics[width=0.13\textwidth]{images/original/mugs_skulls.jpeg} &
        \includegraphics[width=0.13\textwidth]{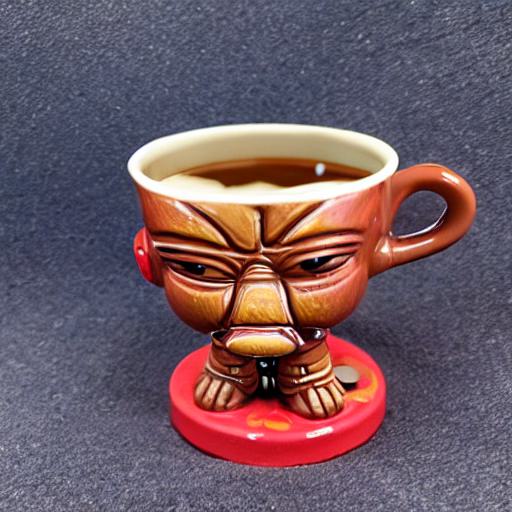} &
        \includegraphics[width=0.13\textwidth]{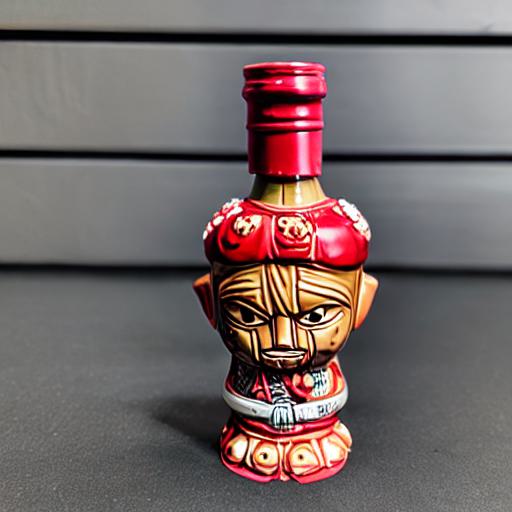} &
        \includegraphics[width=0.13\textwidth]{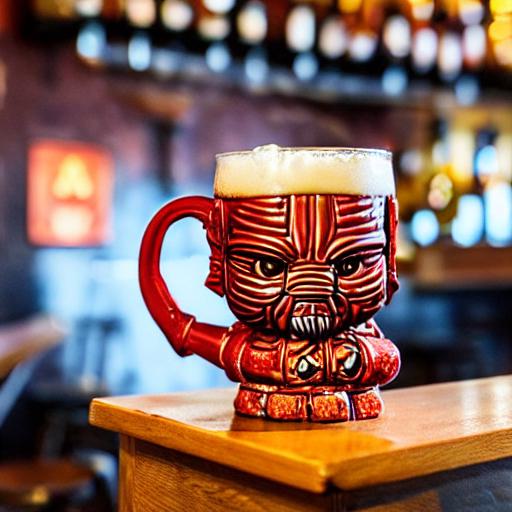} &
        \includegraphics[width=0.13\textwidth]{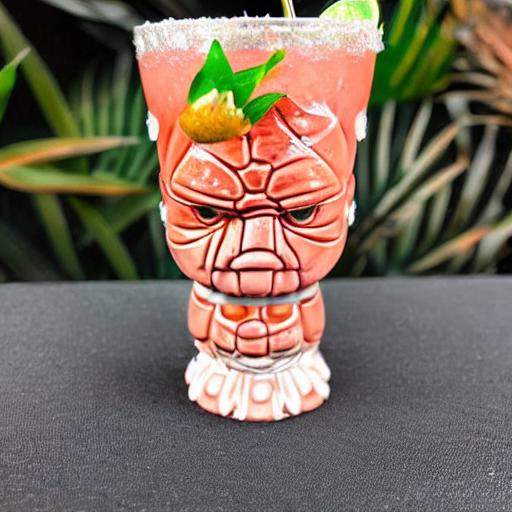} &
        \includegraphics[width=0.13\textwidth]{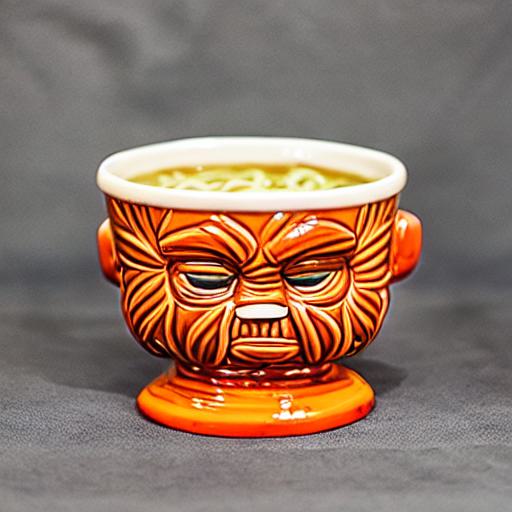} \\

        Real Sample &
        \begin{tabular}{c} ``A photo of a $S_*$ \\ espresso cup'' \end{tabular} &
        \begin{tabular}{c} ``A photo of a $S_*$ \\ wine bottle'' \end{tabular} &
        \begin{tabular}{c} ``A $S_*$ beer \\ pitcher in a bar'' \end{tabular} &
        \begin{tabular}{c} ``A $S_*$ tropical \\ cocktail glass'' \end{tabular} &
        \begin{tabular}{c} ``Chicken noodle soup \\ served in a $S_*$ bowl'' \end{tabular} \\

    \\[-0.4cm]        
    \end{tabular}
    }
    \vspace{-0.15cm}
    \caption{Sample text-guided personalized generation results obtained with NeTI.}
    \label{fig:our_results}
    \vspace{-0.1cm}
\end{figure*}
\begin{figure*}
    \centering
    \renewcommand{\arraystretch}{0.3}
    \setlength{\tabcolsep}{0.1pt}
    {\footnotesize
    \hspace*{-0.3cm}
    \begin{tabular}{c@{\hspace{0.3cm}} c c @{\hspace{0.3cm}} c c @{\hspace{0.3cm}} c c @{\hspace{0.3cm}} c c @{\hspace{0.3cm}}}

        \begin{tabular}{c} Real Sample \end{tabular} &
        \multicolumn{2}{c}{Textual Inversion (TI)} &
        \multicolumn{2}{c}{DreamBooth} &
        \multicolumn{2}{c}{Extended Textual Inversion} &
        \multicolumn{2}{c}{NeTI} \\

        \includegraphics[width=0.09325\textwidth]{images/original/rainbow_cat.jpeg} &
        \hspace{0.2cm}
        \includegraphics[width=0.09325\textwidth]{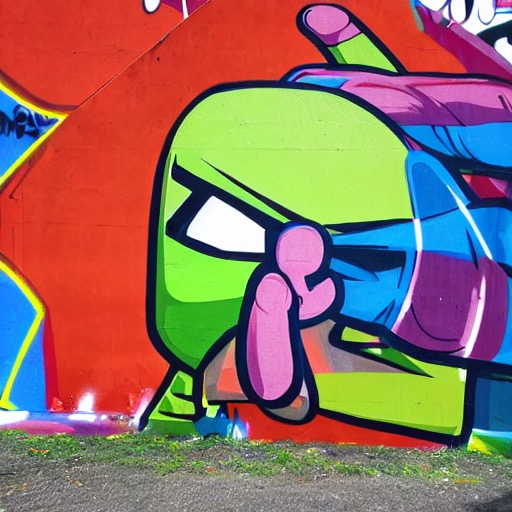} &
        \includegraphics[width=0.09325\textwidth]{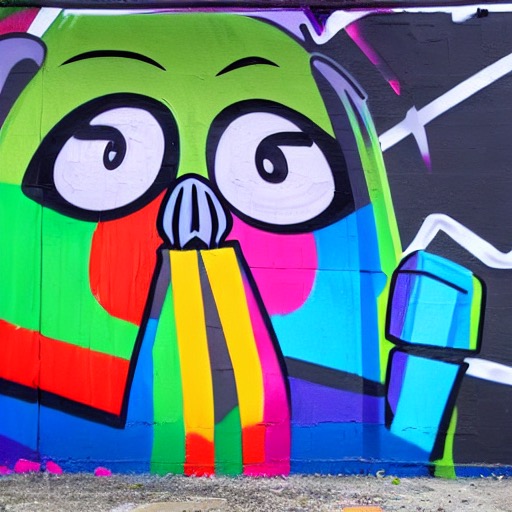} &
        \hspace{0.2cm}
        \includegraphics[width=0.09325\textwidth]{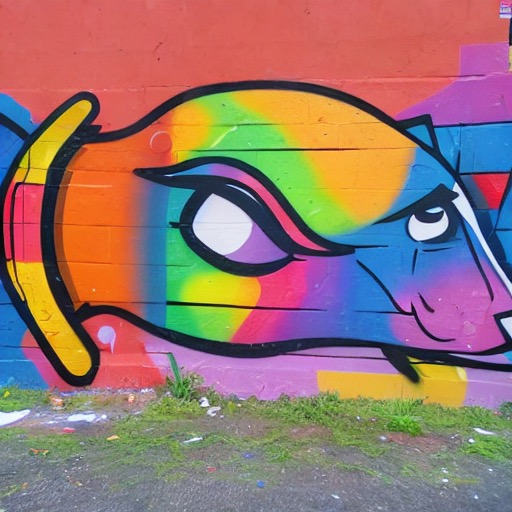} &
        \includegraphics[width=0.09325\textwidth]{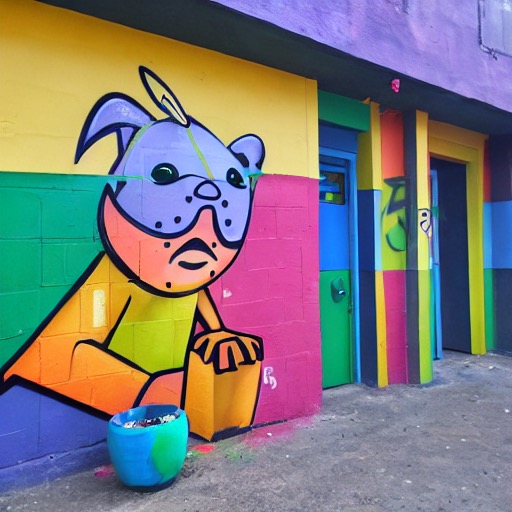} &
        \hspace{0.2cm}
        \includegraphics[width=0.09325\textwidth]{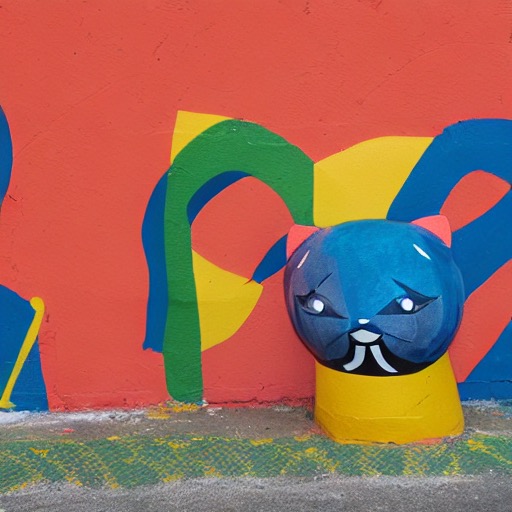} &
        \includegraphics[width=0.09325\textwidth]{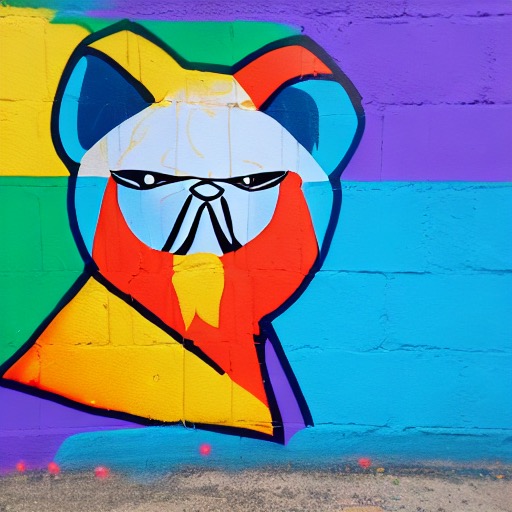} &
        \hspace{0.2cm}
        \includegraphics[width=0.09325\textwidth]{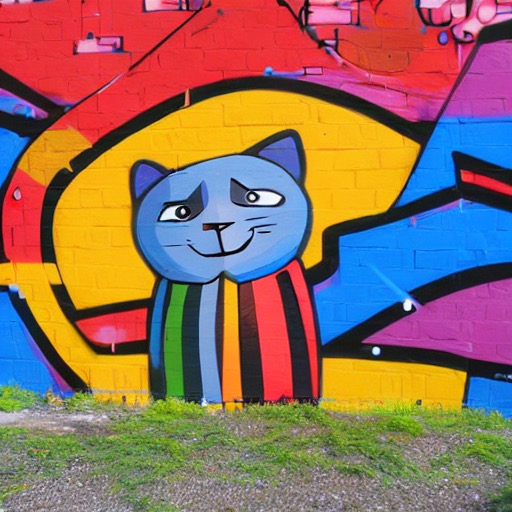} &
        \includegraphics[width=0.09325\textwidth]{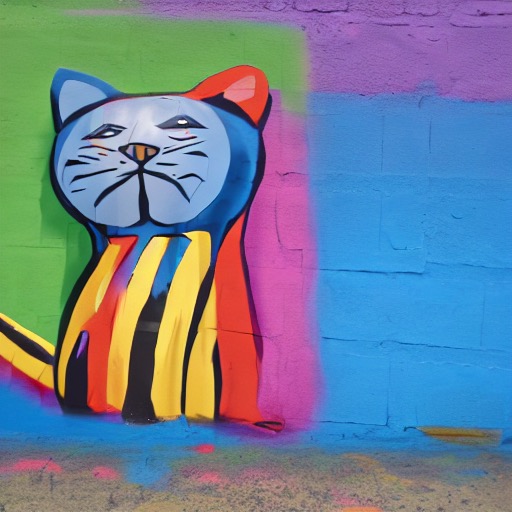} \\

        \raisebox{0.3in}{\begin{tabular}{c} ``A colorful \\ \\[-0.05cm] graffiti \\ \\[-0.05cm] of $S_*$''\end{tabular}} &
        \hspace{0.2cm}
        \includegraphics[width=0.09325\textwidth]{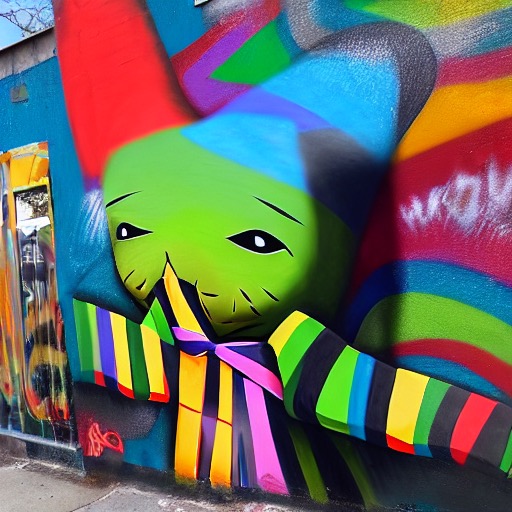} &
        \includegraphics[width=0.09325\textwidth]{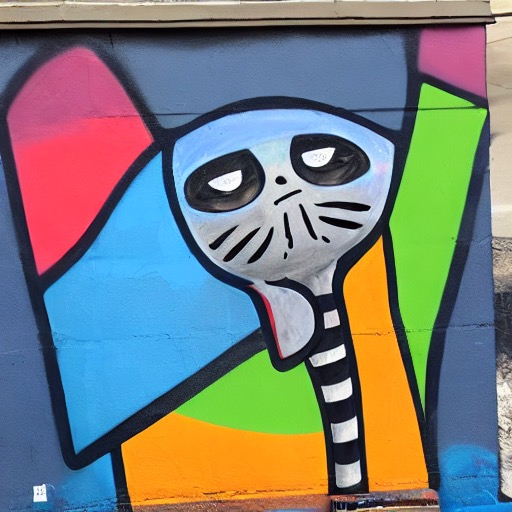} &
        \hspace{0.2cm}
        \includegraphics[width=0.09325\textwidth]{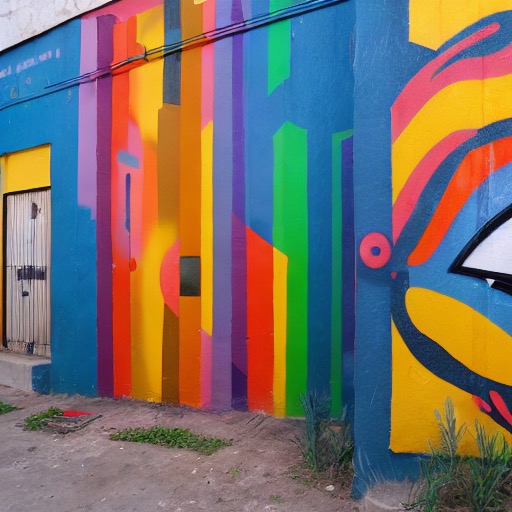} &
        \includegraphics[width=0.09325\textwidth]{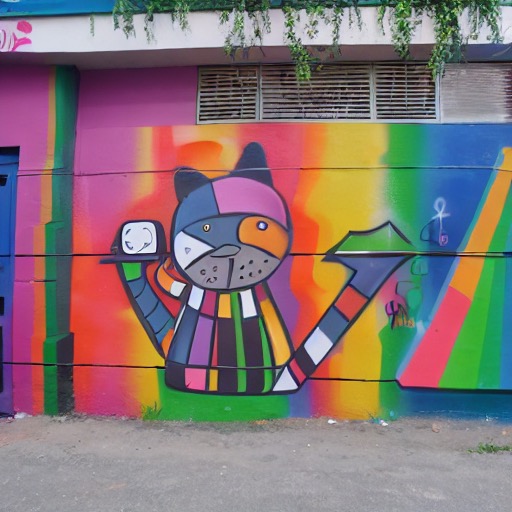} &
        \hspace{0.2cm}
        \includegraphics[width=0.09325\textwidth]{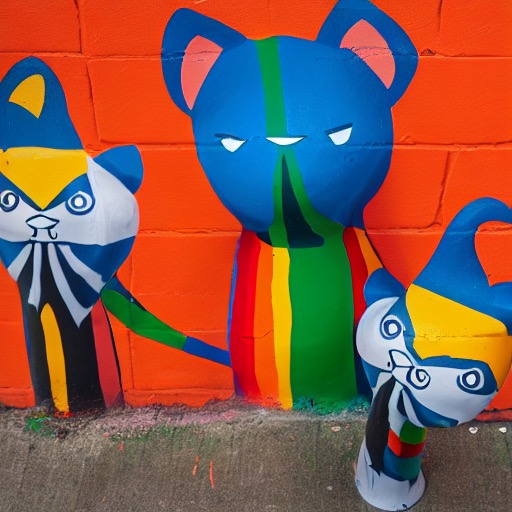} &
        \includegraphics[width=0.09325\textwidth]{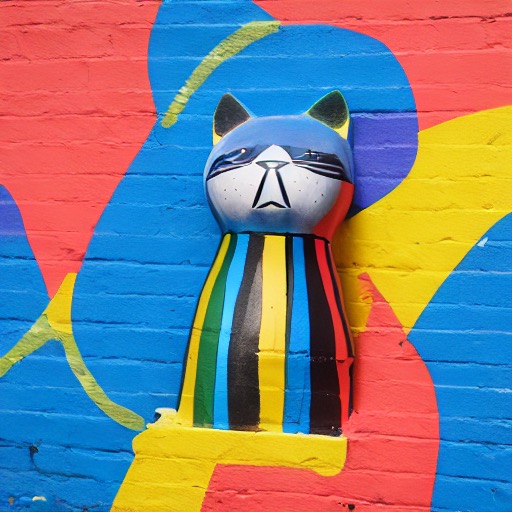} &
        \hspace{0.2cm}
        \includegraphics[width=0.09325\textwidth]{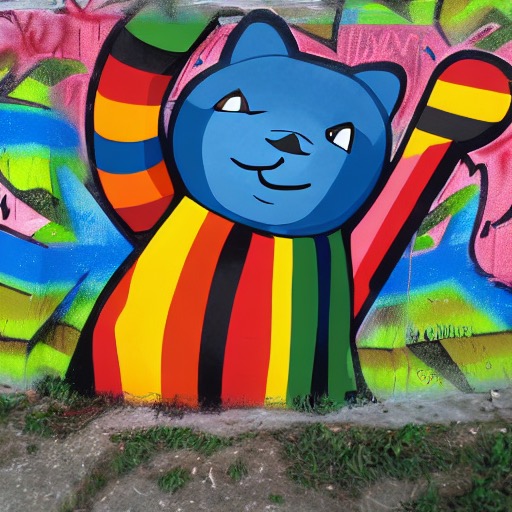} &
        \includegraphics[width=0.09325\textwidth]{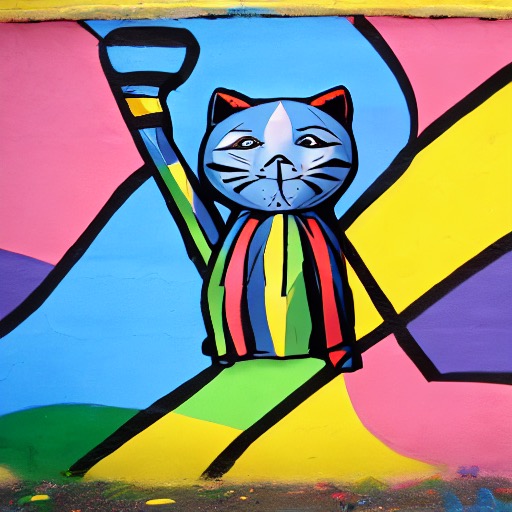} \\ \\
        
        \includegraphics[width=0.09325\textwidth]{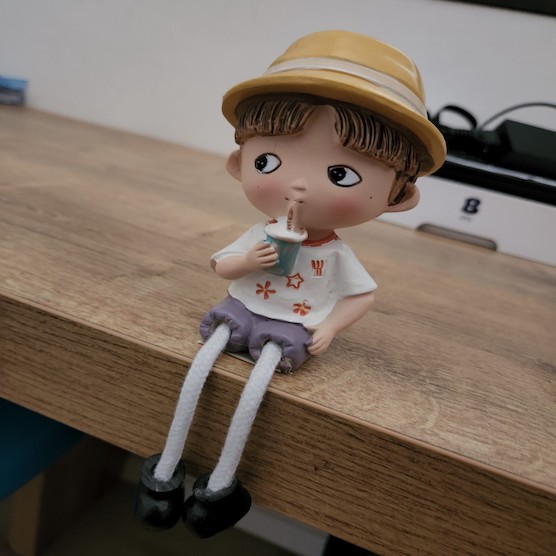} &
        \hspace{0.2cm}
        \includegraphics[width=0.09325\textwidth]{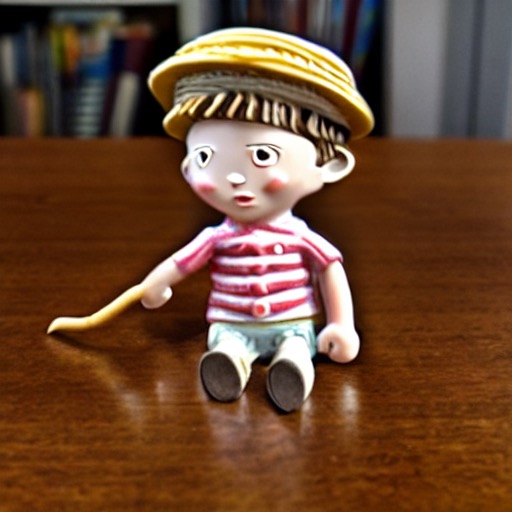} &
        \includegraphics[width=0.09325\textwidth]{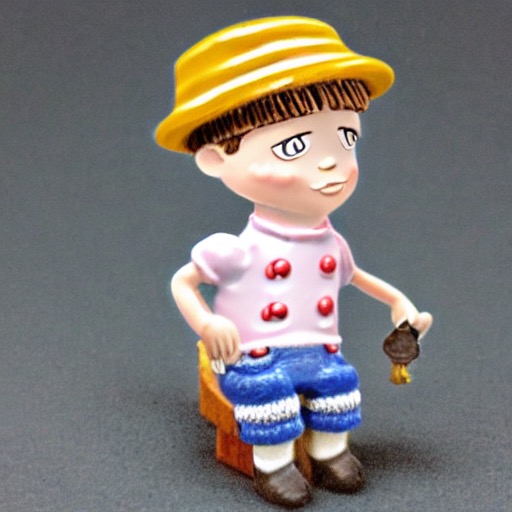} &
        \hspace{0.2cm}
        \includegraphics[width=0.09325\textwidth]{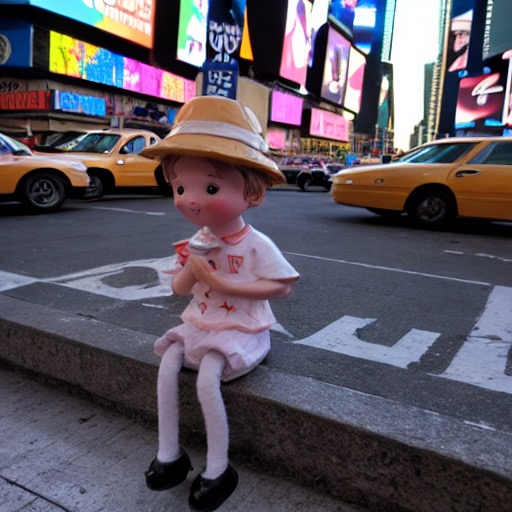} &
        \includegraphics[width=0.09325\textwidth]{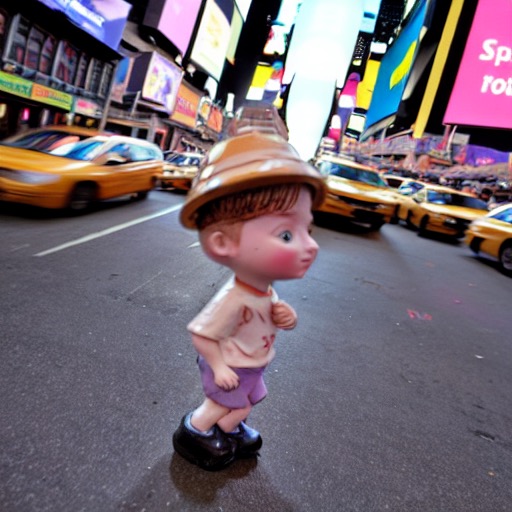} &
        \hspace{0.2cm}
        \includegraphics[width=0.09325\textwidth]{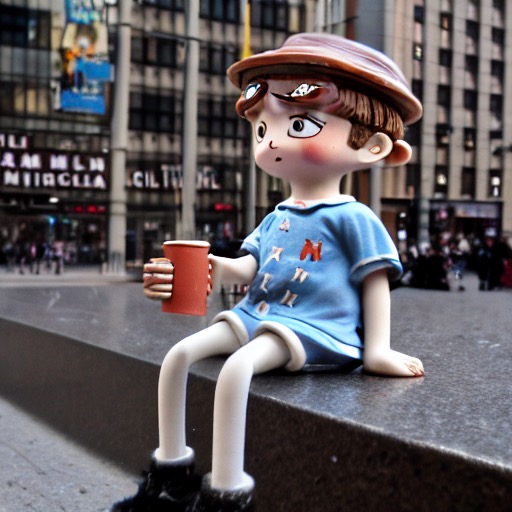} &
        \includegraphics[width=0.09325\textwidth]{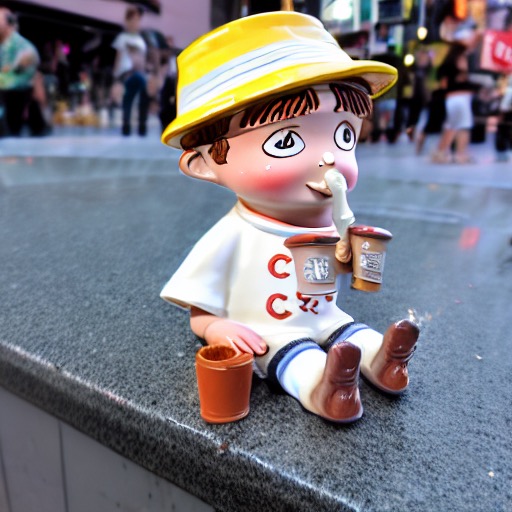} &
        \hspace{0.2cm}
        \includegraphics[width=0.09325\textwidth]{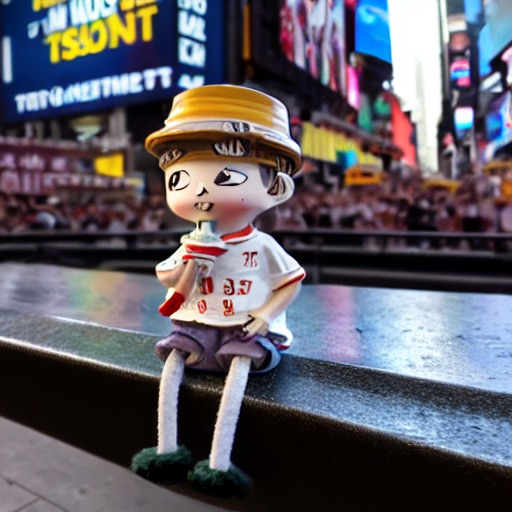} &
        \includegraphics[width=0.09325\textwidth]{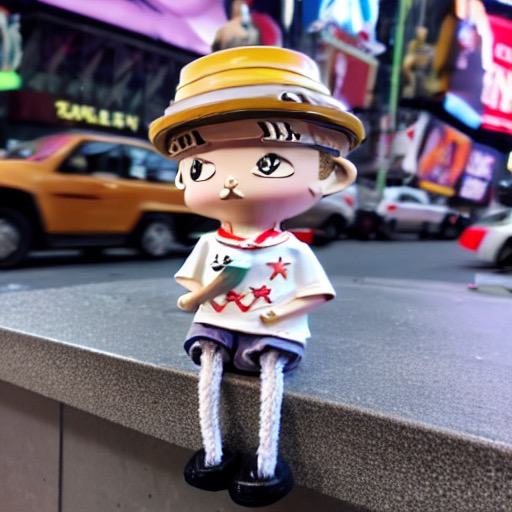} \\

        \raisebox{0.3in}{\begin{tabular}{c} ``A photo of \\ \\[-0.05cm] $S_*$ in Times \\ \\[-0.05cm] Square''\end{tabular}} &
        \hspace{0.2cm}
        \includegraphics[width=0.09325\textwidth]{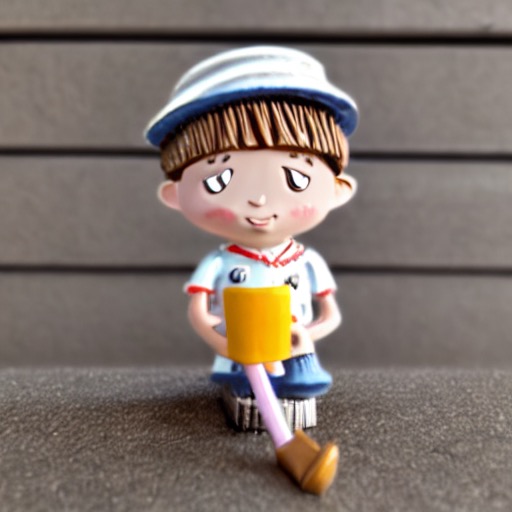} &
        \includegraphics[width=0.09325\textwidth]{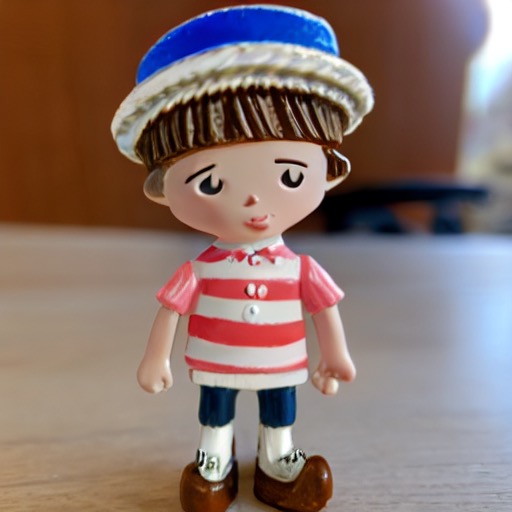} &
        \hspace{0.2cm}
        \includegraphics[width=0.09325\textwidth]{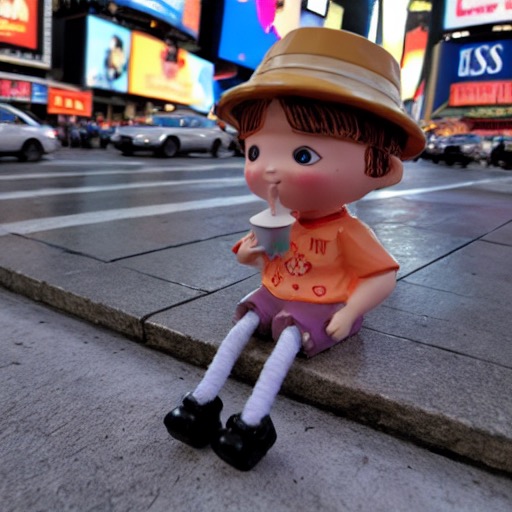} &
        \includegraphics[width=0.09325\textwidth]{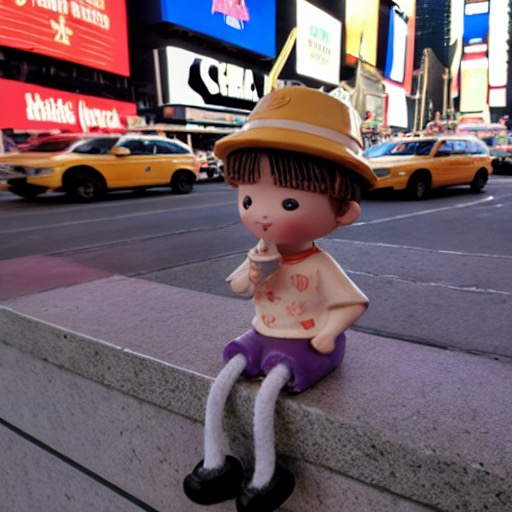} &
        \hspace{0.2cm}
        \includegraphics[width=0.09325\textwidth]{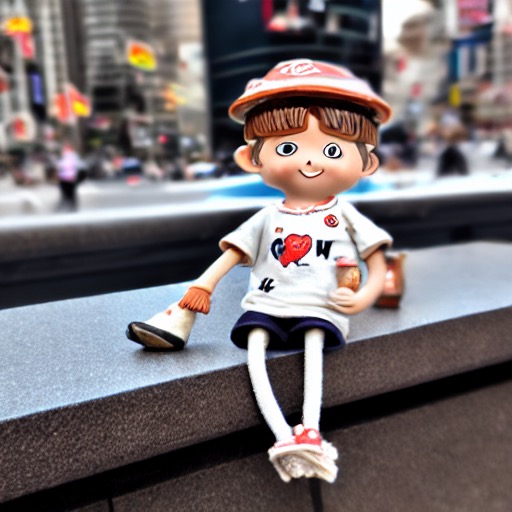} &
        \includegraphics[width=0.09325\textwidth]{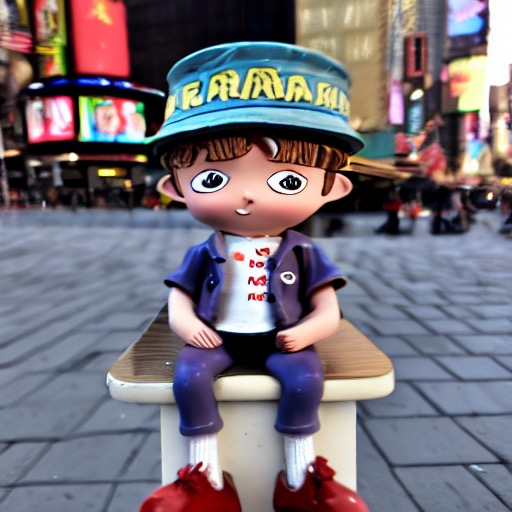} &
        \hspace{0.2cm}
        \includegraphics[width=0.09325\textwidth]{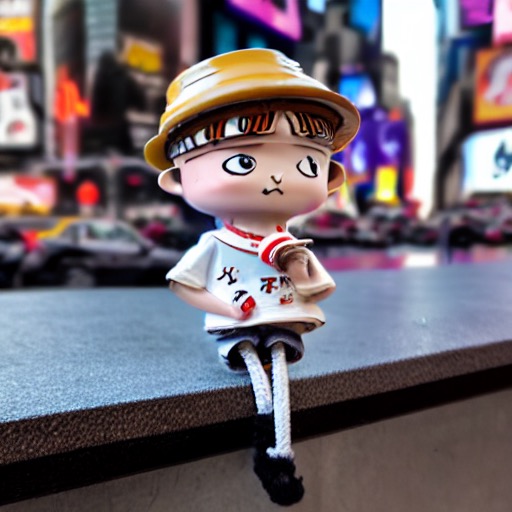} &
        \includegraphics[width=0.09325\textwidth]{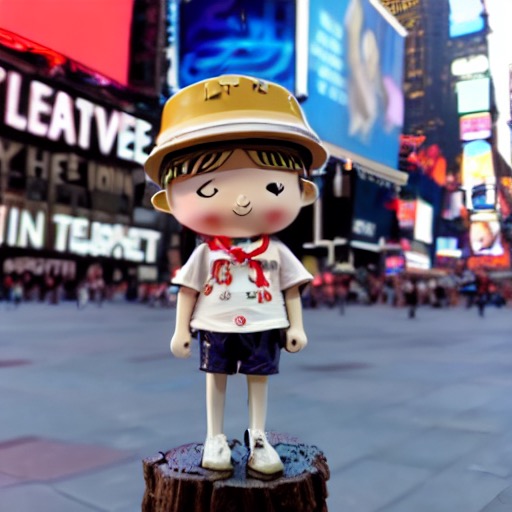} \\ \\

        \includegraphics[width=0.09325\textwidth]{images/original/headless_statue.jpeg} &
        \hspace{0.2cm}
        \includegraphics[width=0.09325\textwidth]{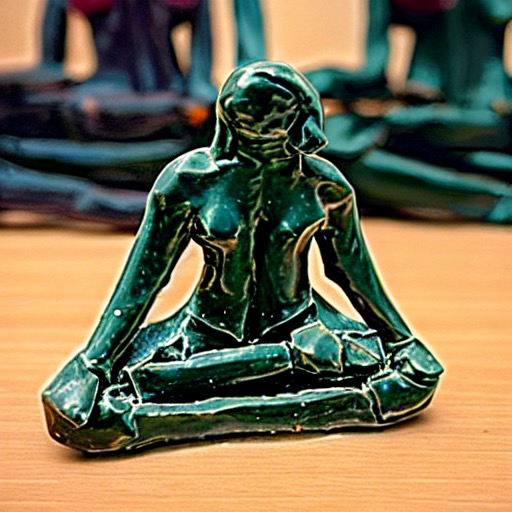} &
        \includegraphics[width=0.09325\textwidth]{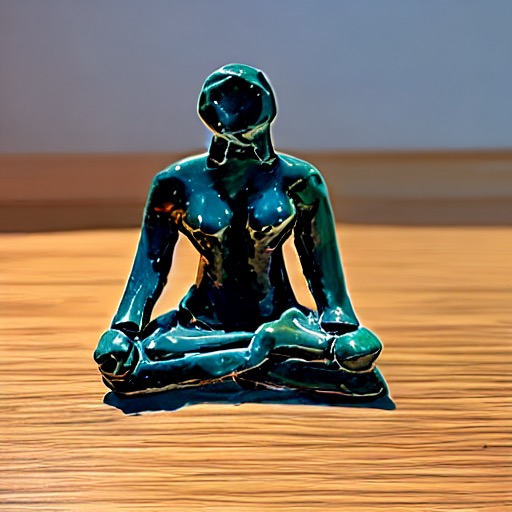} &
        \hspace{0.2cm}
        \includegraphics[width=0.09325\textwidth]{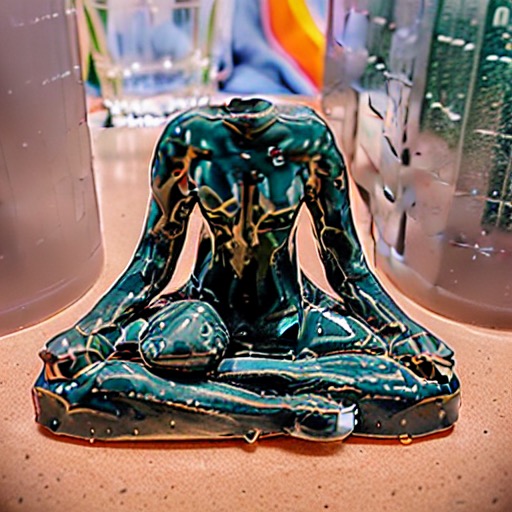} &
        \includegraphics[width=0.09325\textwidth]{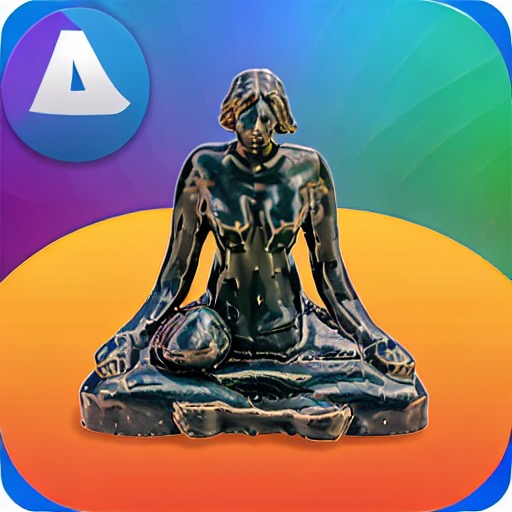} &
        \hspace{0.2cm}
        \includegraphics[width=0.09325\textwidth]{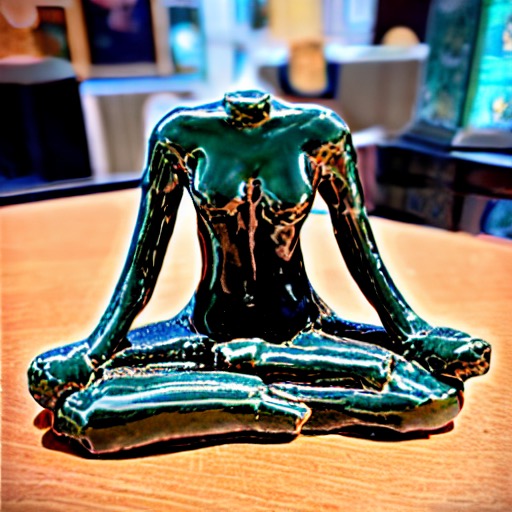} &
        \includegraphics[width=0.09325\textwidth]{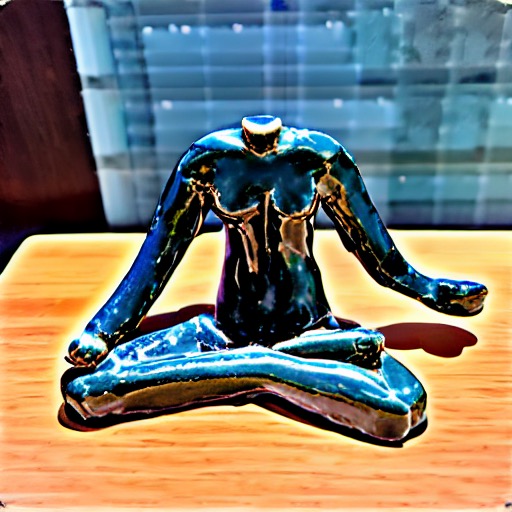} &
        \hspace{0.2cm}
        \includegraphics[width=0.09325\textwidth]{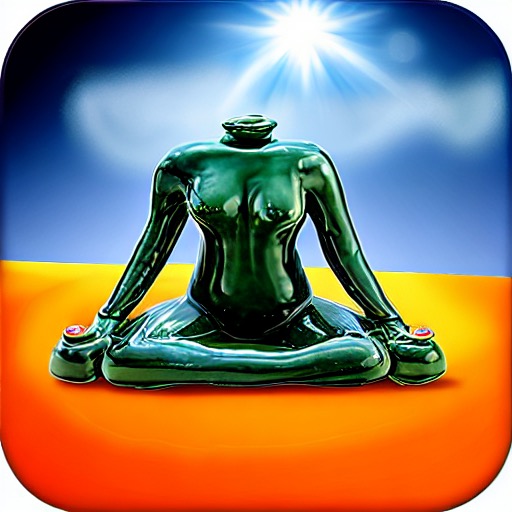} &
        \includegraphics[width=0.09325\textwidth]{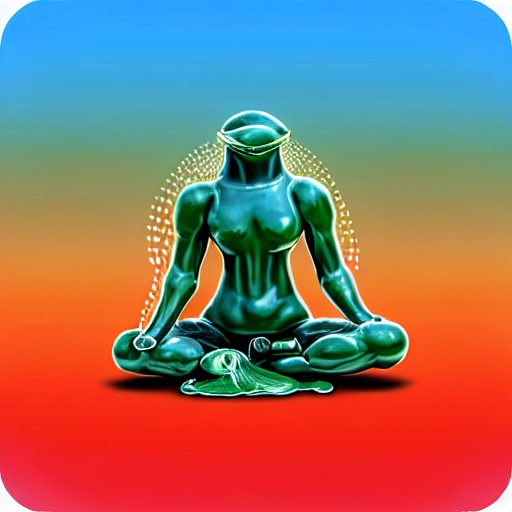} \\

        \raisebox{0.3in}{\begin{tabular}{c} ``An app \\ \\[-0.05cm] icon of \\ \\[-0.05cm] $S_*$''\end{tabular}} &
        \hspace{0.2cm}
        \includegraphics[width=0.09325\textwidth]{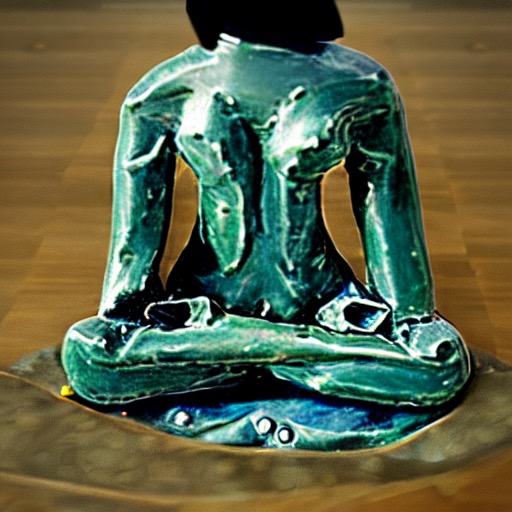} &
        \includegraphics[width=0.09325\textwidth]{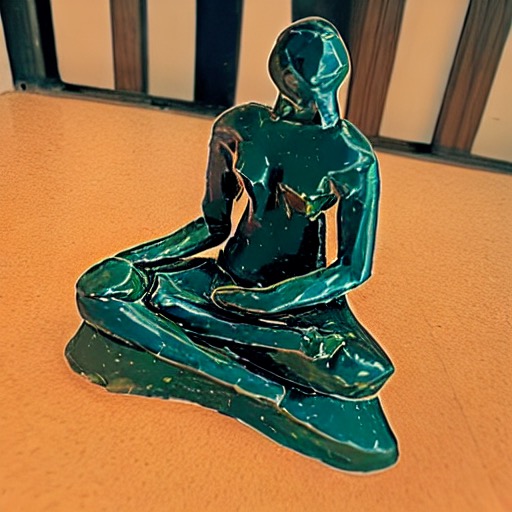} &
        \hspace{0.2cm}
        \includegraphics[width=0.09325\textwidth]{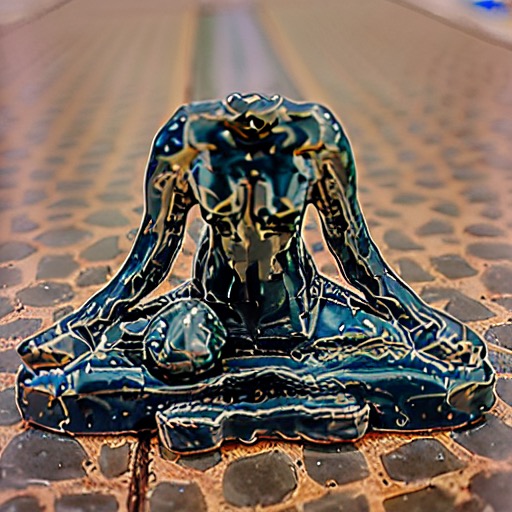} &
        \includegraphics[width=0.09325\textwidth]{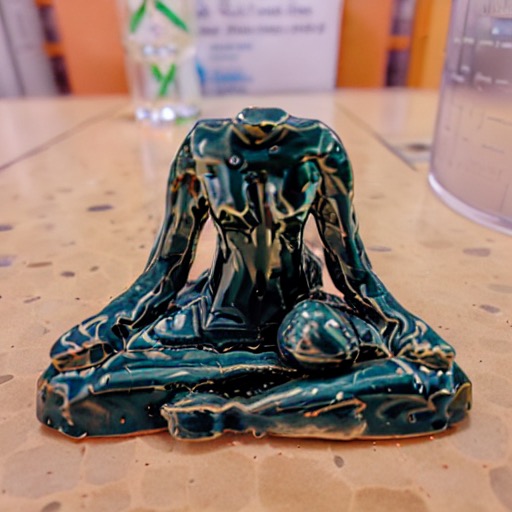} &
        \hspace{0.2cm}
        \includegraphics[width=0.09325\textwidth]{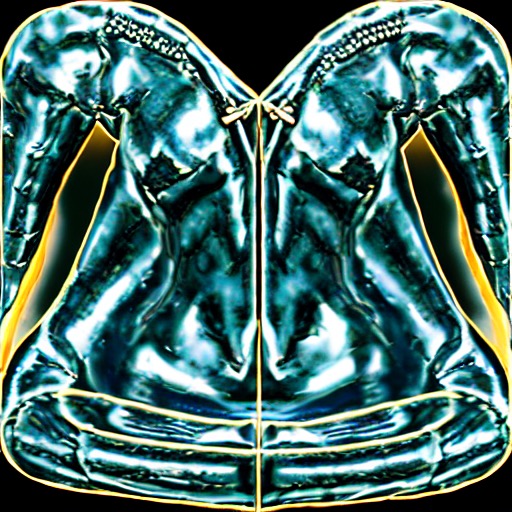} &
        \includegraphics[width=0.09325\textwidth]{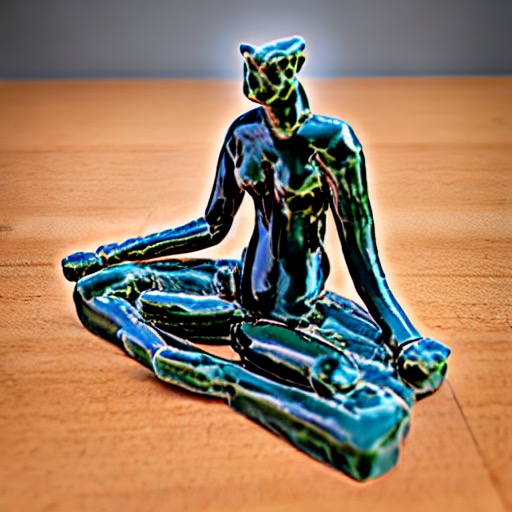} &
        \hspace{0.2cm}
        \includegraphics[width=0.09325\textwidth]{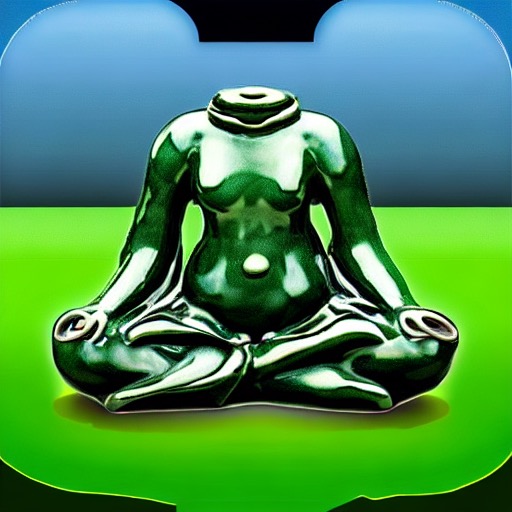} &
        \includegraphics[width=0.09325\textwidth]{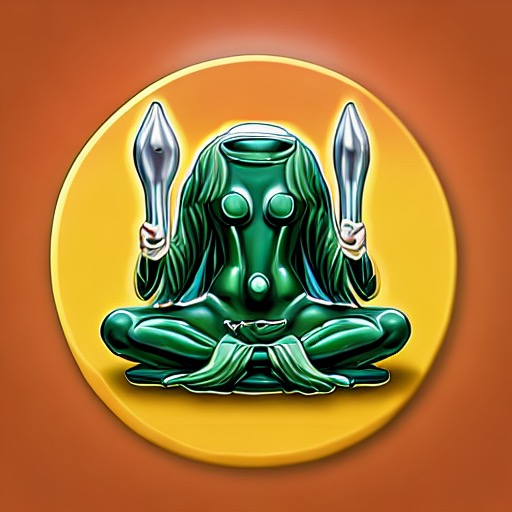} \\ \\

    \end{tabular}
    \\[-0.35cm]
    }
    \caption{Qualitative comparisons. For each concept, we show four images generated by each method using the same set of random seeds. Results for TI are obtained after $5,000$ optimization steps while the remaining methods are all trained for $500$ steps.}
    \label{fig:qualitative_comparison}
    \vspace{-0.4cm}
\end{figure*}

\vspace*{-0.025cm}
\section{Results}
In the following section, we demonstrate the effectiveness of NeTI and the appealing properties of our \pstar text-conditioning space. 

\vspace*{-0.025cm}
\subsection{Evaluations and Comparisons}

\vspace{-0.05cm}
\paragraph{\textbf{Evaluation Setup.}} We evaluate NeTI with respect to state-of-the-art inversion methods (Textual Inversion (TI)~\cite{gal2023image}, Extended Textual Inversion (XTI)~\cite{voynov2023p}) and fine-tuning approaches 
(DreamBooth~\cite{ruiz2022dreambooth}, CustomDiffusion~\cite{kumari2022customdiffusion}). We consider $10$ concepts taken from TI and $6$ concepts from CustomDiffusion. To quantitatively evaluate each method, we construct a set of $15$ text prompts ranging from background modifications (e.g., ``A photo of $S_*$ on the beach'') and artistic styles (e.g., ``A manga drawing of $S_*$'') to more abstract compositions (e.g., ``App icon of $S_*$''). 
For each concept and prompt, we generate $32$ images using $32$ random seeds shared across all methods. 
For a fair comparison, and unless otherwise noted, we do \textit{not} apply Nested Dropout on the results obtained by NeTI. 
Additional details are provided in~\Cref{sec:additional_details}.

\vspace{-0.2cm}
\paragraph{\textbf{Qualitative Evaluation.}}
First, in~\Cref{fig:our_results}, we demonstrate NeTI's ability to compose learned concepts in novel compositions. These generations range from placing the concepts in new scenes to capturing their key semantics and forming new creations inspired by them.  

In Figure~\ref{fig:qualitative_comparison}, we present a visual comparison of new compositions of various concepts. As can be seen, TI, which operates in the relatively small \p space, fails to capture the exact characteristics of the concept or compose the concept in novel scenes. 
By tuning the model, DreamBooth is able to achieve higher-fidelity reconstructions, such as that of the statue in the third row. However, this comes at a heavy cost of high storage requirements and reduced editability. 
For example, in the first row, DreamBooth fails to place the cat statue in the scene when performing a complex edit. 
Moreover, even when tuning the model, DreamBooth may still fail to reconstruct the concept as seen in the second row. 
By operating over \pplus, XTI achieves improved reconstructions and editability but still fails to capture concept-specific details.
Our approach, NeTI, attains high-fidelity reconstructions while staying faithful to the provided text prompt, e.g., the string-like legs of the toy in the second row.
Notably, these results are obtained after $500$ training steps, the same number used for DreamBooth and XTI. 

In~\Cref{sec:custom_comparison} and~\Cref{fig:additional_qualitative_comparison,fig:additional_qualitative_comparison_2} we provide additional qualitative results and comparisons and investigate NeTI under a single-image training setting. In~\Cref{sec:ablation_study} we perform an ablation study of our key design choices.
Finally, \Cref{fig:ours_results_supplementary} presents additional results of our method on a variety of concepts and prompts.

\begin{figure}
    \centering
    \includegraphics[width=0.475\textwidth]{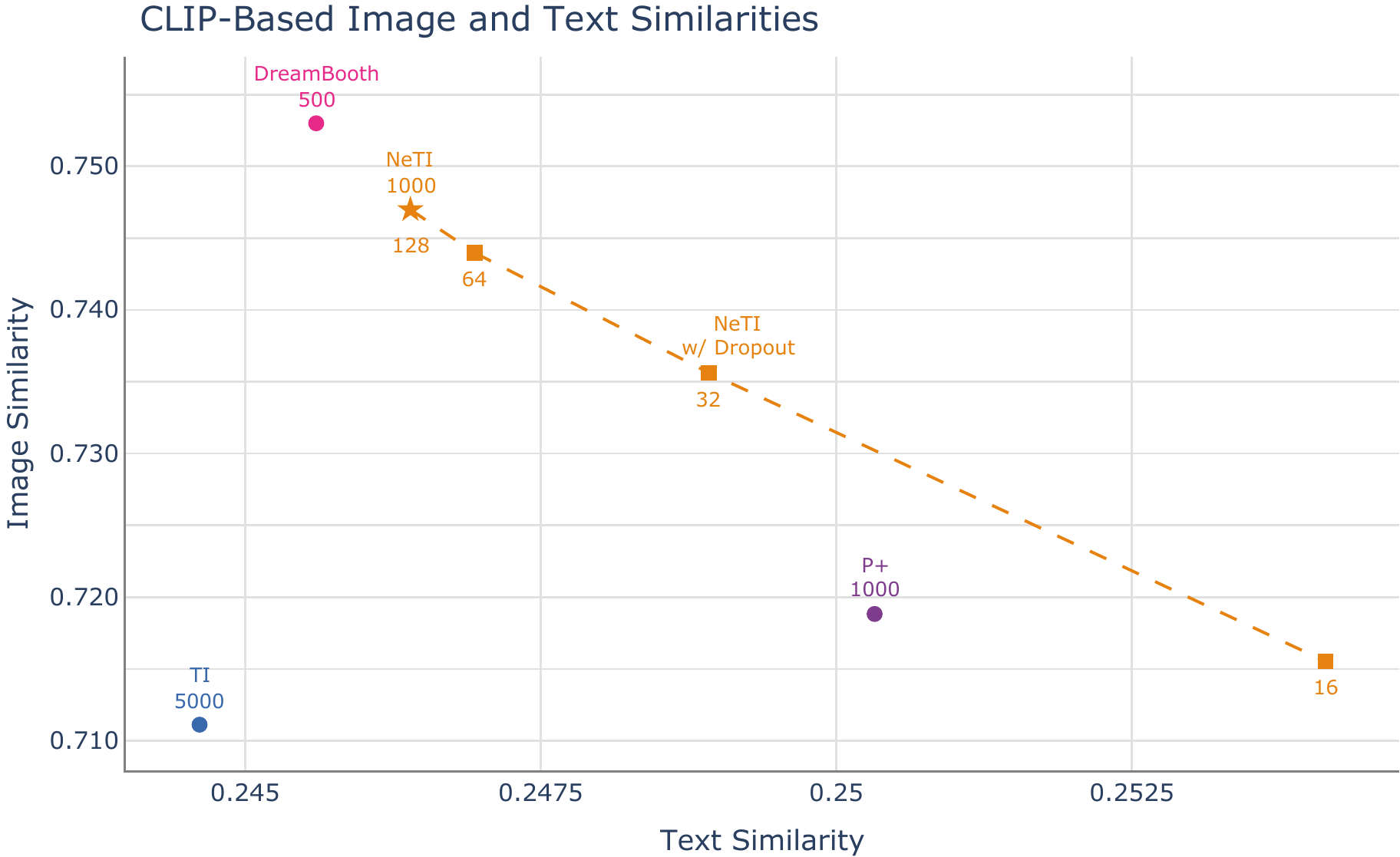}
    \\[-0.2cm]
    \caption{Quantitative evaluation. Below each method, we indicate the number of optimization steps performed during training. 
    }
    \label{fig:reconstruction_editability}
    \vspace{-0.2cm}
\end{figure}

\vspace{-0.25cm}
\paragraph{\textbf{Reconstruction-Editability.}}
We follow the evaluation setup from TI and evaluate the performance of each method in their ability to (1) reconstruct the target concept and (2) synthesize novel compositions containing the concept. For evaluating reconstruction, we generate $32$ images for each text prompt and compute the CLIP-space similarity between the generated images and the training images. 
To measure editability, for each text prompt, we calculate the CLIP-space similarity between the embeddings of the generated images and the embedding of the text prompt where we omit the placeholder $S_*$ from the prompt. 
Results are presented in~\Cref{fig:reconstruction_editability}. We first note that NeTI is able to achieve comparable results to those of DreamBooth, without requiring any tuning of the model. While this does require additional training time, our models require $\sim$2MB of disk space while DreamBooth requires several GBs. 

As marked using the dashed line, when more units are dropped from our hidden representation at inference time, we can shift along this curve, reaching higher editability at the cost of reduced visual fidelity.
Finally, when compared to XTI trained for the same number of steps, NeTI achieves both improved reconstruction and editability across this dropout curve.
In~\Cref{sec:custom_comparison}, we additionally compare NeTI and XTI after $250$ and $500$ training steps.

\vspace{-0.25cm}
\paragraph{\textbf{User Study.}}
We additionally conduct a user study to analyze all approaches. 
For each concept, a text prompt was randomly selected from one of our $15$ prompts and used to generate $2$ images for each method using the same random seeds. Respondents were asked to rate the images based on their (1) similarity to the concept's training images and (2) similarity to the text prompt.

Note that previous personalization studies asked respondents to rate each of the above properties independently. 
However, consider a case where a method is tasked with generating a painting of a concept, but omits the concept from the painting entirely. In this case, it would still score favorably along the text similarity if measured independently. Yet, this is clearly not a desirable result. 
Therefore, we asked respondents to consider both aspects \textit{together} and rate the overall image \textit{and} text similarity on a scale from $1$ to $5$.
Results are shown in~\Cref{tb:user_study}. In total, we had 35 respondents, for a total of $560$ ratings per method. 
As shown, NeTI outperforms other inversion methods and remains competitive with DreamBooth without requiring model tuning.

\begin{table}
\small
\centering
\setlength{\tabcolsep}{3pt}   
\renewcommand{\arraystretch}{0.3}
\caption{User Study.
We asked respondents to rate each method based on their faithfulness to the original images and prompt on a scale of $1$ to $5$.
\\[-0.75cm]} 
\begin{tabular}{l c c c c} 
    \toprule
    & TI & DreamBooth & XTI & \textbf{NeTI}  \\
    \midrule
    \multirow{2}{*}{Avg. Rating ($\uparrow$)} & 2.77 & 3.71 & 3.15 & \textbf{3.97} \\[0.1cm]
    & ($\pm$ 1.20) & ($\pm$ 1.13) & ($\pm$ 1.09) &  ($\pm$ \textbf{1.12}) \\
    \bottomrule
\end{tabular}
\label{tb:user_study}
\vspace{-0.35cm}
\end{table}

\vspace{0.2cm}
\subsection{Time for Some Analysis}~\label{sec:analysis}

\begin{figure}[b]
    \centering
    \renewcommand{\arraystretch}{0.3}
    \setlength{\tabcolsep}{0.5pt}
    \vspace{-0.2cm}
    {\footnotesize

    \begin{tabular}{c@{\hspace{0.2cm}} c c c c }

        \begin{tabular}{c} Real \end{tabular} &
        \hspace{0.05cm}
        \begin{tabular}{c} No Time \end{tabular} &
        \begin{tabular}{c} No Space \end{tabular} &
        \begin{tabular}{c} Neither \end{tabular} &
        \begin{tabular}{c} Both \end{tabular} \\

        \includegraphics[width=0.085\textwidth]{images/original/rainbow_cat.jpeg} &
        \hspace{0.05cm}
        \includegraphics[width=0.085\textwidth]{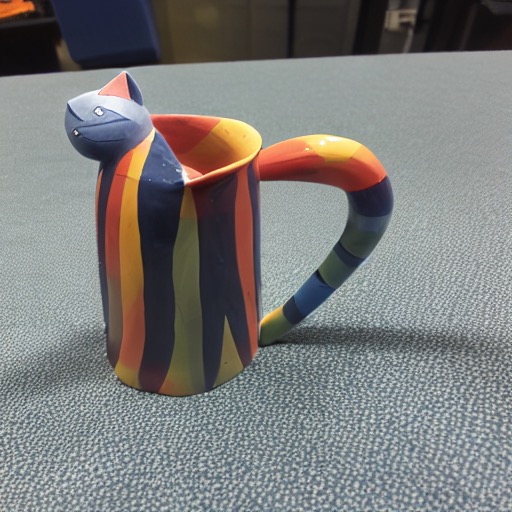} &
        \includegraphics[width=0.085\textwidth]{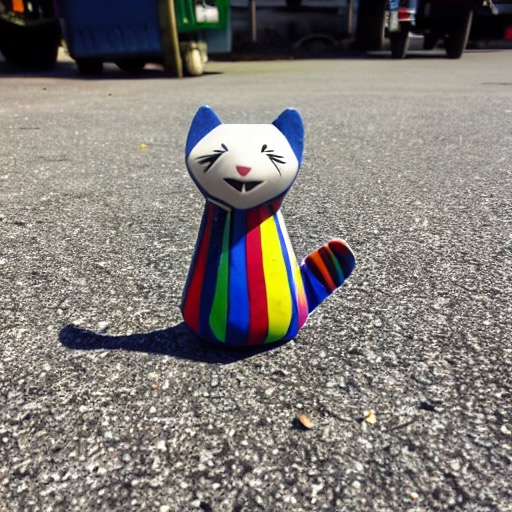} &
        \includegraphics[width=0.085\textwidth]{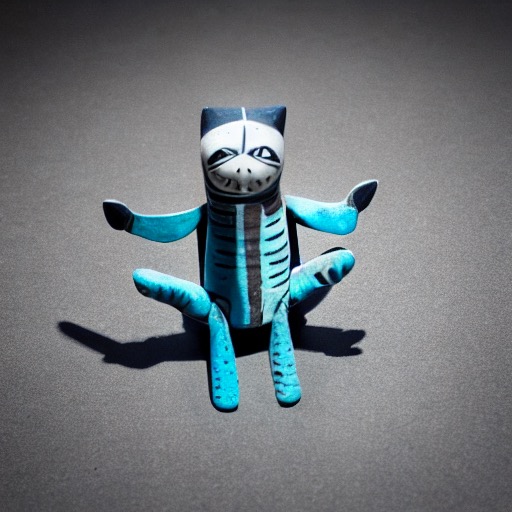} &
        \includegraphics[width=0.085\textwidth]{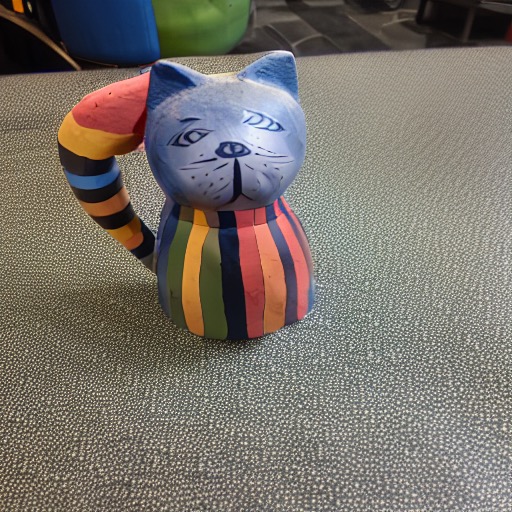} \\ \\

        \includegraphics[width=0.085\textwidth]{images/original/colorful_teapot.jpg} &
        \hspace{0.05cm}
        \includegraphics[width=0.085\textwidth]{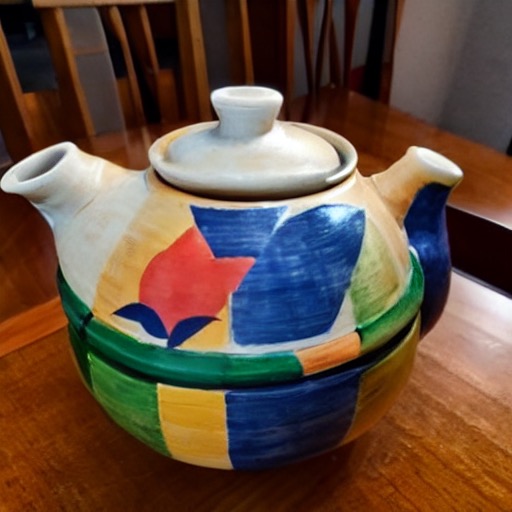} &
        \includegraphics[width=0.085\textwidth]{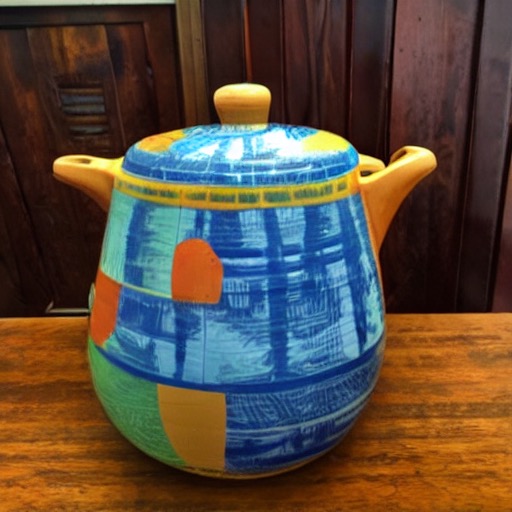} &
        \includegraphics[width=0.085\textwidth]{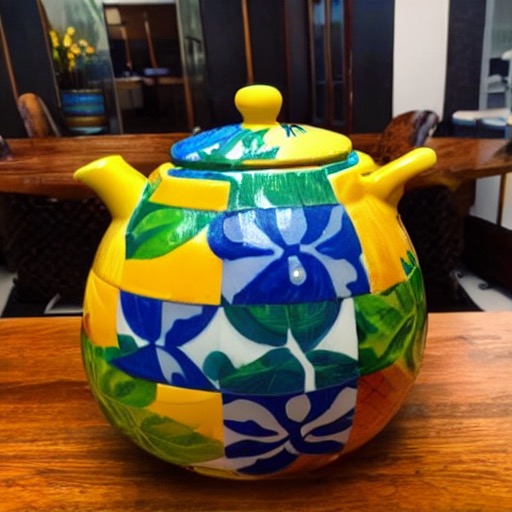} &
        \includegraphics[width=0.085\textwidth]{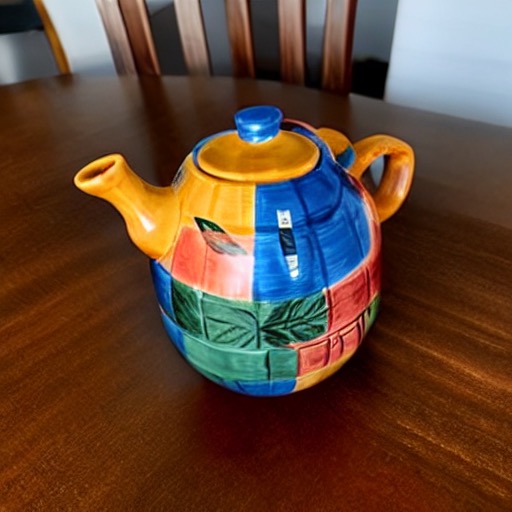} \\ \\

    \end{tabular}
    \\[-0.4cm]
    }
    \caption{Importance of time-space conditioning. 
    As can be seen, the combination of both time and space is essential for attaining high visual fidelity.}
    \label{fig:space_time_conditioning_main}
    \vspace{-0.4cm}
\end{figure}

\vspace{-0.9cm}
\paragraph{\textbf{Time and Space.}}
We begin by analyzing the use of both time and space and validate the importance of conditioning our mapper on both input types.
Specifically, we consider three variants of our mapper: (1) a space-conditioned mapper; (2) a time-conditioned mapper; and (3) a mapper that is not conditioned on either.
For the final variant, we simply pass a fixed input to the neural network and optimize the network parameters as is done in our standard training scheme. Sample reconstructions of each variant using the prompt ``A photo of $S_*$'' are provided in~\Cref{fig:space_time_conditioning_main}. As can be seen, conditioning our neural mapper on both time and space is crucial for capturing fine-level details.

\begin{figure}
    \centering
    \setlength{\tabcolsep}{0.5pt}
    \renewcommand{\arraystretch}{0.3}
    \addtolength{\belowcaptionskip}{-12.5pt}
    {\footnotesize
    \begin{tabular}{c@{\hspace{0.1cm}} c@{\hspace{0.1cm}} c c c c}

        \includegraphics[width=0.075\textwidth]{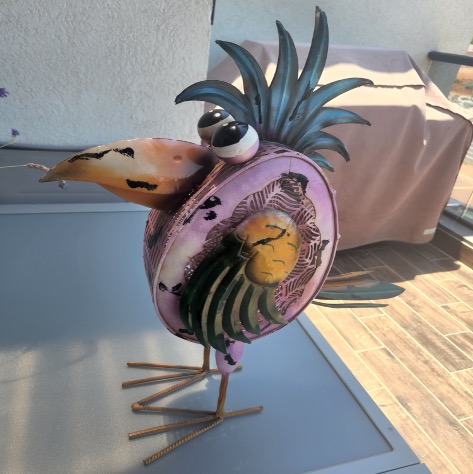} &
        \raisebox{0.5cm}{\begin{tabular}{c} ``Colorful \\ \\[-0.05cm] grafitti of \\ \\[-0.05cm] $S_*$'' \end{tabular}} &
        \includegraphics[width=0.075\textwidth]{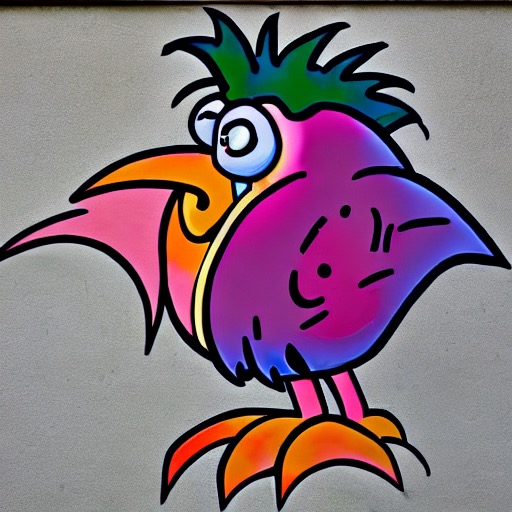} &
        \includegraphics[width=0.075\textwidth]{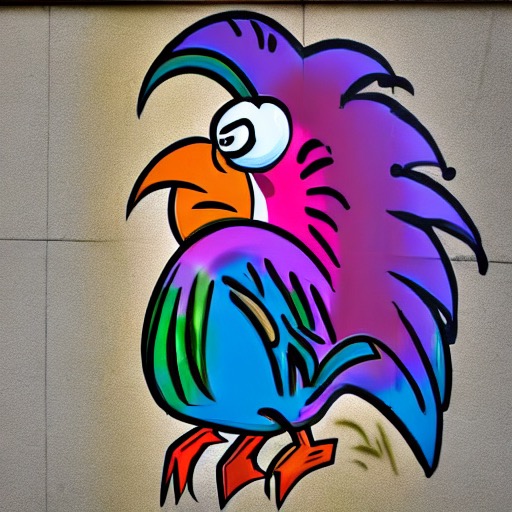} &
        \includegraphics[width=0.075\textwidth]{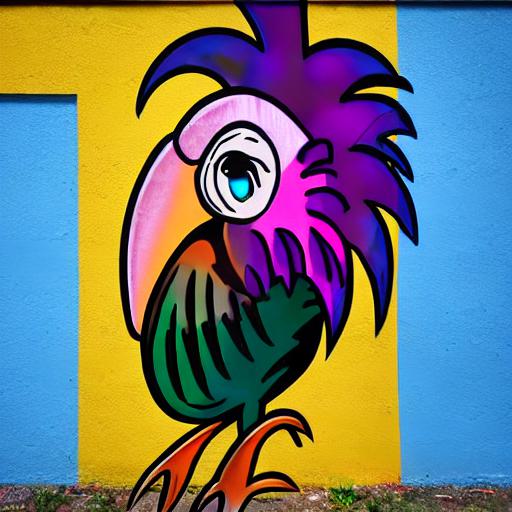} &
        \includegraphics[width=0.075\textwidth]{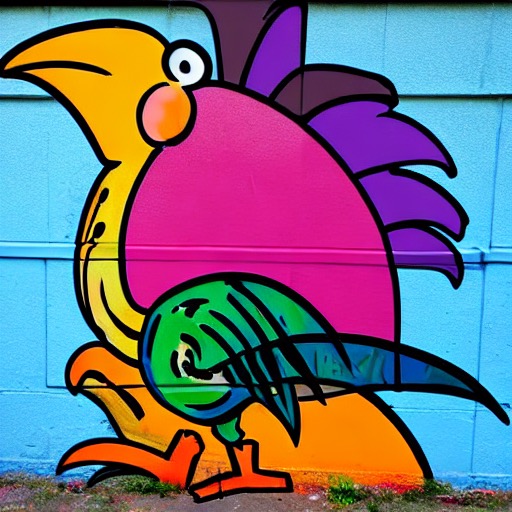} \\ \\

        \includegraphics[width=0.075\textwidth]{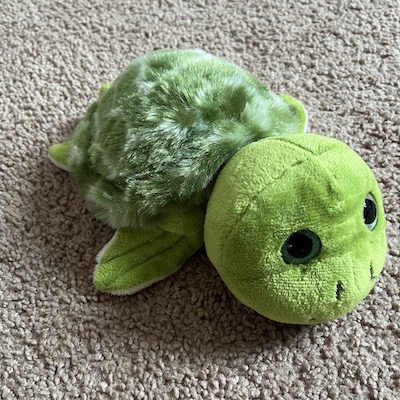} &
        \raisebox{0.5cm}{\begin{tabular}{c} ``Pokemon \\ \\[-0.05cm] in the style \\ \\[-0.05cm] of $S_*$'' \end{tabular}} &
        \includegraphics[width=0.075\textwidth]{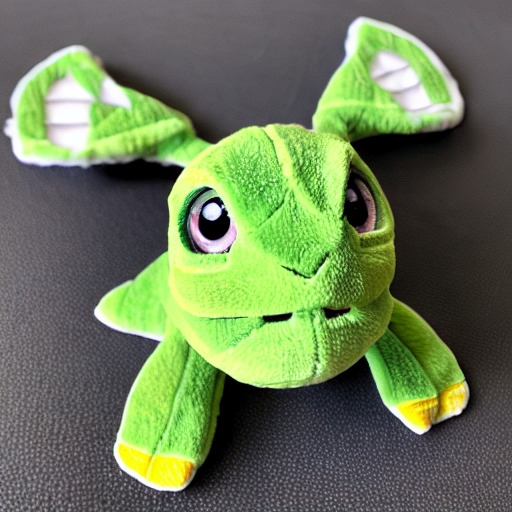} &
        \includegraphics[width=0.075\textwidth]{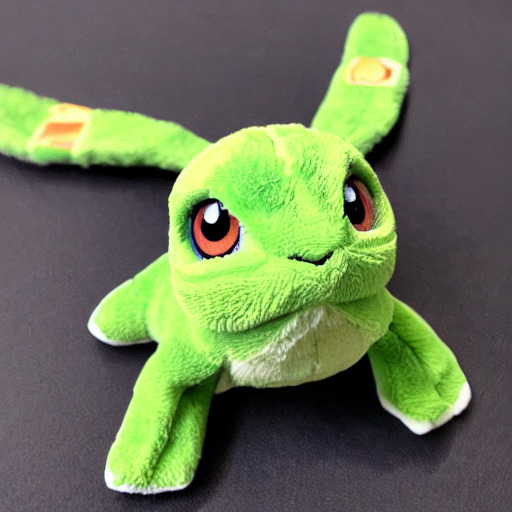} &
        \includegraphics[width=0.075\textwidth]{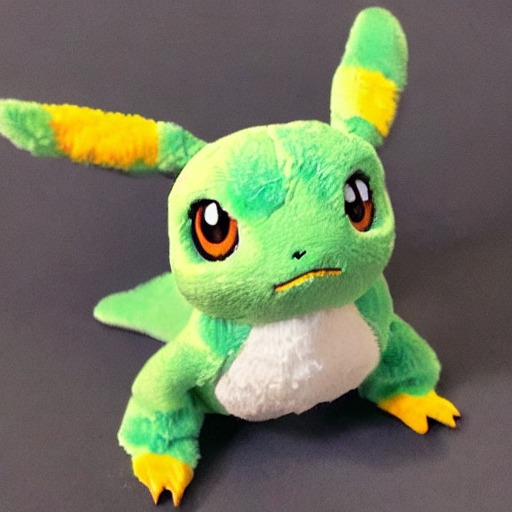} &
        \includegraphics[width=0.075\textwidth]{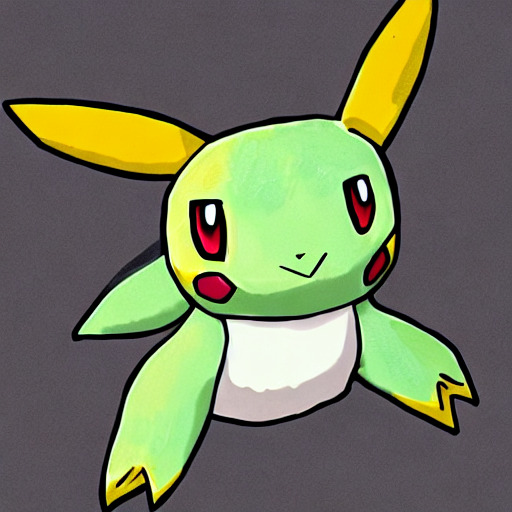} \\ \\

        \includegraphics[width=0.075\textwidth,height=0.075\textwidth]{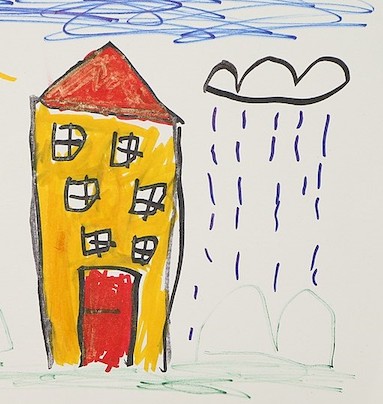} &
        \raisebox{0.225in}{\begin{tabular}{c} ``A $S_*$ \\ \\[-0.05cm] painting of \\ \\[-0.05cm] a Medieval \\ \\[-0.05cm] castle'' \end{tabular}} &
        \includegraphics[width=0.075\textwidth]{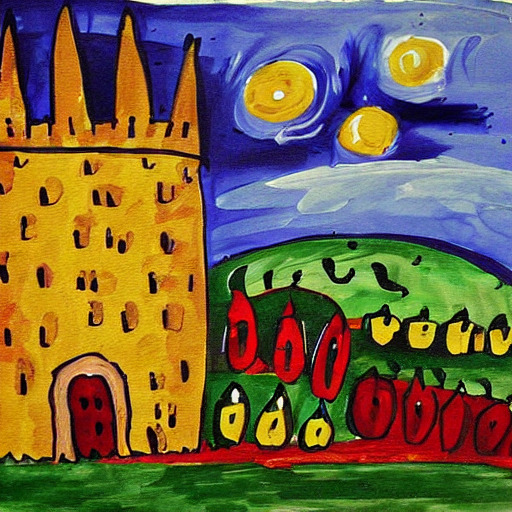} &
        \includegraphics[width=0.075\textwidth]{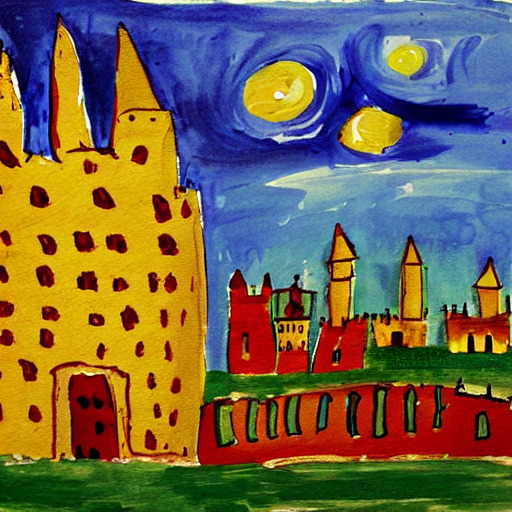} &
        \includegraphics[width=0.075\textwidth]{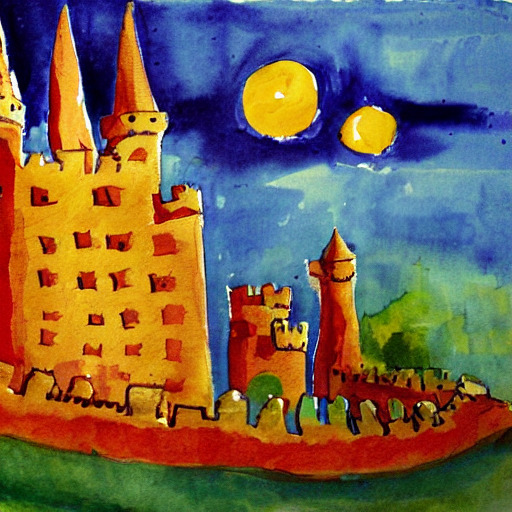} &
        \includegraphics[width=0.075\textwidth]{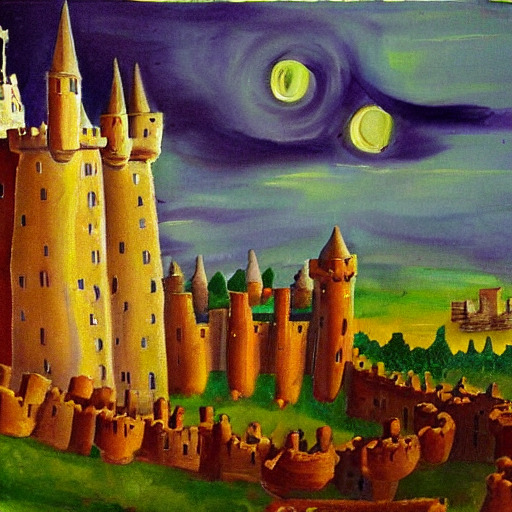} \\

        Real & & \multicolumn{4}{c}{\raisebox{0.05cm}{\limitarrowmain{}}} \\

    \\[-0.3cm]
    \end{tabular}
    }
    \caption{Controlling editability with Nested Dropout.
    During inference, we gradually vary the dropout strength to form a series of plausible generations.
    }
    \vspace{0.2cm}
    \label{fig:nested_dropout}
\end{figure}

\vspace{-0.25cm}
\paragraph{\textbf{Controlling Editability.}}
Thanks to our importance-based ordering over our mapper's hidden representation, we can control our dimensionality at inference time. In~\Cref{fig:nested_dropout} we gradually change the strength of our dropout to show how this affects the generated image's visual and text fidelity. When a stronger dropout is applied we get a more coarse/semantic representation of our concept that is more amenable to edits and new compositions. This inference-time control enables users to dynamically choose the dropout strength that best suits their target concept and prompt without having to train multiple models.

\begin{figure}
    \centering
    \renewcommand{\arraystretch}{0.3}
    \setlength{\tabcolsep}{0.5pt}
    {\small
    \begin{tabular}{c c c c c c}
        
        \includegraphics[width=0.075\textwidth]{images/original/colorful_teapot.jpg} &        
        \includegraphics[width=0.075\textwidth]{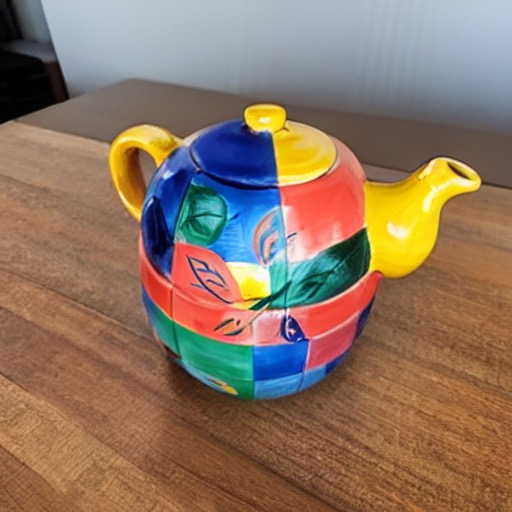} &        
        \includegraphics[width=0.075\textwidth]{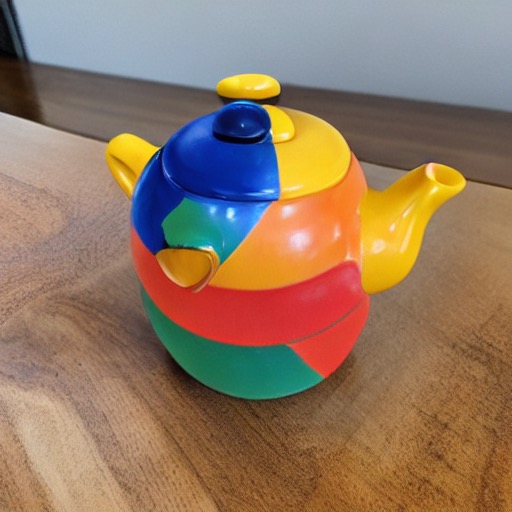} &   
        \includegraphics[width=0.075\textwidth]{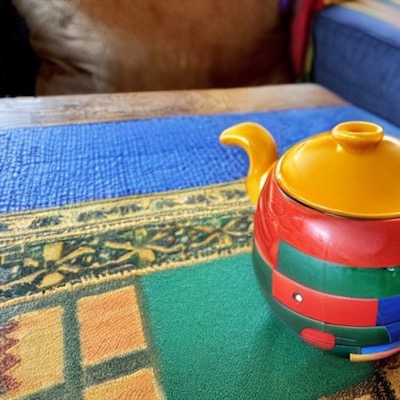} &   
        \includegraphics[width=0.075\textwidth]{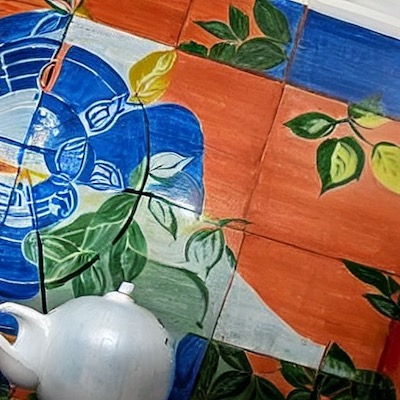} &  
        \includegraphics[width=0.075\textwidth]{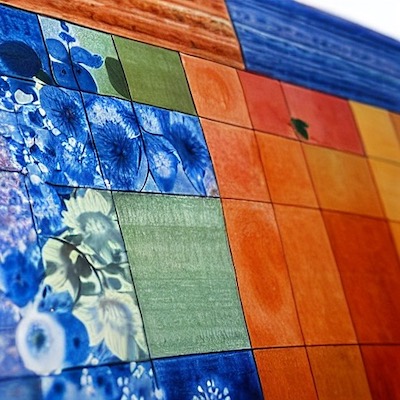} \\ \\ 
        
        \includegraphics[width=0.075\textwidth]{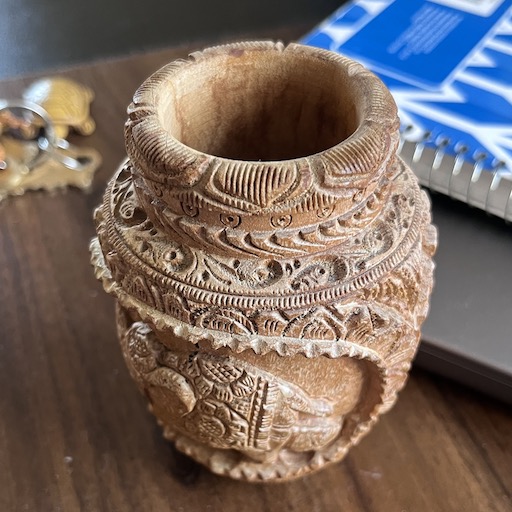} &        
        \includegraphics[width=0.075\textwidth]{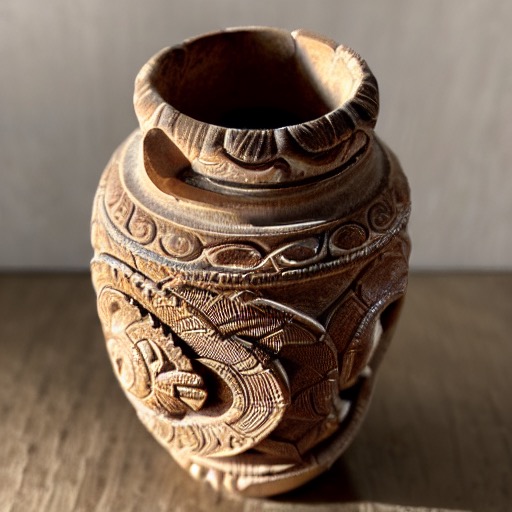} &        
        \includegraphics[width=0.075\textwidth]{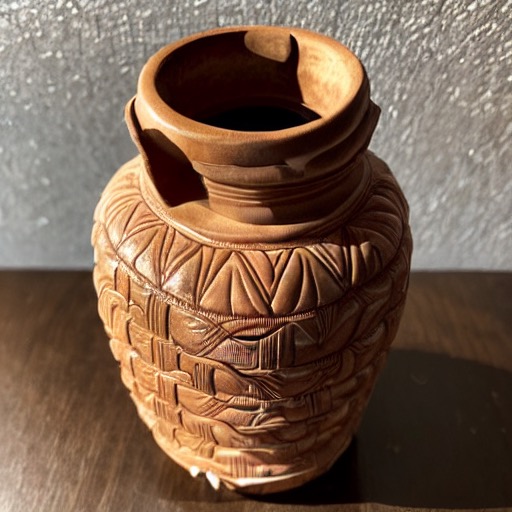} &   
        \includegraphics[width=0.075\textwidth]{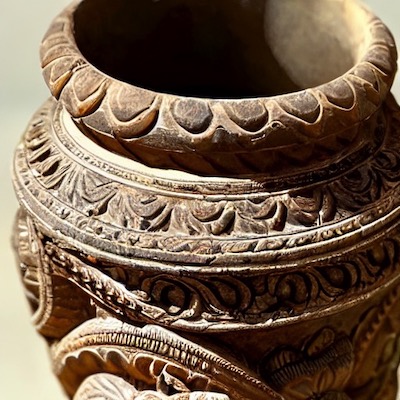} &   
        \includegraphics[width=0.075\textwidth]{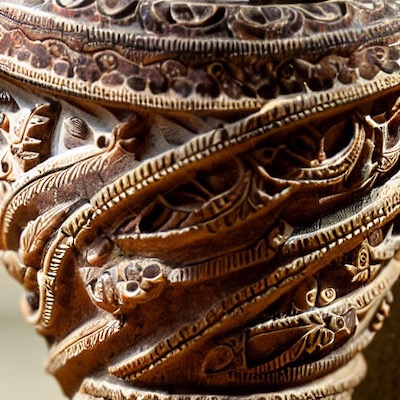} &   
        \includegraphics[width=0.075\textwidth]{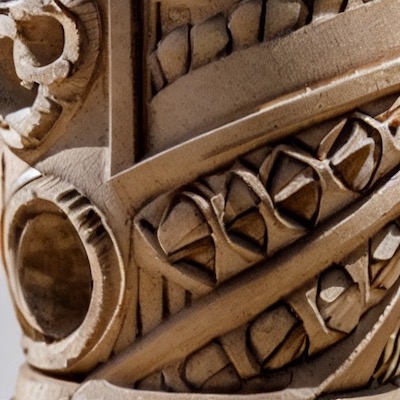} \\ \\

        \includegraphics[width=0.075\textwidth]{images/original/rainbow_cat.jpeg} &
        \includegraphics[width=0.075\textwidth]{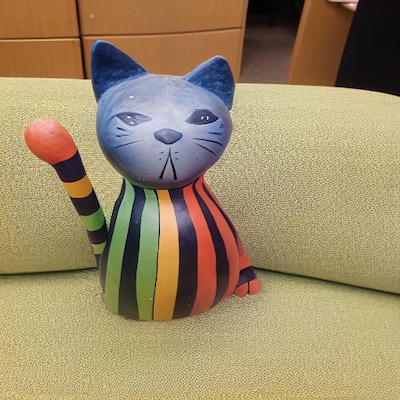} &
        \includegraphics[width=0.075\textwidth]{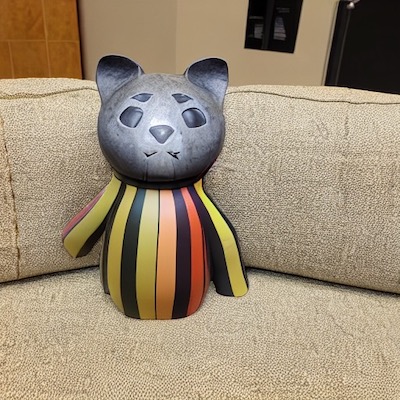} &
        \includegraphics[width=0.075\textwidth]{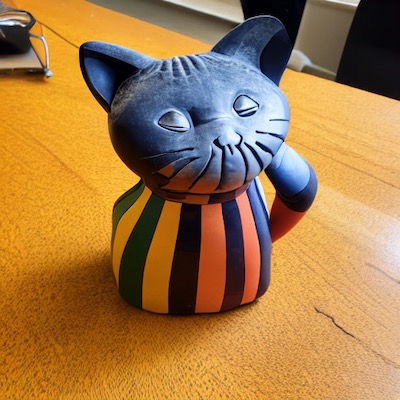} &
        \includegraphics[width=0.075\textwidth]{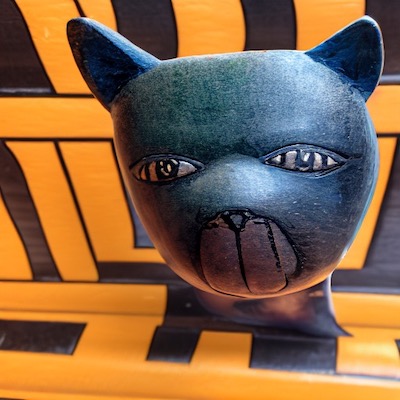} &
        \includegraphics[width=0.075\textwidth]{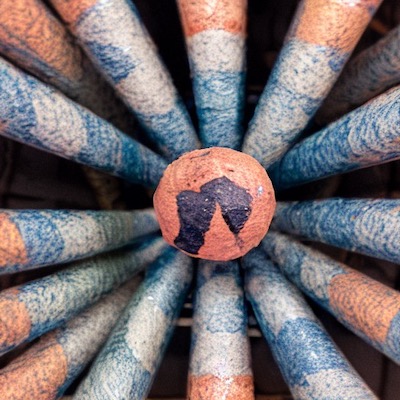} \\ \\

        Real & All & $t=999$ & $t=666$ & $t=333$ & $t=50$
        
    \\[-0.2cm]     
    \end{tabular}
    
    }
    \caption{Analyzing the different timesteps with \pstar and NeTI. We illustrate which concept-specific details are captured at different denoising timesteps. 
    }
    \label{fig:timestep_analysis}
    \vspace{-0.3cm}
\end{figure}

\vspace{-0.125cm}
\paragraph{\textbf{Per-Timestep Decomposition.}}
We now turn to analyze what aspects of the personalized concept are captured at different timesteps. To do so, we consider a single timestep $t$ and query our network using all combinations of $t$ and $\{\ell_1,\dots,\ell_{16}\}$. We then apply the resulting token embeddings across \textit{all} timesteps. 
In~\Cref{fig:timestep_analysis} we perform the above process for timesteps spanning different stages of the denoising process. As shown, at the early timesteps, (e.g., $t=999$) NeTI learns to capture coarse details such as the concept's general structure and color scheme. 
Yet fine-grained details are missing, such as the exact pattern on the teapot or the engraving of the pot.
As we continue along the denoising process, more concept-specific details are added.
Importantly, no single timestep is able to capture all the concept-specific details while a combination of all timesteps attains high-fidelity reconstructions.
This behavior of learning more global aspects (e.g., structure) followed by local aspects (e.g., style) also aligns with previous observations of the denoising process~\cite{chefer2023attendandexcite,patashnik2023localizing}.

\section{Conclusions}
We introduced a new text-conditioning space \pstar that considers both the time-dependent nature of the denoising process and the different attention layers of the denoising network.
We then presented NeTI that implicitly represents concepts in \pstar via a simple neural mapper.
While we have demonstrated the effectiveness of NeTI, there are limitations that should be considered. First, our dropout-based method for traversing the reconstruction-editability tradeoff still requires a dedicated inference pass for each dropout value we check.
We also still optimize each concept independently, requiring hundreds of steps per concept. 
An interesting avenue could be exploring pairing our approach with faster encoder-based approaches~\cite{gal2023encoderbased,wei2023elite}. 
We hope that \pstar and NeTI will help encourage further advancements in personalization methods.

\section*{Acknowledgements}
We would like the thank Or Patashnik, Rinon Gal, and Yael Vinker for their insightful discussions and early feedback. This work was supported by the Israel Science Foundation under Grant No. 2366/16 and Grant No. 2492/20.

\bibliographystyle{ieee_fullname}
\bibliography{main}

\clearpage
\appendix
\appendixpage
\setlength{\abovedisplayskip}{5pt}
\setlength{\belowdisplayskip}{5pt}

\section{Additional Details}~\label{sec:additional_details}

\vspace*{-0.6cm}
\subsection{Implementation Details}

We operate over the official Stable Diffusion v1.4 text-to-image model that uses the pretrained text encoder from the CLIP ViT-L/14 model~\cite{radford2021learning}. 

\vspace{-0.3cm}
\paragraph{\textbf{Input Representation.}}
Our timesteps $t$ range from $0$ to 1,000, as in the standard Stable Diffusion training scheme. For choosing the U-Net layers, we follow Voynov~\etal~\shortcite{voynov2023p} and consider $16$ different cross-attention layers. Our positional encoding maps the pair of scalars $(t,\ell)$ into a $160$-dimensional vector. Our $160$ uniformly-spaced $(t,\ell)$ anchors are defined using $16$ U-Net layers and $10$ time-based anchors corresponding to $t=0,100,200,\dots,900$. 
Recall, that the output of our positional encoding is given by: 
\begin{equation*}
e_{t,\ell} = E \times f(t,\ell) \in \mathbb{R}^{160}
\end{equation*}
where
\begin{equation*}
E =
\begin{pmatrix}
    -  & f(0,0) & - \\
    -  & f(0,1) & - \\
       & ... &  \\
       & f(0,16) & - \\
    -  & f(100,0) & - \\
    -  & f(100,1) & - \\
       & ... &  \\
    -  & f(900,16) & - \\
\end{pmatrix}_{160 \times 2048},
\end{equation*}
For the variance of the Fourier Features, we set $\sigma_t=0.03$ and $\sigma_\ell=2$ for the time and layer encoding. This introduces an inductive bias where close timesteps obtain a similar encoding, see~\Cref{fig:positional_encoding_visualization}.

\vspace{-0.25cm}
\paragraph{\textbf{Network Architecture.}}
The encoded input $e_{t,\ell}$ is mapped to a $128$-dimensional vector via two fully connected layers. After each layer, we apply LayerNorm~\cite{ba2016layer} followed by a LeakyReLU activation. During training, we apply Nested Dropout~\cite{rippel2014learning} with a probability of $0.5$ and sample the truncation index as $t\sim U[0,128)$. Finally, an additional fully-connected layer maps the $128$-dimensional vector to a $768$-dimension vector, matching the dimension of the standard \p token embedding space. 

\vspace{-0.25cm}
\paragraph{\textbf{Training \& Inference Scheme.}}
Training is performed on a single GPU using a batch size of $2$ with a gradient accumulation of $4$ for an effective batch size of $8$. When applying our textual bypass technique, we perform \textit{up to} $1000$ optimization steps. Importantly, we found that good results are also possible with far fewer steps (see~\Cref{fig:neti_convergence}).
We use a base learning rate of $0.001$, which is scaled to an effective learning rate of $0.008$. At inference time, we apply a guidance scale of $7.5$ for $50$ denoising steps.

\subsection{Evaluation Setup}
\paragraph{\textbf{Baseline Methods.}} 
For Textual Inversion~\cite{gal2023image}, we follow the original paper and train for $5,000$ optimization steps using a batch size of $8$ using the unofficial implementation from the diffusers~\cite{diffusers} library. For XTI~\cite{voynov2023p}, we use the implementation provided by the authors and follow the official hyperparameters specified in the paper, training with a batch size of $8$ for $500$ optimization steps and a learning rate of 5e-3. 
For a fair comparison, we also quantitatively evaluate their results at $1,000$ optimization steps, see~\Cref{tb:p_plus_convergence_comparison}. 

For Dreambooth~\cite{ruiz2022dreambooth}, we use the diffusers implementation and tune only the denoiser’s U-Net with no prior preservation loss. We perform $500$ fine-tuning steps using a learning rate of 5e-6 and batch size of $4$. Finally, we compare to CustomDiffusion~\cite{kumari2022customdiffusion} using their six released models available in their official implementation. Inference was performed using their released implementation and default guidance scale parameters.

\begin{table}
\small
\centering
\setlength{\tabcolsep}{3pt}   
\renewcommand{\arraystretch}{1.2}
\caption{A list of the $15$ prompts used in our evaluation protocol.\\[-0.3cm]}
\begin{tabular}{c | c}
    \bottomrule
    ``A photo of a $S_*$'' & ``A photo of $S_*$ in the jungle'' \\
    \hline
    ``A photo of $S_*$ on a beach'' & ``A photo of $S_*$ in Times Square'' \\
    \hline
    ``A photo of $S_*$ in the moon'' & \begin{tabular}{c}``A painting of $S_*$ \\ in the style of Monet''\end{tabular} \\
    \hline
    ``Oil painting of $S_*$'' & ``A Marc Chagall painting of $S_*$'' \\
    \hline
    ``A manga drawing of $S_*$'' & ``A watercolor painting of $S_*$'' \\
    \hline
    ``A statue of $S_*$'' & ``App icon of $S_*$'' \\
    \hline
    ``A sand sculpture of $S_*$'' & ``Colorful graffiti of $S_*$'' \\
    \hline
    \begin{tabular}{c}``A photograph of two \\ $S_*$ on a table'' \end{tabular} & \mbox{} \\
    \bottomrule
\end{tabular}
\label{tb:prompts}
\vspace{-0.3cm}
\end{table}

\begin{figure*}
    \centering
    \renewcommand{\arraystretch}{0.3}
    \setlength{\tabcolsep}{0.5pt}

    {\small

    \begin{tabular}{c@{\hspace{0.2cm}} c c @{\hspace{0.2cm}} c c @{\hspace{0.2cm}} c c @{\hspace{0.2cm}} c c }

        \begin{tabular}{c} Real Sample \\ \& Prompt \end{tabular} &
        \multicolumn{2}{c}{No Time Conditioning} &
        \multicolumn{2}{c}{No Space Conditioning} &
        \multicolumn{2}{c}{\begin{tabular}{c} No Space nor \\ Time Conditioning \end{tabular}} &
        \multicolumn{2}{c}{\begin{tabular}{c} Both Space and \\ Time Conditioning \end{tabular}} \\

        \includegraphics[width=0.0975\textwidth]{images/original/rainbow_cat.jpeg} &
        \includegraphics[width=0.0975\textwidth]{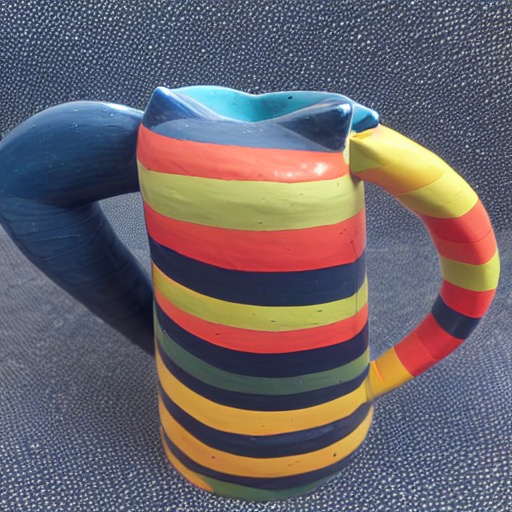} &
        \includegraphics[width=0.0975\textwidth]{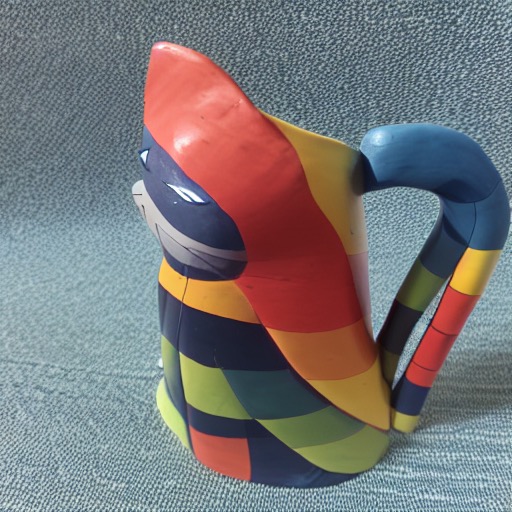} &
        \hspace{0.05cm}
        \includegraphics[width=0.0975\textwidth]{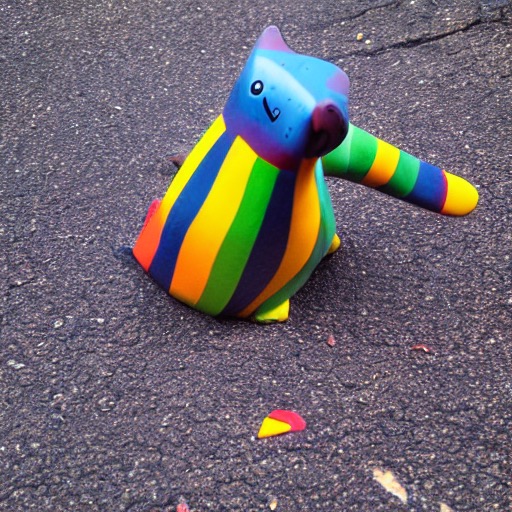} &
        \includegraphics[width=0.0975\textwidth]{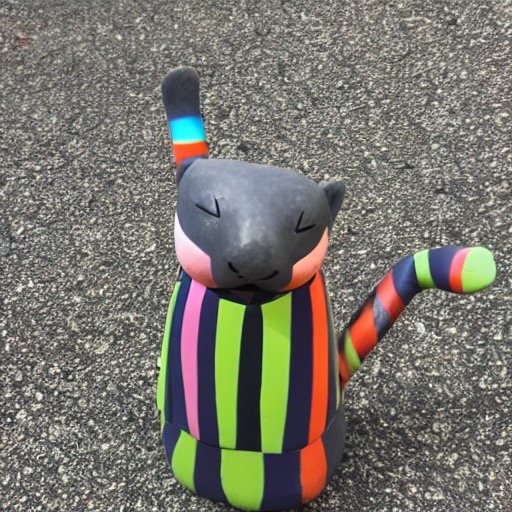} &
        \hspace{0.05cm}
        \includegraphics[width=0.0975\textwidth]{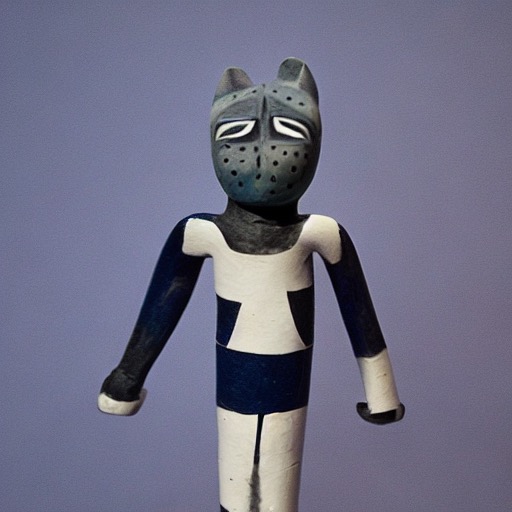} &
        \includegraphics[width=0.0975\textwidth]{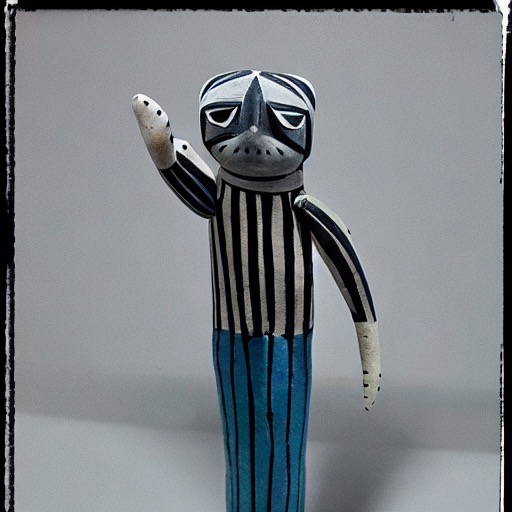} &
        \hspace{0.05cm}
        \includegraphics[width=0.0975\textwidth]{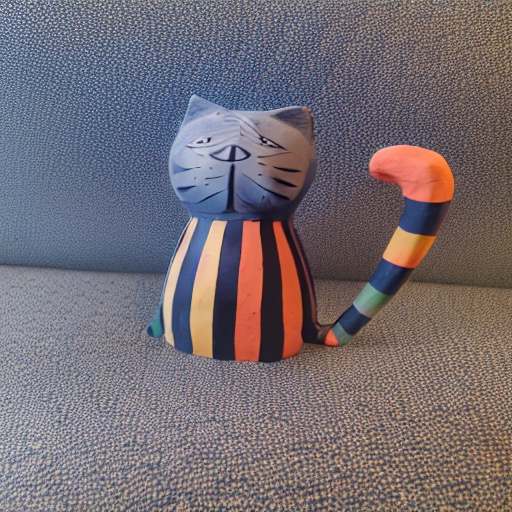} &
        \includegraphics[width=0.0975\textwidth]{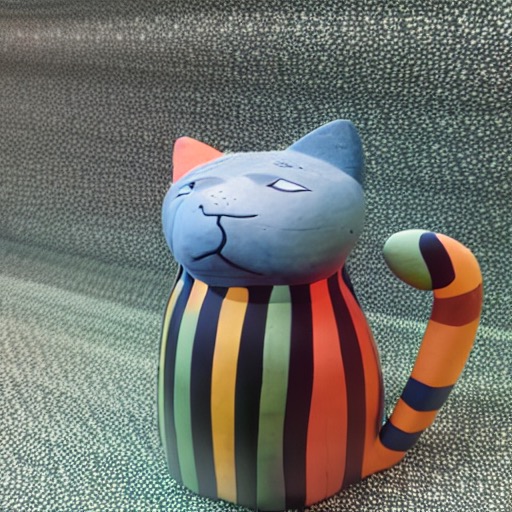} \\

        \raisebox{0.3in}{\begin{tabular}{c} ``A photo \\ of $S_*$''\end{tabular}} &
        \includegraphics[width=0.0975\textwidth]{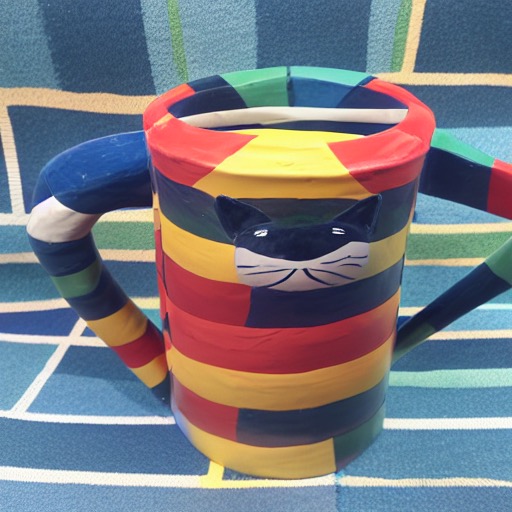} &
        \includegraphics[width=0.0975\textwidth]{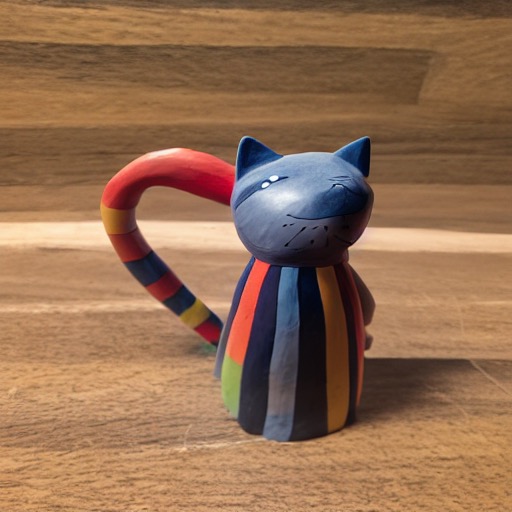} &
        \hspace{0.05cm}
        \includegraphics[width=0.0975\textwidth]{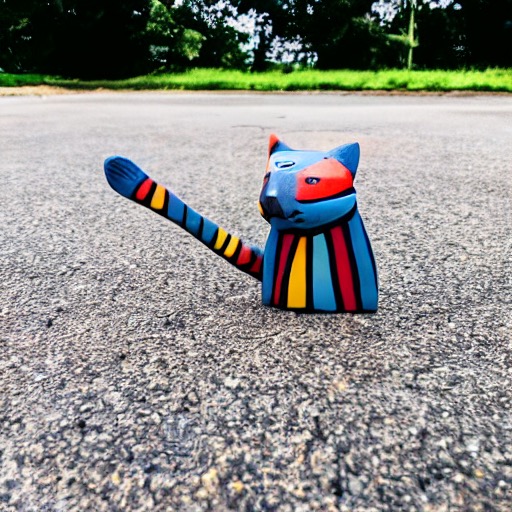} &
        \includegraphics[width=0.0975\textwidth]{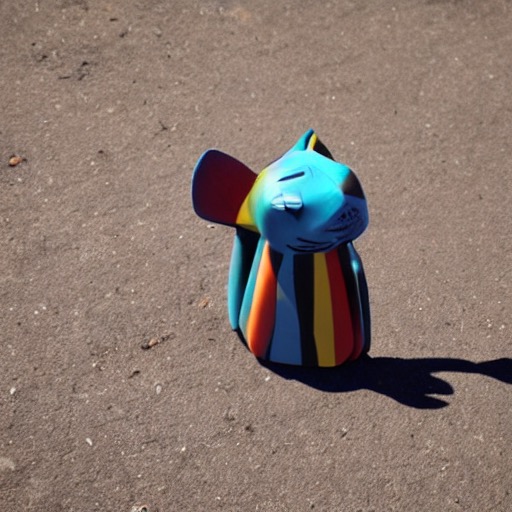} &
        \hspace{0.05cm}
        \includegraphics[width=0.0975\textwidth]{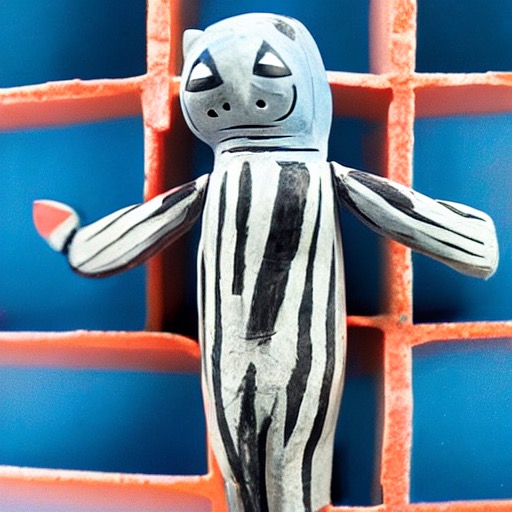} &
        \includegraphics[width=0.0975\textwidth]{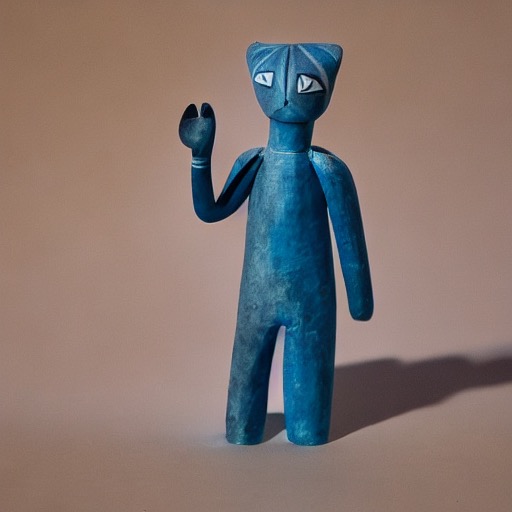} &
        \hspace{0.05cm}
        \includegraphics[width=0.0975\textwidth]{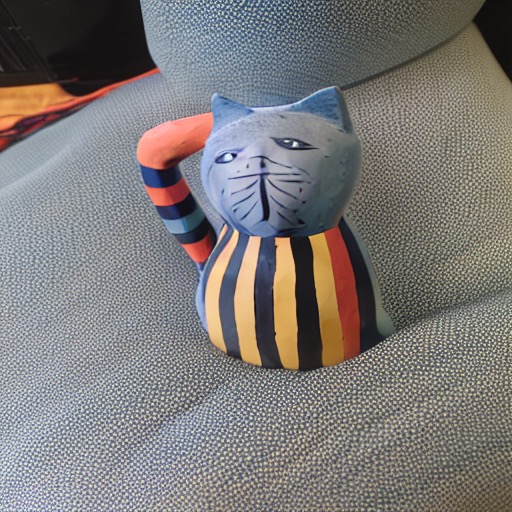} &
        \includegraphics[width=0.0975\textwidth]{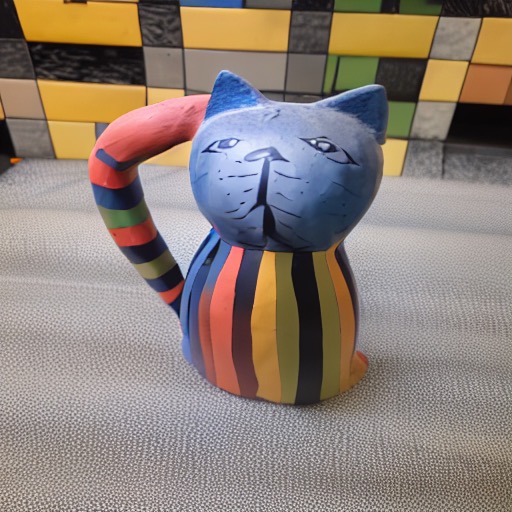} \\ \\

        \includegraphics[width=0.0975\textwidth]{images/original/colorful_teapot.jpg} &
        \includegraphics[width=0.0975\textwidth]{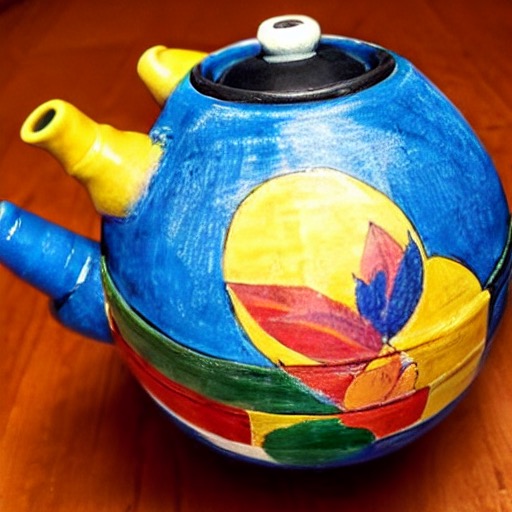} &
        \includegraphics[width=0.0975\textwidth]{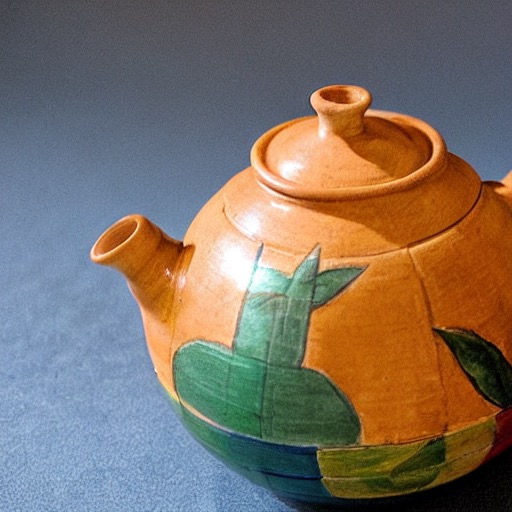} &
        \hspace{0.05cm}
        \includegraphics[width=0.0975\textwidth]{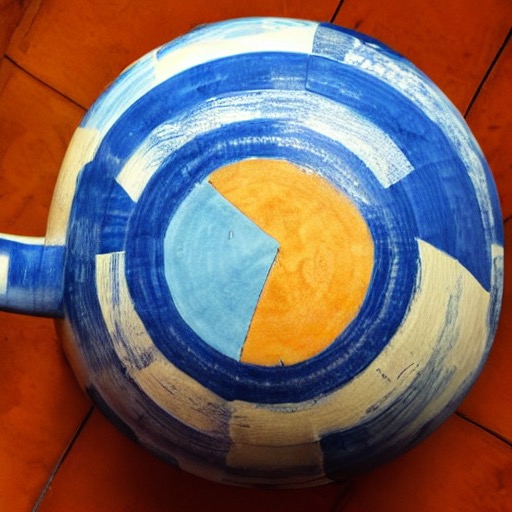} &
        \includegraphics[width=0.0975\textwidth]{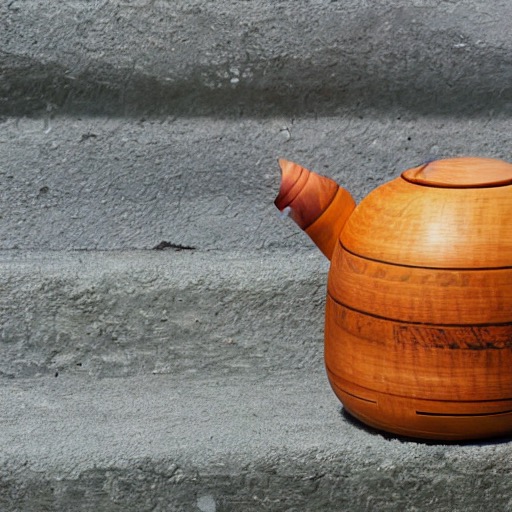} &
        \hspace{0.05cm}
        \includegraphics[width=0.0975\textwidth]{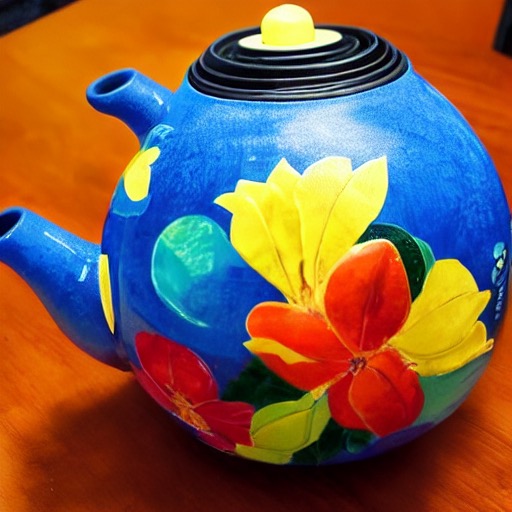} &
        \includegraphics[width=0.0975\textwidth]{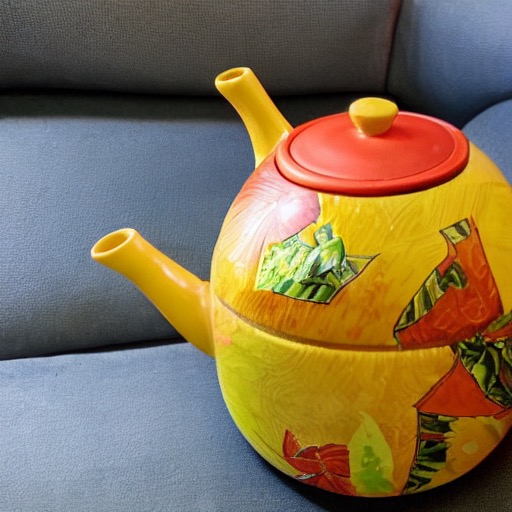} &
        \hspace{0.05cm}
        \includegraphics[width=0.0975\textwidth]{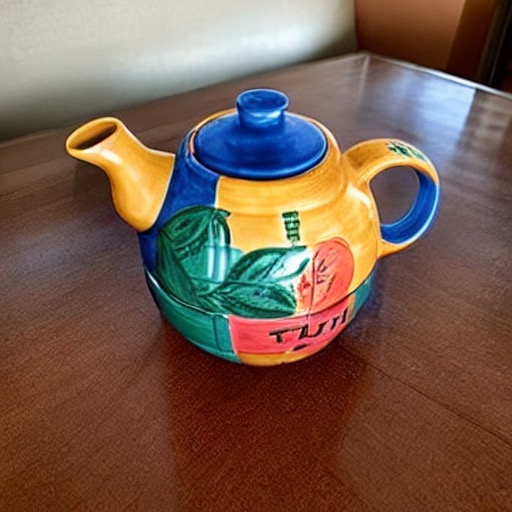} &
        \includegraphics[width=0.0975\textwidth]{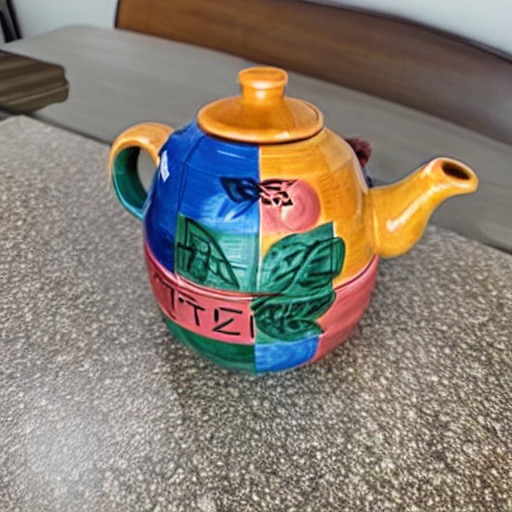} \\

        \raisebox{0.3in}{\begin{tabular}{c} ``A photo \\ of $S_*$''\end{tabular}} &
        \includegraphics[width=0.0975\textwidth]{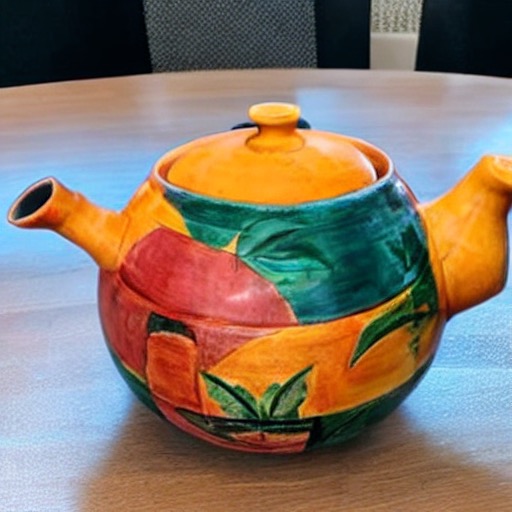} &
        \includegraphics[width=0.0975\textwidth]{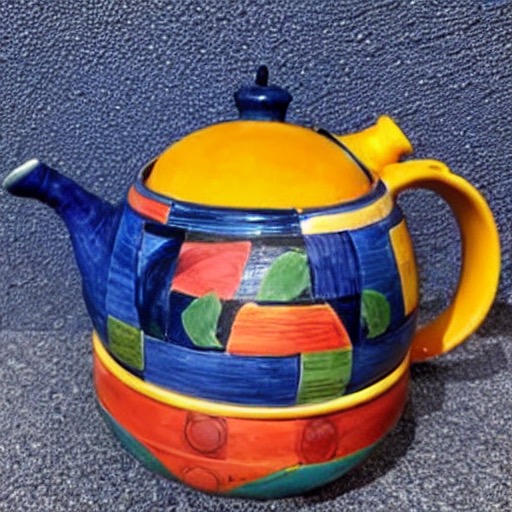} &
        \hspace{0.05cm}
        \includegraphics[width=0.0975\textwidth]{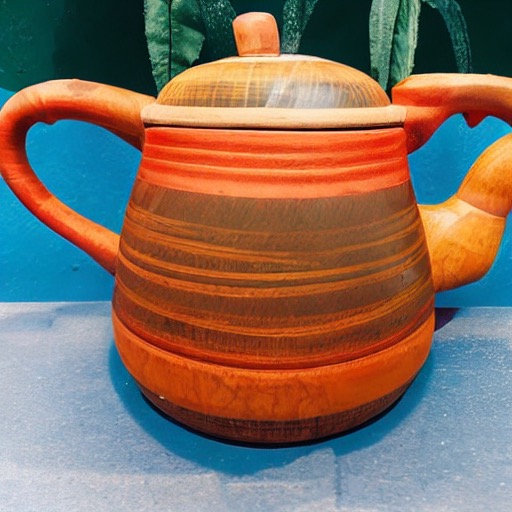} &
        \includegraphics[width=0.0975\textwidth]{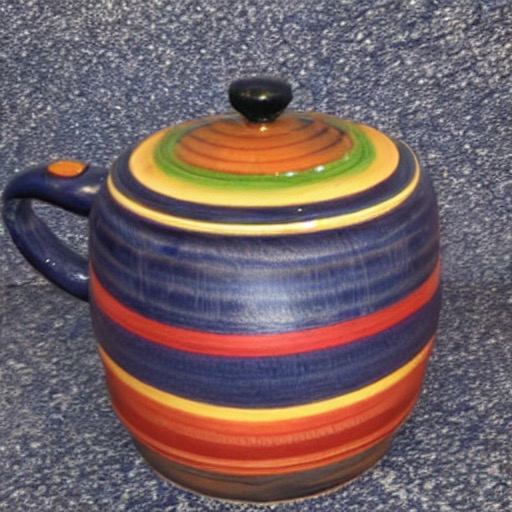} &
        \hspace{0.05cm}
        \includegraphics[width=0.0975\textwidth]{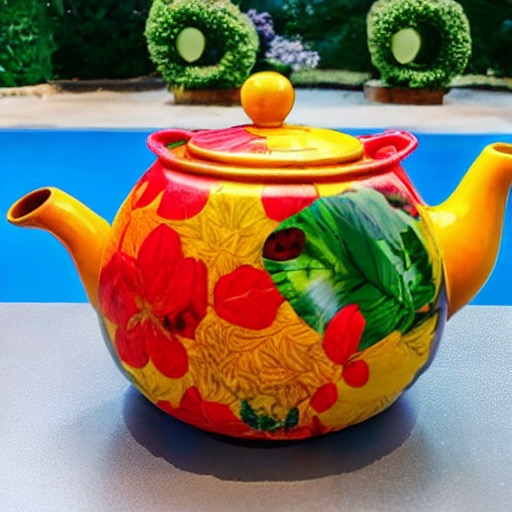} &
        \includegraphics[width=0.0975\textwidth]{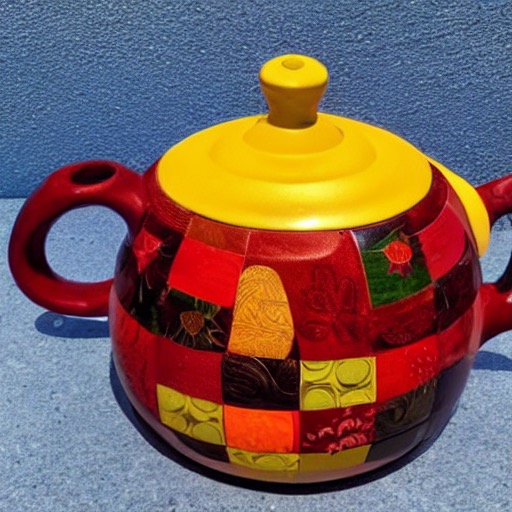} &
        \hspace{0.05cm}
        \includegraphics[width=0.0975\textwidth]{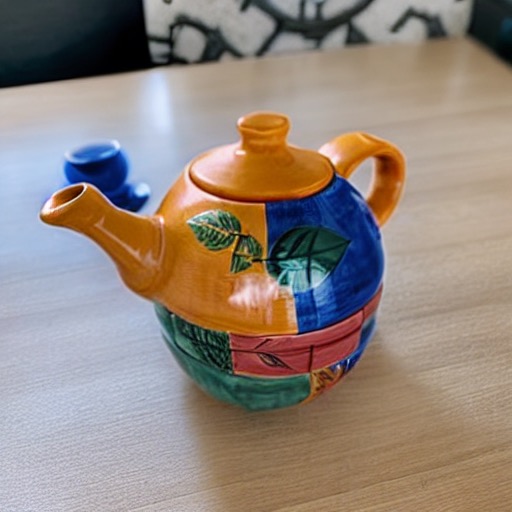} &
        \includegraphics[width=0.0975\textwidth]{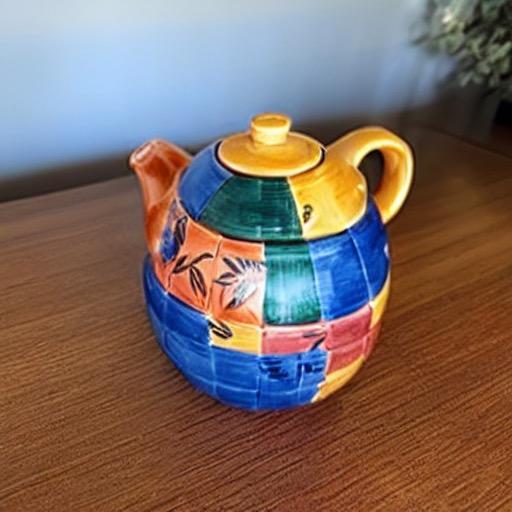} \\ \\

    \end{tabular}
    \\[-0.4cm]
    }
    \caption{Additional results validating our space-time conditioning of NeTI. We train NeTI with and without our time and space conditioning. All models are trained for the same number of optimization steps. As can be seen, the combination of both time and space is essential for attaining high visual fidelity.}
    \label{fig:ablation_space_time_conditioning}
    \vspace{-0.3cm}
\end{figure*}

\vspace{-0.25cm}
\paragraph{\textbf{Concepts and Text Prompts.}}
Below, we list the set of concepts used across all of our evaluations. From Gal~\etal~\shortcite{gal2023image}, we use the following $10$ concepts: \\[-0.1cm]

\begin{minipage}[t]{.45\linewidth}
    \begin{itemize}
        \item Clock
        \item Colorful Teapot
        \item Dangling Child
        \item Elephant
        \item Fat Stone Bird 
    \end{itemize}
\end{minipage}%
\begin{minipage}[t]{.45\linewidth}
    \begin{itemize}
        \item Headless Statue
        \item Metal Bird
        \item Mugs Skulls 
        \item Rainbow Cat Statue
        \item Red Bowl \\
    \end{itemize}
\end{minipage}

From Kumari~\etal~\shortcite{kumari2022customdiffusion}, we use $6$ concepts: \\[-0.2cm]

\begin{minipage}[t]{.175\linewidth}
    \begin{itemize}
        \item Barn
        \item Cat
    \end{itemize}
\end{minipage}
\begin{minipage}[t]{.275\linewidth}
    \begin{itemize}
        \item Dog
        \item Teddybear 
    \end{itemize}
\end{minipage}
\begin{minipage}[t]{.475\linewidth}
    \begin{itemize}
        \item Tortoise Plushy
        \item Wooden Pot \\
    \end{itemize}
\end{minipage}

All models are trained on the same training set and initialization token, when applicable. For a list of all $15$ text prompts considered in the evaluation protocol, please refer to~\Cref{tb:prompts}.

\vspace{0.3cm}
\section{Storage Requirements}
When applying our textual bypass, our mapper networks contain approximately $560,000$ learnable parameters. When textual bypass is not applied, this reduces to approximately $460,000$ trainable parameters. This amounts to 2.2MB and 1.86MB of disk space required to store each learned concept, respectively. 

Thanks to our use of Nested Dropout, we can further compress the representation of the concept by dropping a significant subset of parameters in the network's final layer. When we reduce the number of units in the final layer from the full $128$ units to $32$ units, the number of parameters decreases to $390,000$ parameters with no textual bypass, a decrease of $15\%$, requiring 1.56MB of disk space per concept. When keeping only the first $8$ units our the final layer, this decreases the $367,000$ parameters or 1.49MB of disk space. 

As a reference to alternative methods, DreamBooth~\cite{ruiz2022dreambooth} requires several GBs of disk space per concept while CustomDiffusion~\cite{kumari2022customdiffusion} requires approximately 73MB of disk space. As shown in the main paper and in~\Cref{sec:custom_comparison}, NeTI attains comparable or better performance with a similar convergence rate while requiring a significantly lower storage footprint.

\section{Ablation Study}~\label{sec:ablation_study}

\vspace{-0.8cm}
\paragraph{\textbf{Conditioning on Both Time and Space.}}
In~\Cref{sec:analysis}, we validated the use of both time and space when conditioning our neural mapper. In~\Cref{fig:ablation_space_time_conditioning}, we provide additional examples of reconstructions obtained when training NeTI with and without each conditioning modality.

\begin{figure*}
    \centering
    \renewcommand{\arraystretch}{0.3}
    \setlength{\tabcolsep}{0.5pt}

    {\small

    \begin{tabular}{c@{\hspace{0.2cm}} c c @{\hspace{0.2cm}} c c @{\hspace{0.2cm}} c c @{\hspace{0.2cm}} c c }

        \\ \\ \\ \\ \\ \\ \\ \\ \\ \\ \\ \\
        \\ \\ \\ \\ \\ \\ \\ \\ \\ \\ \\ \\
        \\ \\ \\ \\

        \begin{tabular}{c} Real Sample \end{tabular} &
        \multicolumn{2}{c}{NeTI w/o PE} &
        \multicolumn{2}{c}{NeTI w/o Nested Dropout} &
        \multicolumn{2}{c}{\begin{tabular}{c} NeTI \end{tabular}} &
        \multicolumn{2}{c}{\begin{tabular}{c} NeTI w/ Textual Bypass \end{tabular}} \\

        \includegraphics[width=0.0975\textwidth]{images/original/teddybear.jpg} &
        \includegraphics[width=0.0975\textwidth]{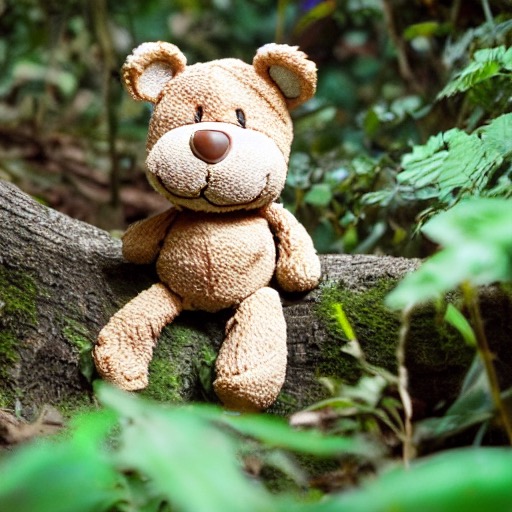} &
        \includegraphics[width=0.0975\textwidth]{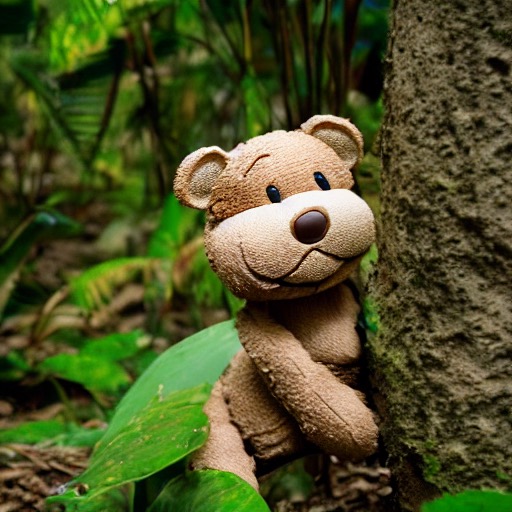} &
        \hspace{0.05cm}
        \includegraphics[width=0.0975\textwidth]{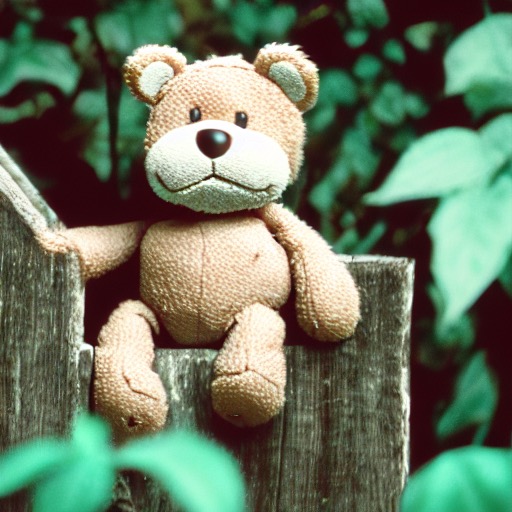} &
        \includegraphics[width=0.0975\textwidth]{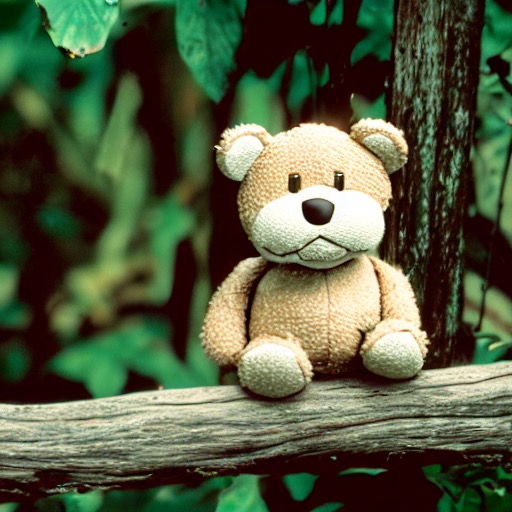} &
        \hspace{0.05cm}
        \includegraphics[width=0.0975\textwidth]{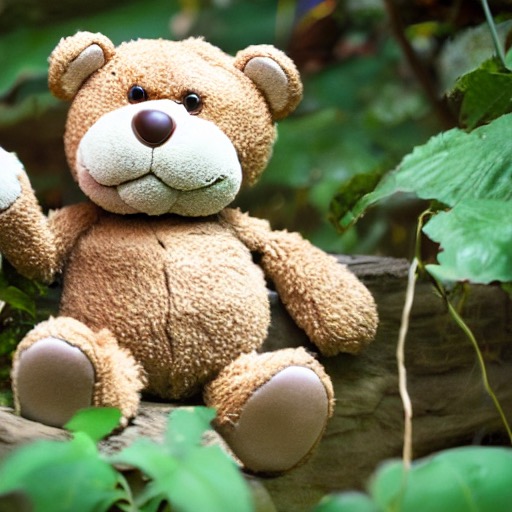} &
        \includegraphics[width=0.0975\textwidth]{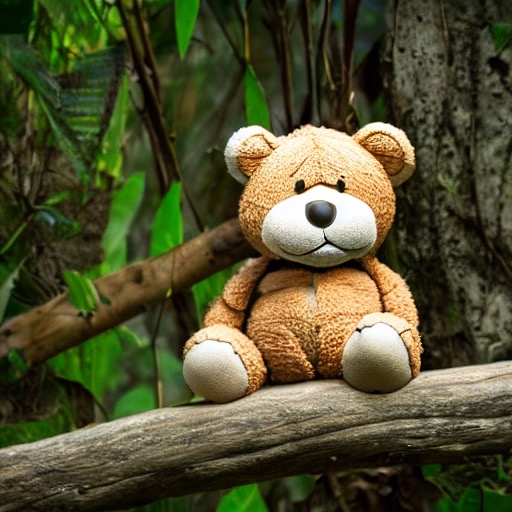} &
        \hspace{0.05cm}
        \includegraphics[width=0.0975\textwidth]{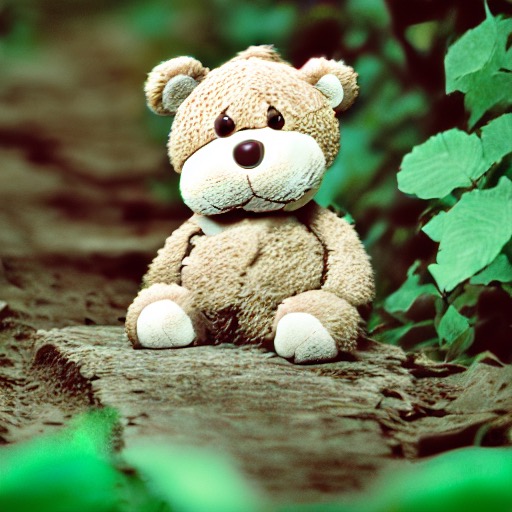} &
        \includegraphics[width=0.0975\textwidth]{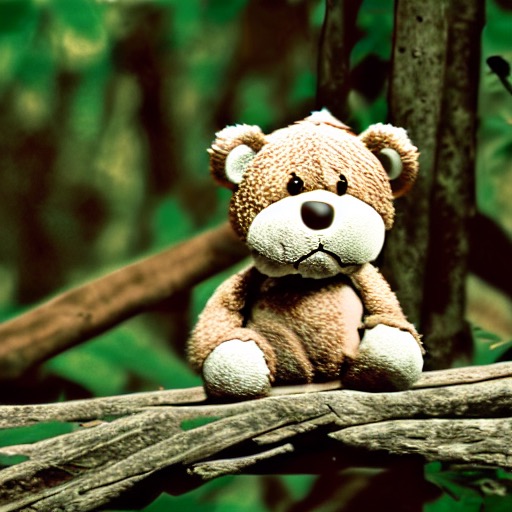} \\

        \raisebox{0.3in}{\begin{tabular}{c} ``A photo \\ \\[-0.05cm] of $S_*$ in \\ \\[-0.05cm] the jungle''\end{tabular}} &
        \includegraphics[width=0.0975\textwidth]{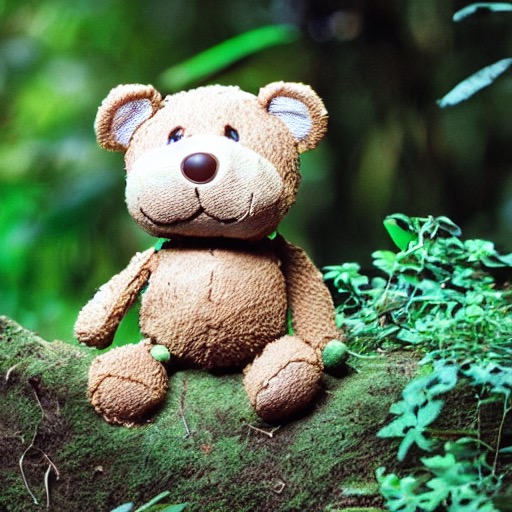} &
        \includegraphics[width=0.0975\textwidth]{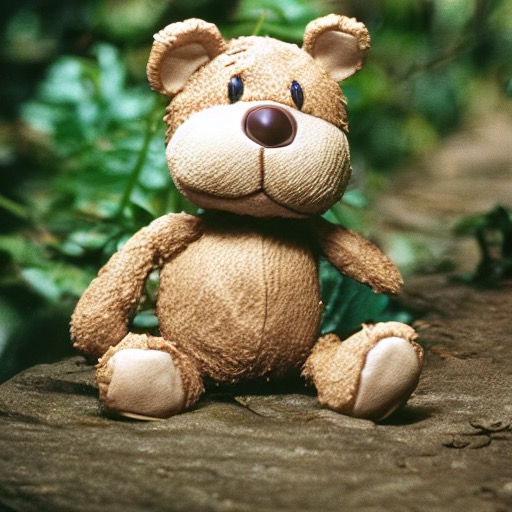} &
        \hspace{0.05cm}
        \includegraphics[width=0.0975\textwidth]{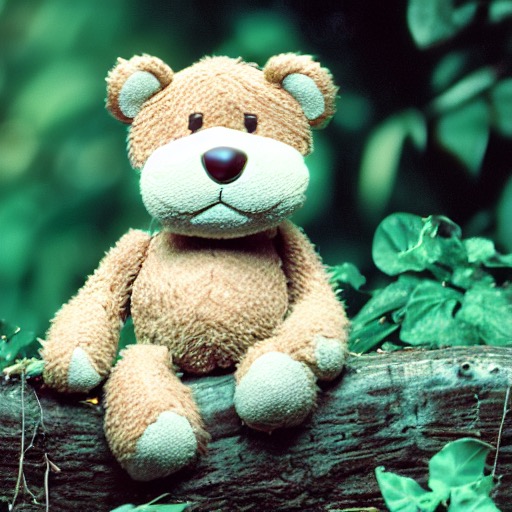} &
        \includegraphics[width=0.0975\textwidth]{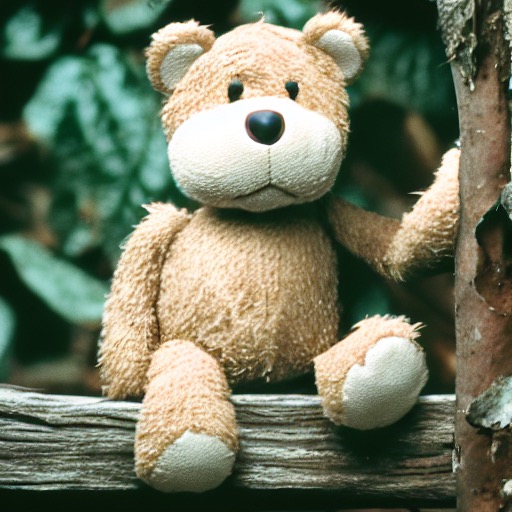} &
        \hspace{0.05cm}
        \includegraphics[width=0.0975\textwidth]{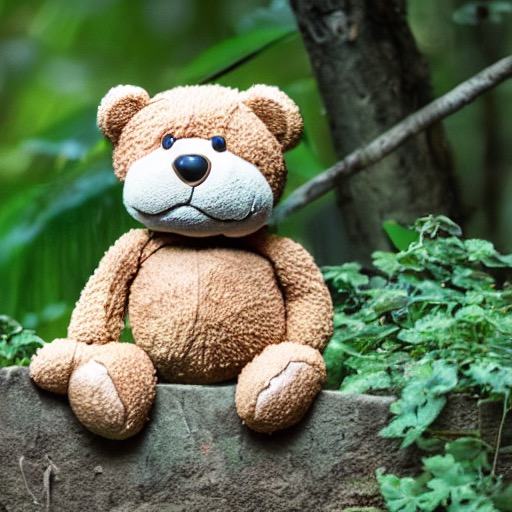} &
        \includegraphics[width=0.0975\textwidth]{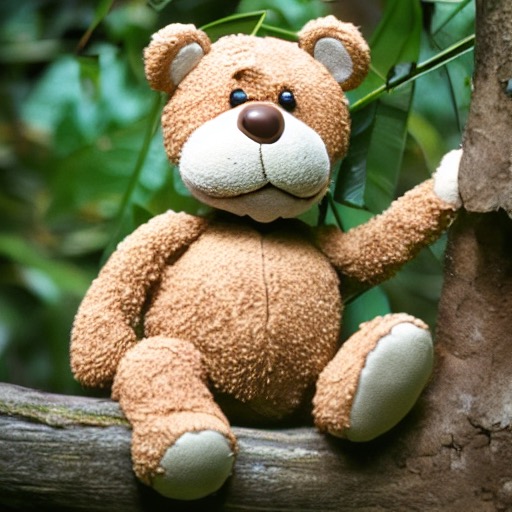} &
        \hspace{0.05cm}
        \includegraphics[width=0.0975\textwidth]{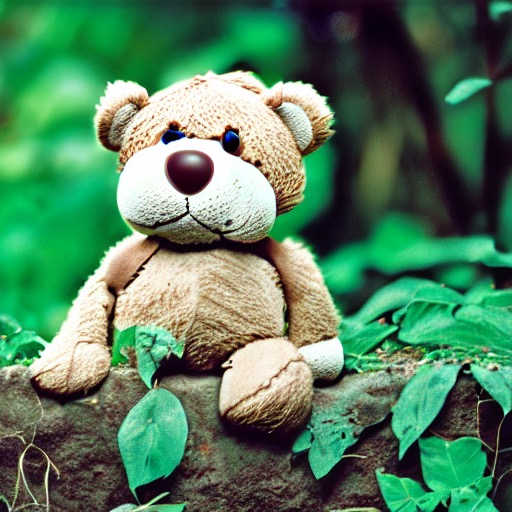} &
        \includegraphics[width=0.0975\textwidth]{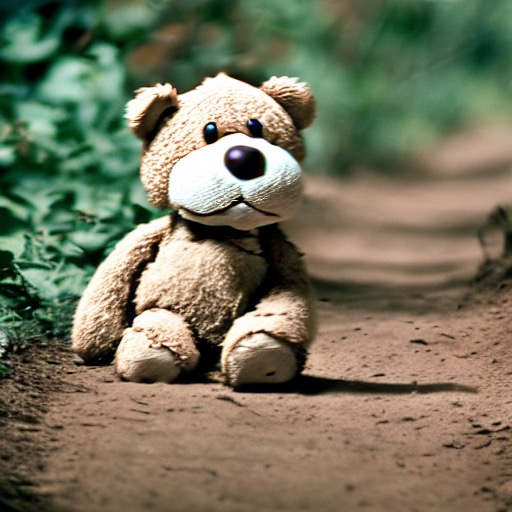} \\ \\

        \includegraphics[width=0.0975\textwidth]{images/original/mugs_skulls.jpeg} &
        \includegraphics[width=0.0975\textwidth]{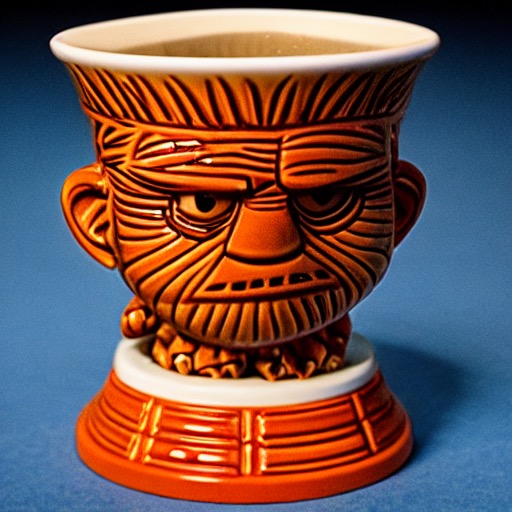} &
        \includegraphics[width=0.0975\textwidth]{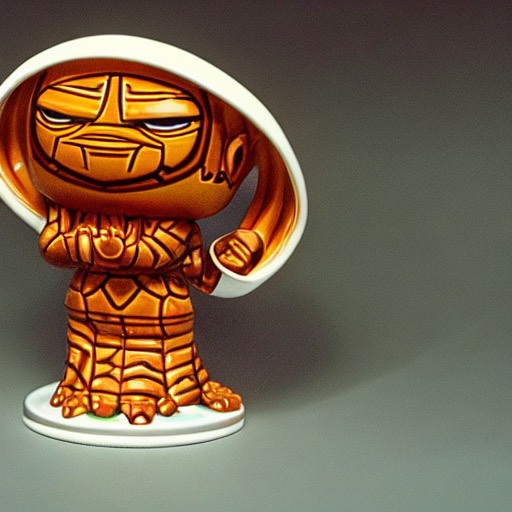} &
        \hspace{0.05cm}
        \includegraphics[width=0.0975\textwidth]{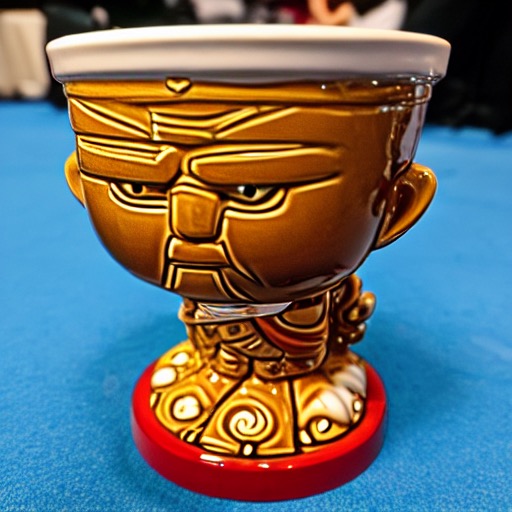} &
        \includegraphics[width=0.0975\textwidth]{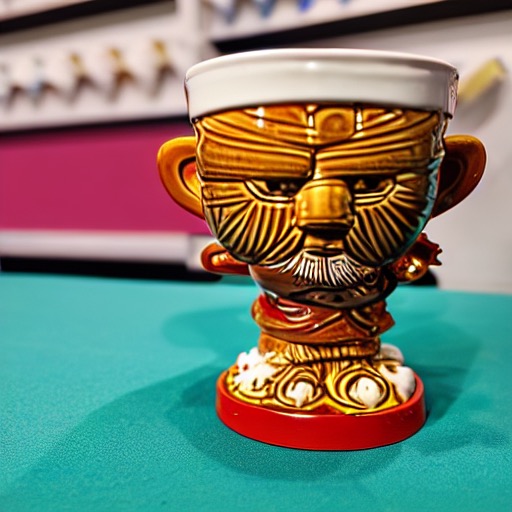} &
        \hspace{0.05cm}
        \includegraphics[width=0.0975\textwidth]{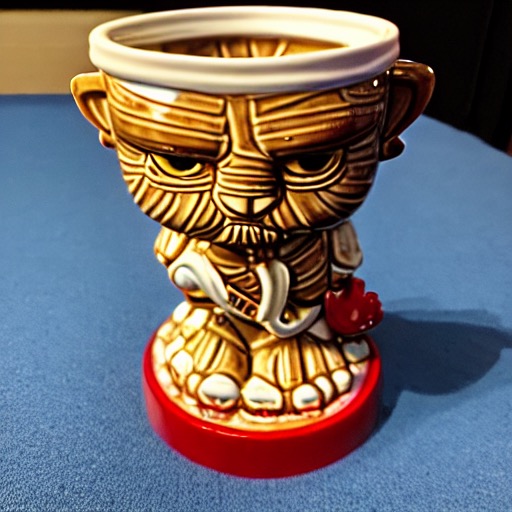} &
        \includegraphics[width=0.0975\textwidth]{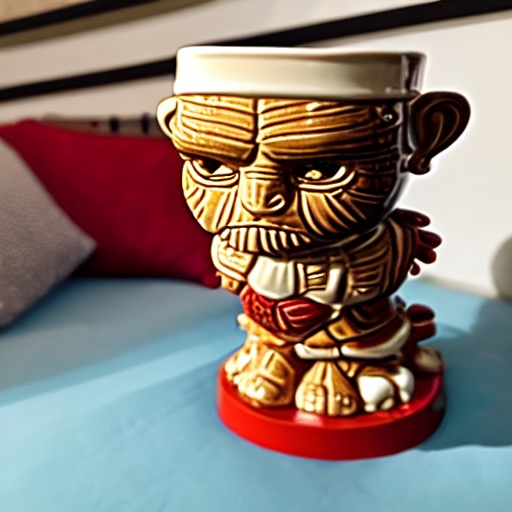} &
        \hspace{0.05cm}
        \includegraphics[width=0.0975\textwidth]{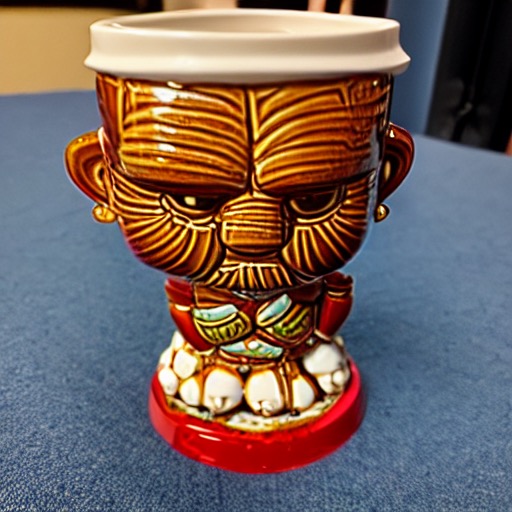} &
        \includegraphics[width=0.0975\textwidth]{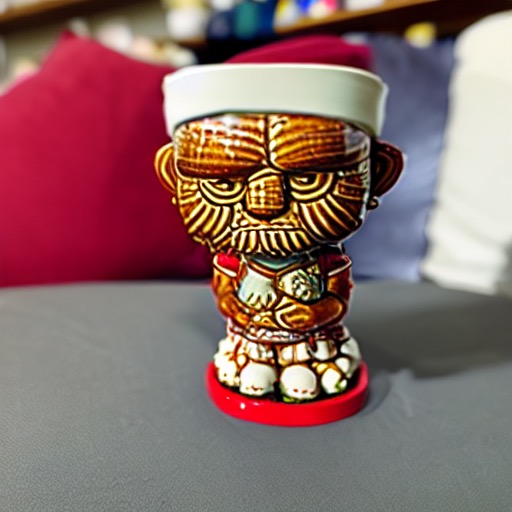} \\

        \raisebox{0.3in}{\begin{tabular}{c} ``A photo \\ \\[-0.05cm] of $S_*$''\end{tabular}} &
        \includegraphics[width=0.0975\textwidth]{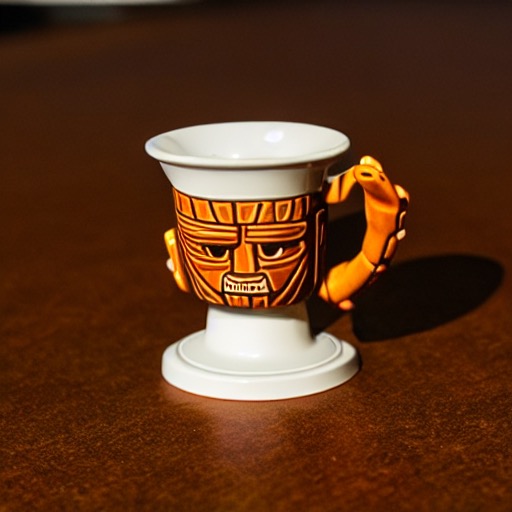} &
        \includegraphics[width=0.0975\textwidth]{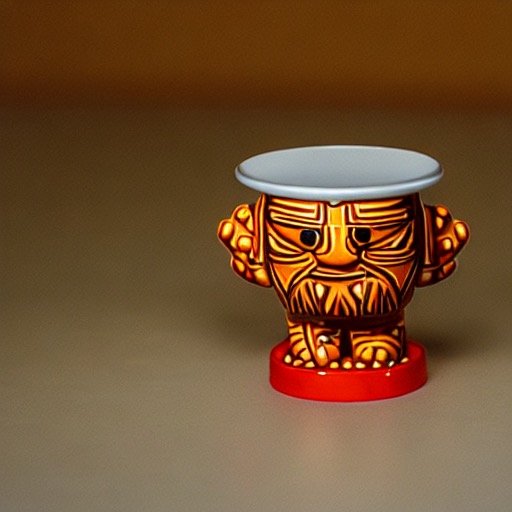} &
        \hspace{0.05cm}
        \includegraphics[width=0.0975\textwidth]{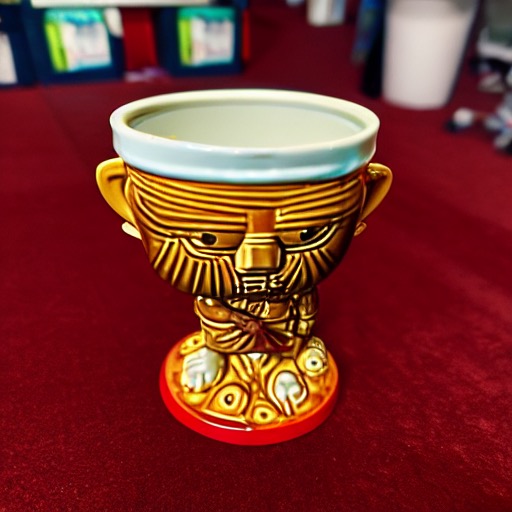} &
        \includegraphics[width=0.0975\textwidth]{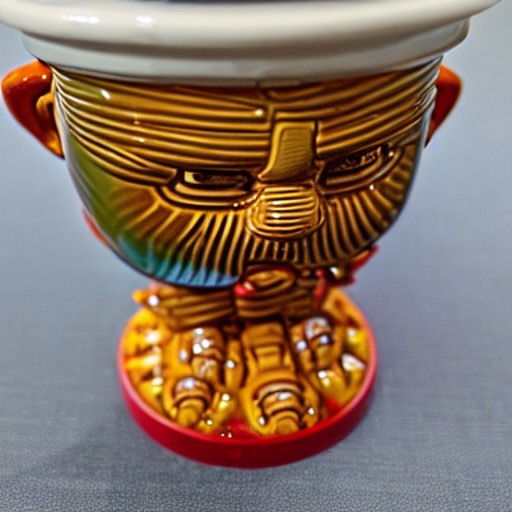} &
        \hspace{0.05cm}
        \includegraphics[width=0.0975\textwidth]{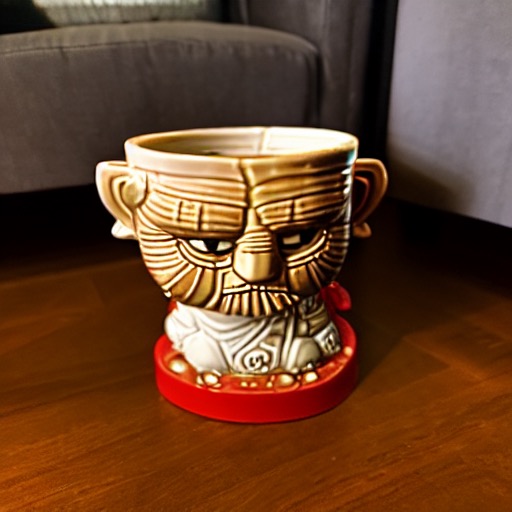} &
        \includegraphics[width=0.0975\textwidth]{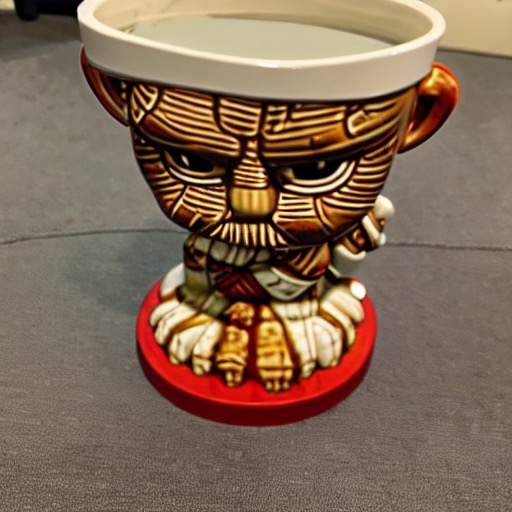} &
        \hspace{0.05cm}
        \includegraphics[width=0.0975\textwidth]{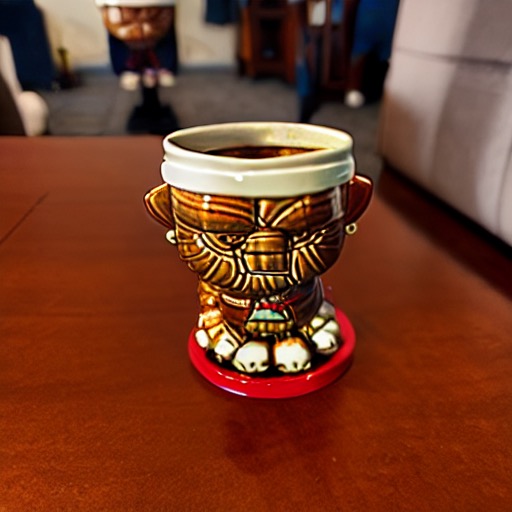} &
        \includegraphics[width=0.0975\textwidth]{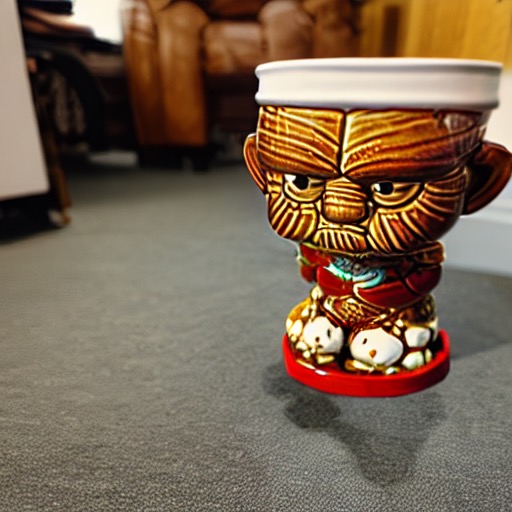} \\ \\

        \includegraphics[width=0.0975\textwidth]{images/original/rainbow_cat.jpeg} &
        \includegraphics[width=0.0975\textwidth]{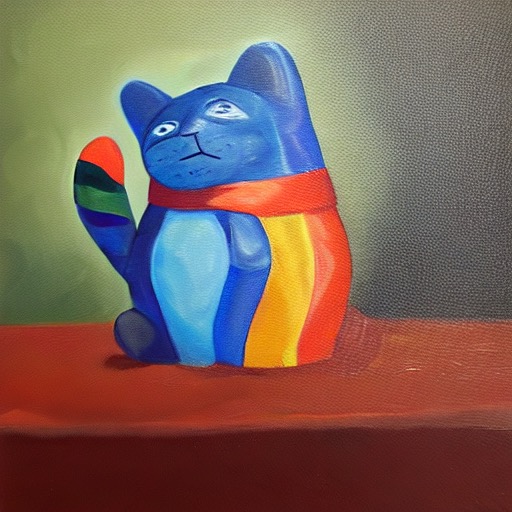} &
        \includegraphics[width=0.0975\textwidth]{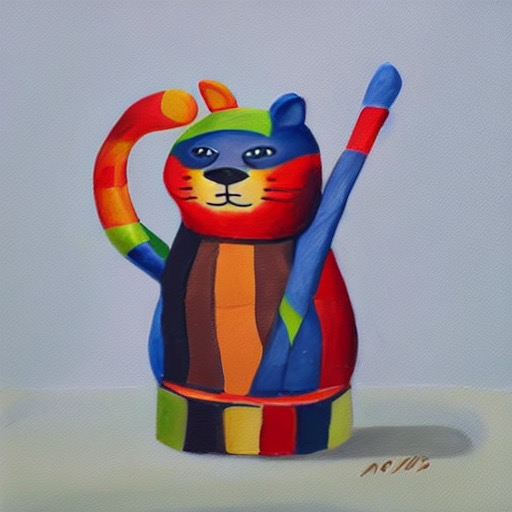} &
        \hspace{0.05cm}
        \includegraphics[width=0.0975\textwidth]{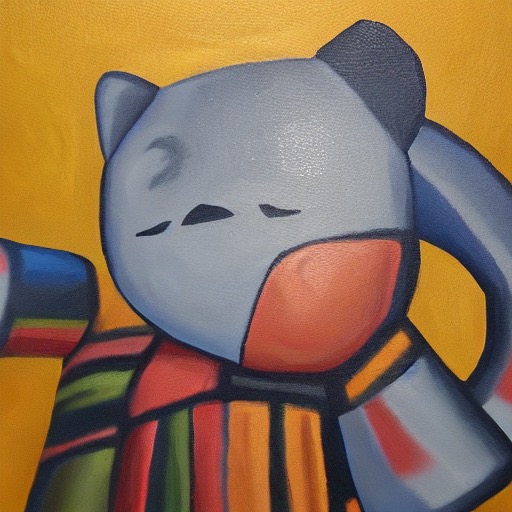} &
        \includegraphics[width=0.0975\textwidth]{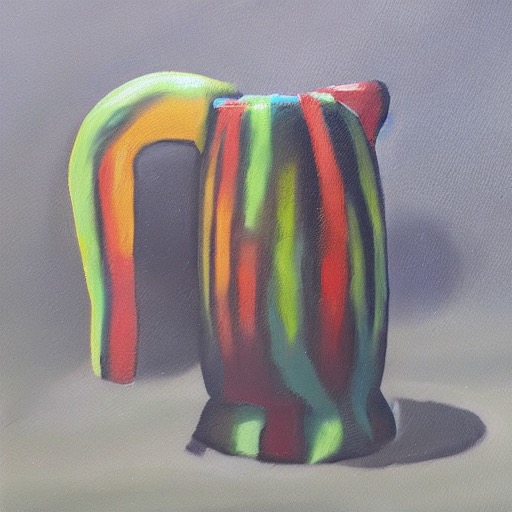} &
        \hspace{0.05cm}
        \includegraphics[width=0.0975\textwidth]{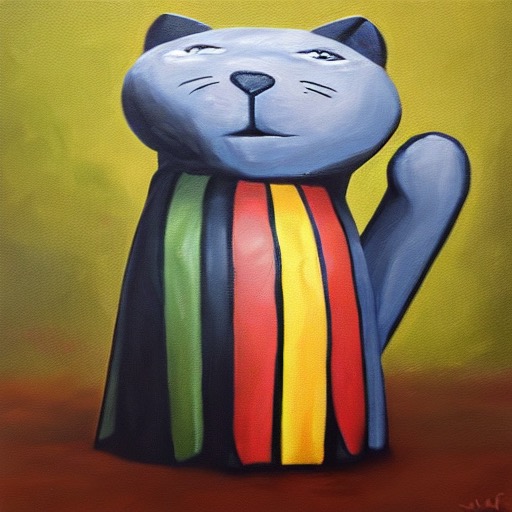} &
        \includegraphics[width=0.0975\textwidth]{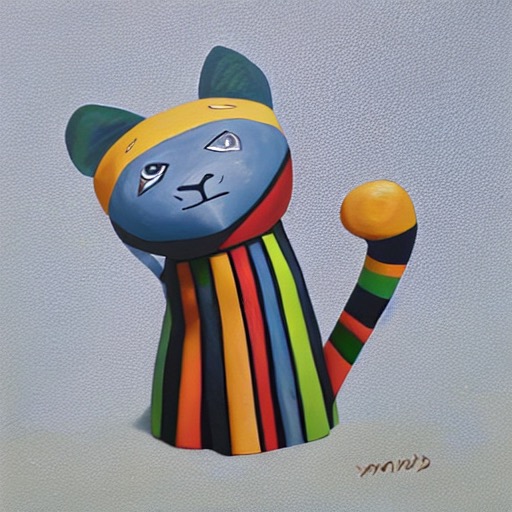} &
        \hspace{0.05cm}
        \includegraphics[width=0.0975\textwidth]{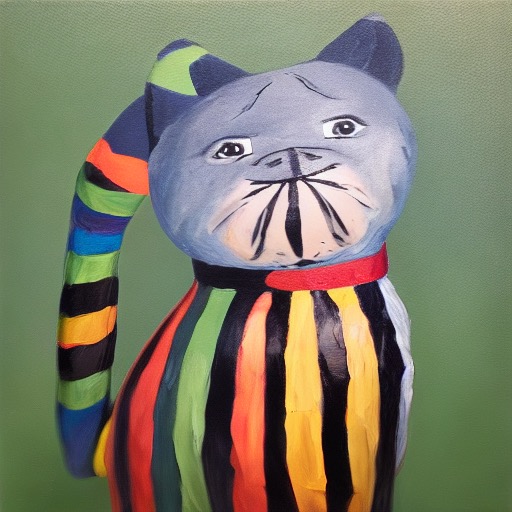} &
        \includegraphics[width=0.0975\textwidth]{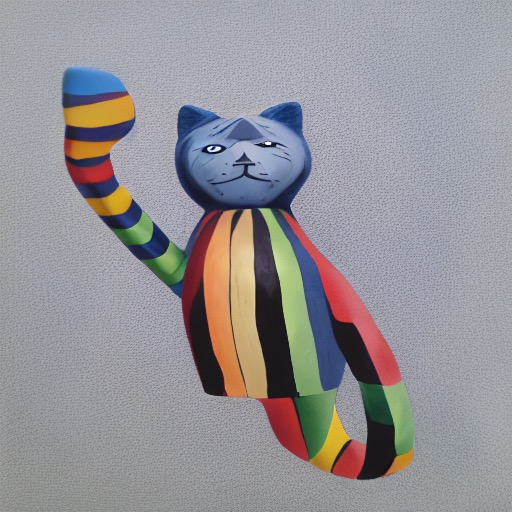} \\

        \raisebox{0.3in}{\begin{tabular}{c} ``An oil \\ \\[-0.05cm] painting of \\ \\[-0.05cm] $S_*$''\end{tabular}} &
        \includegraphics[width=0.0975\textwidth]{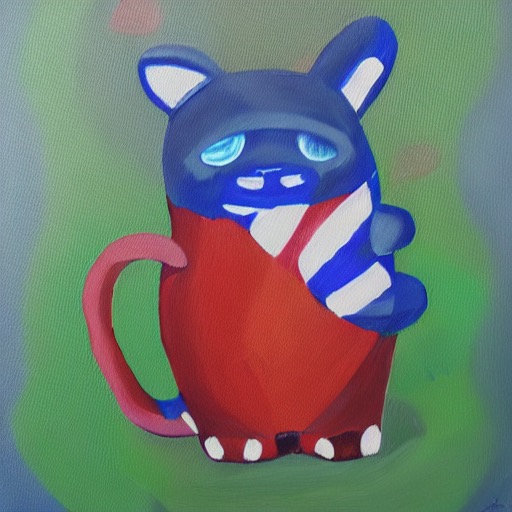} &
        \includegraphics[width=0.0975\textwidth]{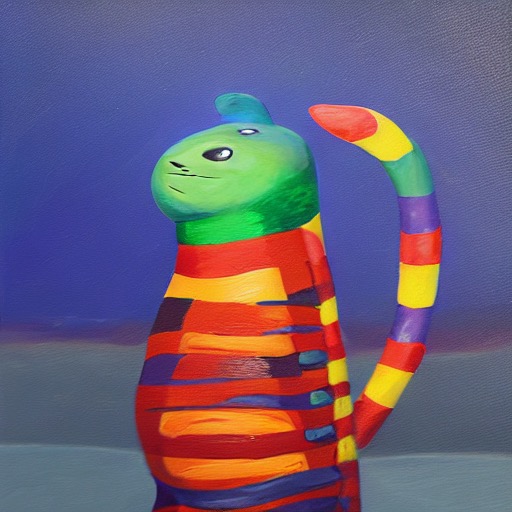} &
        \hspace{0.05cm}
        \includegraphics[width=0.0975\textwidth]{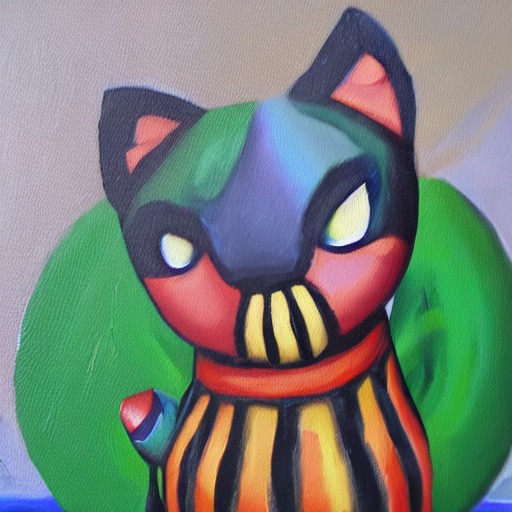} &
        \includegraphics[width=0.0975\textwidth]{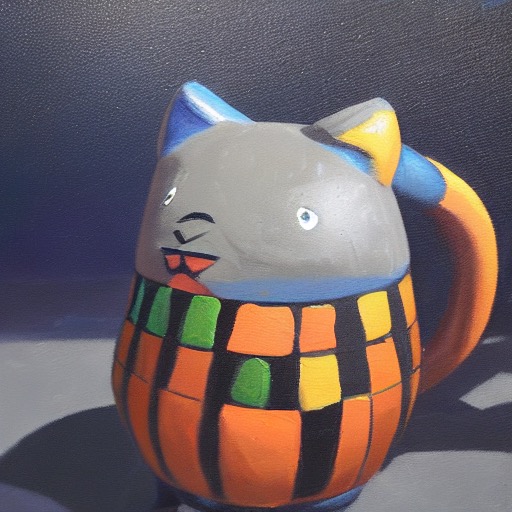} &
        \hspace{0.05cm}
        \includegraphics[width=0.0975\textwidth]{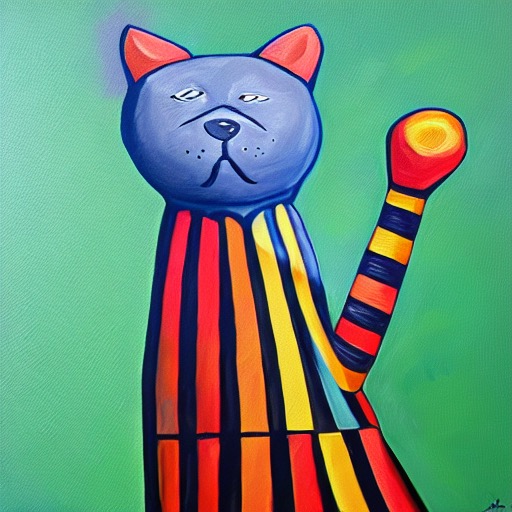} &
        \includegraphics[width=0.0975\textwidth]{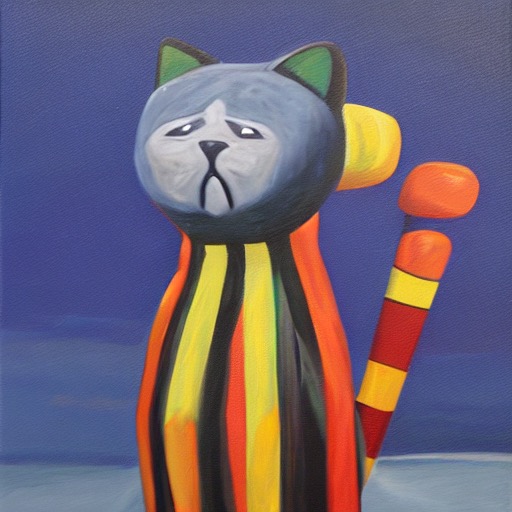} &
        \hspace{0.05cm}
        \includegraphics[width=0.0975\textwidth]{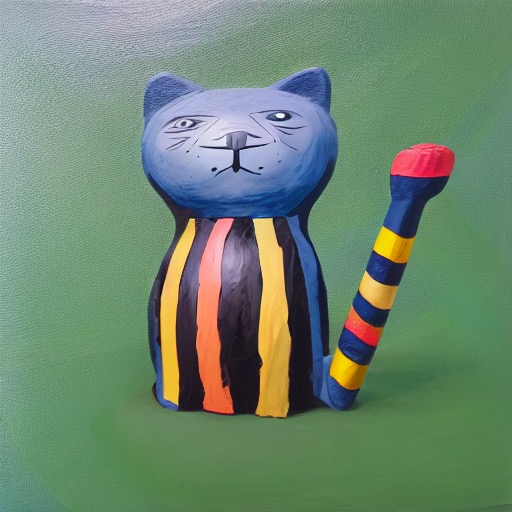} &
        \includegraphics[width=0.0975\textwidth]{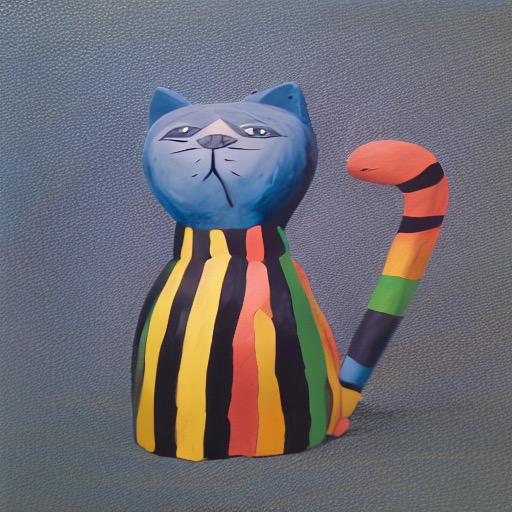} \\ \\

        \includegraphics[width=0.0975\textwidth]{images/original/dangling_child.jpg} &
        \includegraphics[width=0.0975\textwidth]{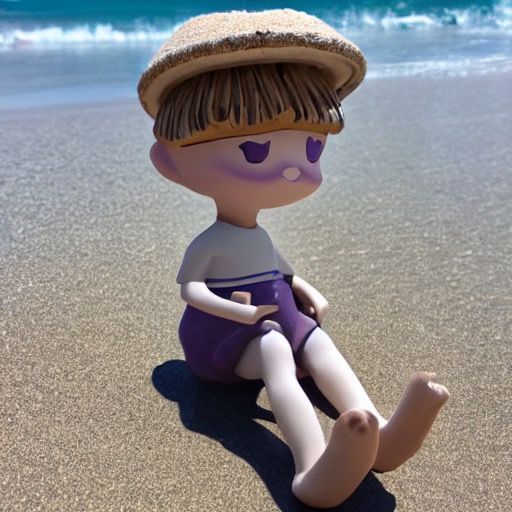} &
        \includegraphics[width=0.0975\textwidth]{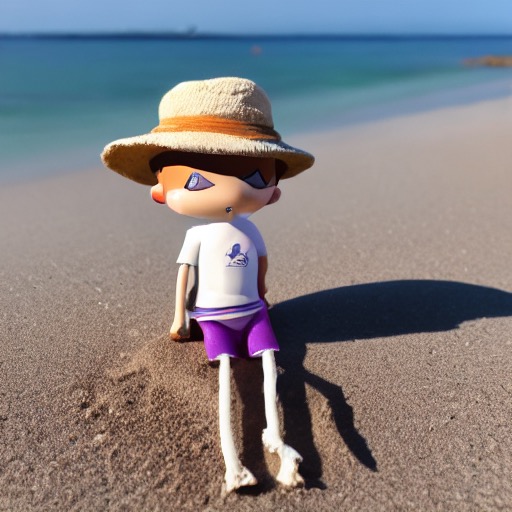} &
        \hspace{0.05cm}
        \includegraphics[width=0.0975\textwidth]{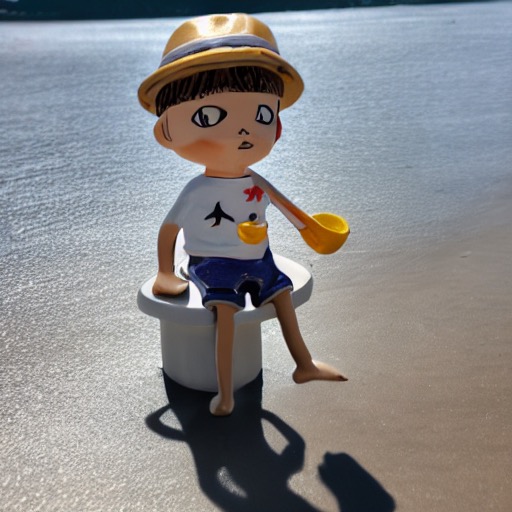} &
        \includegraphics[width=0.0975\textwidth]{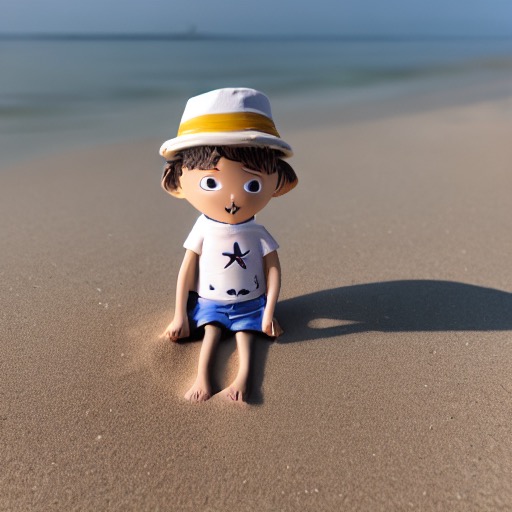} &
        \hspace{0.05cm}
        \includegraphics[width=0.0975\textwidth]{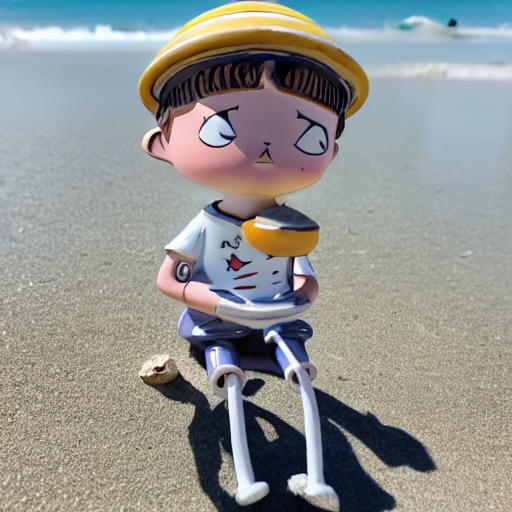} &
        \includegraphics[width=0.0975\textwidth]{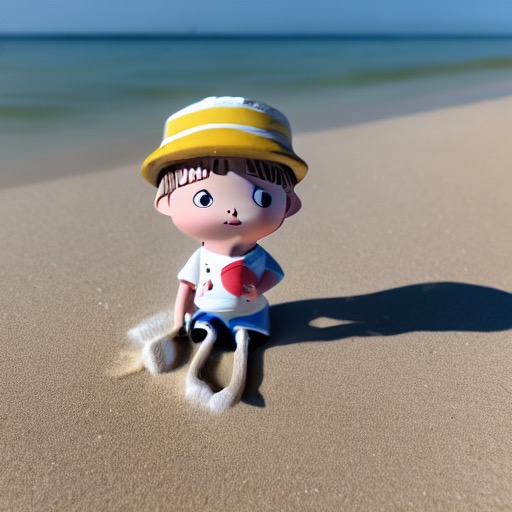} &
        \hspace{0.05cm}
        \includegraphics[width=0.0975\textwidth]{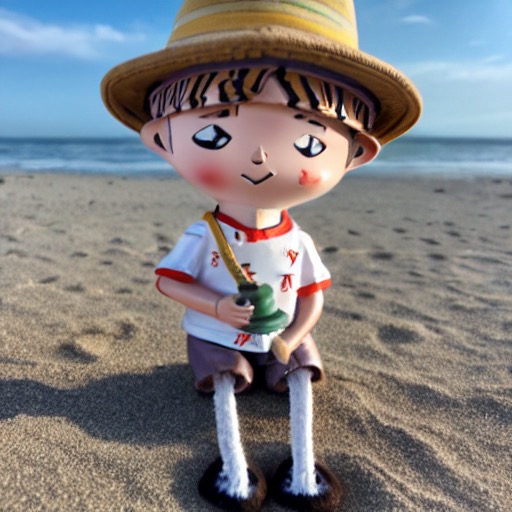} &
        \includegraphics[width=0.0975\textwidth]{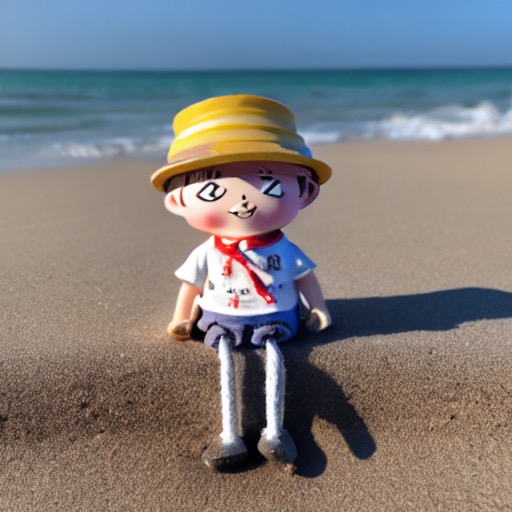} \\
        
        \raisebox{0.3in}{\begin{tabular}{c} ``A photo of \\ \\[-0.05cm] $S_*$ on  \\ \\[-0.05cm] the beach''\end{tabular}} &
        \includegraphics[width=0.0975\textwidth]{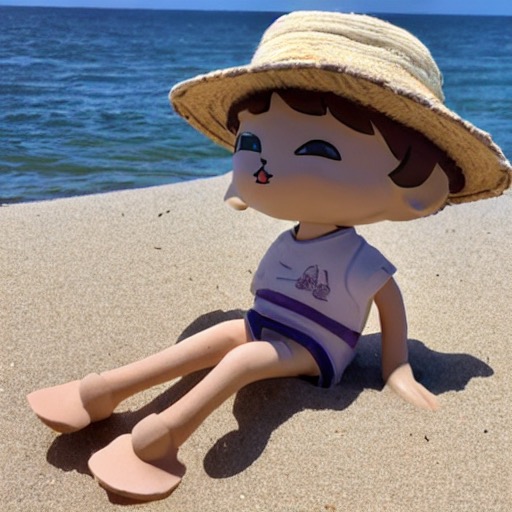} &
        \includegraphics[width=0.0975\textwidth]{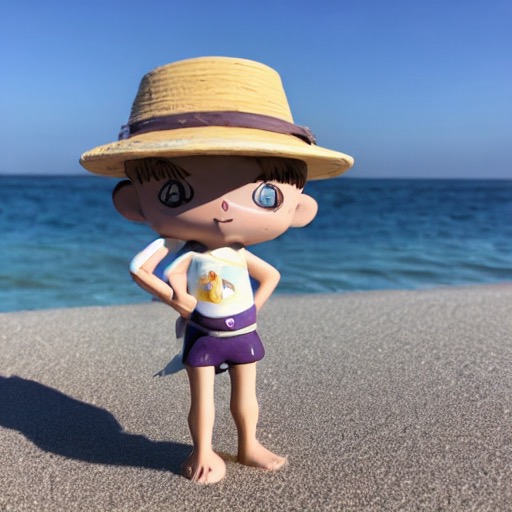} &
        \hspace{0.05cm}
        \includegraphics[width=0.0975\textwidth]{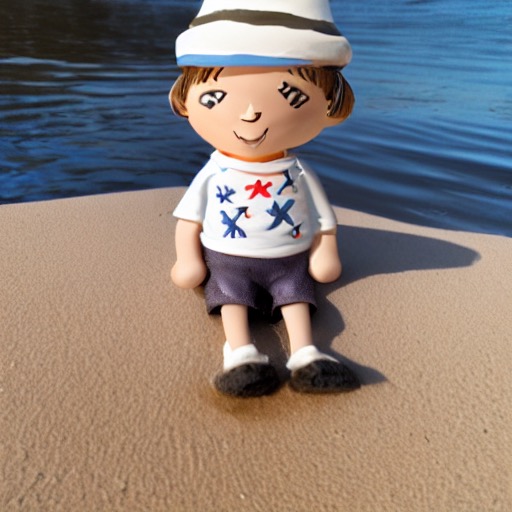} &
        \includegraphics[width=0.0975\textwidth]{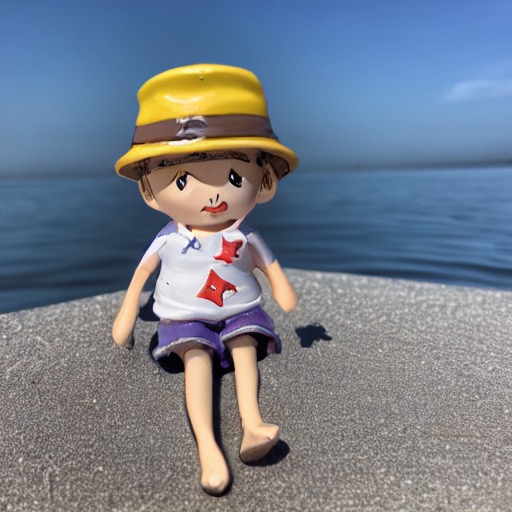} &
        \hspace{0.05cm}
        \includegraphics[width=0.0975\textwidth]{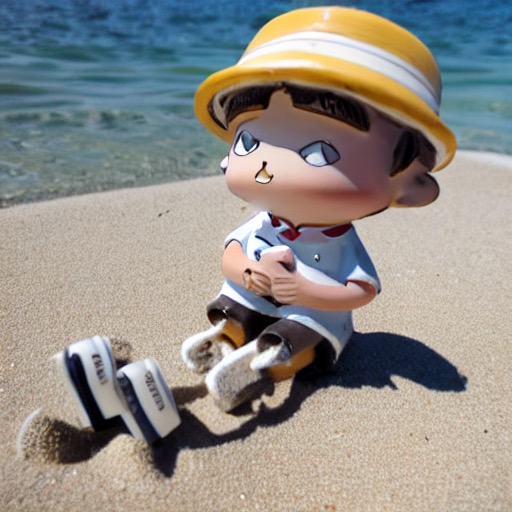} &
        \includegraphics[width=0.0975\textwidth]{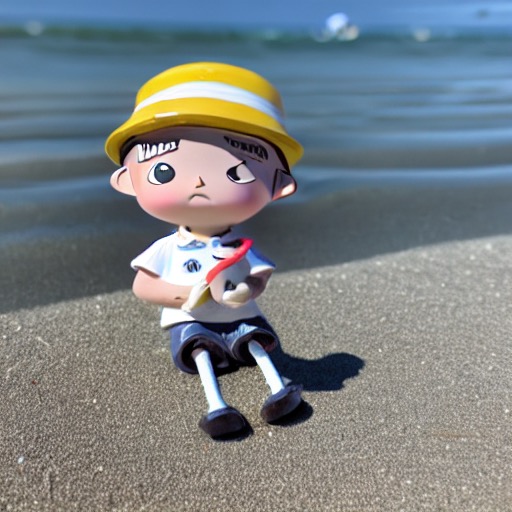} &
        \hspace{0.05cm}
        \includegraphics[width=0.0975\textwidth]{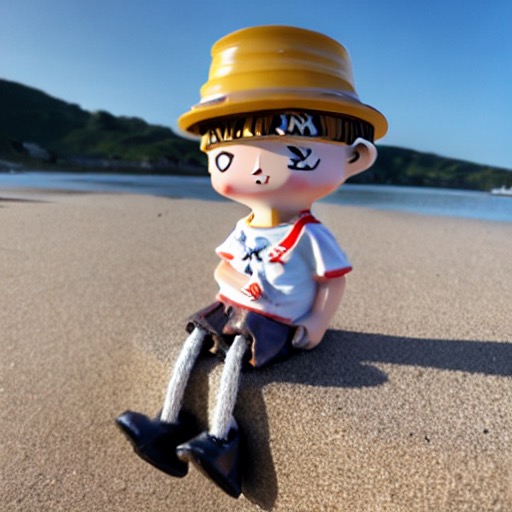} &
        \includegraphics[width=0.0975\textwidth]{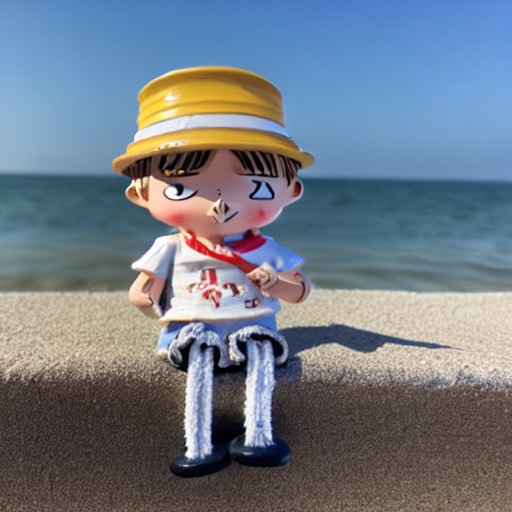} \\ \\
        
    \end{tabular}
    \\[-0.3cm]
    }
    \caption{Ablation study. We compare our NeTI models trained without our positional encoding function, without Nested Dropout applied during training, and without our textual bypass technique. As can be seen, all three components are essential for achieving high-fidelity reconstructions of the input while remaining faithful to the input text prompt, provided on the left.}
    \label{fig:ablation_network_architecture}
    \vspace{-0.2cm}
\end{figure*}

\paragraph{\textbf{Network Architecture and Training Scheme.}}
Having demonstrated the importance of our space-time representation, we now turn to analyze three key components of our network and training scheme: (1) the use of a positional encoding on the input pair $(t,\ell)$; (2) the use of Nested Dropout~\cite{rippel2014learning} before the final output layer of our mapper; and (3) our textual bypass technique. We train all models using the same number of optimization steps with the same set of hyperparameters, when applicable.

In~\Cref{fig:ablation_network_architecture}, we provide a qualitative comparison of the results obtained by each variant. First, observe that when no positional encoding is applied on our scalar inputs, the resulting model is unable to adequately capture the concept's visual details. This is most notable in the second and third rows where the model fails to capture the mug's shape or the cat statue's unique colorful stripes. When we omit the use of Nested Dropout during training, we get reconstructions comparable to those of our NeTI models, as seen with the mug example in the second row. However, we find that models trained without Nested Dropout are less editable than those trained with the dropout. For example, in the third row, NeTI without Nested Dropout is unable to achieve a plausible edit of the cat statue. We attribute this to the fact that Nested Dropout can be viewed as a form of regularization during training. This allows the model to focus on capturing the concept-specific features while ignoring spurious details in the training images. In doing so, the resulting models tend to be more amenable to edits at inference time, allowing us to create more accurate, novel compositions of the concept. 

\vspace{0.3cm}

Finally, in the third and fourth columns, we compare NeTI with and without our textual bypass technique. Observe the improved reconstructions that arise when we are able to leverage the additional bypass vector passed directly to the output space of the text encoder. For example, notice the skulls on the bottom of the mug in the second example, the tail of the cat in the third row, or the accurate reconstructions of the child's string-like legs in the final row. In a sense, the bypass vector is able to ``fill in'' the missing details that were not captured by the text encoder. Importantly, we find that using the textual bypass does not harm editability, especially with more complex concepts such as those shown here.  

\clearpage
\newpage

\begin{figure}
    \centering
    \renewcommand{\arraystretch}{0.3}
    \setlength{\tabcolsep}{0.5pt}

    {\small

    \begin{tabular}{c@{\hspace{0.15cm}} c c @{\hspace{0.1cm}} c c}

        \begin{tabular}{c} Real Sample \end{tabular} &
        \multicolumn{2}{c}{Results for $v_{base}$} &
        \multicolumn{2}{c}{Results with $v_{pass}$} \\ \\

        \includegraphics[width=0.0875\textwidth]{images/original/metal_bird.jpg} &
        \includegraphics[width=0.0875\textwidth]{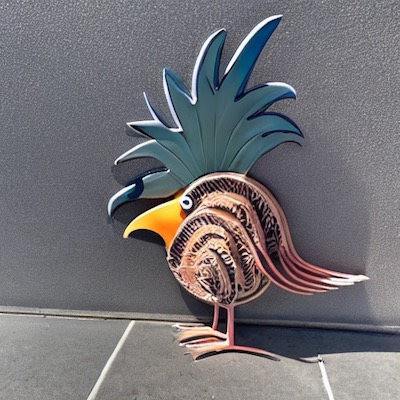} &
        \includegraphics[width=0.0875\textwidth]{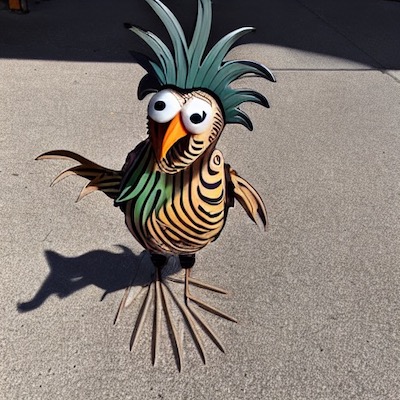} &
        \hspace{0.015cm}
        \includegraphics[width=0.0875\textwidth]{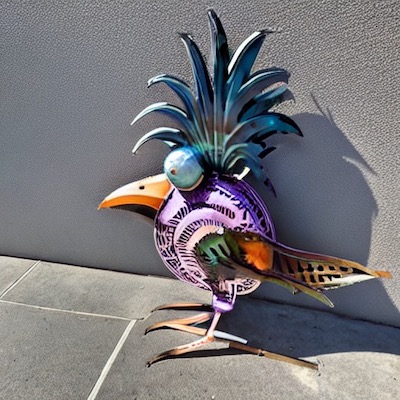} &
        \includegraphics[width=0.0875\textwidth]{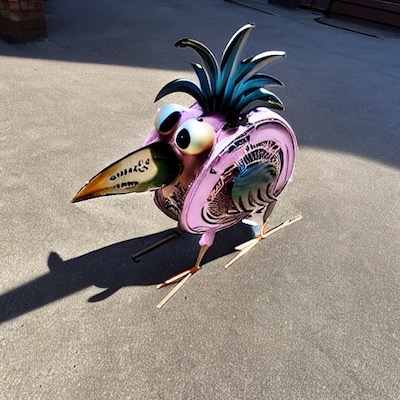} \\ \\

        \includegraphics[width=0.0875\textwidth]{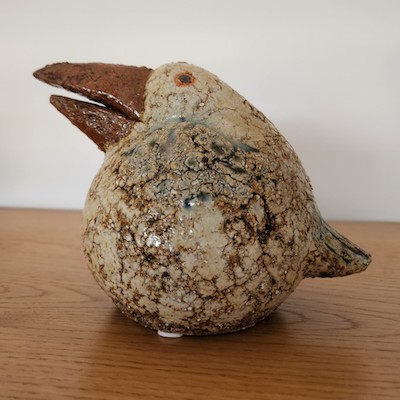} &
        \includegraphics[width=0.0875\textwidth]{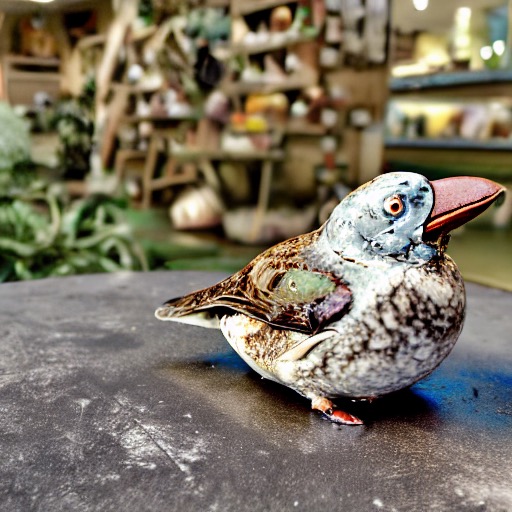} &
        \includegraphics[width=0.0875\textwidth]{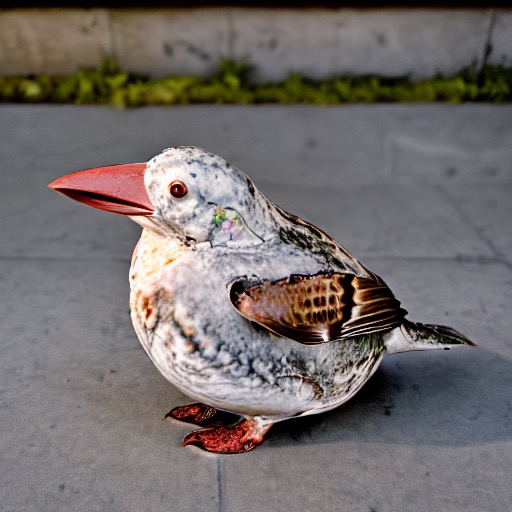} &
        \hspace{0.015cm}
        \includegraphics[width=0.0875\textwidth]{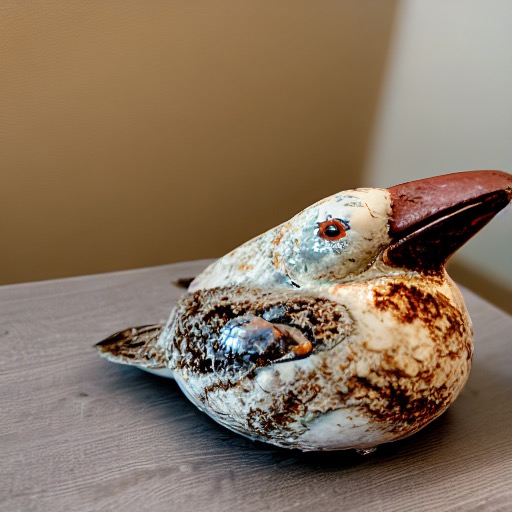} &
        \includegraphics[width=0.0875\textwidth]{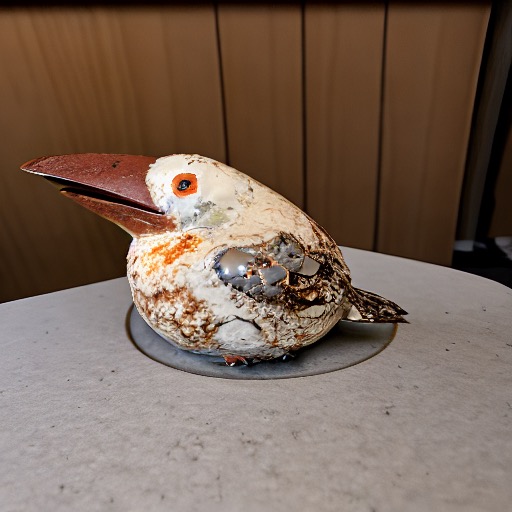} \\ \\
        
        \includegraphics[width=0.0875\textwidth]{images/original/dangling_child.jpg} &
        \includegraphics[width=0.0875\textwidth]{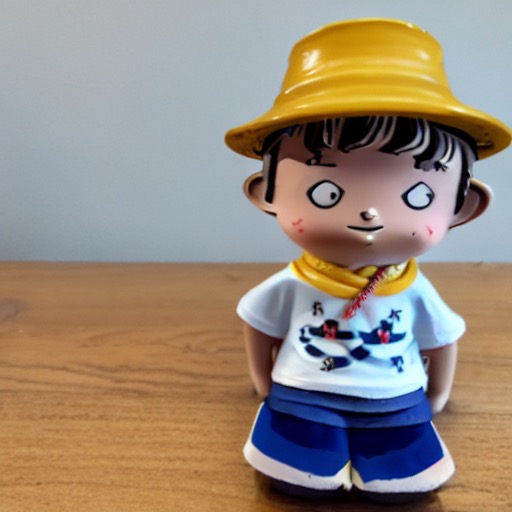} &
        \includegraphics[width=0.0875\textwidth]{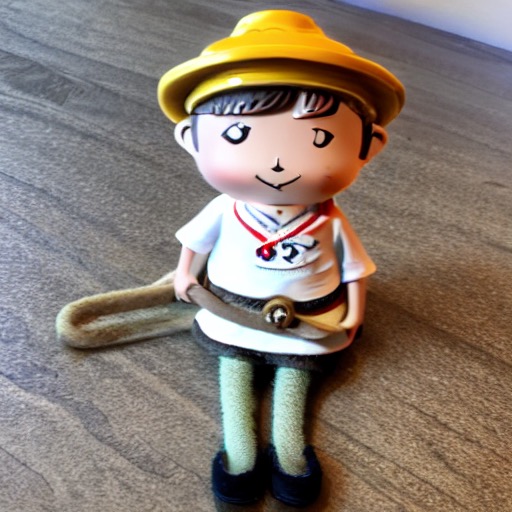} &
        \hspace{0.015cm}
        \includegraphics[width=0.0875\textwidth]{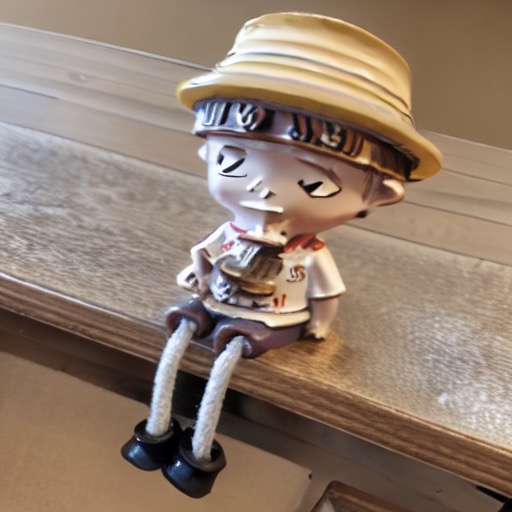} &
        \includegraphics[width=0.0875\textwidth]{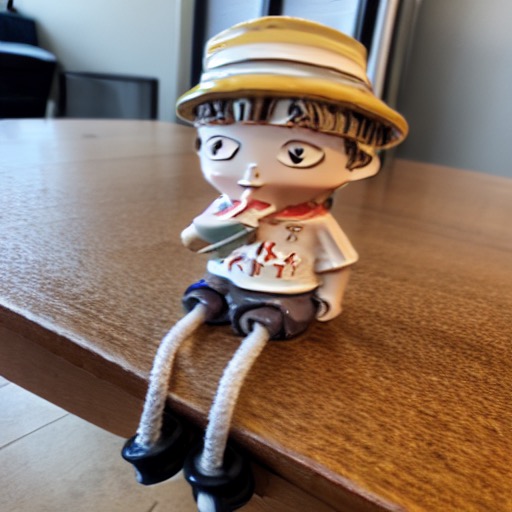} \\ \\

        \includegraphics[width=0.0875\textwidth]{images/original/rainbow_cat.jpeg} &
        \includegraphics[width=0.0875\textwidth]{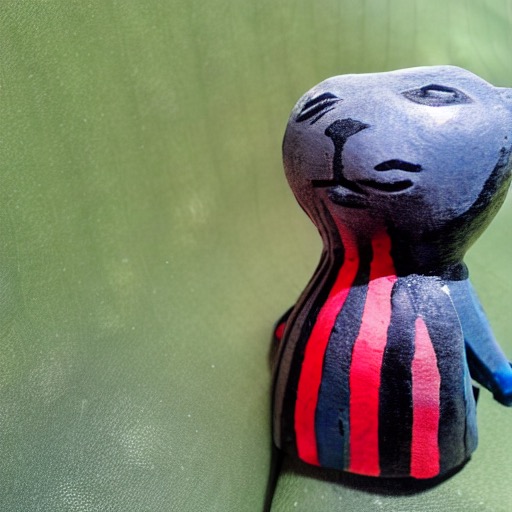} &
        \includegraphics[width=0.0875\textwidth]{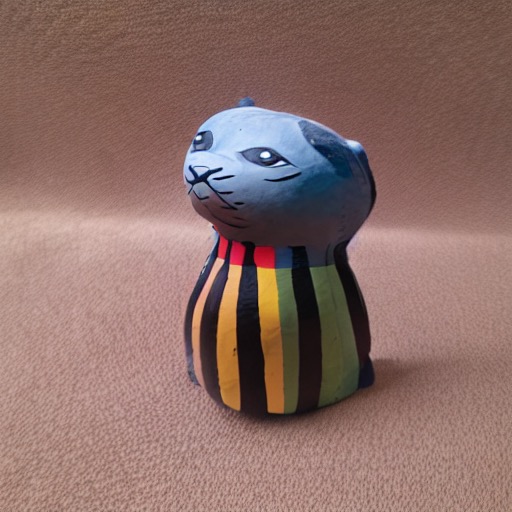} &
        \hspace{0.015cm}
        \includegraphics[width=0.0875\textwidth]{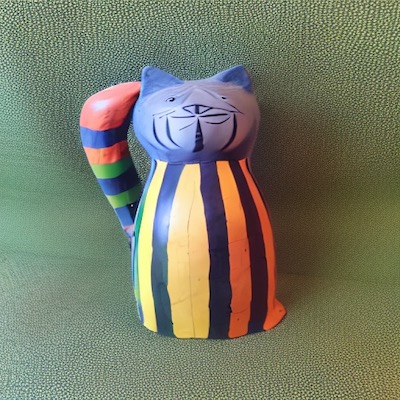} &
        \includegraphics[width=0.0875\textwidth]{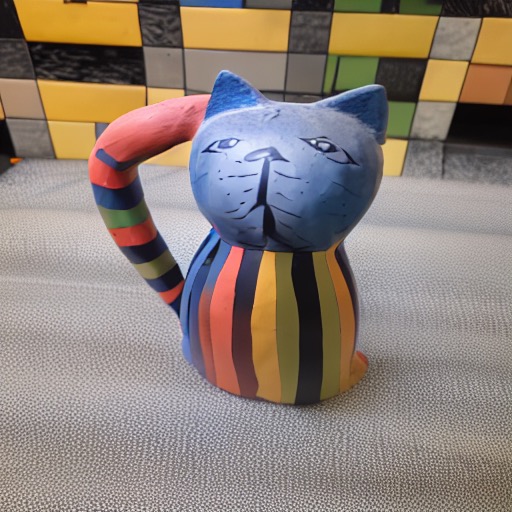} \\ \\

        \includegraphics[width=0.0875\textwidth]{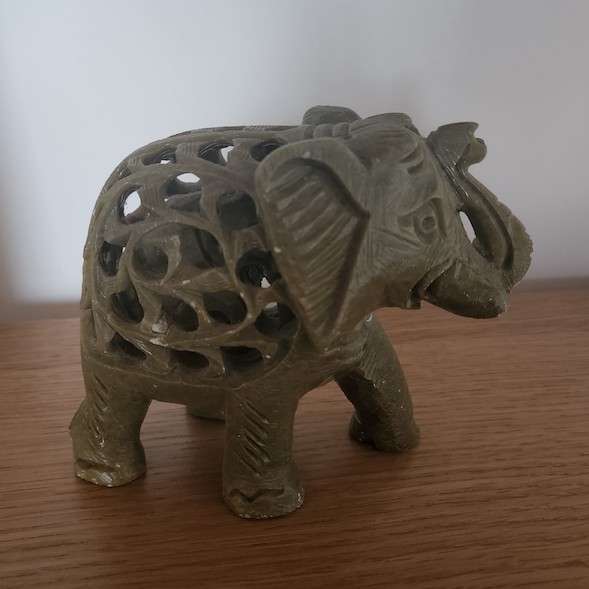} &
        \includegraphics[width=0.0875\textwidth]{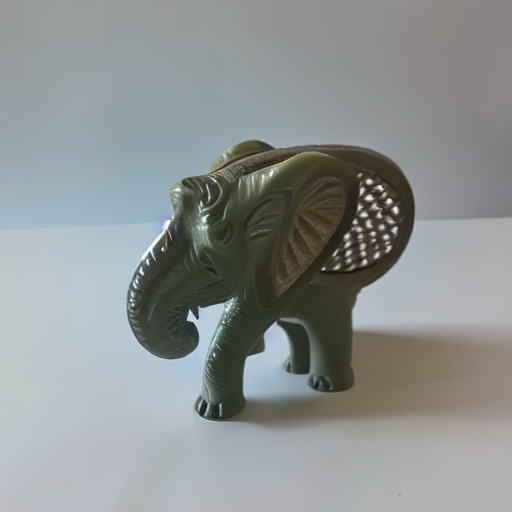} &
        \includegraphics[width=0.0875\textwidth]{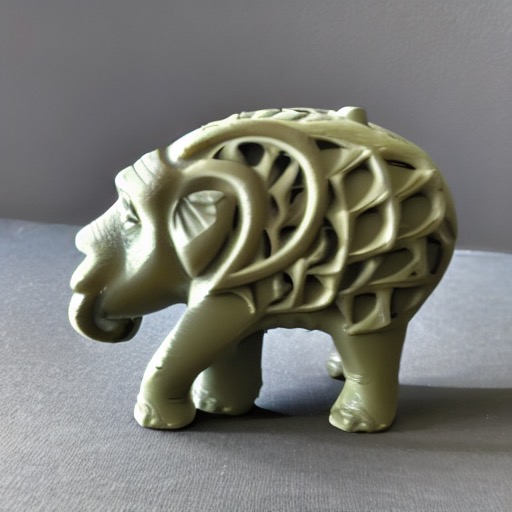} &
        \hspace{0.015cm}
        \includegraphics[width=0.0875\textwidth]{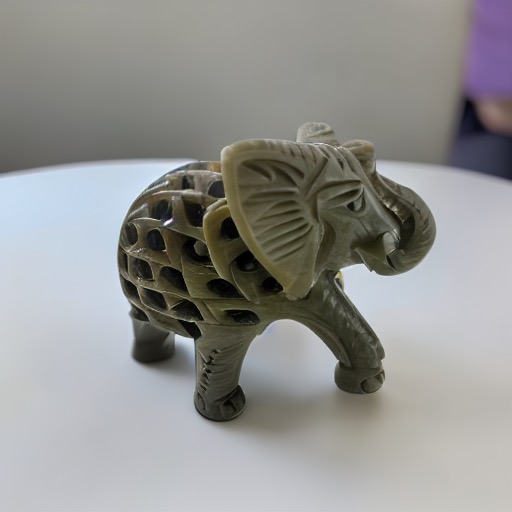} &
        \includegraphics[width=0.0875\textwidth]{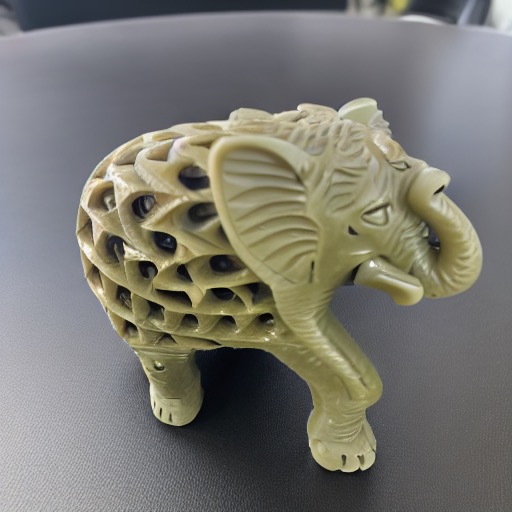} \\ \\

    \end{tabular}
    \\[-0.4cm]
    }
    \caption{Additional results of our learned textual bypass representations. When trained with \textit{textual bypass}, $v_{base}$ learns to reconstruct the coarse concept with the bypass $v_{pass}$ further refining the results.}
    \vspace{-0.1cm}
    \label{fig:textual_bypass_ablation}
\end{figure}

\paragraph{\textbf{The Role of the Textual Bypass.}}
In the main paper, we demonstrated that applying our textual bypass technique results in improved visual fidelity. We also demonstrate that the base vector $v_{base}$ learns to capture the coarse-level details of the concept while the additional bypass vector $v_{pass}$ complements this by adding finer-level details such as texture and further refining the concept's structure. In~\Cref{fig:textual_bypass_ablation} we provide additional examples to complement those provided in the main paper (\Cref{fig:textual_bypass_ablation_main_paper}). As can be seen, by adding $v_{pass}$ we are able to better capture the shape of the metal bird in the first row and the elephant in the final row. Interestingly, in the second row, $v_{base}$ generates images resembling that of a real bird, and when we add $v_{pass}$ to the denoising process, this bird is transformed to more accurately depict the stone-like structure of the real concept. 

Please note that the results provided here for $v_{base}$ do not represent the optimal results that can be achieved without our textual bypass technique. This visualization serves to provide insights as to what aspects the network chose to encode into each one of its output vectors. For an ablation study of our textual bypass technique, we refer the reader to~\Cref{sec:ablation_study}.

\begin{figure}
    \centering
    \renewcommand{\arraystretch}{0.3}
    \setlength{\tabcolsep}{0.5pt}

    {\small

    \begin{tabular}{c@{\hspace{0.15cm}} c c c c}

        \begin{tabular}{c} Real Sample \end{tabular} & & & & \\

        \includegraphics[width=0.0875\textwidth]{images/original/rainbow_cat.jpeg} &
        \hspace{0.05cm}
        \includegraphics[width=0.0875\textwidth]{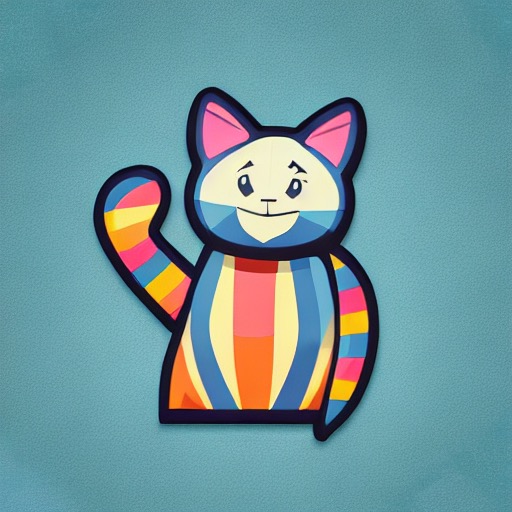} &
        \includegraphics[width=0.0875\textwidth]{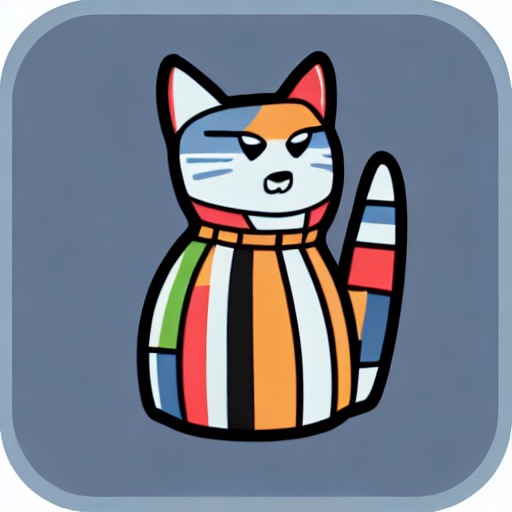} &
        \includegraphics[width=0.0875\textwidth]{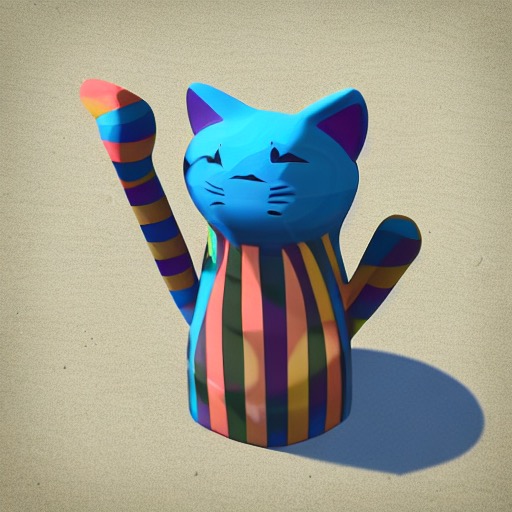} &
        \includegraphics[width=0.0875\textwidth]{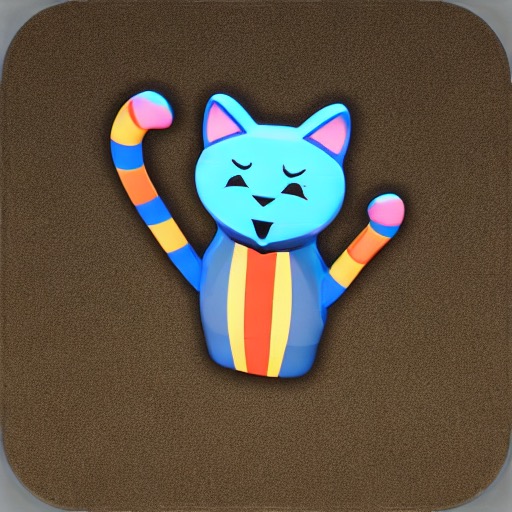} \\ \\
        & \multicolumn{4}{c}{\begin{tabular}{c} ``An app icon of $S_*$''\end{tabular}} \\ \\

        \includegraphics[width=0.0875\textwidth]{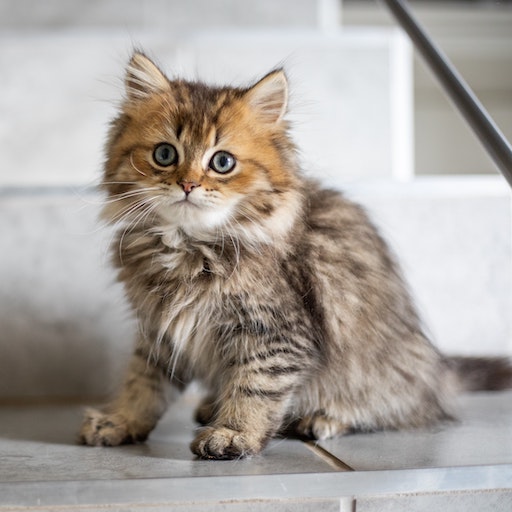} &
        \hspace{0.05cm}
        \includegraphics[width=0.0875\textwidth]{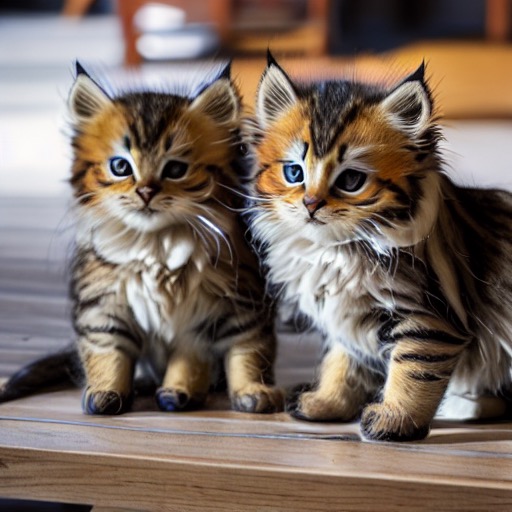} &
        \includegraphics[width=0.0875\textwidth]{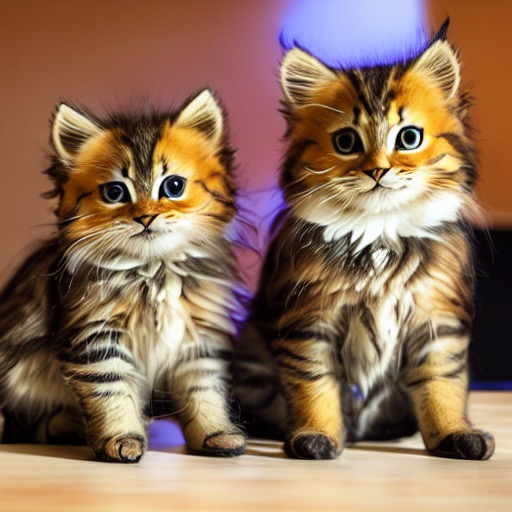} &
        \includegraphics[width=0.0875\textwidth]{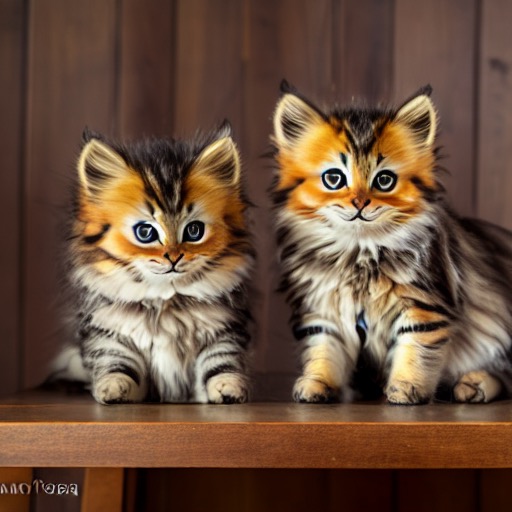} &
        \includegraphics[width=0.0875\textwidth]{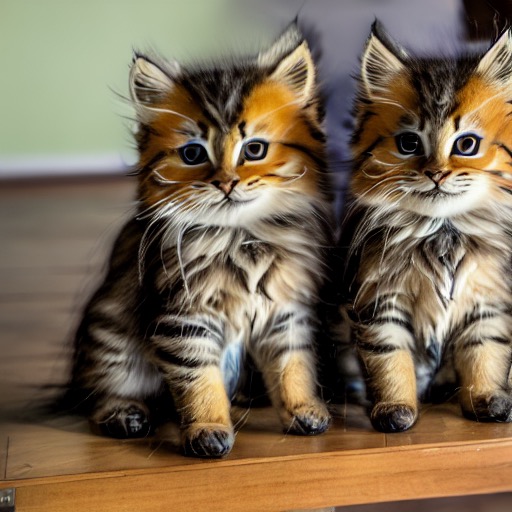} \\ \\
        & \multicolumn{4}{c}{\begin{tabular}{c} ``A photograph of two $S_*$ on a table''\end{tabular}} \\ \\

        \includegraphics[width=0.0875\textwidth]{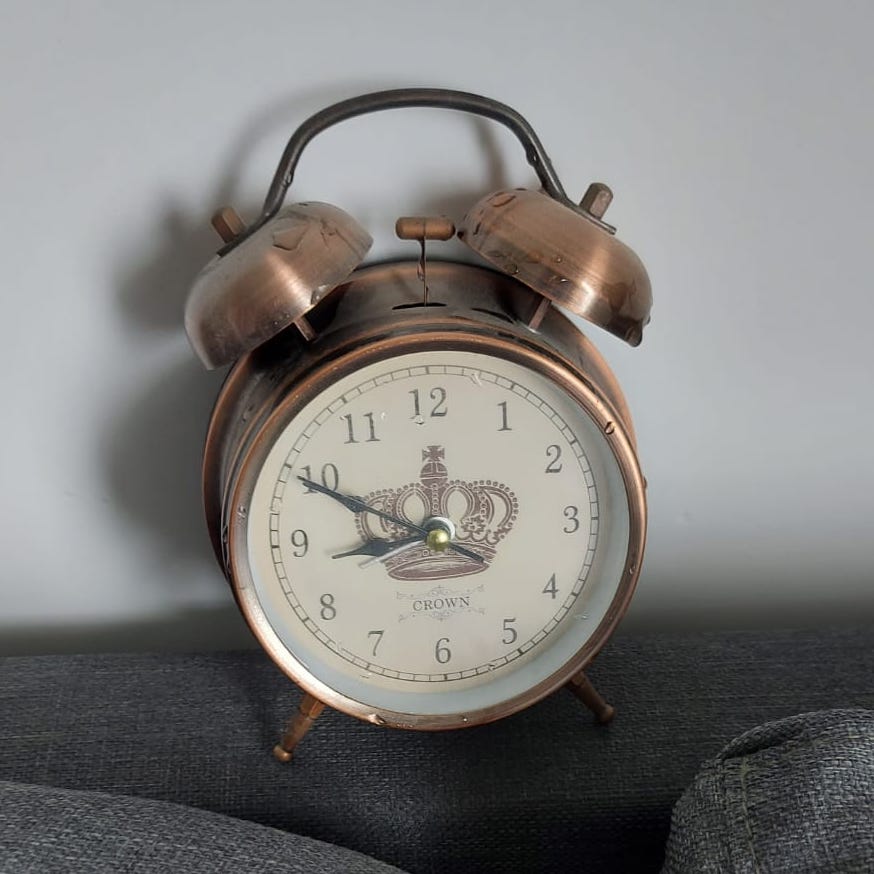} &
        \hspace{0.05cm}
        \includegraphics[width=0.0875\textwidth]{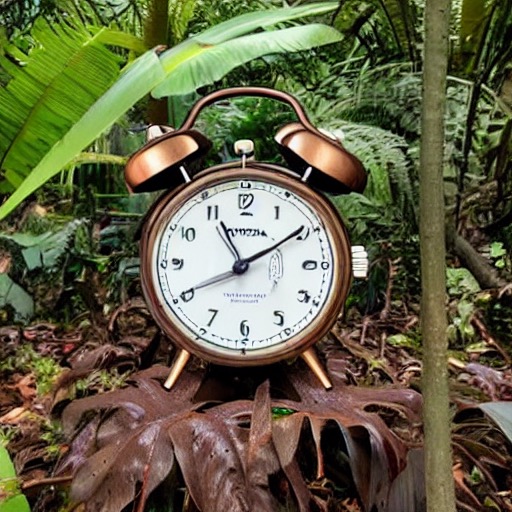} &
        \includegraphics[width=0.0875\textwidth]{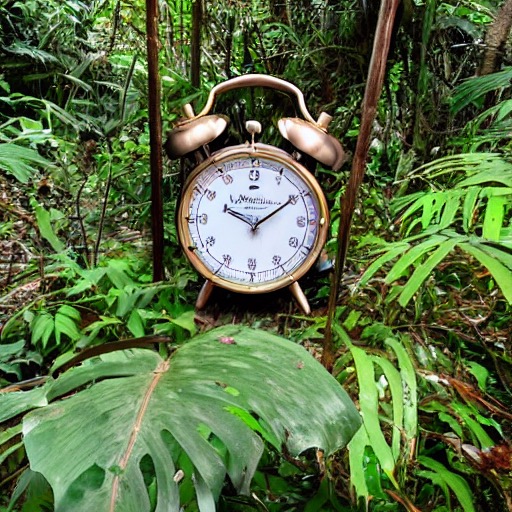} &
        \includegraphics[width=0.0875\textwidth]{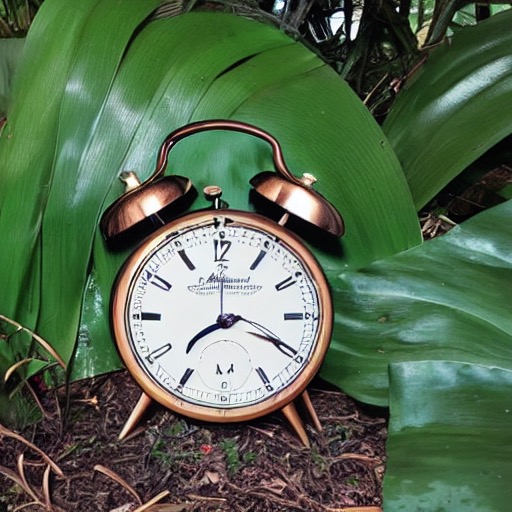} &
        \includegraphics[width=0.0875\textwidth]{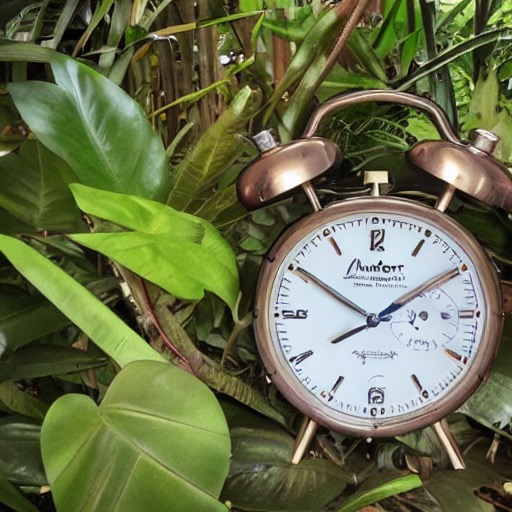} \\ \\
        & \multicolumn{4}{c}{\begin{tabular}{c} ``A photo of $S_*$ in the jungle''\end{tabular}} \\ \\

        \includegraphics[width=0.0875\textwidth]{images/original/colorful_teapot.jpg} &
        \hspace{0.05cm}
        \includegraphics[width=0.0875\textwidth]{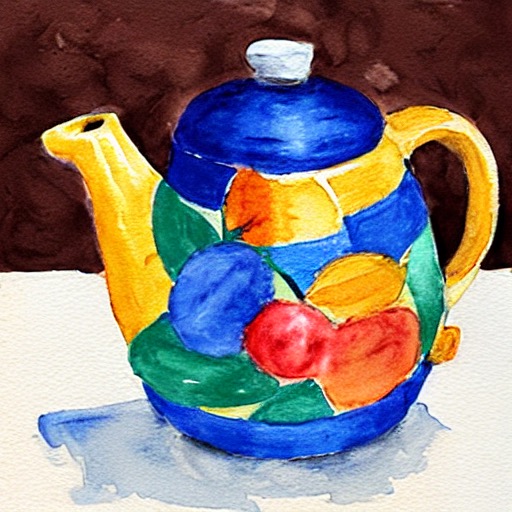} &
        \includegraphics[width=0.0875\textwidth]{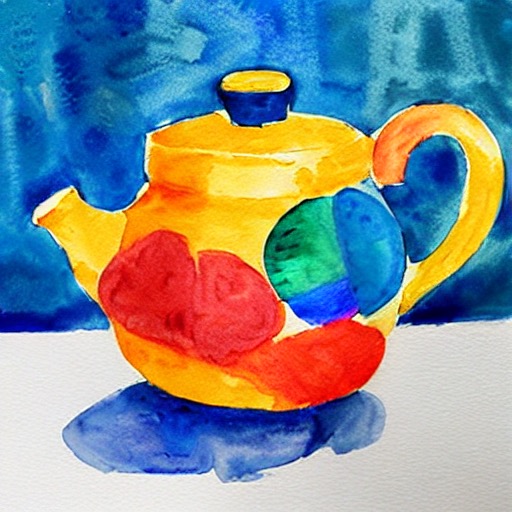} &
        \includegraphics[width=0.0875\textwidth]{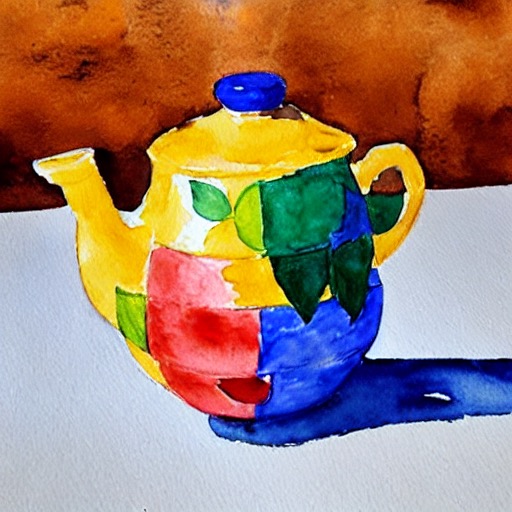} &
        \includegraphics[width=0.0875\textwidth]{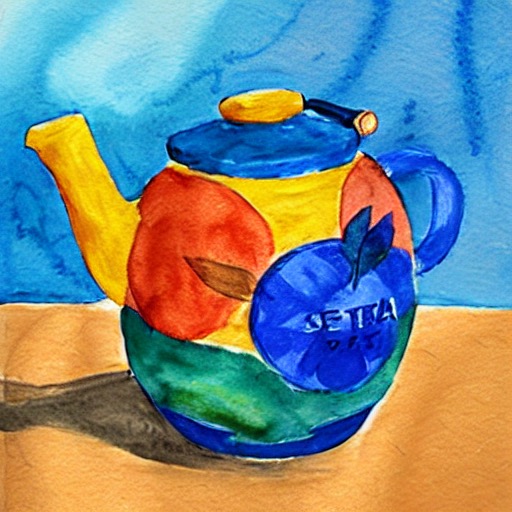} \\ \\
        & \multicolumn{4}{c}{\begin{tabular}{c} ``A watercolor painting of $S_*$''\end{tabular}} \\ \\

        \includegraphics[width=0.0875\textwidth]{images/original/fat_stone_bird.jpg} &
        \hspace{0.05cm}
        \includegraphics[width=0.0875\textwidth]{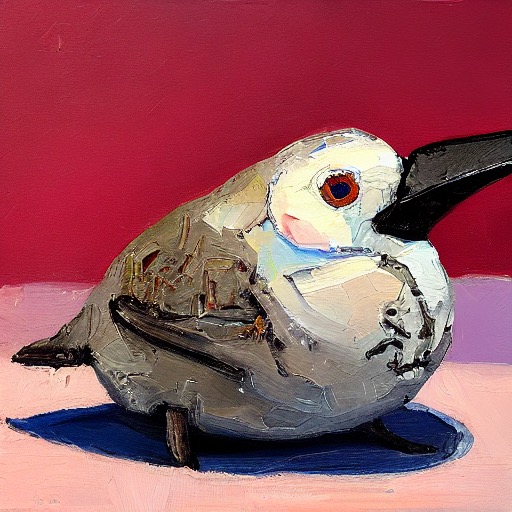} &
        \includegraphics[width=0.0875\textwidth]{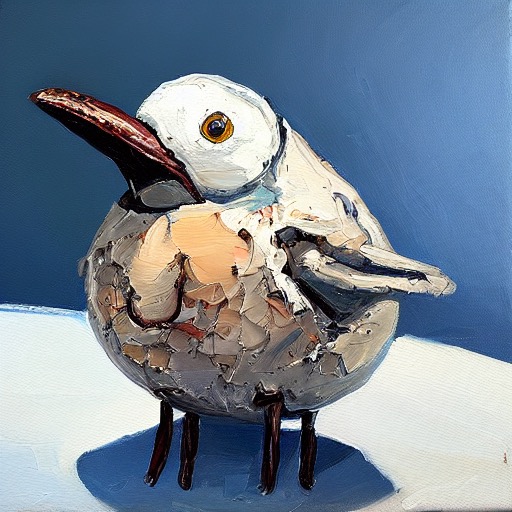} &
        \includegraphics[width=0.0875\textwidth]{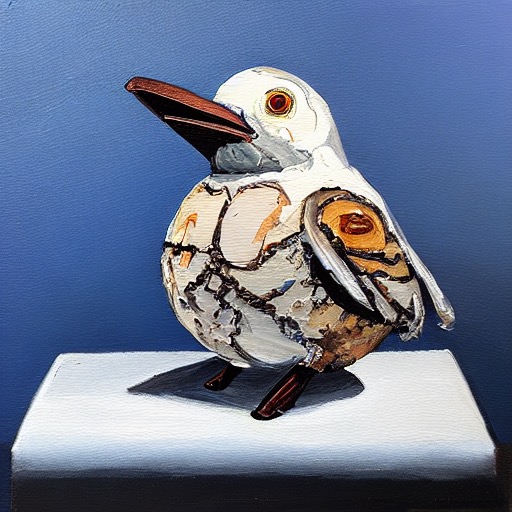} &
        \includegraphics[width=0.0875\textwidth]{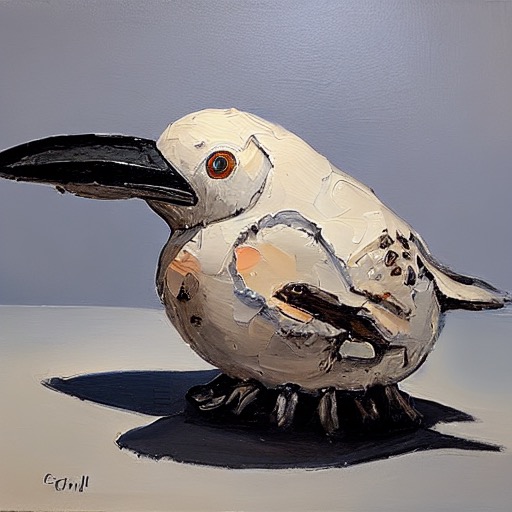} \\ \\
        & \multicolumn{4}{c}{\begin{tabular}{c} ``Oil painting of $S_*$''\end{tabular}} \\ \\

        \includegraphics[width=0.0875\textwidth]{images/original/elephant.jpg} &
        \hspace{0.05cm}
        \includegraphics[width=0.0875\textwidth]{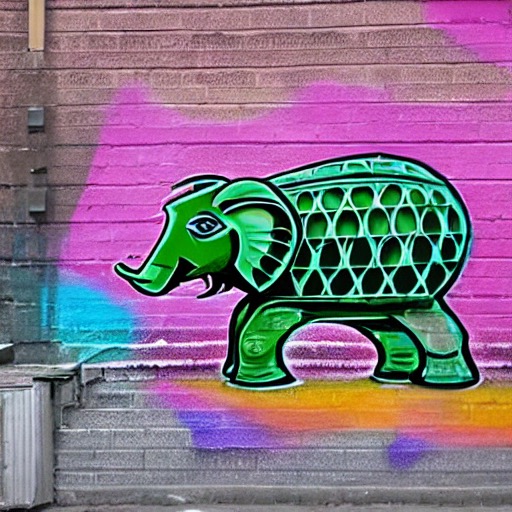} &
        \includegraphics[width=0.0875\textwidth]{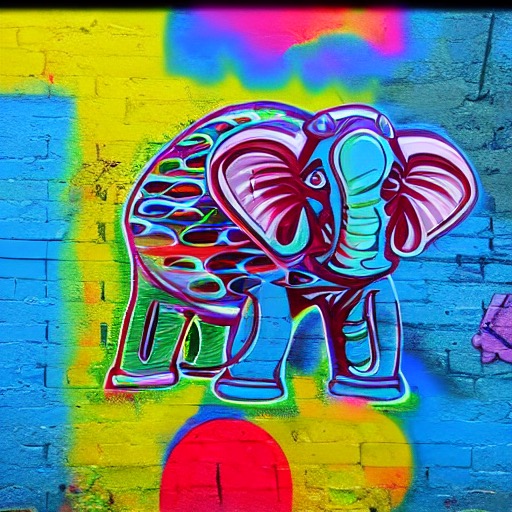} &
        \includegraphics[width=0.0875\textwidth]{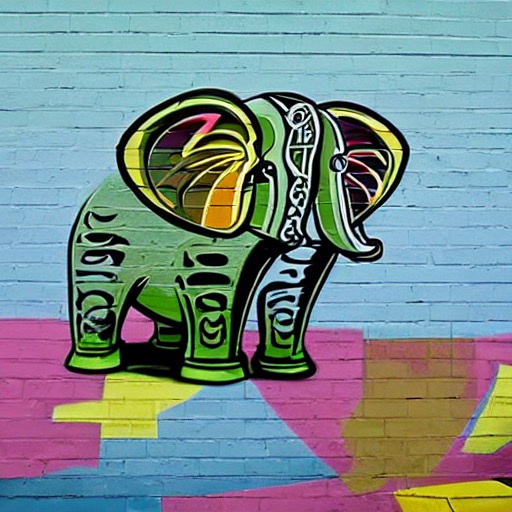} &
        \includegraphics[width=0.0875\textwidth]{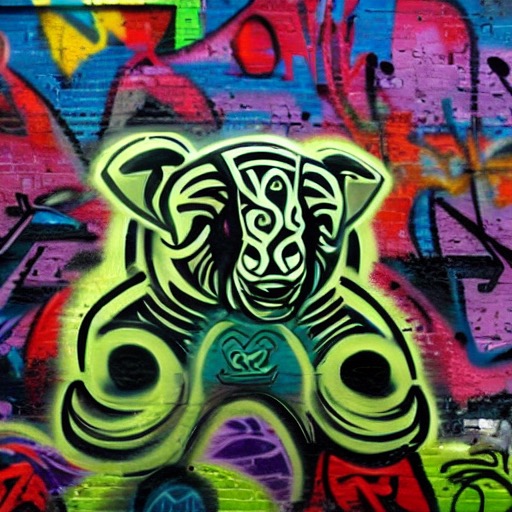} \\ \\
        & \multicolumn{4}{c}{\begin{tabular}{c} ``A colorful graffiti of $S_*$''\end{tabular}} \\ \\

    \end{tabular}
    \\[-0.4cm]
    }
    \caption{Image generations with NeTI under a single-image training setting.}
    \label{fig:single_image}
    \vspace{-0.2cm}
\end{figure}

\section{Additional Comparisons}~\label{sec:custom_comparison}
In this section, we provide additional evaluations and comparisons of our NeTI scheme.

\paragraph{\textbf{Single Image Personalization.}}
Here, we evaluate NeTI when only a single image is used during training. We apply the same training scheme as used in our other evaluations and train our models for $500$ optimization steps without textual bypass. In~\Cref{fig:single_image} we provide visual examples of results obtained by NeTI under this single-image training setting. As can be seen, even under this more challenging setting, NeTI performs well in terms of both reconstructing the target concept and remaining consistent with the provided text prompt. We do note that NeTI may still fail to fully reconstruct the concept in some cases. For example, in the first row, NeTI generates images with two tails and in the third row, we fail to generate fine-level details of the clock. We hope to further explore this challenging setting in the future.

\clearpage
\newpage

\vspace{-0.1cm}
\paragraph{\textbf{Training Convergence.}}
We now turn to compare the convergence speed of NeTI when compared to XTI~\cite{voynov2023p}. In~\Cref{tb:p_plus_convergence_comparison}, we provide quantitative metrics computed over all $16$ concepts following our evaluation protocol described in the main paper. As can be seen, NeTI with our textual bypass attains comparable performance to XTI, even when trained with $4\times$ fewer optimization steps (i.e., $250$ steps for NeTI compared to $1000$ steps for XTI). Moreover, when we continue training NeTI for $1000$ steps, we achieve a significant improvement compared to XTI in our image similarity metrics with only a small decrease in text similarity.

\begin{table}
\small
\setlength{\tabcolsep}{6.5pt}
\centering
\caption{Quantitative comparison with XTI~\cite{voynov2023p} after a varying number of optimization steps. Observe that NeTI with our textual bypass at 250 training steps outperforms XTI when trained for $1000$ steps in both the image and text similarities. Note, the higher the better. \\[-0.65cm]} 
\begin{tabular}{l c c c c c c} 
    \toprule
    Steps & \multicolumn{2}{c}{250} & \multicolumn{2}{c}{500} & \multicolumn{2}{c}{1000} \\
    \cline{2-7}
    & Image & Text & Image & Text & Image & Text \\
    \midrule
    XTI            & - & - & 0.711 & 0.260 & 0.722 & 0.255 \\
    NeTI & 0.729 & 0.262 & 0.751 & 0.253 & 0.767 & 0.247 \\
    \bottomrule
\end{tabular}
\label{tb:p_plus_convergence_comparison}
\vspace{-0.1cm}
\end{table}

\begin{figure}
    \centering
    \renewcommand{\arraystretch}{0.3}
    \setlength{\tabcolsep}{0.5pt}
    {\small
    \begin{tabular}{c@{\hspace{0.1cm}} c c c c}

        \begin{tabular}{c} Real Sample \end{tabular} &
        \hspace{0.015cm}
        25 & 50 & 100 & 150 \\

        \includegraphics[width=0.0875\textwidth]{images/original/cat.jpg} &
        \hspace{0.015cm}
        \includegraphics[width=0.0875\textwidth]{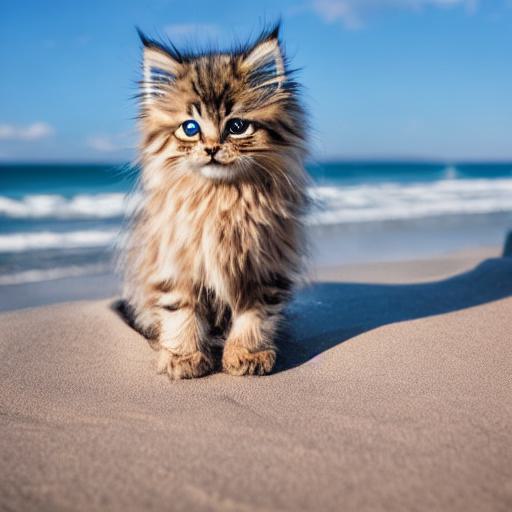} &
        \includegraphics[width=0.0875\textwidth]{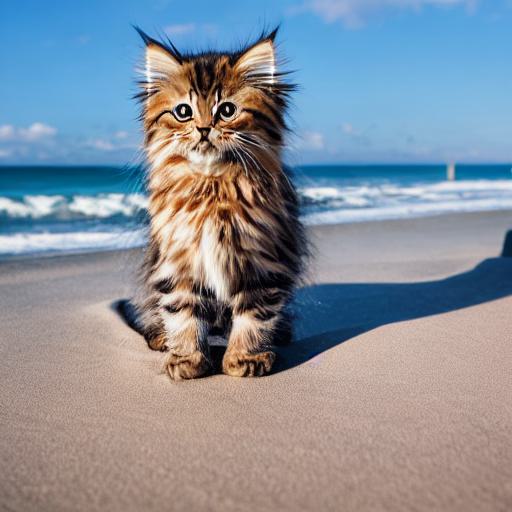} &
        \includegraphics[width=0.0875\textwidth]{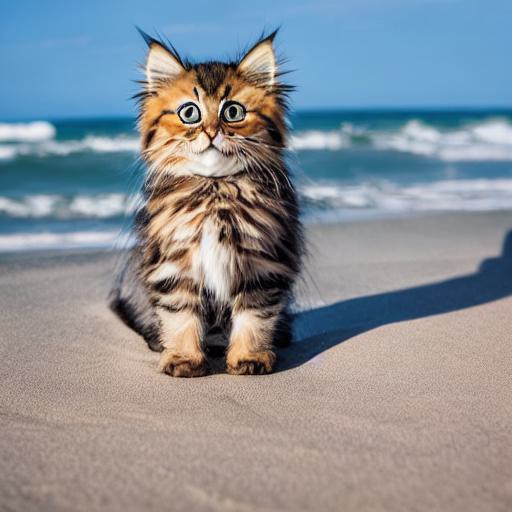} &
        \includegraphics[width=0.0875\textwidth]{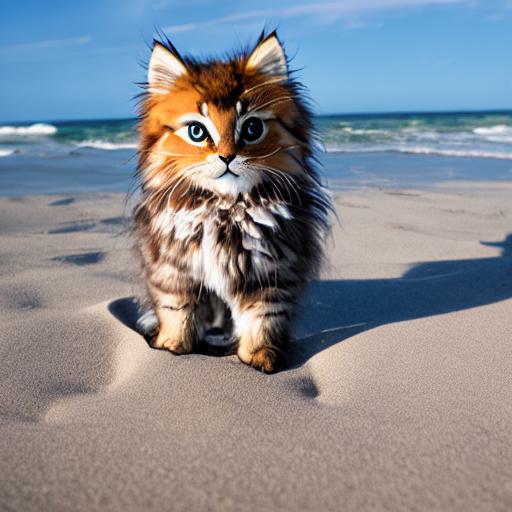} \\
        
        & \multicolumn{4}{c}{\begin{tabular}{c} ``A photo of $S_*$ on the beach'' \end{tabular}} \\

        \includegraphics[width=0.0875\textwidth]{images/original/tortoise_plushy.jpg} &
        \hspace{0.015cm}
        \includegraphics[width=0.0875\textwidth]{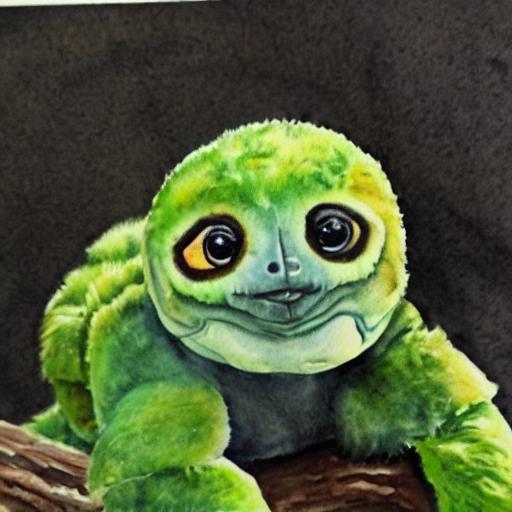} &
        \includegraphics[width=0.0875\textwidth]{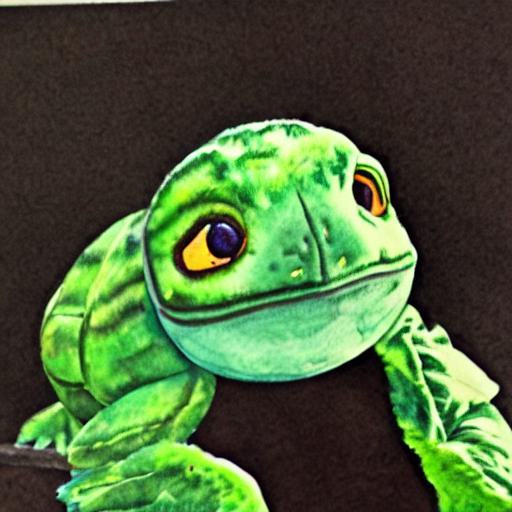} &
        \includegraphics[width=0.0875\textwidth]{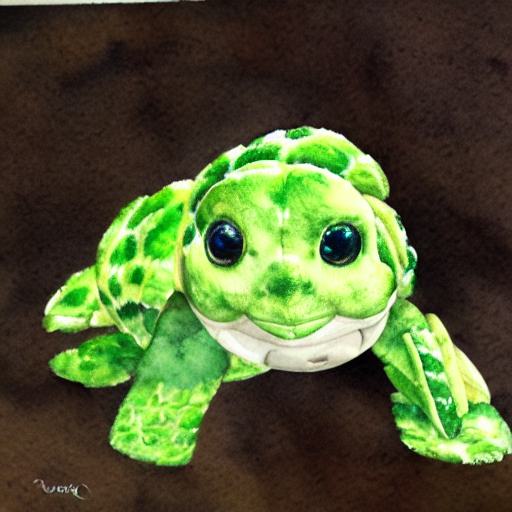} &
        \includegraphics[width=0.0875\textwidth]{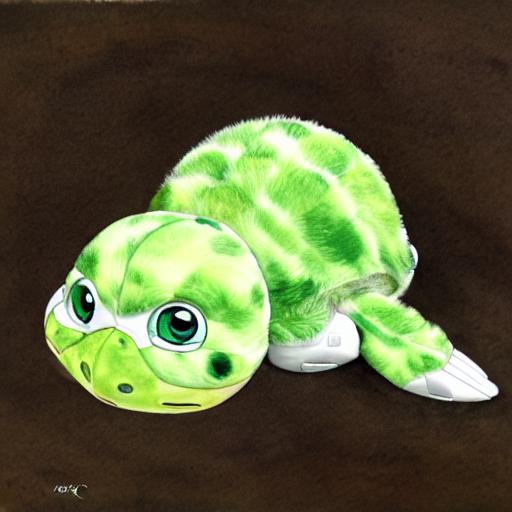} \\
        & \multicolumn{4}{c}{\begin{tabular}{c} ``A watercolor painting of $S_*$'' \end{tabular}} \\

        \includegraphics[width=0.0875\textwidth]{images/original/teddybear.jpg} &
        \hspace{0.015cm}
        \includegraphics[width=0.0875\textwidth]{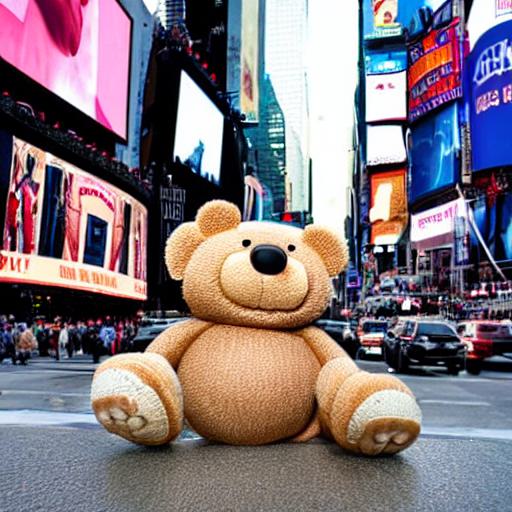} &
        \includegraphics[width=0.0875\textwidth]{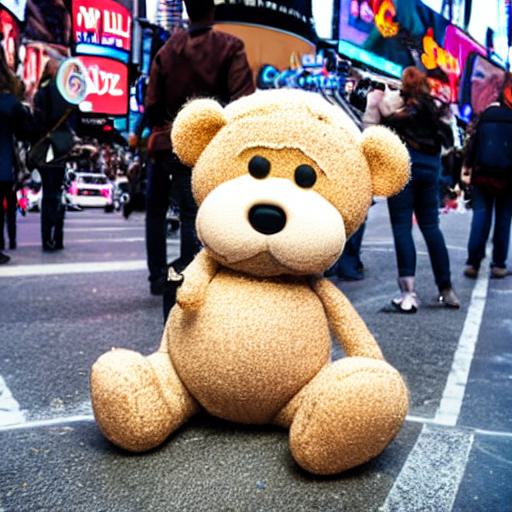} &
        \includegraphics[width=0.0875\textwidth]{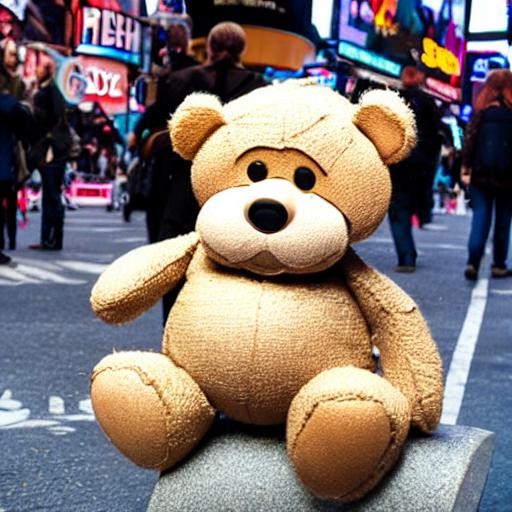} &
        \includegraphics[width=0.0875\textwidth]{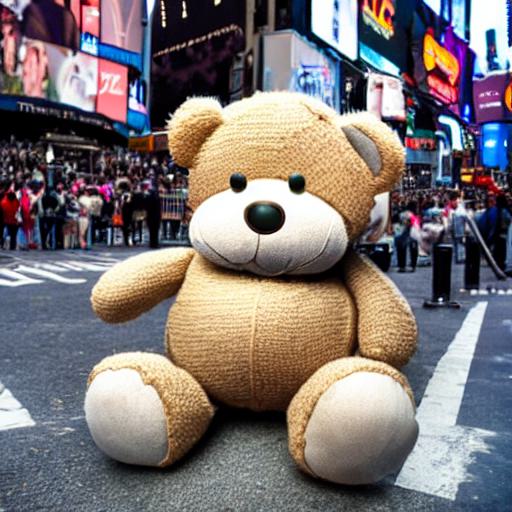} \\
        & \multicolumn{4}{c}{\begin{tabular}{c} ``A photo of $S_*$ in Times Square'' \end{tabular}}

    \end{tabular}
    \\[-0.2cm]
    }
    \caption{Generation results obtained with NeTI after a varying number of steps. Observe that even after a small number of training steps, concept-specific details are already captured by NeTI.}
    \label{fig:neti_convergence}
    \vspace{-0.25cm}
\end{figure}

Next, in~\Cref{fig:neti_convergence} we show results obtained by NeTI when trained for a small number of optimization steps. As can be seen, even after training for a very small number of steps, e.g., as few as 25 steps, NeTI is able to capture the core concept-specific details such as the color of the fur of the cat in the first row or the shape of the tortoise in the second row. Although most results presented in the paper are obtained after $500$ training steps, this further highlights that some concepts, depending on their complexity, may converge much faster using our neural mapper and \pstar space.

\vspace{-0.1cm}
\paragraph{\textbf{Comparison to CustomDiffusion.}}
Finally, we provide a comparison to CustomDiffusion~\cite{kumari2022customdiffusion}. We choose to compare our approach with their six official publicly-released datasets (see~\Cref{sec:additional_details}). We evaluated their models with our $15$ text prompts, generating $32$ images for each prompt using the same $32$ random seeds as used to evaluate all methods in the main paper. 

\begin{figure}
    \centering
    \renewcommand{\arraystretch}{0.3}
    \setlength{\tabcolsep}{0.5pt}
    {\small
    \begin{tabular}{c@{\hspace{0.1cm}} c c @{\hspace{0.1cm}} c c}

        \begin{tabular}{c} Real Sample \\ \& Prompt \end{tabular} &
        \multicolumn{2}{c}{CustomDiffusion} &
        \multicolumn{2}{c}{NeTI w/ Textual} \\

        \includegraphics[width=0.085\textwidth]{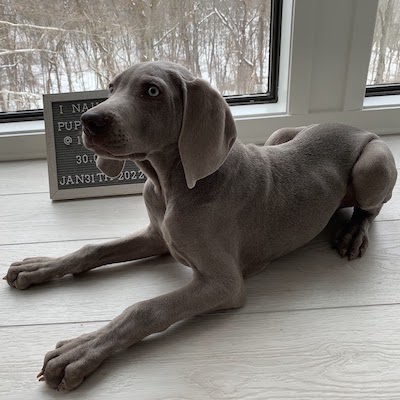} &
        \includegraphics[width=0.085\textwidth]{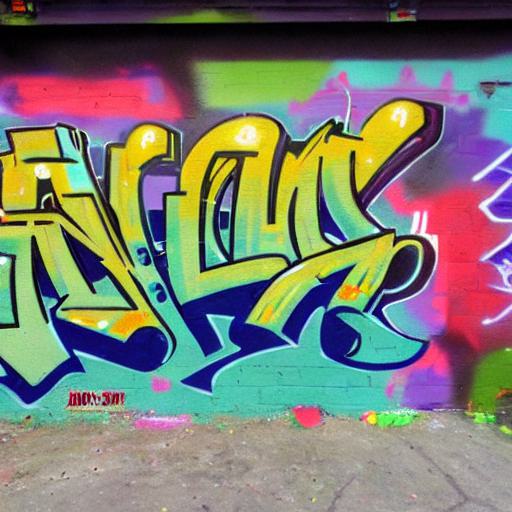} &
        \includegraphics[width=0.085\textwidth]{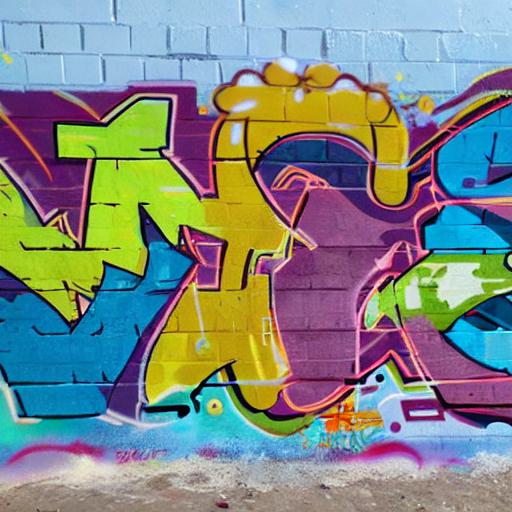} &
        \hspace{0.015cm}
        \includegraphics[width=0.085\textwidth]{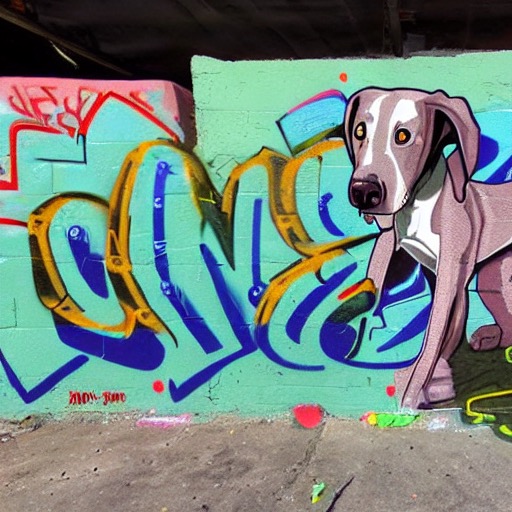} &
        \includegraphics[width=0.085\textwidth]{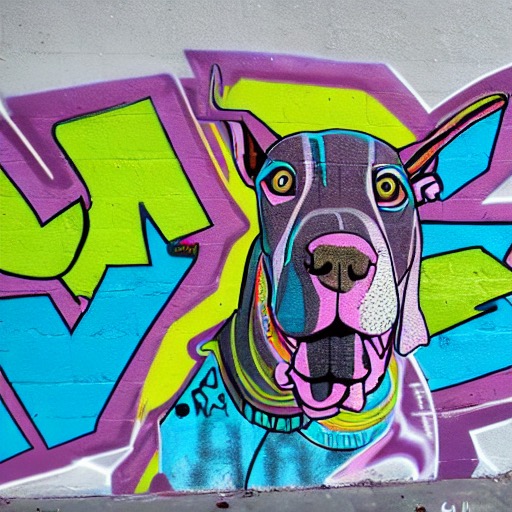} \\
        
        \raisebox{0.25in}{\begin{tabular}{c} ``A colorful \\ \\[-0.05cm] graffiti of \\ \\[-0.05cm] $S_*$''\end{tabular}} &
        \includegraphics[width=0.085\textwidth]{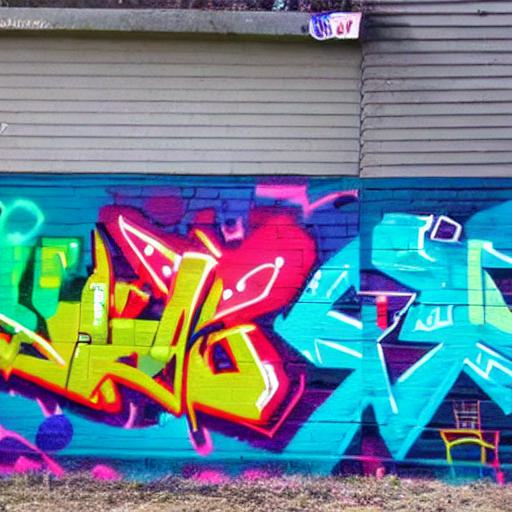} &
        \includegraphics[width=0.085\textwidth]{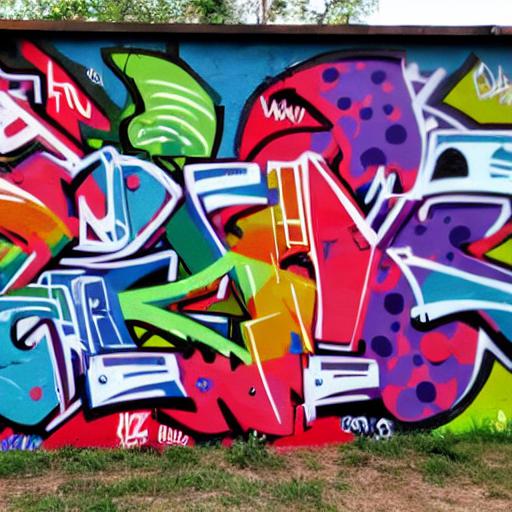} &
        \hspace{0.015cm}
        \includegraphics[width=0.085\textwidth]{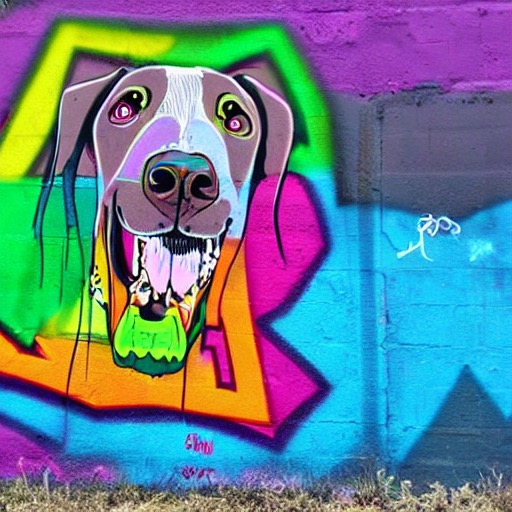} &
        \includegraphics[width=0.085\textwidth]{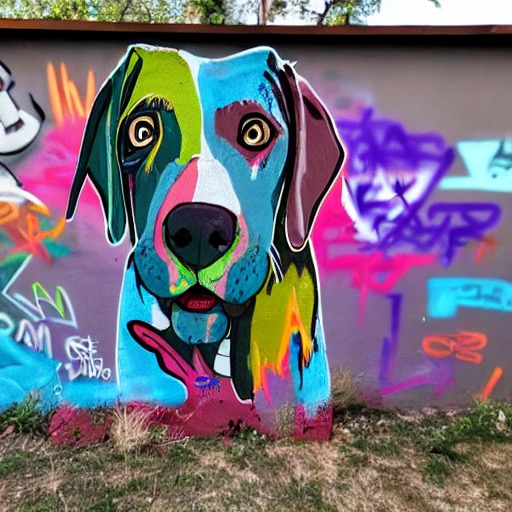} \\ \\

        \includegraphics[width=0.085\textwidth]{images/original/tortoise_plushy.jpg} &
        \includegraphics[width=0.085\textwidth]{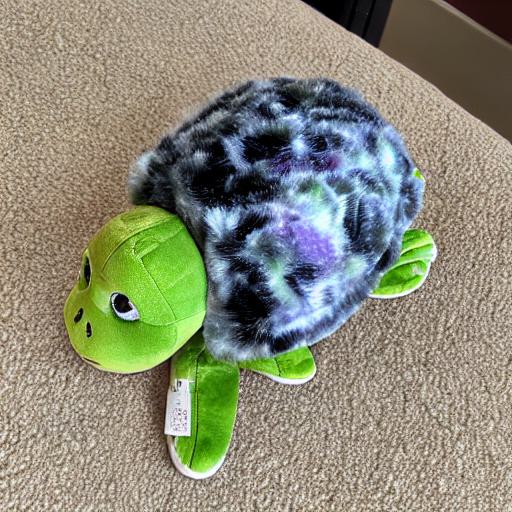} &
        \includegraphics[width=0.085\textwidth]{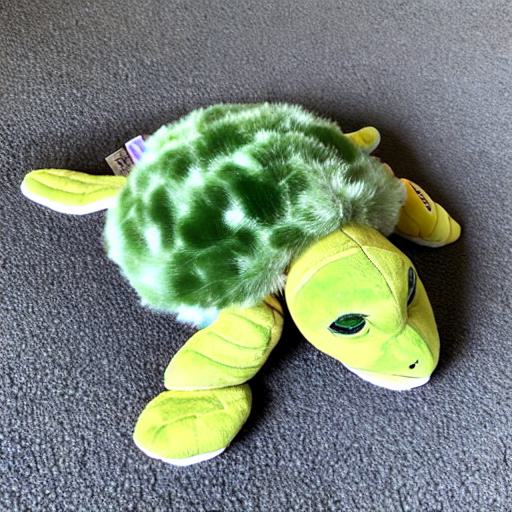} &
        \hspace{0.015cm}
        \includegraphics[width=0.085\textwidth]{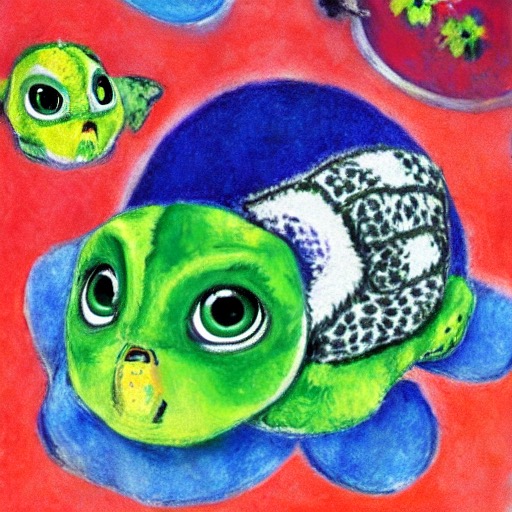} &
        \includegraphics[width=0.085\textwidth]{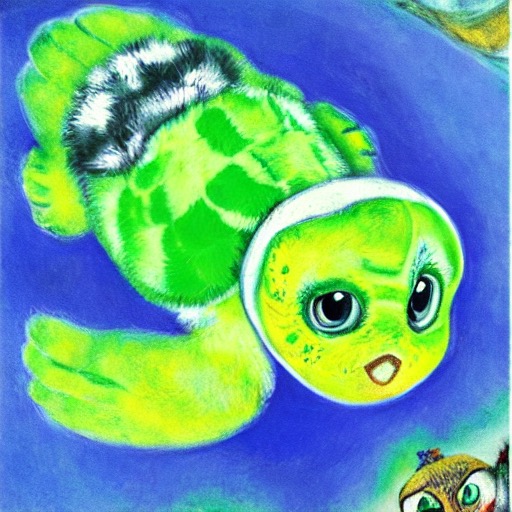} \\

        \raisebox{0.25in}{\begin{tabular}{c} ``Marc \\ \\[-0.05cm] Chagall \\ \\[-0.05cm] painting \\ \\[-0.05cm] of $S_*$''\end{tabular}} &
        \includegraphics[width=0.085\textwidth]{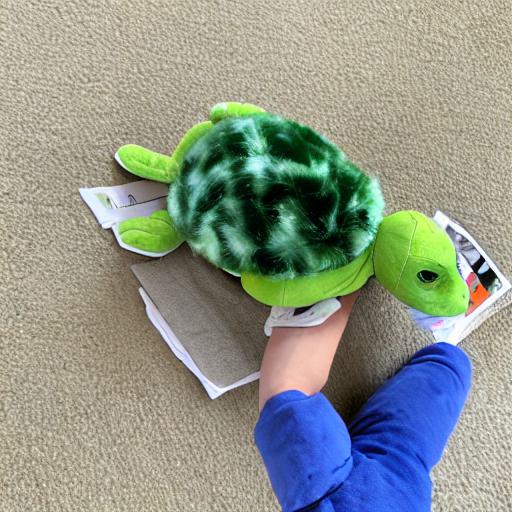} &
        \includegraphics[width=0.085\textwidth]{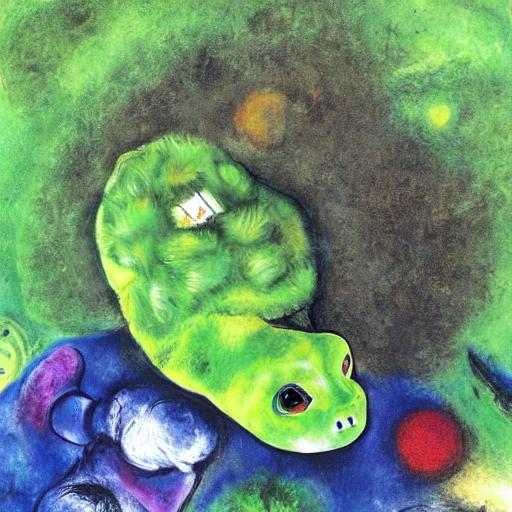} &
        \hspace{0.015cm}
        \includegraphics[width=0.085\textwidth]{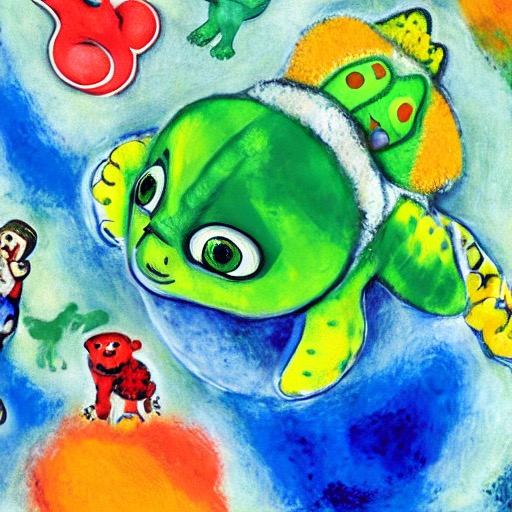} &
        \includegraphics[width=0.085\textwidth]{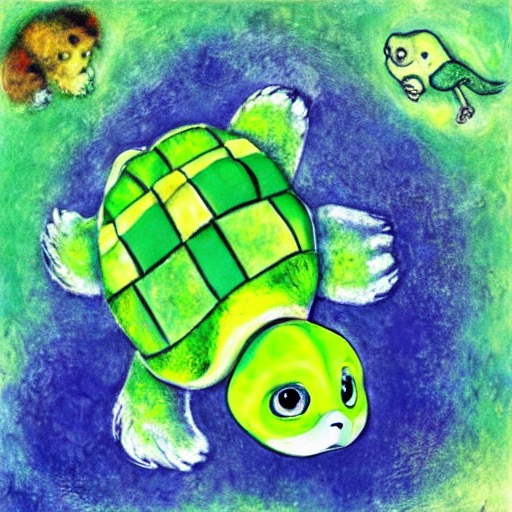} \\ \\

        \includegraphics[width=0.085\textwidth]{images/original/teddybear.jpg} &
        \includegraphics[width=0.085\textwidth]{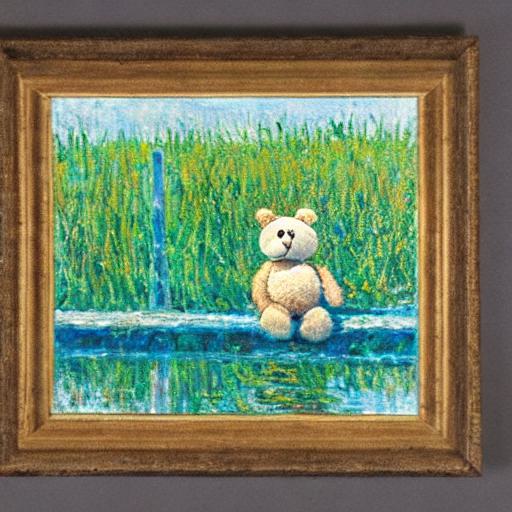} &
        \includegraphics[width=0.085\textwidth]{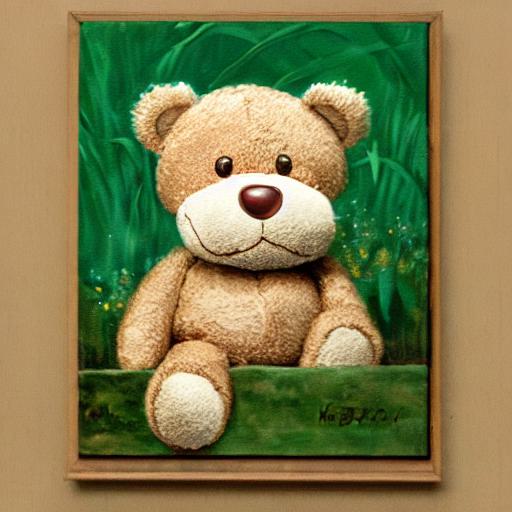} &
        \hspace{0.015cm}
        \includegraphics[width=0.085\textwidth]{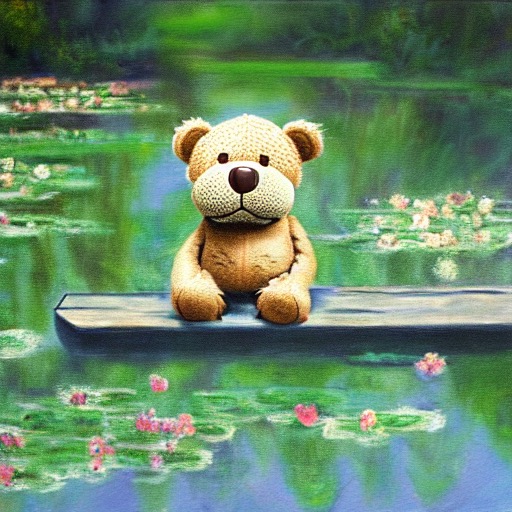} &
        \includegraphics[width=0.085\textwidth]{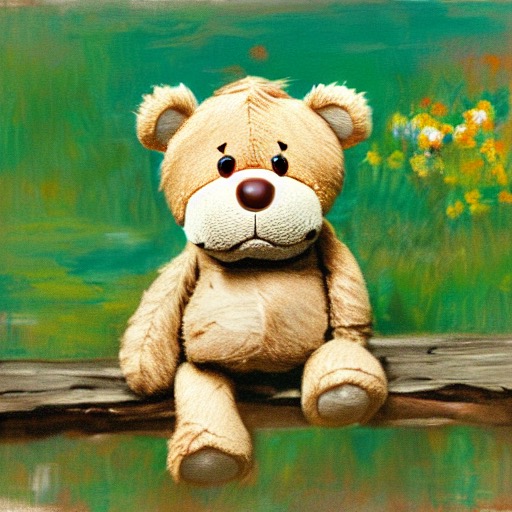} \\

        \raisebox{0.25in}{\begin{tabular}{c} ``A painting \\ \\[-0.05cm]  of $S_*$ in the \\ \\[-0.05cm] style of \\ \\[-0.05cm] Monet''\end{tabular}} &
        \includegraphics[width=0.085\textwidth]{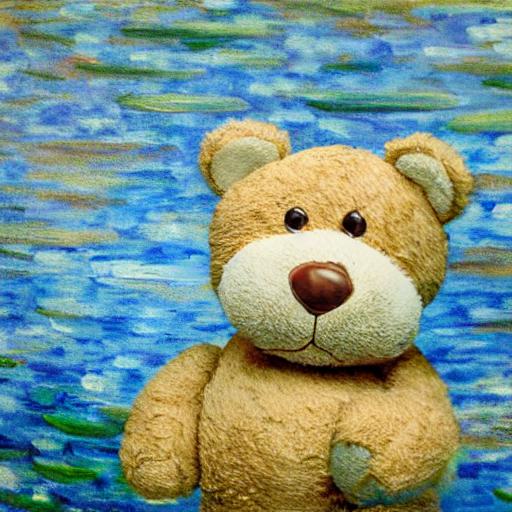} &
        \includegraphics[width=0.085\textwidth]{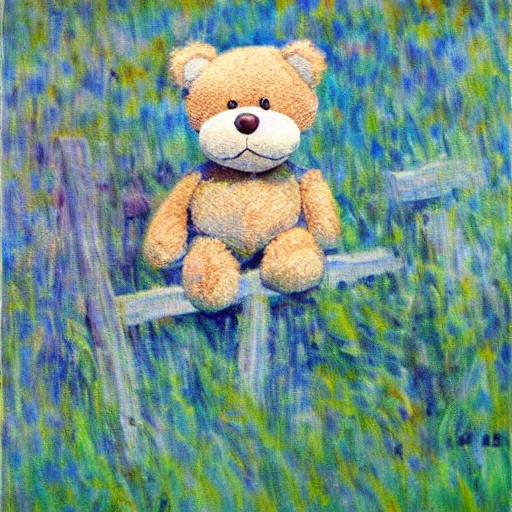} &
        \hspace{0.015cm}
        \includegraphics[width=0.085\textwidth]{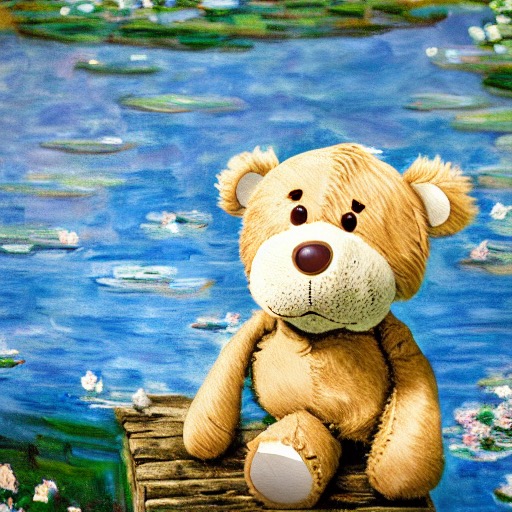} &
        \includegraphics[width=0.085\textwidth]{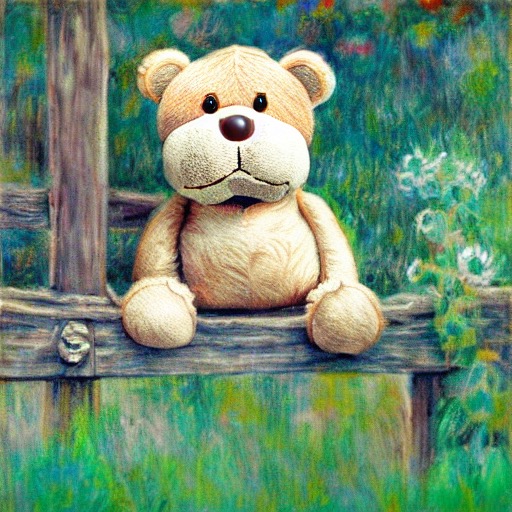} \\ \\

        \includegraphics[width=0.085\textwidth]{images/original/wooden_pot.jpg} &
        \includegraphics[width=0.085\textwidth]{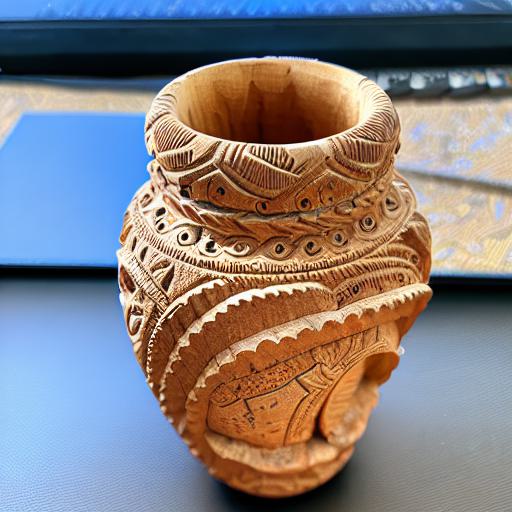} &
        \includegraphics[width=0.085\textwidth]{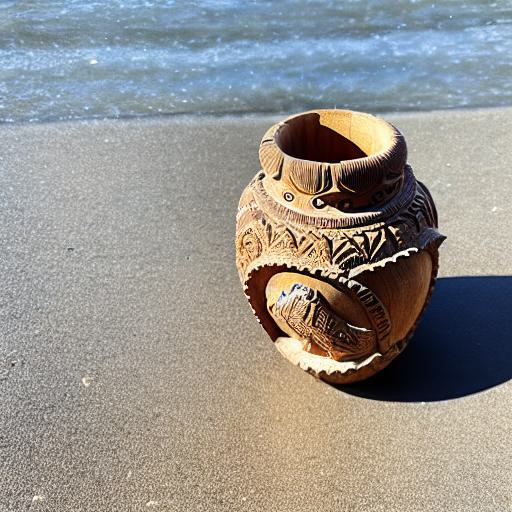} &
        \hspace{0.015cm}
        \includegraphics[width=0.085\textwidth]{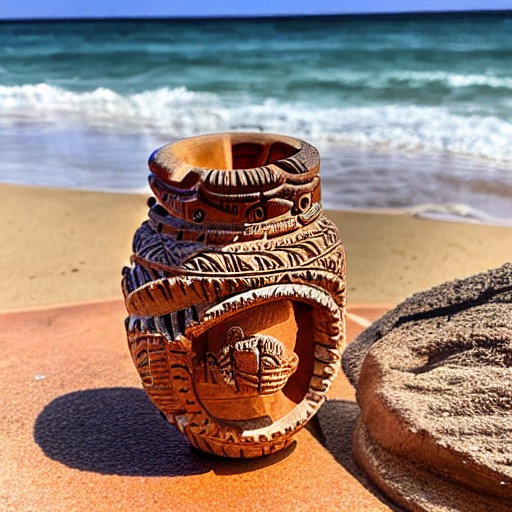} &
        \includegraphics[width=0.085\textwidth]{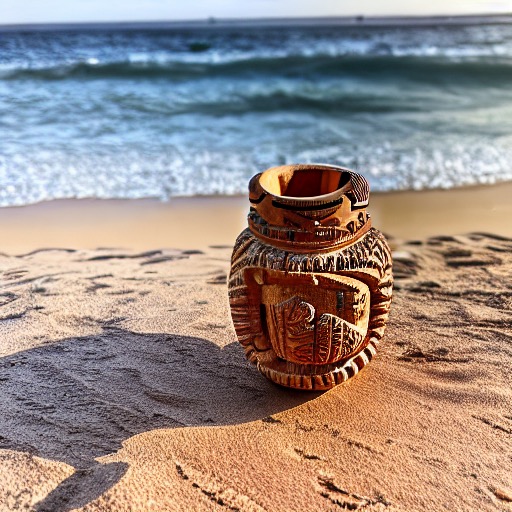} \\

        \raisebox{0.25in}{\begin{tabular}{c} ``A photo of \\ \\[-0.05cm] $S_*$ on the \\ \\[-0.05cm] beach''\end{tabular}} &
        \includegraphics[width=0.085\textwidth]{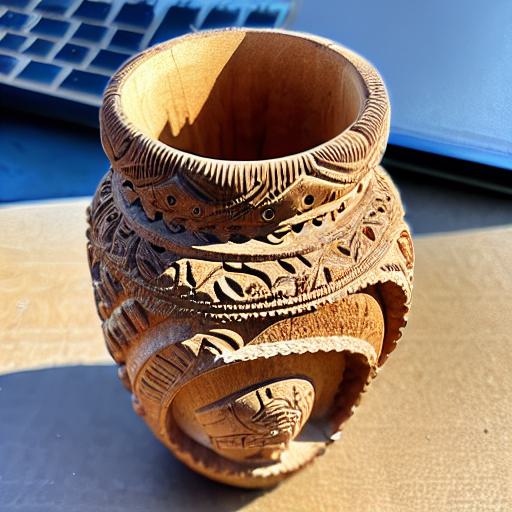} &
        \includegraphics[width=0.085\textwidth]{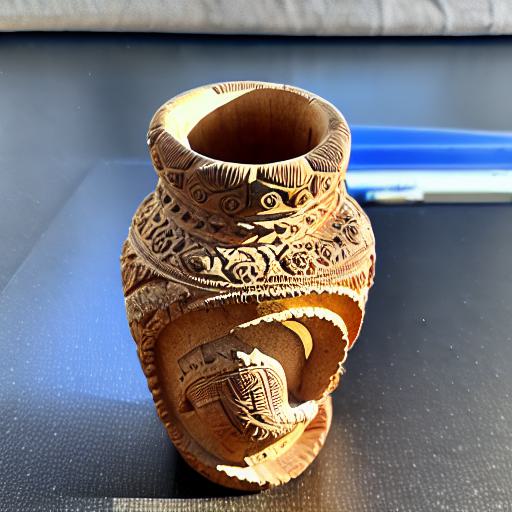} &
        \hspace{0.015cm}
        \includegraphics[width=0.085\textwidth]{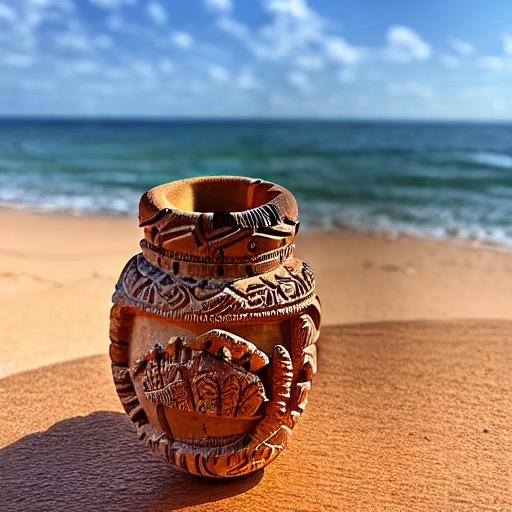} &
        \includegraphics[width=0.085\textwidth]{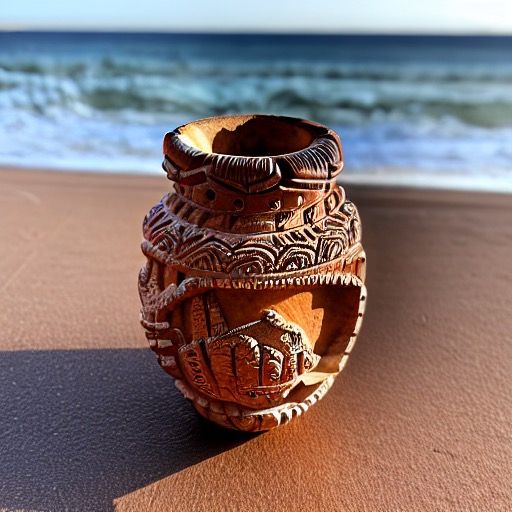} \\ \\

    \end{tabular}
    \\[-0.4cm]
    }
    \caption{Comparison to CustomDiffusion. Both models were trained for $500$ steps and images for each concept were generated using the same seeds.}
    \label{fig:customdiffusion_comparison}
\end{figure}

In~\Cref{fig:customdiffusion_comparison} we provide a visual comparison over various concepts and text prompts obtained by both methods after $500$ training steps.
As can be seen, CustomDiffusion, which finetunes the diffusion model, often fails to place the learned concept in new compositions. We attribute this to the fact that the model tends to overfit the original training images, as can be seen in the wooden pot example in the final row. There, CustomDiffusion leaks details appearing in the training images and fails to place the pot on the beach as specified in the prompt. The leakage is also present in the plush tortoise example in the second row, where the rug texture can be found in the generated images. In contrast, NeTI more accurately reconstructs the concepts while placing them in new scenes.

\begin{figure}
    \centering
    \includegraphics[width=0.475\textwidth]{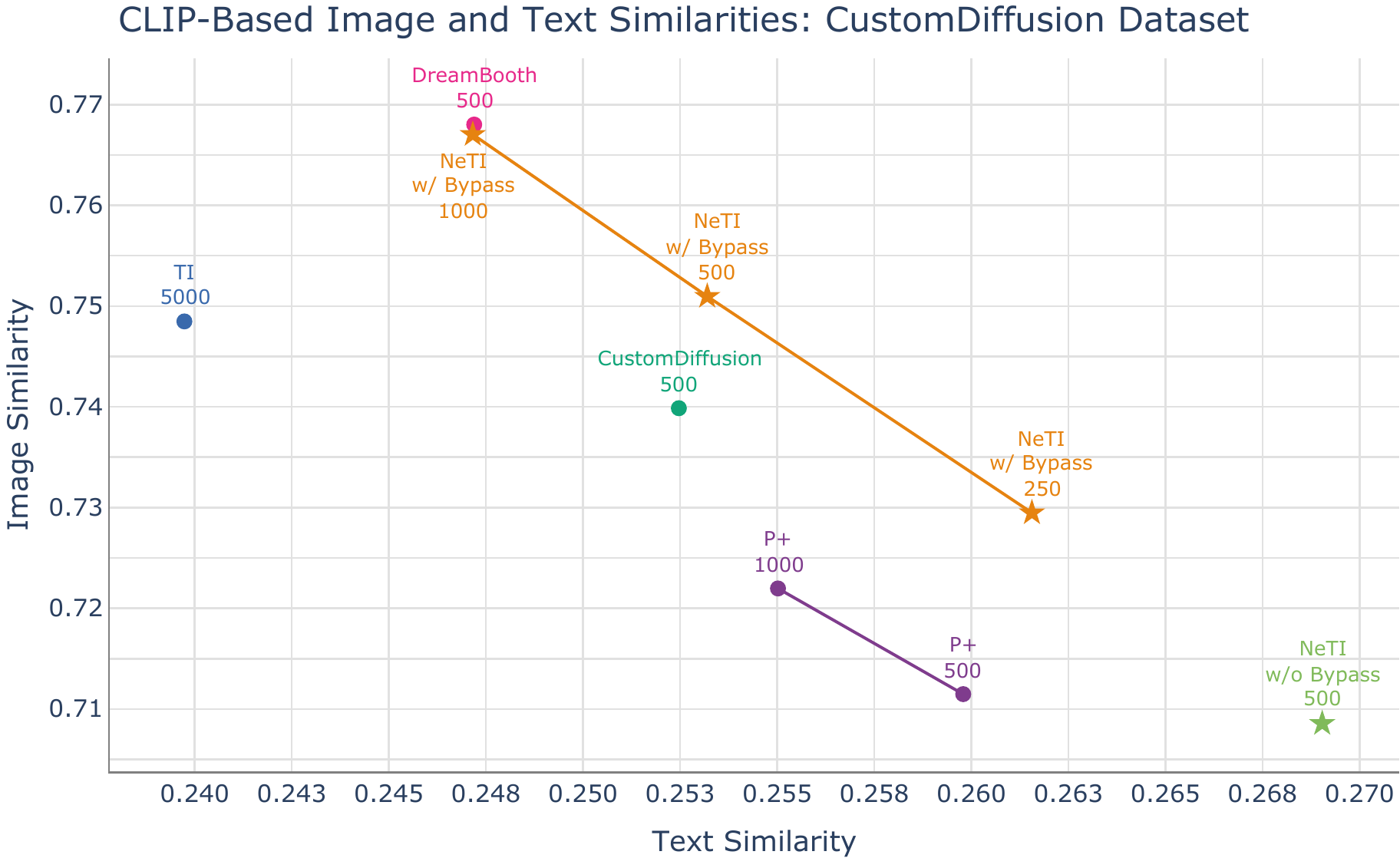}
    \\[-0.35cm]
    \caption{Quantitative Metrics over the $6$ concepts from CustomDiffusion. NeTI with our textual bypass outperform CustomDiffusion along both fronts when trained for the same number of steps. 
    When trained for 1,000 steps, our models become competitive with DreamBooth~\cite{ruiz2022dreambooth}. 
    }
    \label{fig:customdiffusion_quantitative}
    \vspace{-0.2cm}
\end{figure}

In~\Cref{fig:customdiffusion_quantitative} we provide a quantitative comparison of their six concepts across all alternative methods and our proposed NeTI variants. First, when trained for $500$ steps our NeTI models with textual bypass outperform CustomDiffusion both in terms of image similarity and text similarity. Moreover, when continuing to train for $1000$ steps, our models reach performance comparable to that of DreamBooth~\cite{ruiz2022dreambooth}, all without requiring tuning the generative model. Importantly, since we do not tune the generative model, our storage footprint is much lower than that of CustomDiffusion, requiring approximately $75\times$ \textit{less} disk space per concept. This further highlights the appealing properties of personalization-by-inversion techniques such as ours.

\section{Style Mixing} 
In Voynov~\etal~\shortcite{voynov2023p} the authors demonstrate that their \pplus latent space allows for mixing the geometry of one concept with the appearance of another concept. They demonstrated that this style mixing capability was made possible since different layers of the denoising U-Net model are responsible for different aspects of the generated image. Here, we investigate whether this style mixing ability holds in our \pstar latent space and neural mapper. Following the notation used in Voynov~\etal~\shortcite{voynov2023p}, given two concepts, the embeddings of the concept whose geometry we want to keep are passed to layers \texttt{(16, down,0),(16,down,1),(8,down,0)} while the embeddings of the appearance concept are passed to the remaining layers. Moreover, we investigate whether performing style mixing at different denoising timesteps can provide us with another axis of control. Therefore, rather than performing the style mixing at all denoising timesteps, we begin mixing at different starting points, such that starting later in the denoising process should preserve more details from the geometry concept.

\begin{figure*}
    \centering
    \setlength{\tabcolsep}{0.5pt}
    \addtolength{\belowcaptionskip}{-12.5pt}
    \begin{tabular}{c c@{\hspace{0.2cm}} c c c c c}

        \\ \\ \\ \\ \\ \\ \\ \\

        \includegraphics[width=0.135\textwidth]{images/original/colorful_teapot.jpg} &
        \includegraphics[width=0.135\textwidth]{images/original/fat_stone_bird.jpg} &
        \hspace{0.05cm}
        \includegraphics[width=0.135\textwidth]{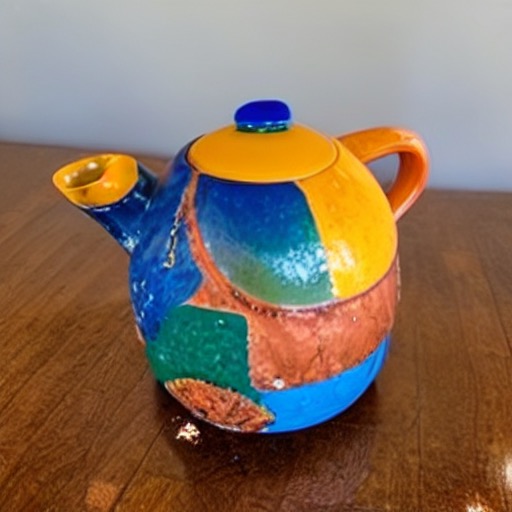} &
        \includegraphics[width=0.135\textwidth]{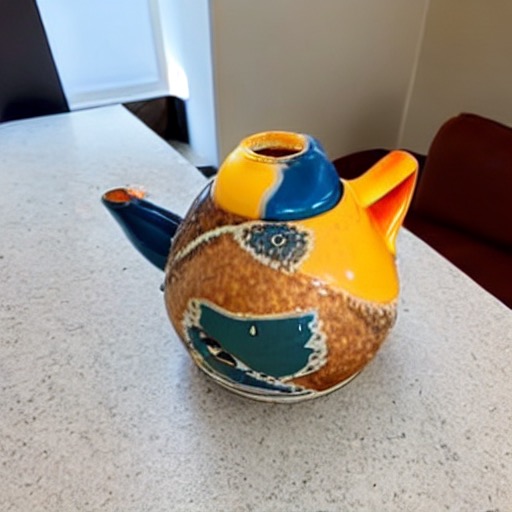} &
        \includegraphics[width=0.135\textwidth]{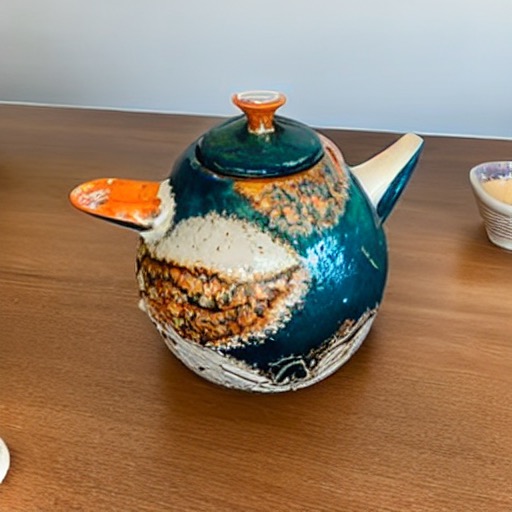} &
        \includegraphics[width=0.135\textwidth]{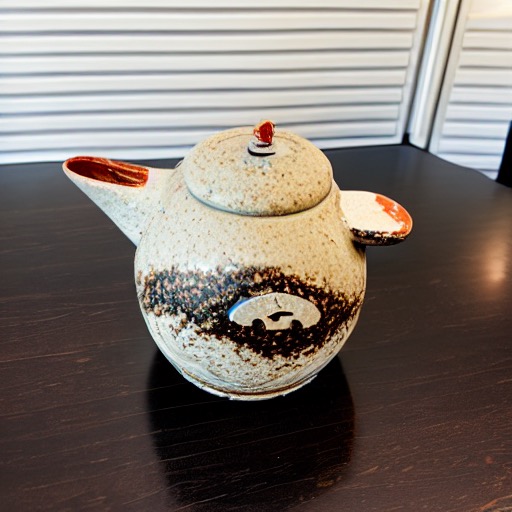} \\

        \includegraphics[width=0.135\textwidth]{images/original/teddybear.jpg} &
        \includegraphics[width=0.135\textwidth]{images/original/rainbow_cat.jpeg} &
        \hspace{0.05cm}
        \includegraphics[width=0.135\textwidth]{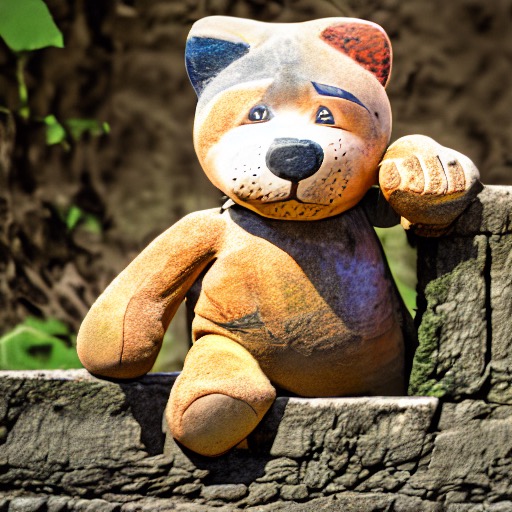} &
        \includegraphics[width=0.135\textwidth]{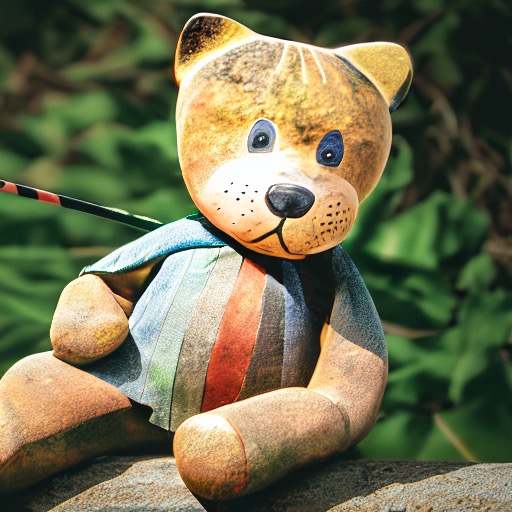} &
        \includegraphics[width=0.135\textwidth]{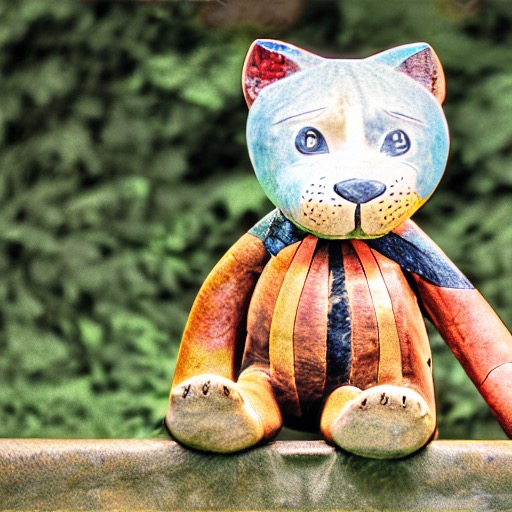} &
        \includegraphics[width=0.135\textwidth]{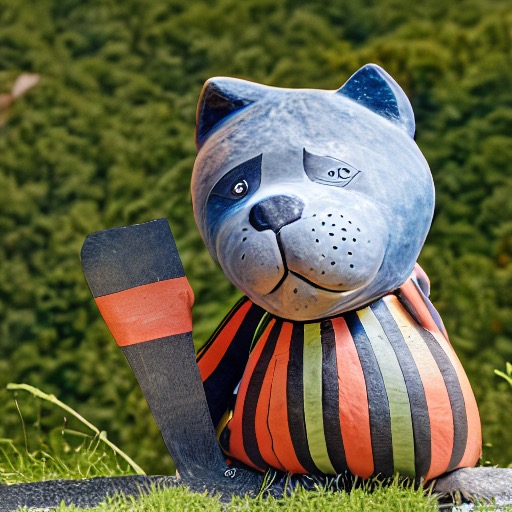} \\

        \includegraphics[width=0.135\textwidth]{images/original/colorful_teapot.jpg} &
        \includegraphics[width=0.135\textwidth]{images/original/rainbow_cat.jpeg} &
        \hspace{0.05cm}
        \includegraphics[width=0.135\textwidth]{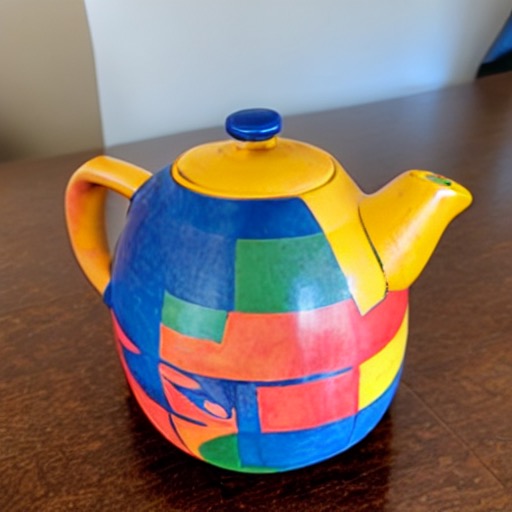} &
        \includegraphics[width=0.135\textwidth]{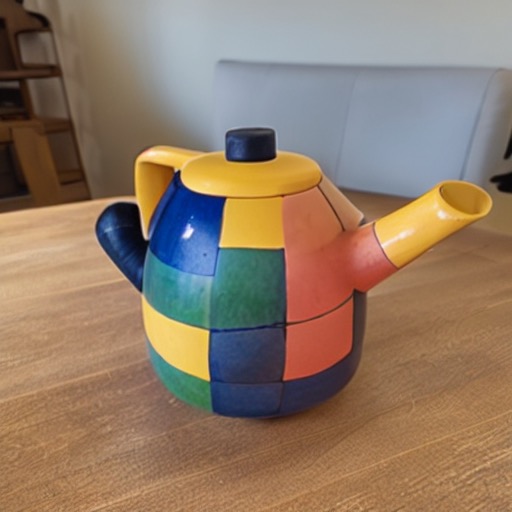} &
        \includegraphics[width=0.135\textwidth]{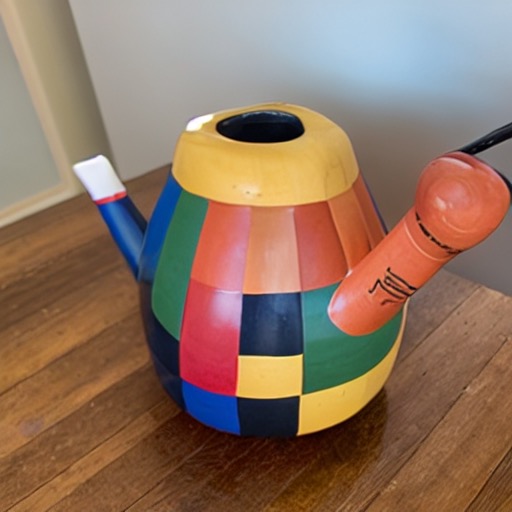} &
        \includegraphics[width=0.135\textwidth]{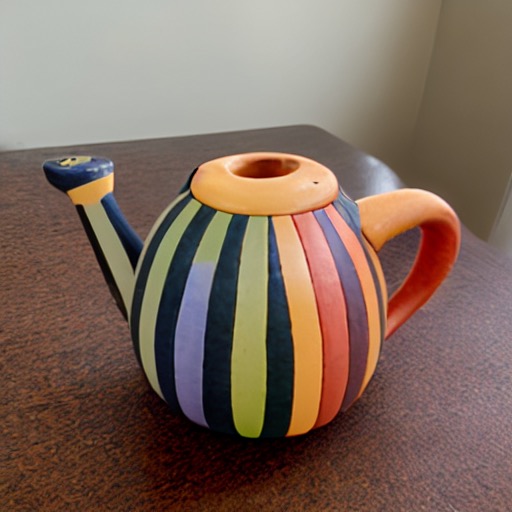} \\

        \includegraphics[width=0.135\textwidth]{images/original/colorful_teapot.jpg} &
        \includegraphics[width=0.135\textwidth]{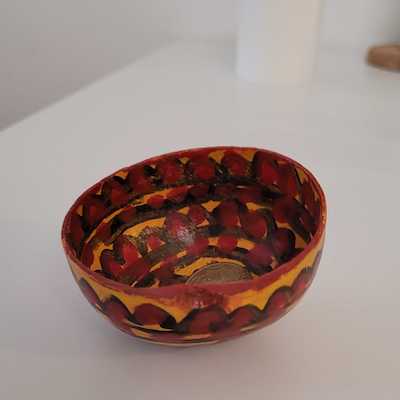} &
        \hspace{0.05cm}
        \includegraphics[width=0.135\textwidth]{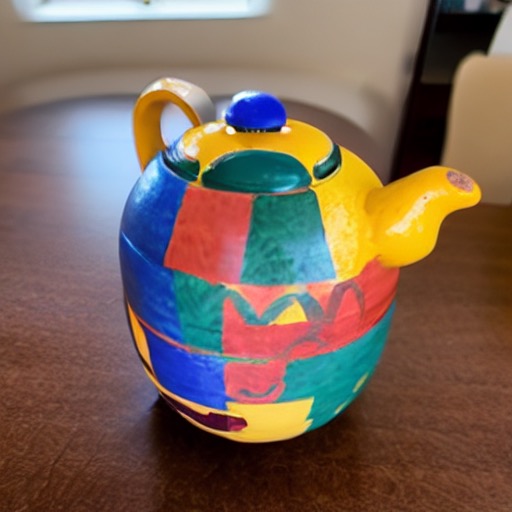} &
        \includegraphics[width=0.135\textwidth]{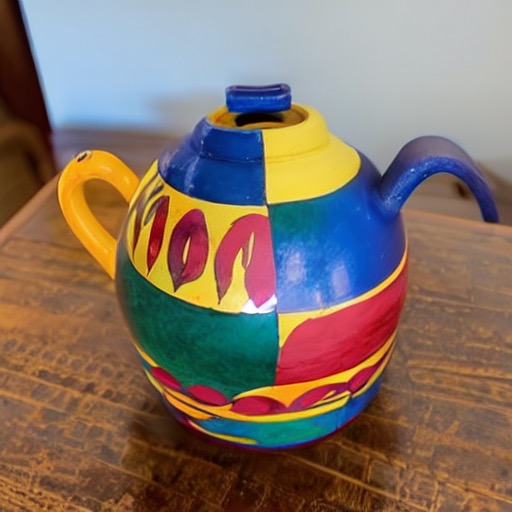} &
        \includegraphics[width=0.135\textwidth]{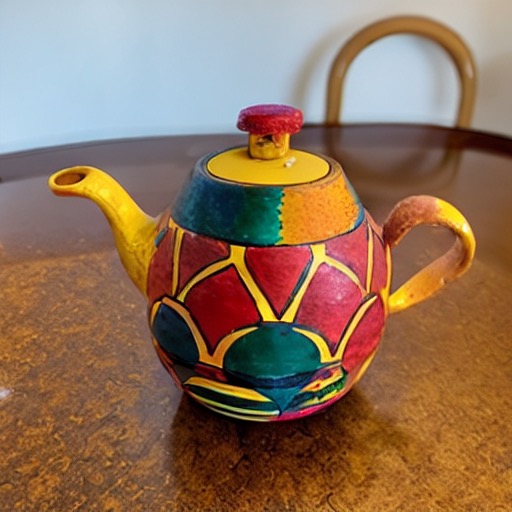} &
        \includegraphics[width=0.135\textwidth]{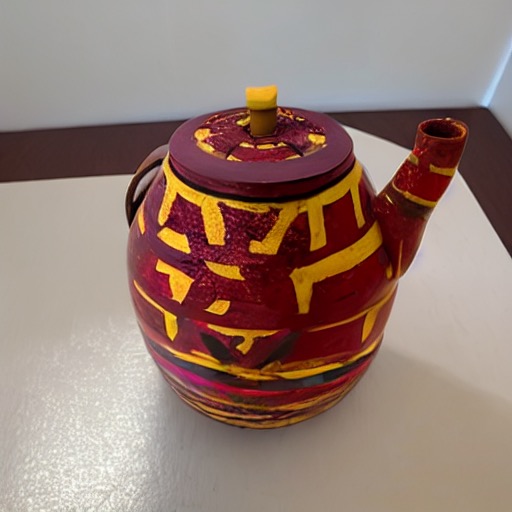} \\

        \includegraphics[width=0.135\textwidth]{images/original/red_bowl.jpg} &
        \includegraphics[width=0.135\textwidth]{images/original/rainbow_cat.jpeg} &
        \hspace{0.05cm}
        \includegraphics[width=0.135\textwidth]{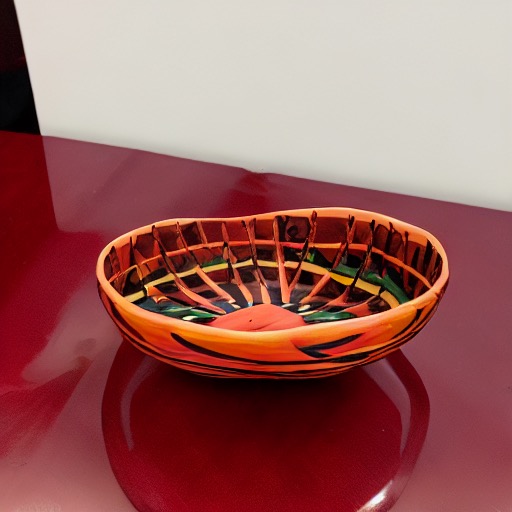} &
        \includegraphics[width=0.135\textwidth]{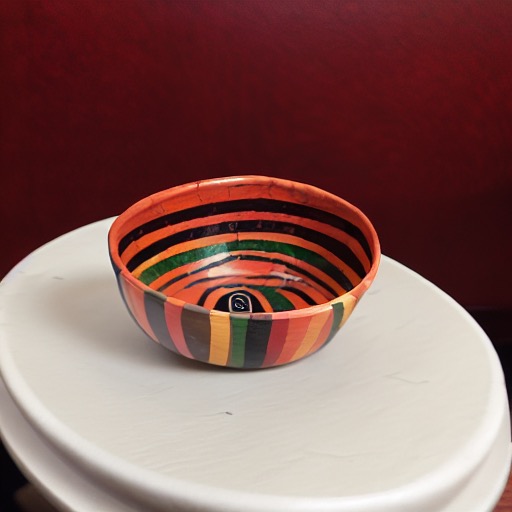} &
        \includegraphics[width=0.135\textwidth]{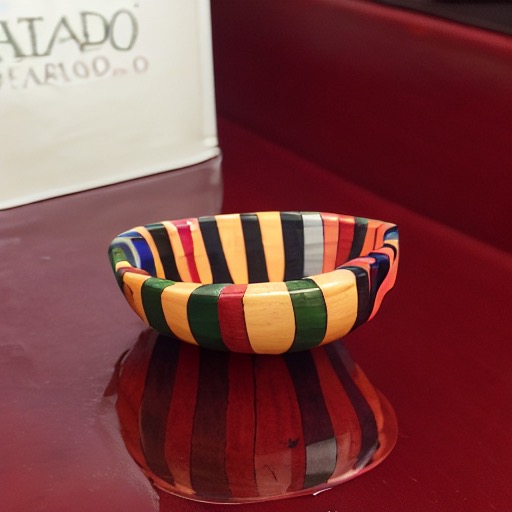} &
        \includegraphics[width=0.135\textwidth]{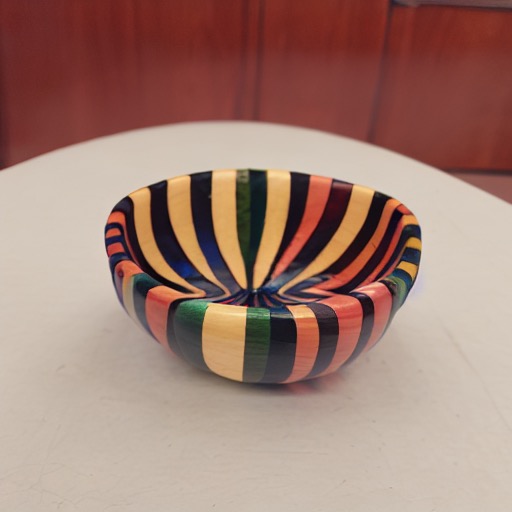} \\
        
        \includegraphics[width=0.135\textwidth]{images/original/mugs_skulls.jpeg} &
        \includegraphics[width=0.135\textwidth]{images/original/rainbow_cat.jpeg} &
        \hspace{0.05cm}
        \includegraphics[width=0.135\textwidth]{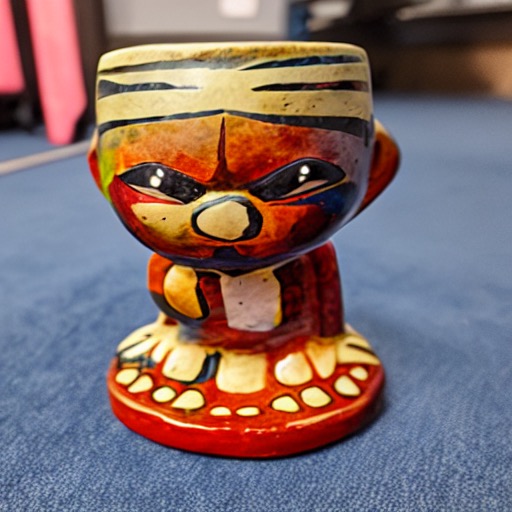} &
        \includegraphics[width=0.135\textwidth]{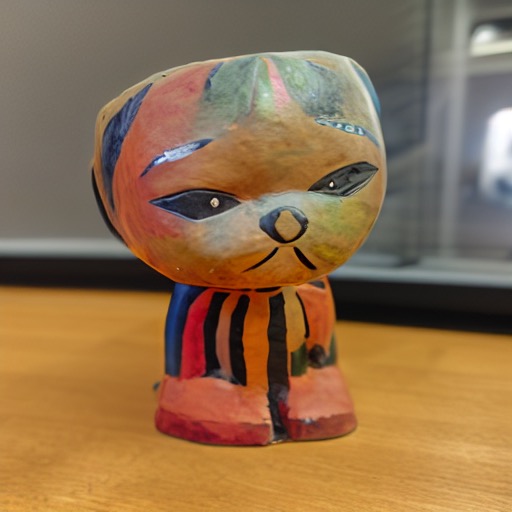} &
        \includegraphics[width=0.135\textwidth]{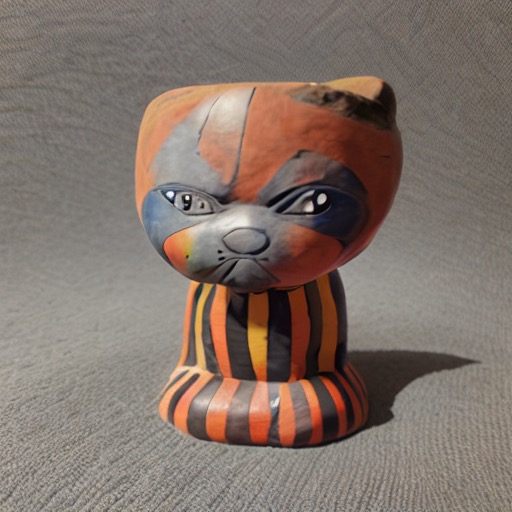} &
        \includegraphics[width=0.135\textwidth]{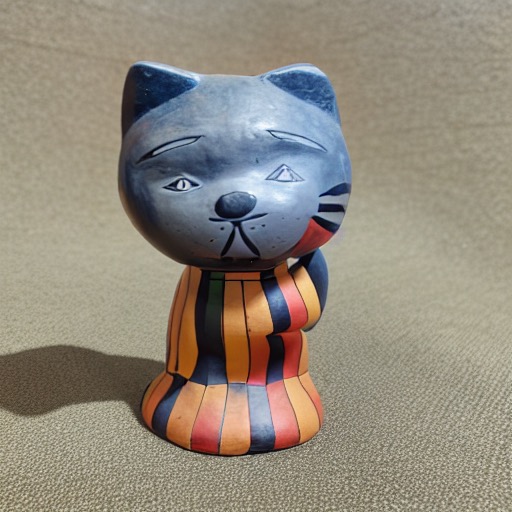} \\
        
        Geometry & Appearance & \multicolumn{4}{c}{Time-Based Mixing Control}

    \end{tabular}
    \vspace{-0.1cm}
    \caption{Style mixing results obtained with NeTI. Given two concepts, the embeddings of the geometry concept are passed to layers \texttt{(16, down,0),(16,down,1),(8,down,0)} while the embeddings of the appearance concept are passed to the remaining U-Net layers. To provide an additional axis of control, we also leverage the time component of \pstar and begin the mixing at various denoising timesteps. From left to right: $t=600,700,800$, and $900$. As can be seen, starting the mixing later, i.e., a smaller value of $t$ passes more information from the geometry concept to the output image.}
    \label{fig:style_mixing}
\end{figure*}

We illustrate this idea in~\Cref{fig:style_mixing}. In each row, we are given a geometry concept (left) and appearance concept (right) and perform the mixing starting at $4$ different timesteps: $t=600,700,800$, and $900$. As can be seen, when we start later in the denoising process, more details from the geometry concept are present in the output image. For example, in the bottom row for $t=600$, the output maintains the white top and color scheme of the original mug. As $t$ increases, i.e., the mixing begins earlier in the denoising process, more of the appearance concept is passed to the output image. In the same example as above, the output now contains colorful stripes and a head resembling that of the original cat statue. This behavior can also be seen in the top row, where the teapot gradually shifts into a rock-like texture, matching the stone bird input.

We do note that style mixing may be challenging over concepts that converge quickly. We believe that this can be attributed to the information sharing present in our neural mapper. Since all embeddings are computed using shared weights, the disentanglement between different U-Net layers may not be as strong as was observed in XTI where each embedding is optimized independently.

\null\newpage
\null\newpage

\begin{figure}
    \centering
    \includegraphics[width=0.45\textwidth]{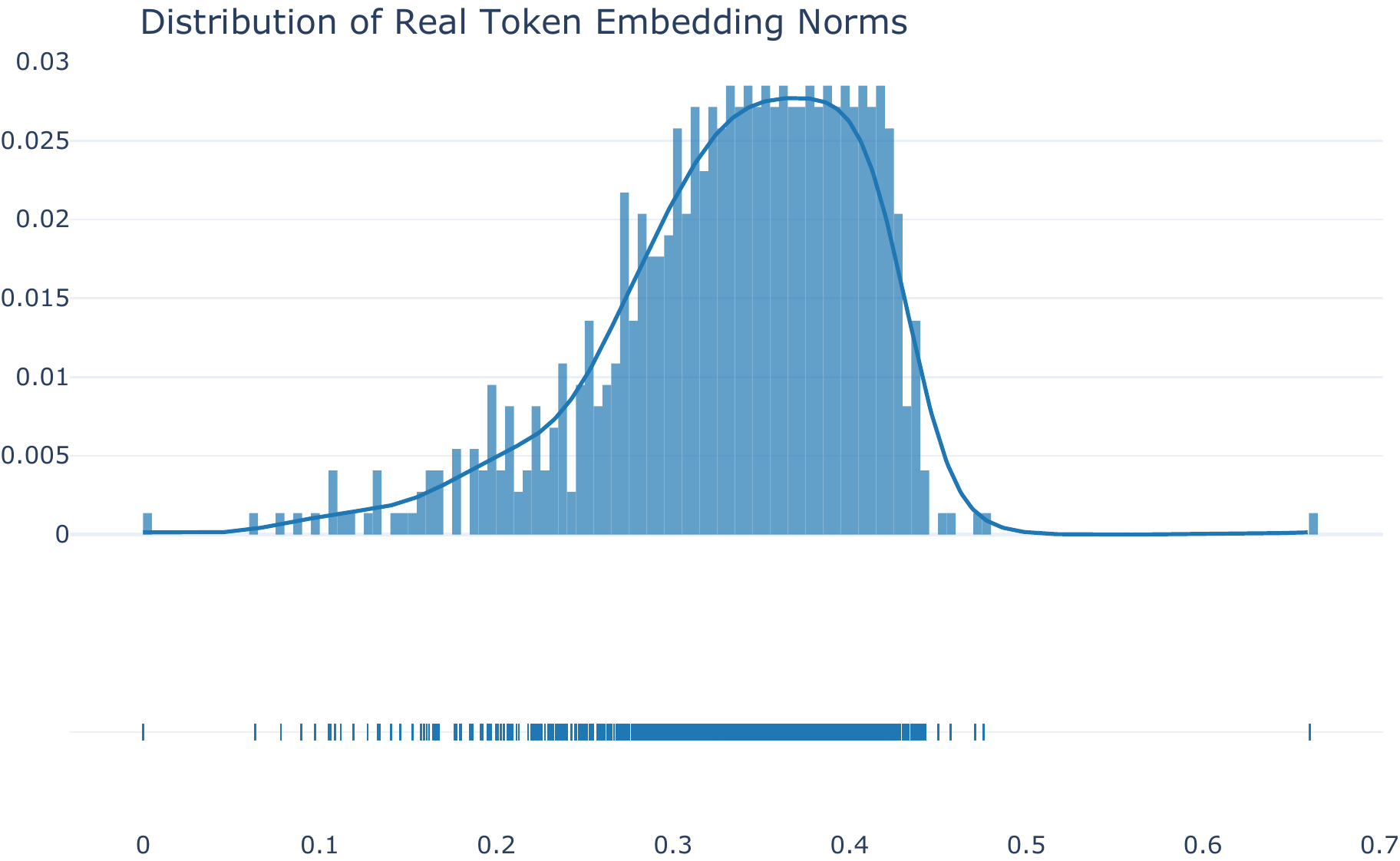}
    \\[-0.2cm]
    \caption{Distribution of the norms of real token embeddings from the pretrained CLIP text encoder~\cite{radford2021learning}.}
    \label{fig:token_norm_distribution}
    \vspace{-0.2cm}
\end{figure}
\begin{figure}
    \centering
    \includegraphics[width=0.45\textwidth]{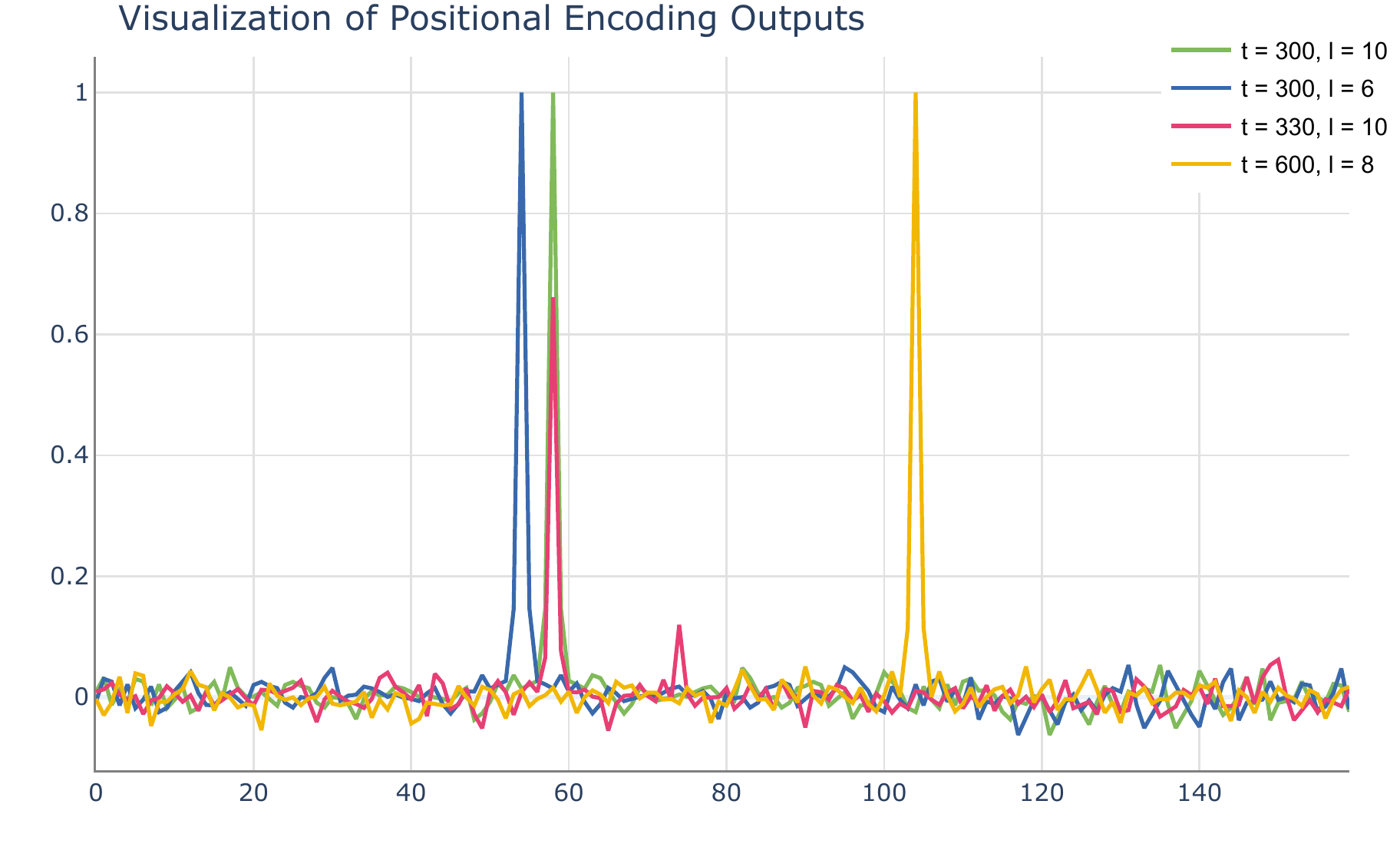}
    \\[-0.45cm]
    \caption{Visualization of our positional encoding for various inputs $(t,\ell)$.}
    \label{fig:positional_encoding_visualization}
    \vspace{-0.2cm}
\end{figure}

\vspace*{1cm}
\section{Additional Analysis}~\label{sec:additional_analysis}

\vspace{-0.5cm}
In this section, we provide additional visualizations and analyses to complement those in the main paper.

\vspace{-0.325cm}
\paragraph{\textbf{Distribution of Token Embedding Norms.}}
First, in~\Cref{fig:token_norm_distribution}, we visualize the distribution of the \textit{norms} of real token embeddings in CLIP's pretrained text encoder~\cite{radford2021learning}. As can be seen, the norms follow a very narrow distribution centered around $0.25 - 0.45$. This distribution provides additional motivation for our rescaling technique described in the main paper. Our embeddings should ideally behave similarly to real token embeddings and by normalizing the norm of our embeddings to match that of real token embeddings, we obtain a representation that behaves similarly to real tokens, thereby improving our editability.

\vspace{-0.325cm}
\paragraph{\textbf{Positional Encoding.}}
In~\Cref{fig:positional_encoding_visualization}, we provide a visualization of various outputs returned by our positional encoding function. As can be seen, the \textcolor{blue}{blue} and \textcolor{darkgreen}{green} curves are well-separated due to their different layer indices. Conversely, the \textcolor{darkgreen}{green} and \textcolor{red}{red} curves share a similar encoding as they both share the same U-Net layer as input, but differ slightly in their timestep. Finally, the \textcolor{yelloworange}{yellow} curve, differs both in its input timestep and layer index, resulting in an encoding that differs significantly from the other encodings.

\section{Additional Qualitative Results}~\label{sec:additional_results}
\begin{figure*}
    \centering
    \setlength{\tabcolsep}{0.5pt}
    \addtolength{\belowcaptionskip}{-12.5pt}
    {\small
    \begin{tabular}{c@{\hspace{0.2cm}} c c c c c}
        
        \includegraphics[width=0.14\textwidth]{images/original/elephant.jpg} &
        \includegraphics[width=0.14\textwidth]{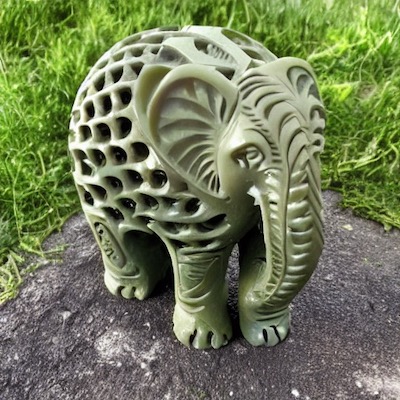} &
        \includegraphics[width=0.14\textwidth]{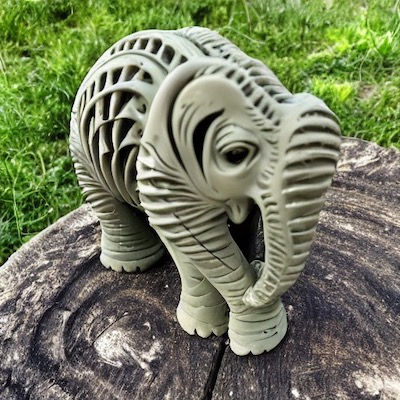} &
        \includegraphics[width=0.14\textwidth]{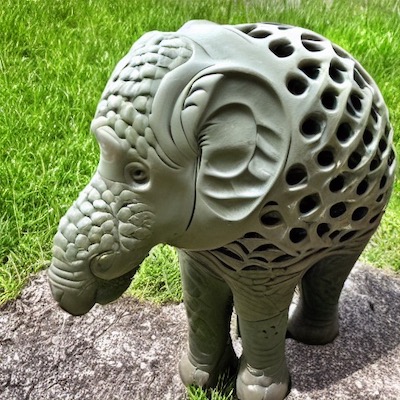} &
        \includegraphics[width=0.14\textwidth]{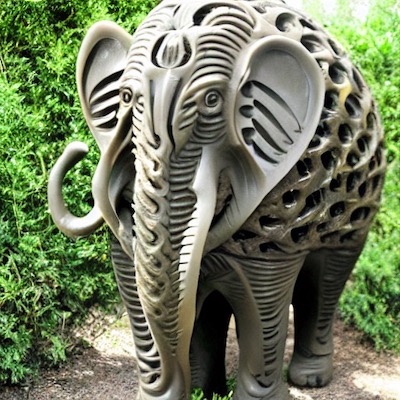} &
        \includegraphics[width=0.14\textwidth]{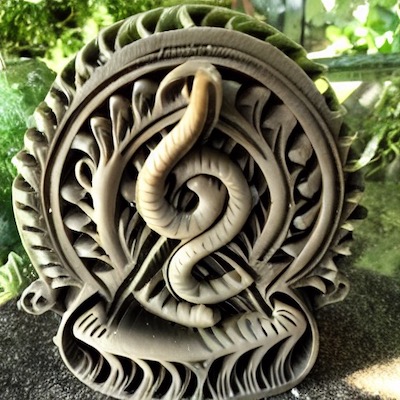} \\
        
        \includegraphics[width=0.14\textwidth]{images/original/fat_stone_bird.jpg} &
        \includegraphics[width=0.14\textwidth]{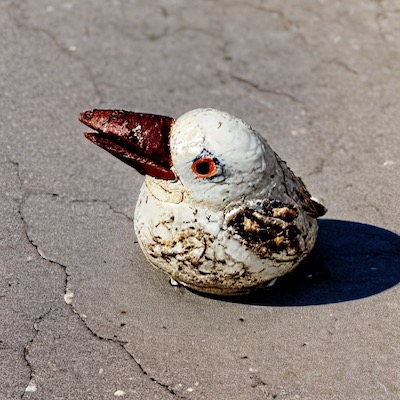} &
        \includegraphics[width=0.14\textwidth]{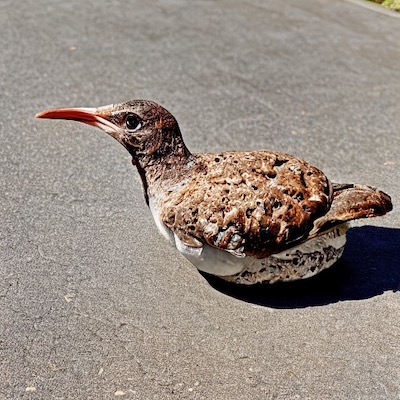} &
        \includegraphics[width=0.14\textwidth]{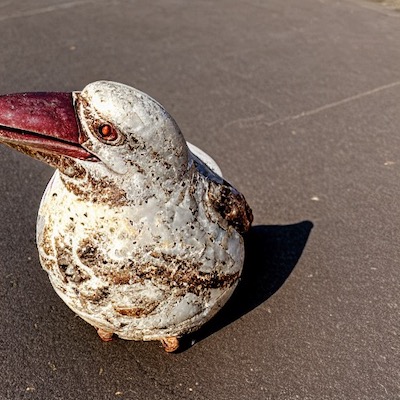} &
        \includegraphics[width=0.14\textwidth]{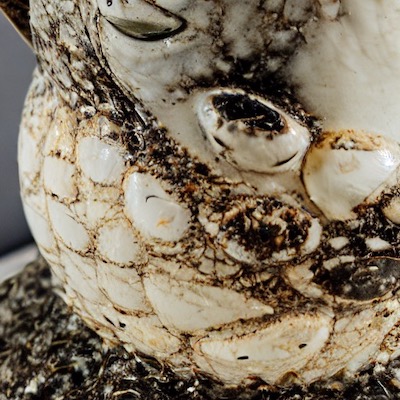} &
        \includegraphics[width=0.14\textwidth]{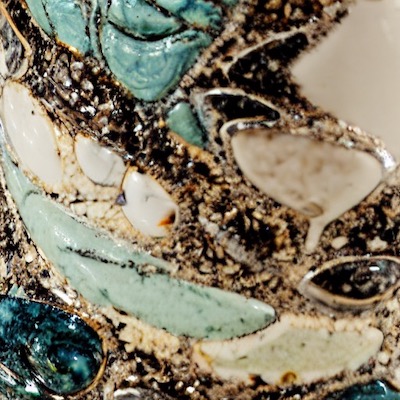} \\
        
        \includegraphics[width=0.14\textwidth]{images/original/mugs_skulls.jpeg} &
        \includegraphics[width=0.14\textwidth]{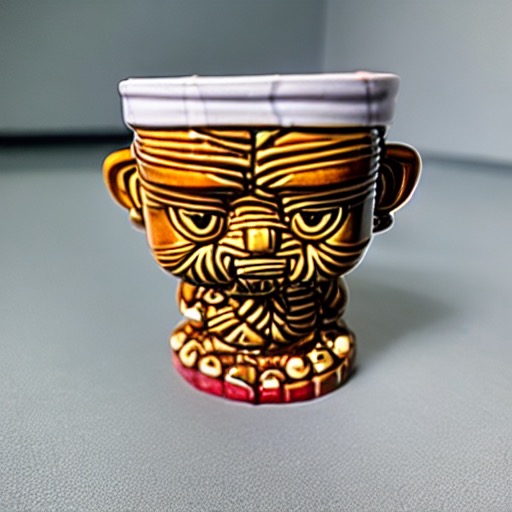} &
        \includegraphics[width=0.14\textwidth]{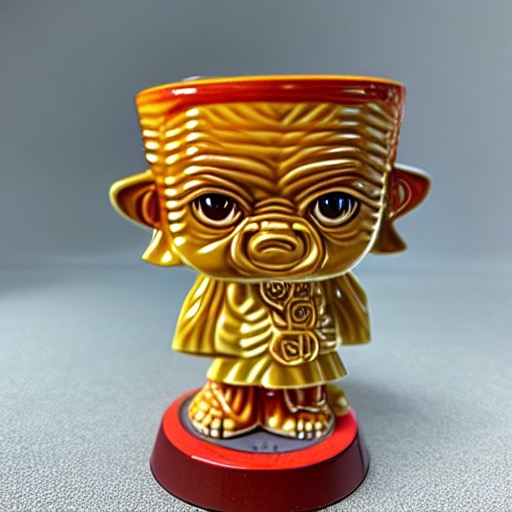} &
        \includegraphics[width=0.14\textwidth]{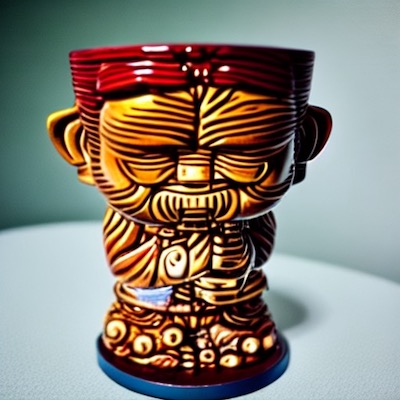} &
        \includegraphics[width=0.14\textwidth]{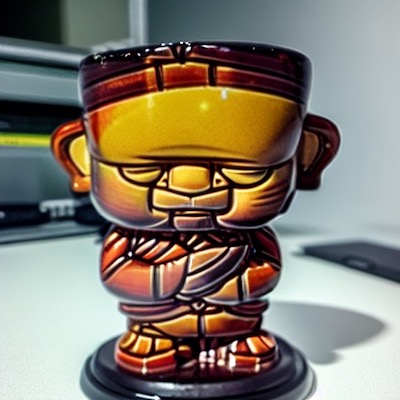} &
        \includegraphics[width=0.14\textwidth]{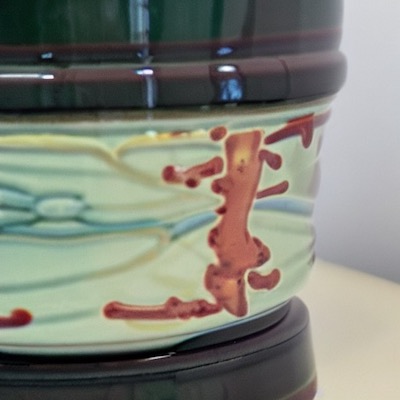} \\

        \includegraphics[width=0.14\textwidth]{images/original/metal_bird.jpg} &
        \includegraphics[width=0.14\textwidth]{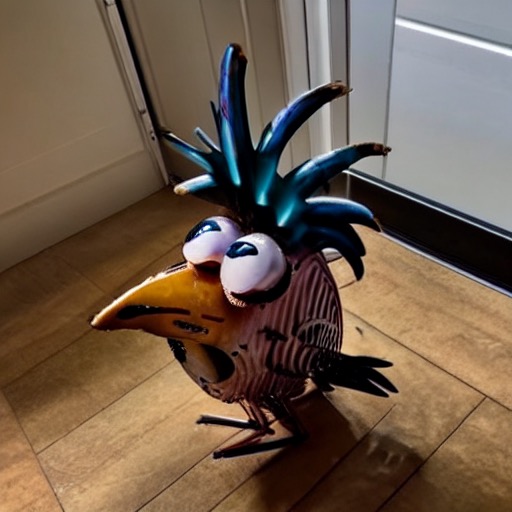} &
        \includegraphics[width=0.14\textwidth]{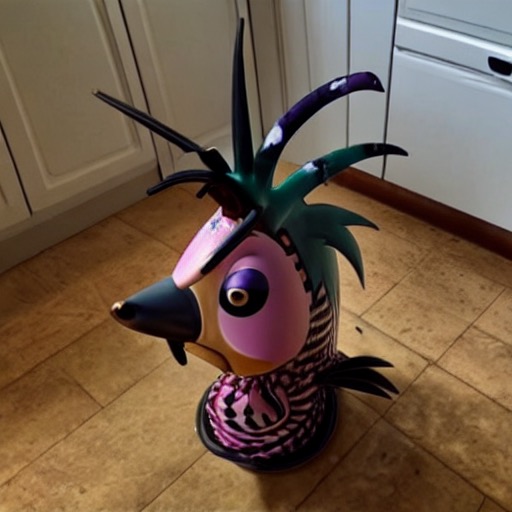} &
        \includegraphics[width=0.14\textwidth]{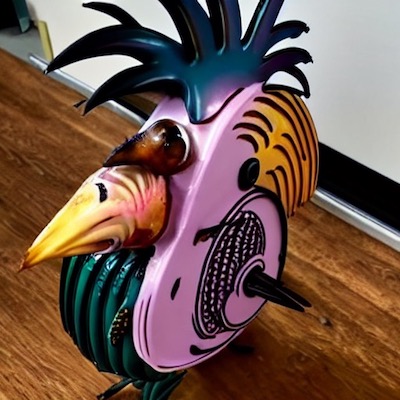} &
        \includegraphics[width=0.14\textwidth]{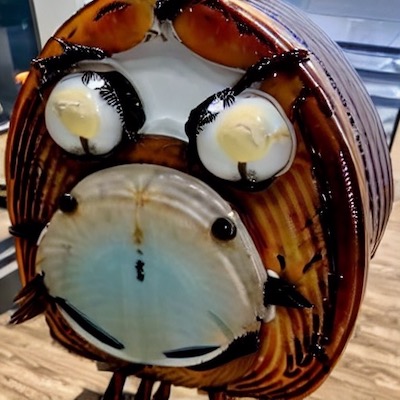} &
        \includegraphics[width=0.14\textwidth]{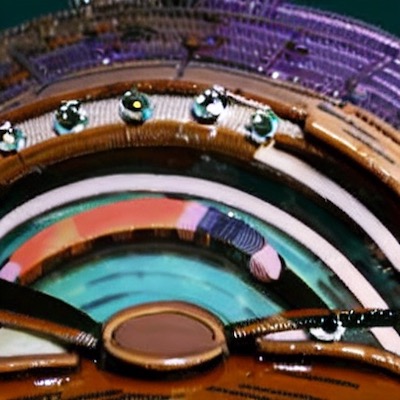} \\

        \includegraphics[width=0.14\textwidth]{images/original/red_bowl.jpg} &
        \includegraphics[width=0.14\textwidth]{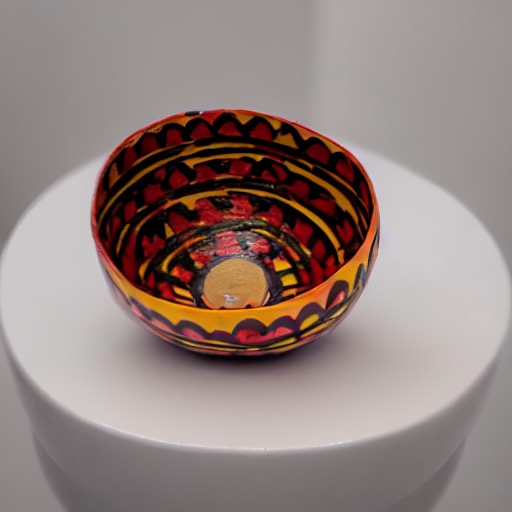} &
        \includegraphics[width=0.14\textwidth]{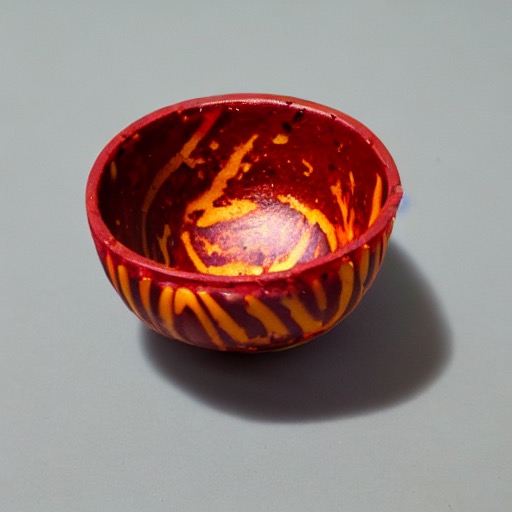} &
        \includegraphics[width=0.14\textwidth]{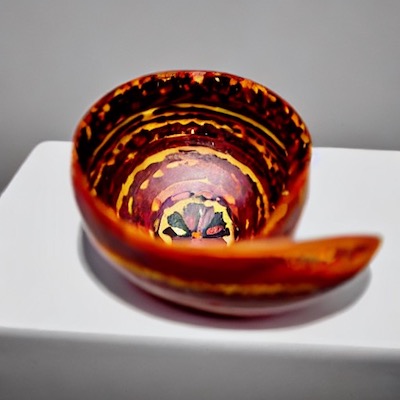} &
        \includegraphics[width=0.14\textwidth]{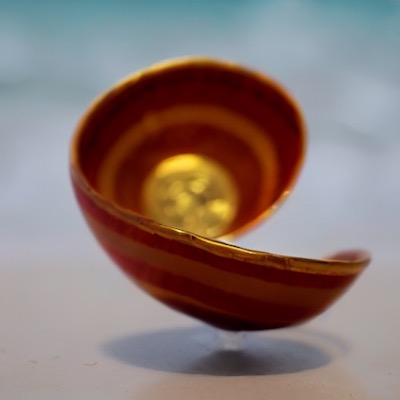} &
        \includegraphics[width=0.14\textwidth]{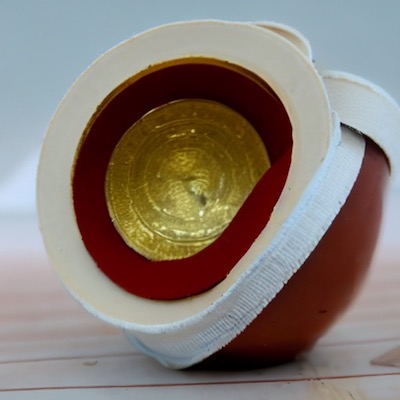} \\

        \includegraphics[width=0.14\textwidth]{images/original/clock.jpeg} &
        \includegraphics[width=0.14\textwidth]{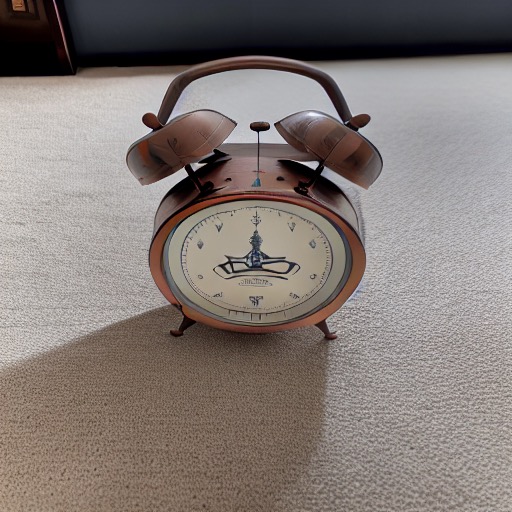} &
        \includegraphics[width=0.14\textwidth]{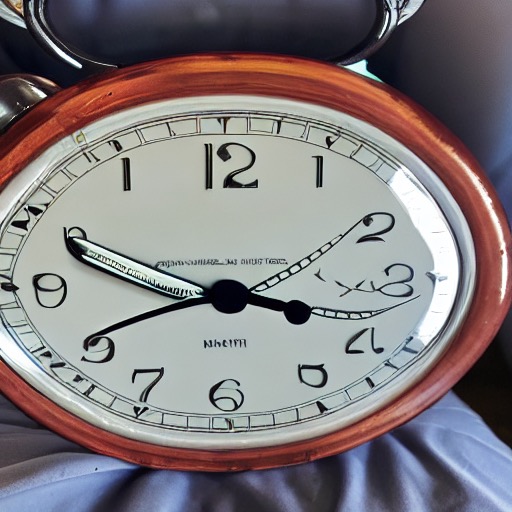} &
        \includegraphics[width=0.14\textwidth]{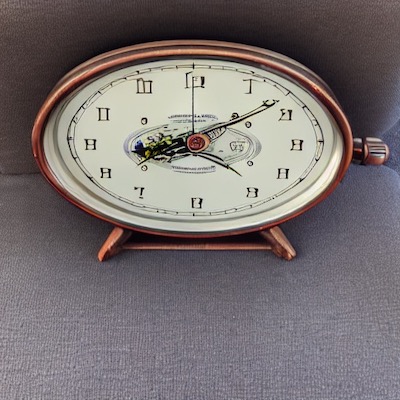} &
        \includegraphics[width=0.14\textwidth]{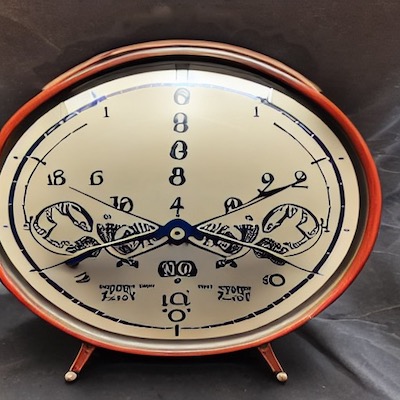} &
        \includegraphics[width=0.14\textwidth]{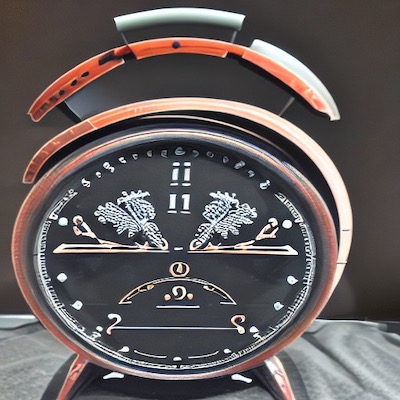} \\

        \includegraphics[width=0.14\textwidth]{images/original/headless_statue.jpeg} &
        \includegraphics[width=0.14\textwidth]{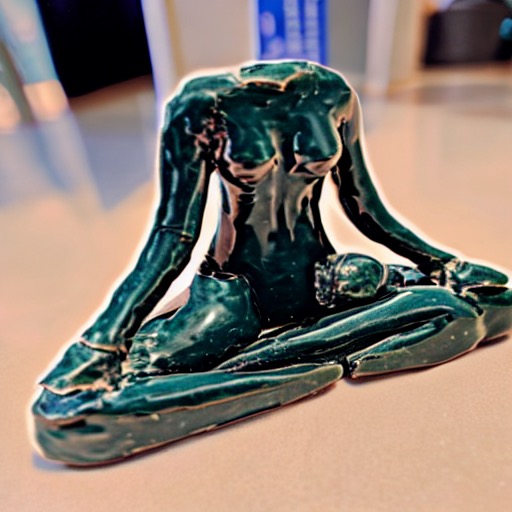} &
        \includegraphics[width=0.14\textwidth]{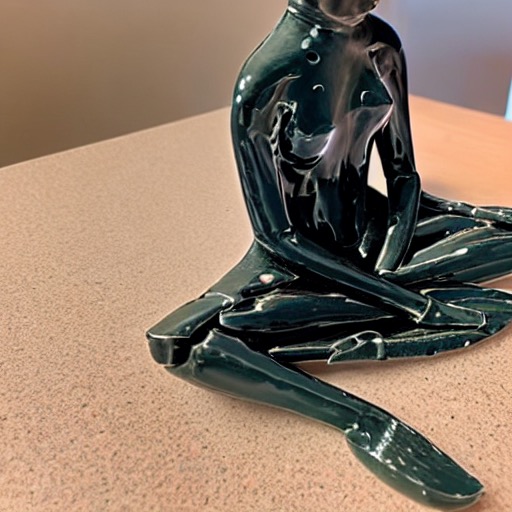} &
        \includegraphics[width=0.14\textwidth]{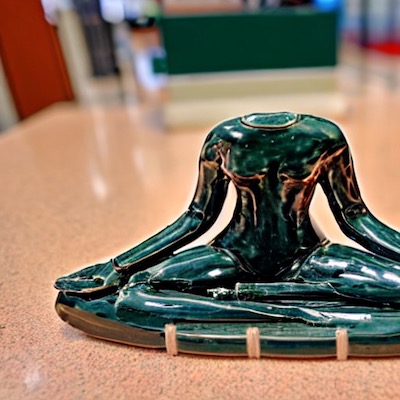} &
        \includegraphics[width=0.14\textwidth]{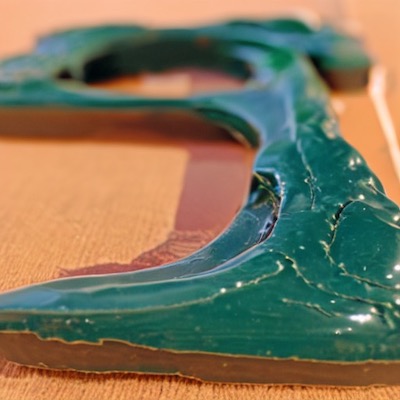} &
        \includegraphics[width=0.14\textwidth]{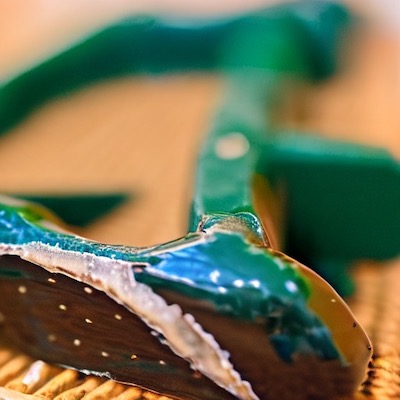} \\

        \includegraphics[width=0.14\textwidth]{images/original/dangling_child.jpg} &
        \includegraphics[width=0.14\textwidth]{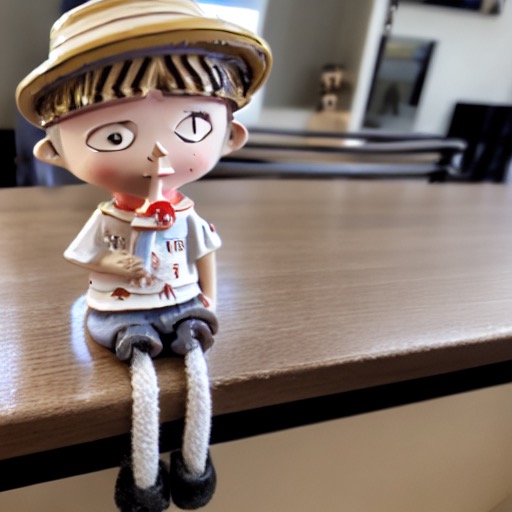} &
        \includegraphics[width=0.14\textwidth]{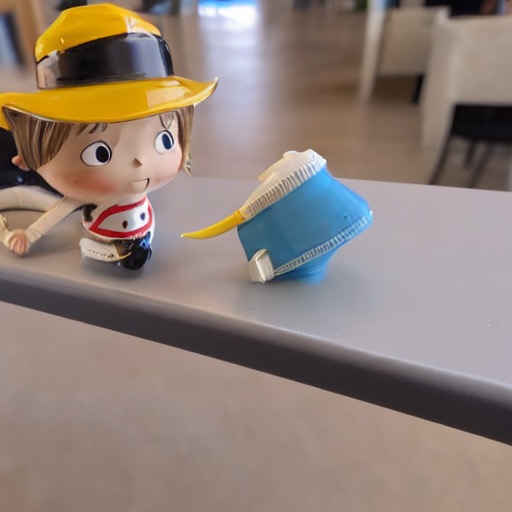} &
        \includegraphics[width=0.14\textwidth]{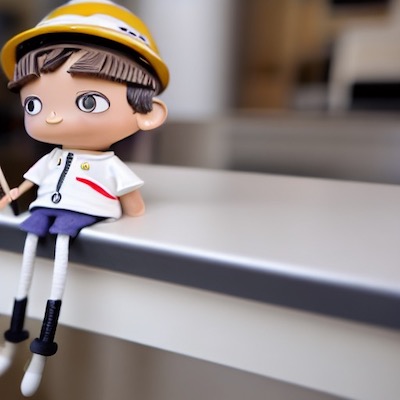} &
        \includegraphics[width=0.14\textwidth]{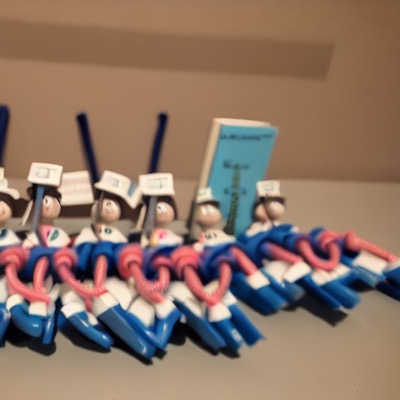} &
        \includegraphics[width=0.14\textwidth]{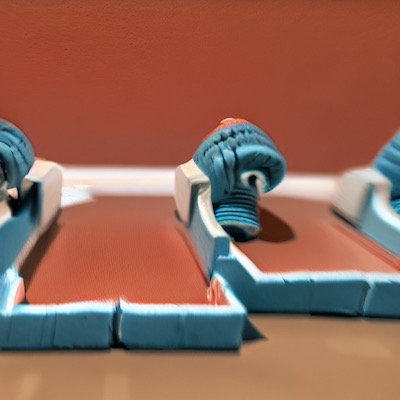} \\
        
        Real & All & $t=999$ & $t=666$ & $t=333$ & $t=50$ \\
        
    \\[-0.2cm]        
    \end{tabular}
    }
    \vspace{-0.3cm}
    \vspace{0.1cm}
    \caption{Additional examples illustrating which concept-specific details are captured at different denoising timesteps using NeTI. The same random seed is used for all images in the same column.}
    \label{fig:timestep_analysis_sup}
\end{figure*}

\begin{figure*}
    \centering
    \setlength{\tabcolsep}{0.5pt}
    \addtolength{\belowcaptionskip}{-12.5pt}
    \begin{tabular}{c@{\hspace{0.2cm}} c@{\hspace{0.2cm}} c@{\hspace{0.2cm}} c@{\hspace{0.2cm}} c@{\hspace{0.2cm}} c@{\hspace{0.2cm}} c}

        \includegraphics[width=0.125\textwidth]{images/original/fat_stone_bird.jpg} &
        \raisebox{0.315in}{\begin{tabular}{c} ``Watercolor \\ painting \\ of $S_*$'' \end{tabular}} &
        \includegraphics[width=0.125\textwidth]{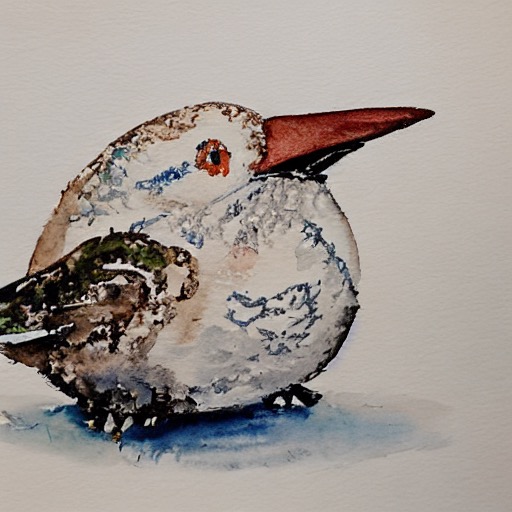} &
        \includegraphics[width=0.125\textwidth]{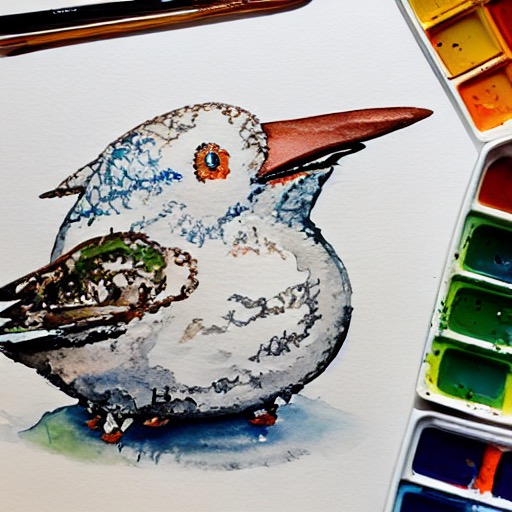} &
        \includegraphics[width=0.125\textwidth]{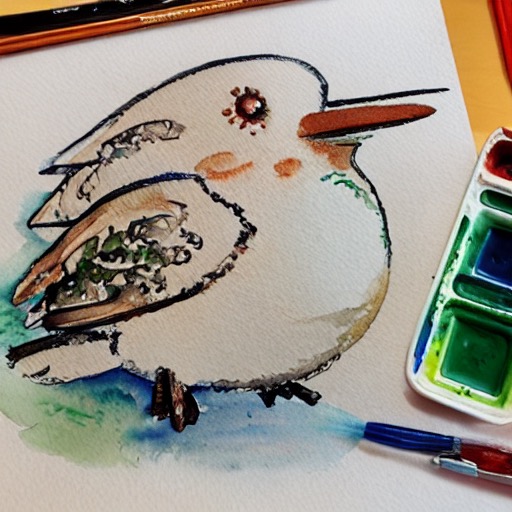} &
        \includegraphics[width=0.125\textwidth]{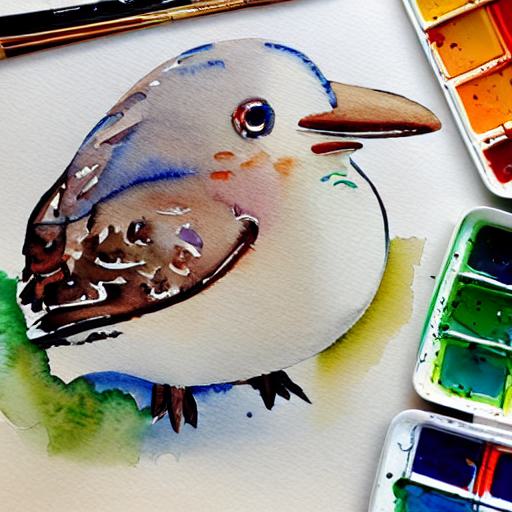} \\
        
        \includegraphics[width=0.125\textwidth]{images/original/colorful_teapot.jpg} &
        \raisebox{0.315in}{\begin{tabular}{c} ``An app \\ icon of $S_*$'' \end{tabular}} &
        \includegraphics[width=0.125\textwidth]{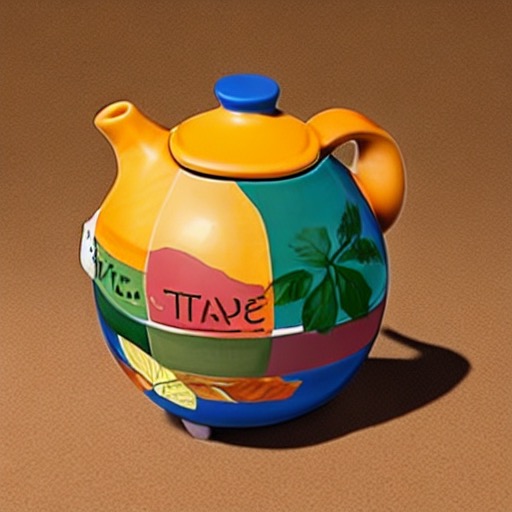} &
        \includegraphics[width=0.125\textwidth]{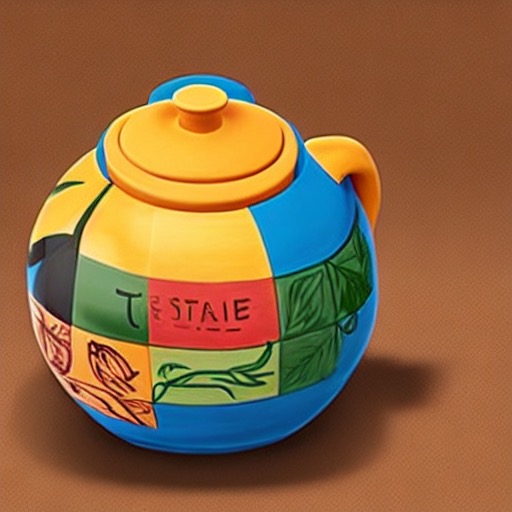} &
        \includegraphics[width=0.125\textwidth]{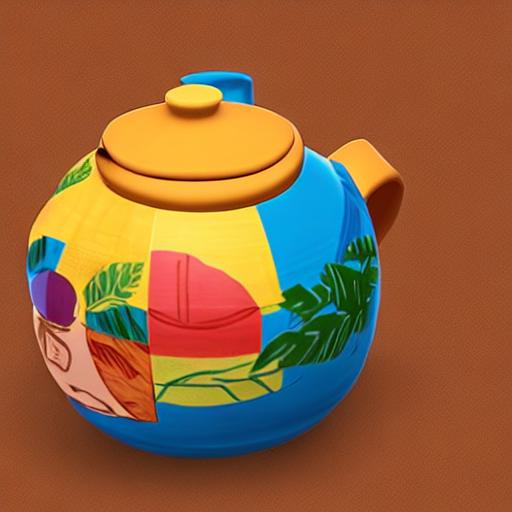} &
        \includegraphics[width=0.125\textwidth]{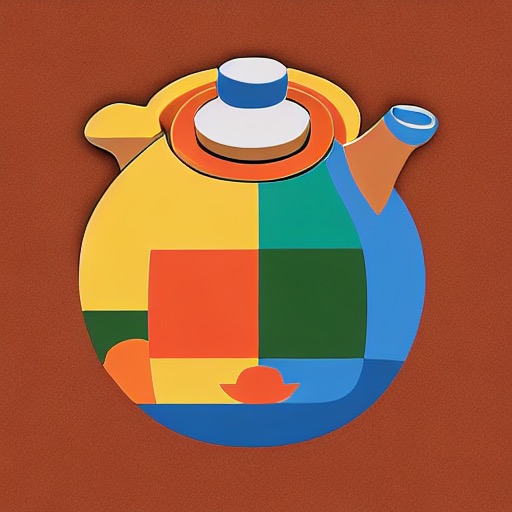} \\
        
        \includegraphics[width=0.125\textwidth]{images/original/teddybear.jpg} &
        \raisebox{0.315in}{\begin{tabular}{c} ``Colorful \\ graffiti \\ of  $S_*$'' \end{tabular}} &
        \includegraphics[width=0.125\textwidth]{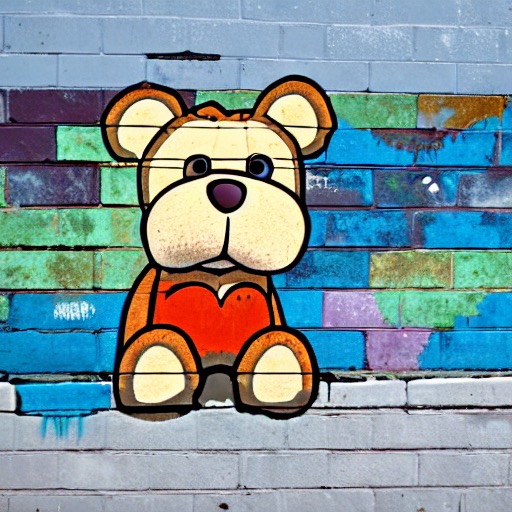} &
        \includegraphics[width=0.125\textwidth]{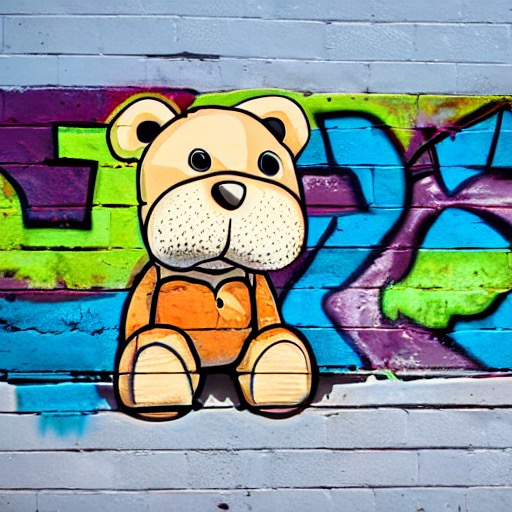} &
        \includegraphics[width=0.125\textwidth]{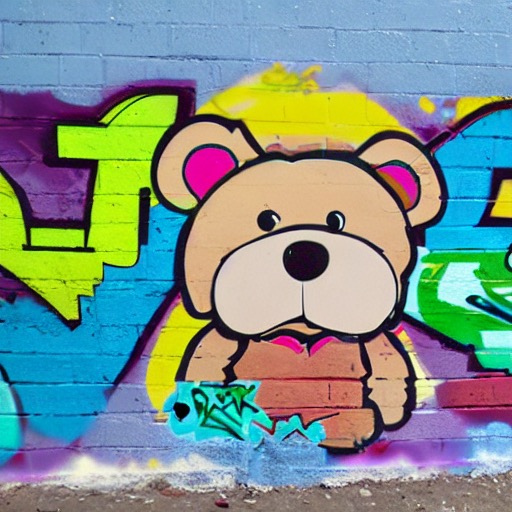} &
        \includegraphics[width=0.125\textwidth]{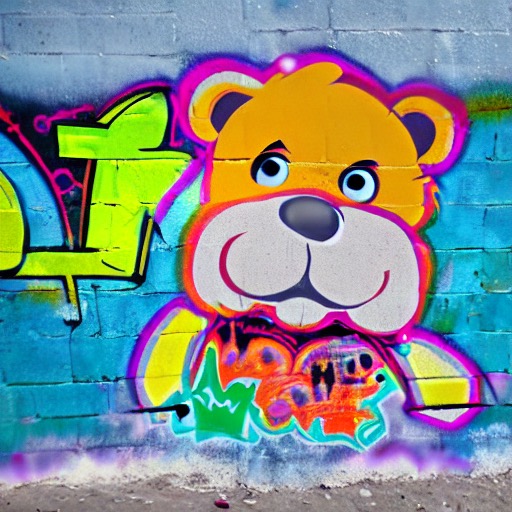} \\

        \includegraphics[width=0.125\textwidth]{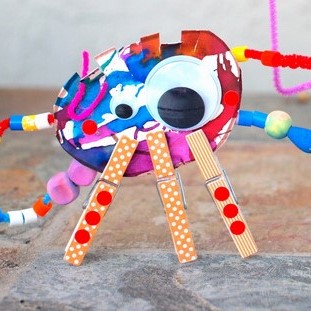} &
        \raisebox{0.315in}{\begin{tabular}{c}``A rabbit $S_*$'' \end{tabular}} &
        \includegraphics[width=0.125\textwidth]{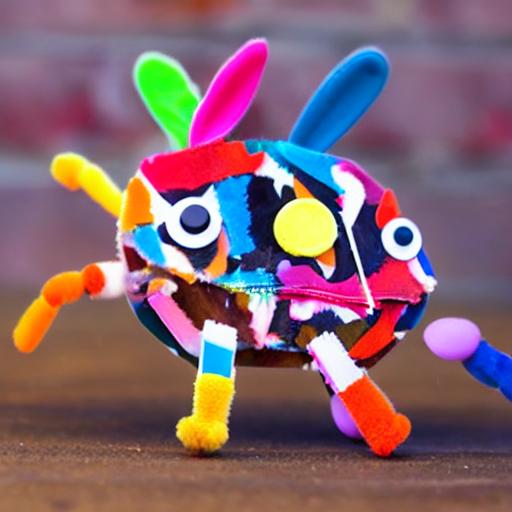} &
        \includegraphics[width=0.125\textwidth]{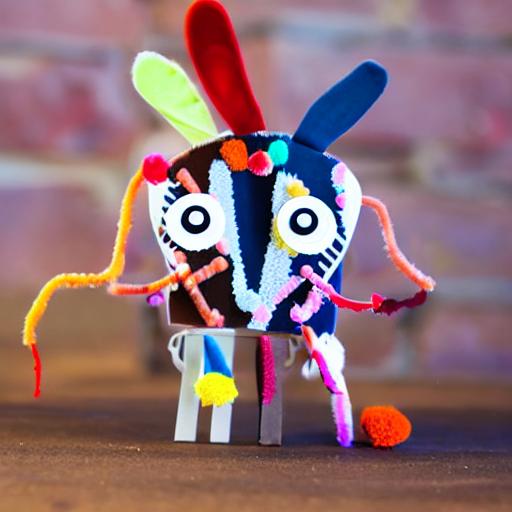} &
        \includegraphics[width=0.125\textwidth]{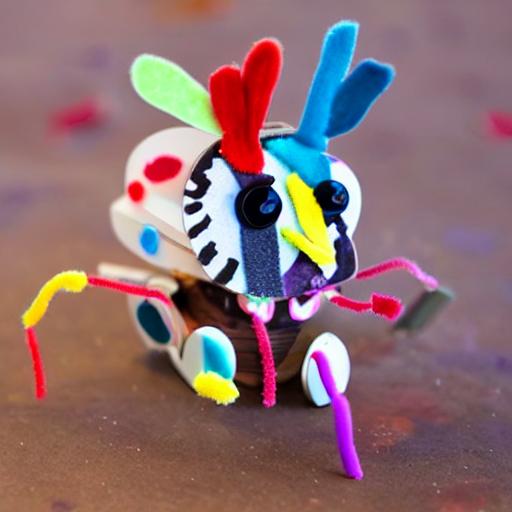} &
        \includegraphics[width=0.125\textwidth]{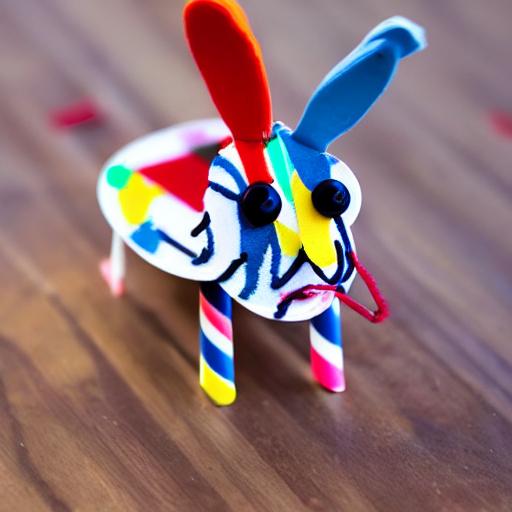} \\

        \includegraphics[width=0.125\textwidth]{images/original/dangling_child.jpg} &
        \raisebox{0.315in}{\begin{tabular}{c}``A painting of $S_*$ \\ as a Samurai \\ holding a katana'' \end{tabular}} &
        \includegraphics[width=0.125\textwidth]{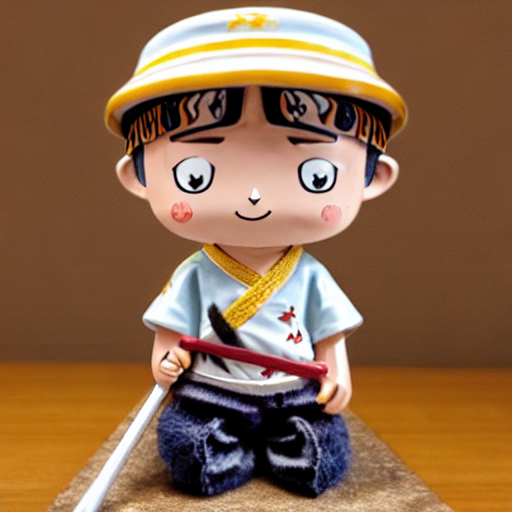} &
        \includegraphics[width=0.125\textwidth]{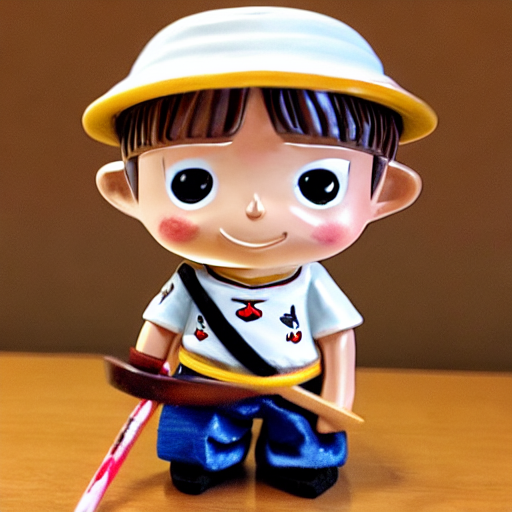} &
        \includegraphics[width=0.125\textwidth]{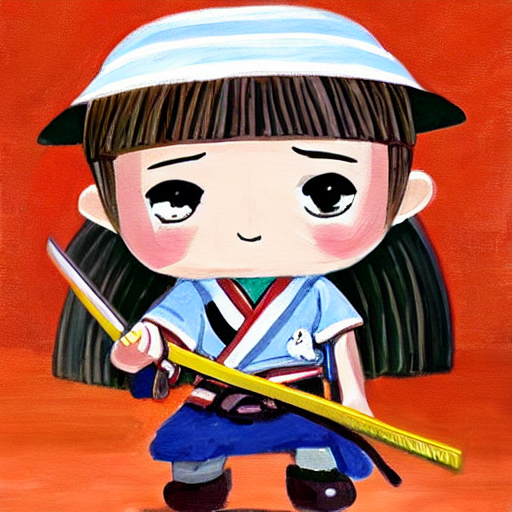} &
        \includegraphics[width=0.125\textwidth]{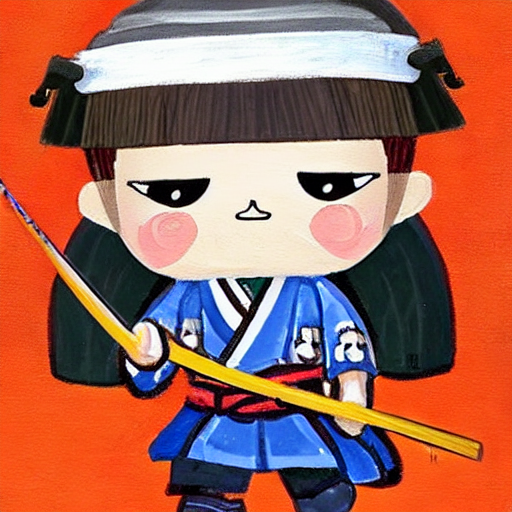} \\

        \includegraphics[width=0.125\textwidth]{images/original/metal_bird.jpg} &
        \raisebox{0.315in}{\begin{tabular}{c}``An oil painting \\ of $S_*$'' \end{tabular}} &
        \includegraphics[width=0.125\textwidth]{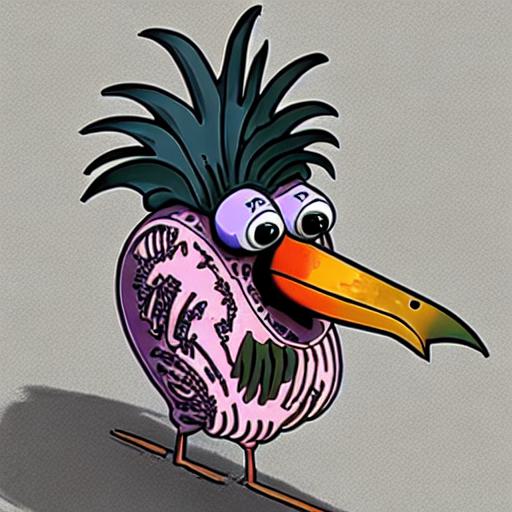} &
        \includegraphics[width=0.125\textwidth]{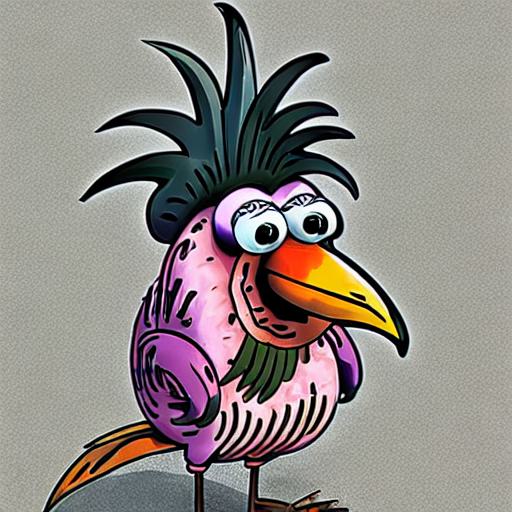} &
        \includegraphics[width=0.125\textwidth]{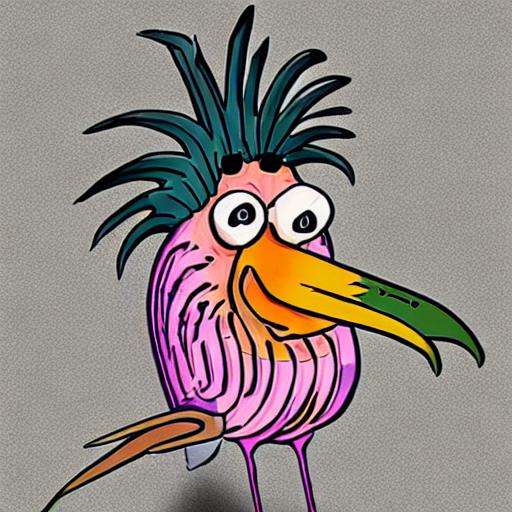} &
        \includegraphics[width=0.125\textwidth]{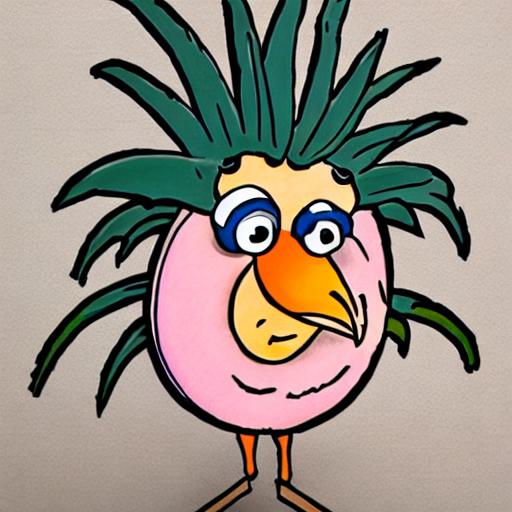} \\

        \includegraphics[width=0.125\textwidth]{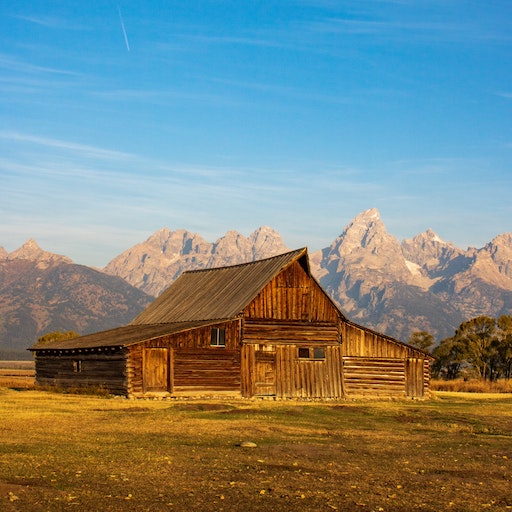} &
        \raisebox{0.315in}{\begin{tabular}{c}``A photo of \\ a $S_*$'' \end{tabular}} &
        \includegraphics[width=0.125\textwidth]{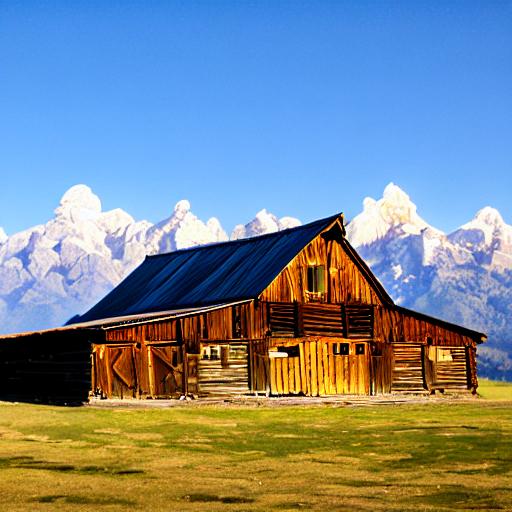} &
        \includegraphics[width=0.125\textwidth]{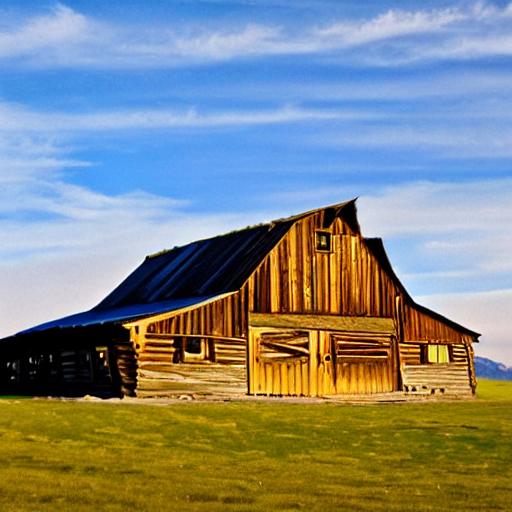} &
        \includegraphics[width=0.125\textwidth]{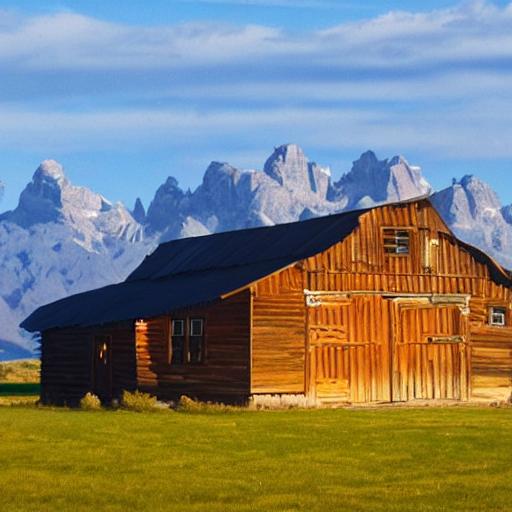} &
        \includegraphics[width=0.125\textwidth]{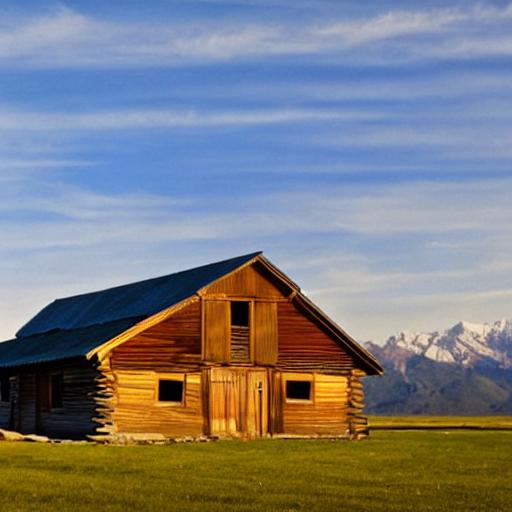} \\

        \includegraphics[width=0.125\textwidth]{images/original/mugs_skulls.jpeg} &
        \raisebox{0.315in}{\begin{tabular}{c}``A photo of \\ a $S_*$'' \end{tabular}} &
        \includegraphics[width=0.125\textwidth]{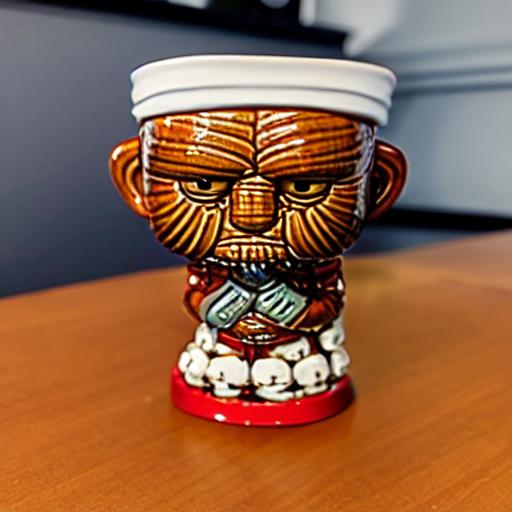} &
        \includegraphics[width=0.125\textwidth]{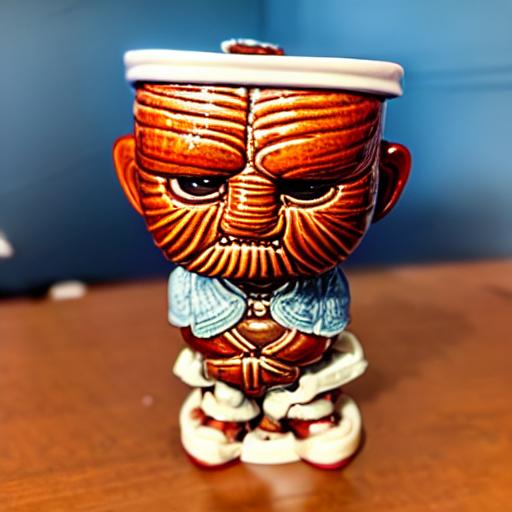} &
        \includegraphics[width=0.125\textwidth]{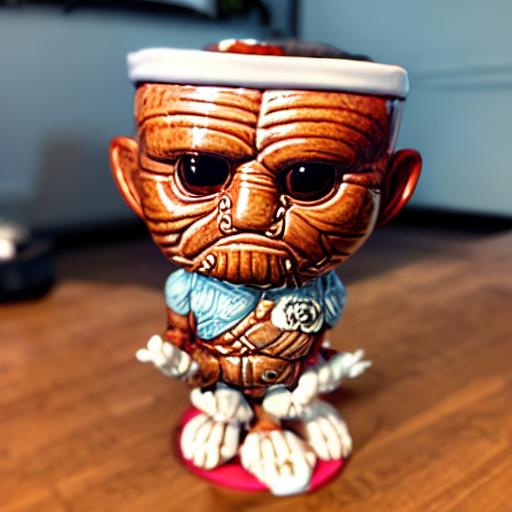} &
        \includegraphics[width=0.125\textwidth]{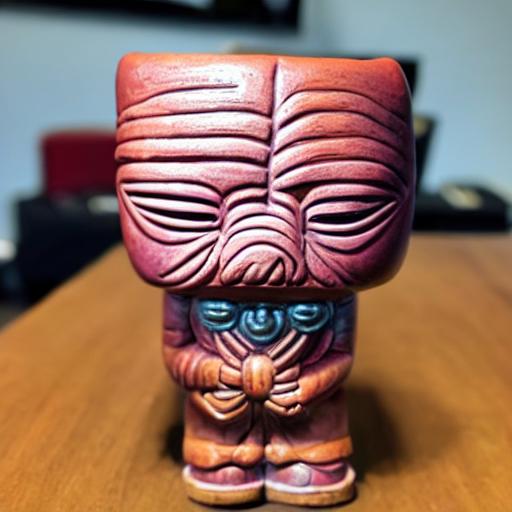} \\

        \includegraphics[width=0.125\textwidth]{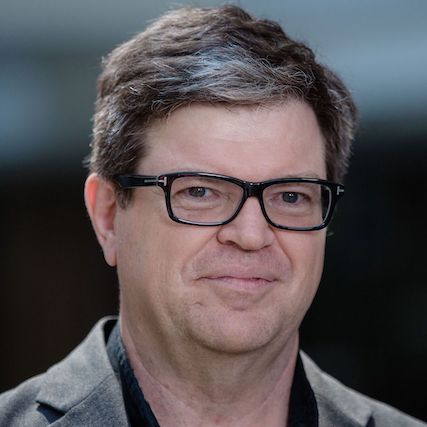} &
        \raisebox{0.315in}{\begin{tabular}{c}``A photo of \\ a $S_*$'' \end{tabular}} &
        \includegraphics[width=0.125\textwidth]{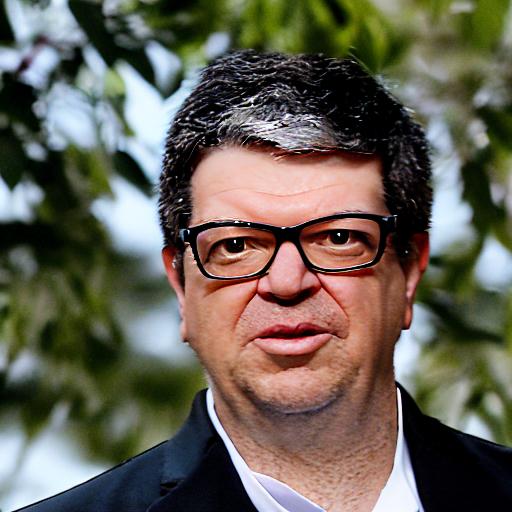} &
        \includegraphics[width=0.125\textwidth]{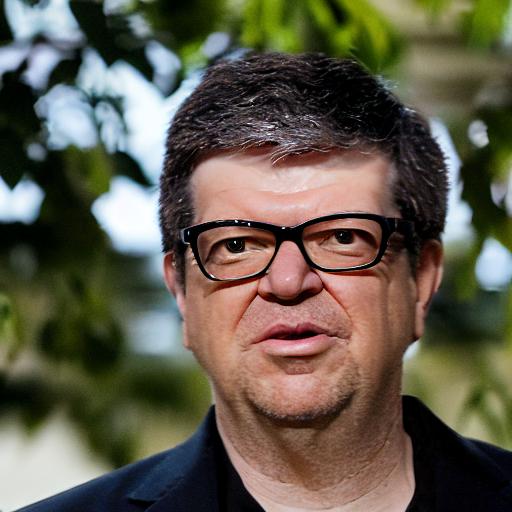} &
        \includegraphics[width=0.125\textwidth]{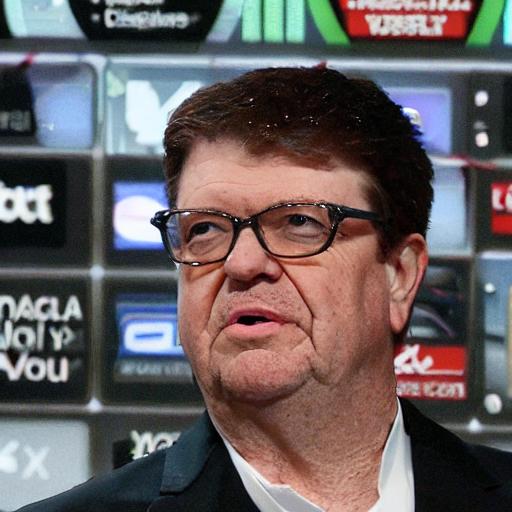} &
        \includegraphics[width=0.125\textwidth]{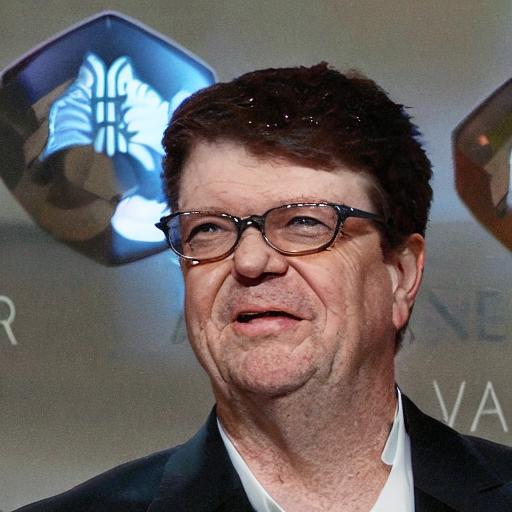} \\

        Real & & \multicolumn{4}{c}{\raisebox{0.1cm}{\limitarrowsup{}}}

    \\[-0.2cm]        
    \end{tabular}
    \caption{Controlling the editability with Nested Dropout. The three bottom examples showcase how dropping out layers affects the reconstruction when no editing is applied. This illustrates how the different vectors in our ordered representation capture different aspects of the concept in an ordered fashion.
    In the second column, we can see that applying dropout is also helpful when some parts of the prompts are not adhered to (in this case ``a painting'').}
    \label{fig:nested_dropout_supplementary}
\end{figure*}

\begin{figure*}
    \centering
    \renewcommand{\arraystretch}{0.3}
    \setlength{\tabcolsep}{0.5pt}

    {\footnotesize
    \begin{tabular}{c@{\hspace{0.2cm}} c c @{\hspace{0.2cm}} c c @{\hspace{0.2cm}} c c @{\hspace{0.2cm}} c c }

        \begin{tabular}{c} Real Sample \\ \\[-0.05cm] \& Prompt \end{tabular} &
        \multicolumn{2}{c}{Textual Inversion (TI)} &
        \multicolumn{2}{c}{DreamBooth} &
        \multicolumn{2}{c}{Extended Textual Inversion} &
        \multicolumn{2}{c}{NeTI} \\

        \includegraphics[width=0.102\textwidth]{images/original/elephant.jpg} &
        \includegraphics[width=0.102\textwidth]{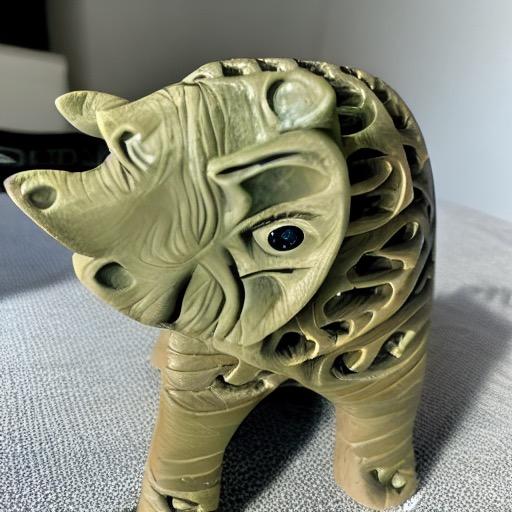} &
        \includegraphics[width=0.102\textwidth]{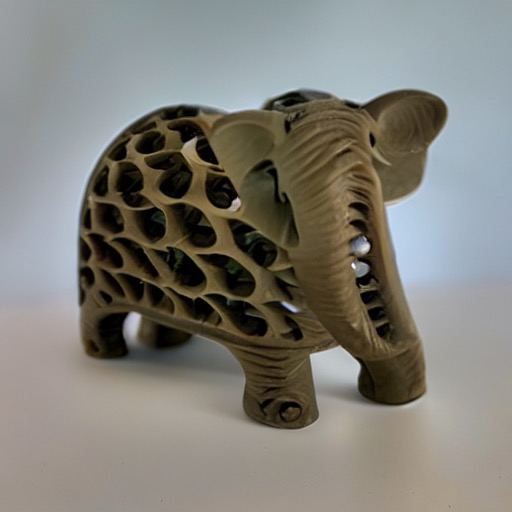} &
        \hspace{0.05cm}
        \includegraphics[width=0.102\textwidth]{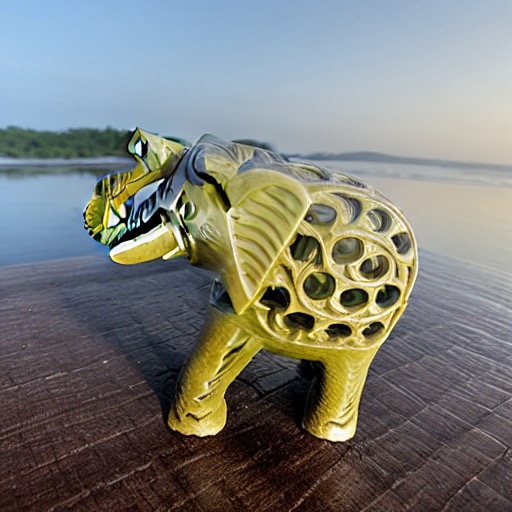} &
        \includegraphics[width=0.102\textwidth]{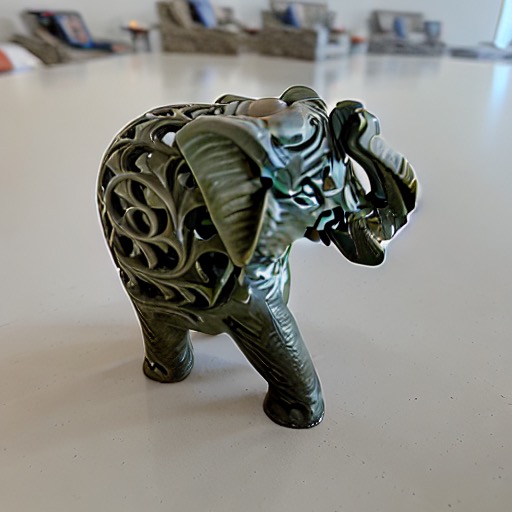} &
        \hspace{0.05cm}
        \includegraphics[width=0.102\textwidth]{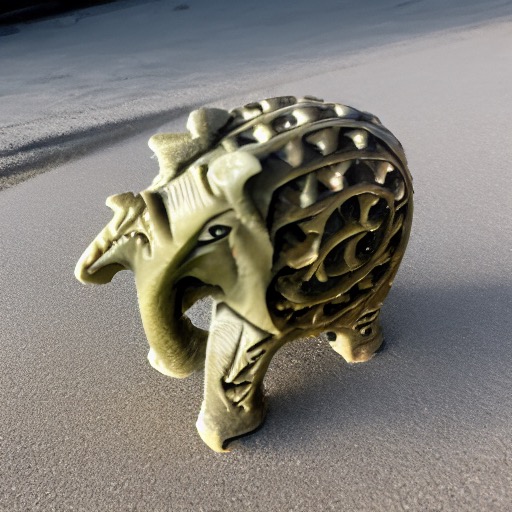} &
        \includegraphics[width=0.102\textwidth]{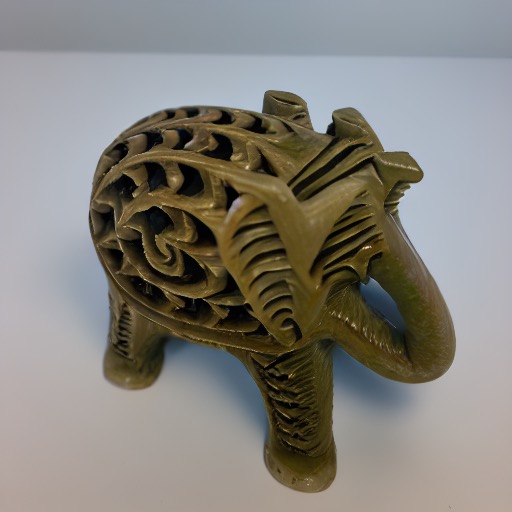} &
        \hspace{0.05cm}
        \includegraphics[width=0.102\textwidth]{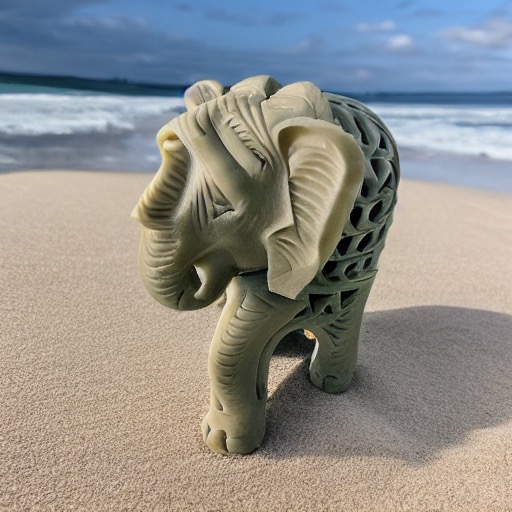} &
        \includegraphics[width=0.102\textwidth]{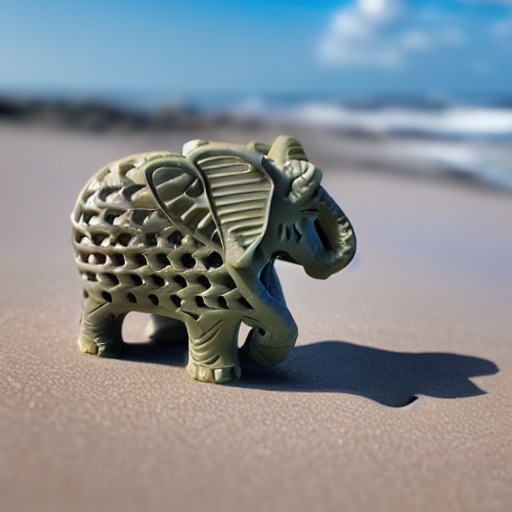} \\
        
        \raisebox{0.325in}{\begin{tabular}{c} ``A photo of $S_*$ \\ \\[-0.05cm] on a beach''\end{tabular}} &
        \includegraphics[width=0.102\textwidth]{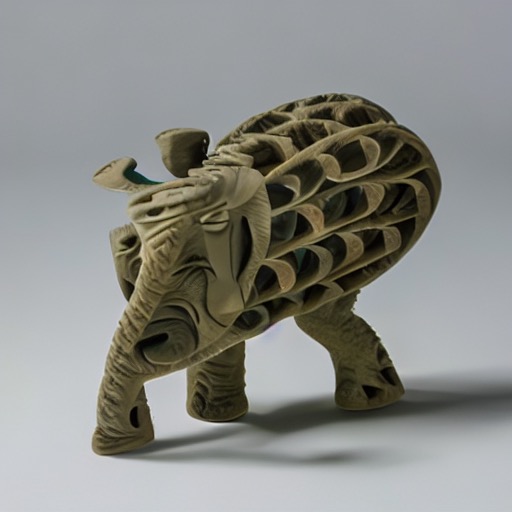} &
        \includegraphics[width=0.102\textwidth]{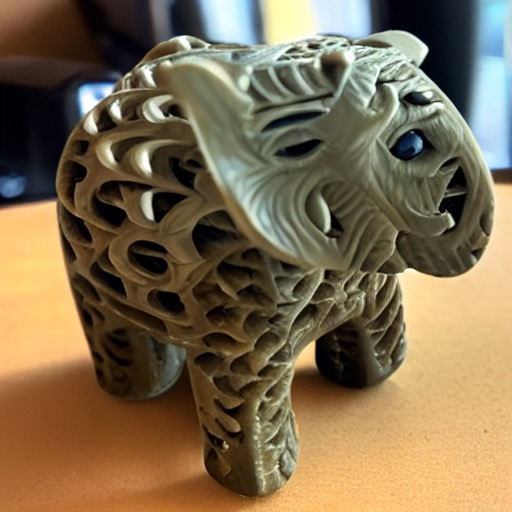} &
        \hspace{0.05cm}
        \includegraphics[width=0.102\textwidth]{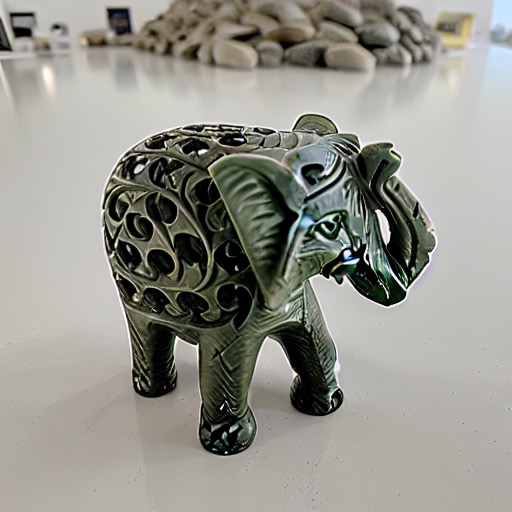} &
        \includegraphics[width=0.102\textwidth]{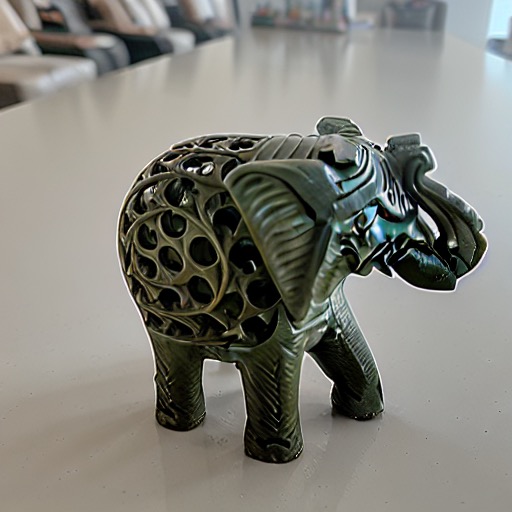} &
        \hspace{0.05cm}
        \includegraphics[width=0.102\textwidth]{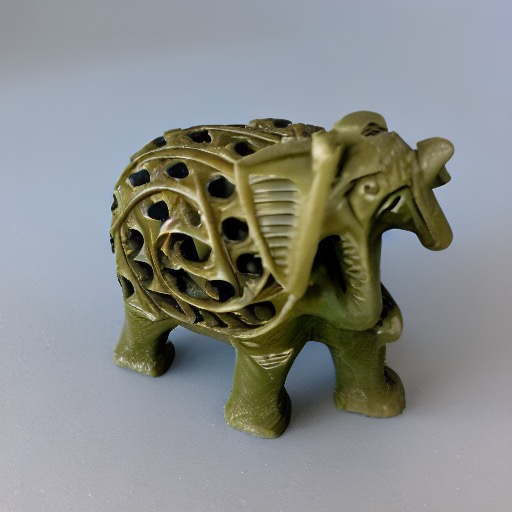} &
        \includegraphics[width=0.102\textwidth]{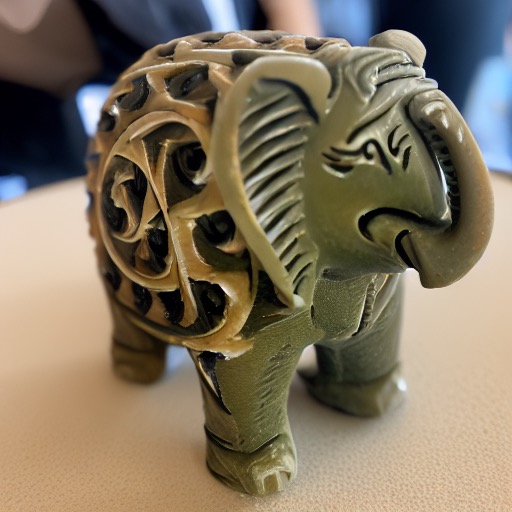} &
        \hspace{0.05cm}
        \includegraphics[width=0.102\textwidth]{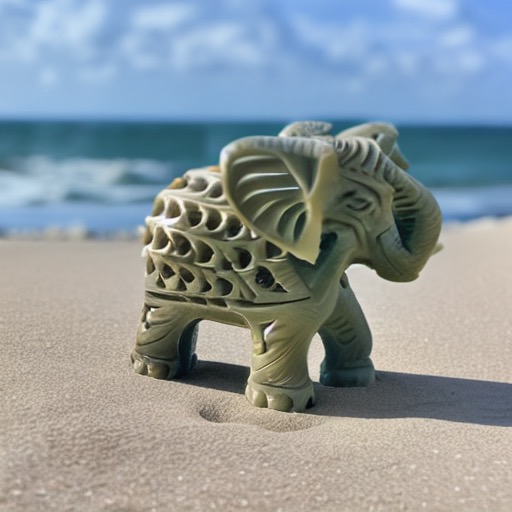} &
        \includegraphics[width=0.102\textwidth]{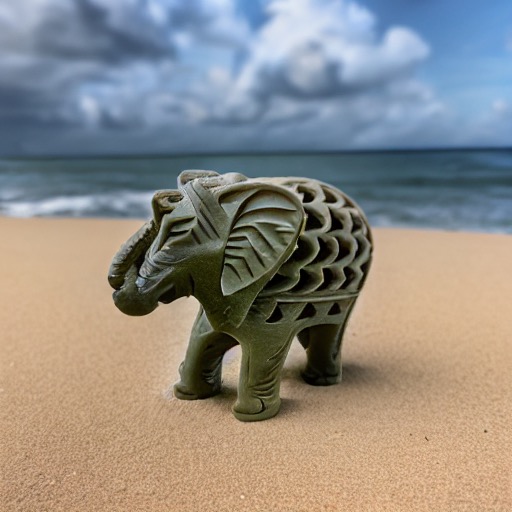} \\ \\

        \includegraphics[width=0.102\textwidth]{images/original/metal_bird.jpg} &
        \includegraphics[width=0.102\textwidth]{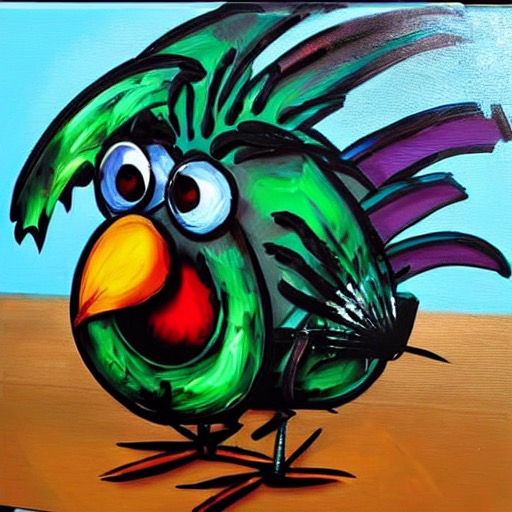} &
        \includegraphics[width=0.102\textwidth]{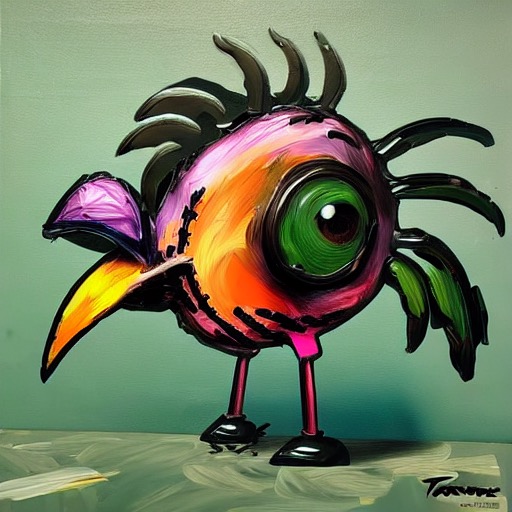} &
        \hspace{0.05cm}
        \includegraphics[width=0.102\textwidth]{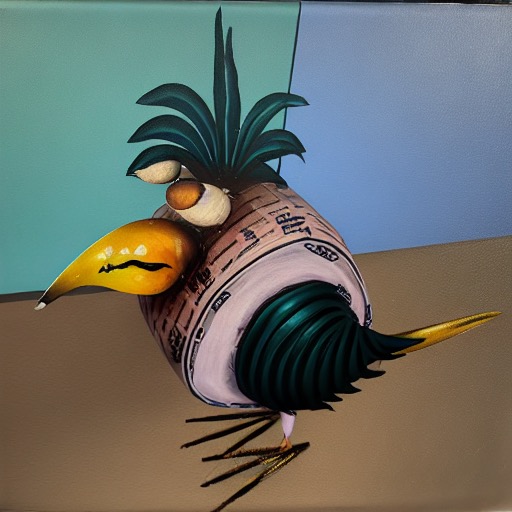} &
        \includegraphics[width=0.102\textwidth]{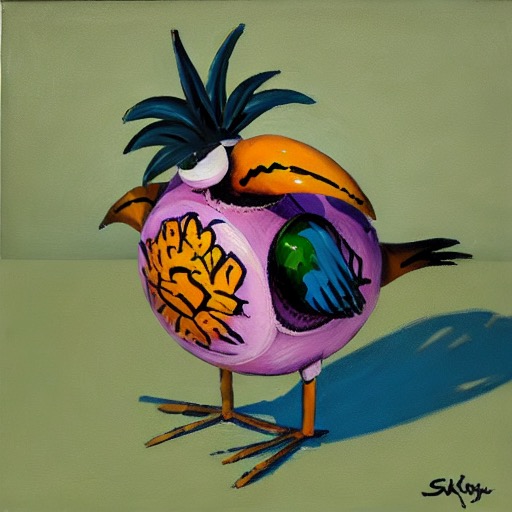} &
        \hspace{0.05cm}
        \includegraphics[width=0.102\textwidth]{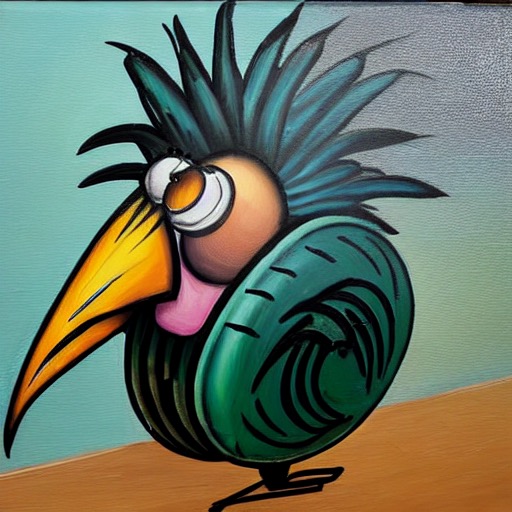} &
        \includegraphics[width=0.102\textwidth]{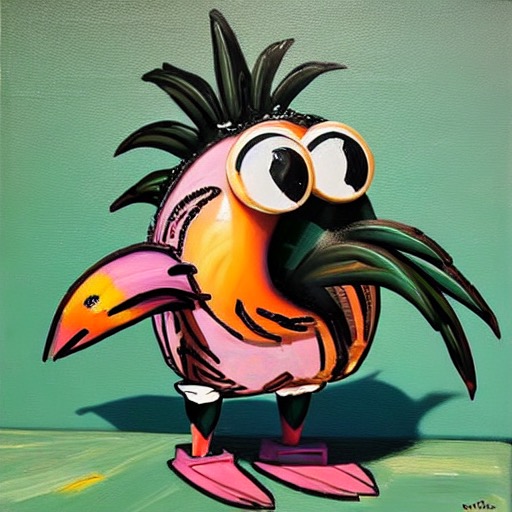} &
        \hspace{0.05cm}
        \includegraphics[width=0.102\textwidth]{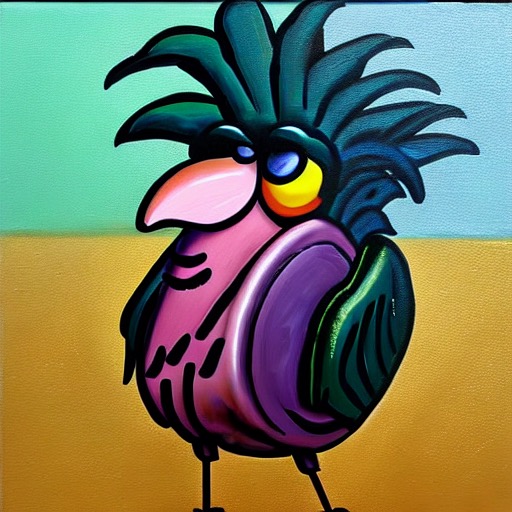} &
        \includegraphics[width=0.102\textwidth]{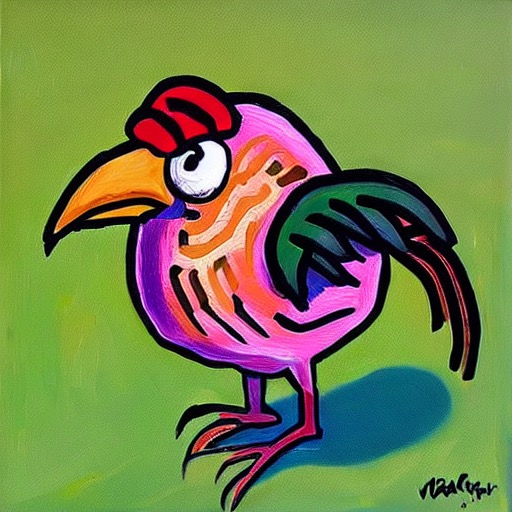} \\
        
        \raisebox{0.325in}{\begin{tabular}{c} ``An oil \\ \\[-0.05cm] painting \\ \\[-0.05cm] of $S_*$''\end{tabular}} &
        \includegraphics[width=0.102\textwidth]{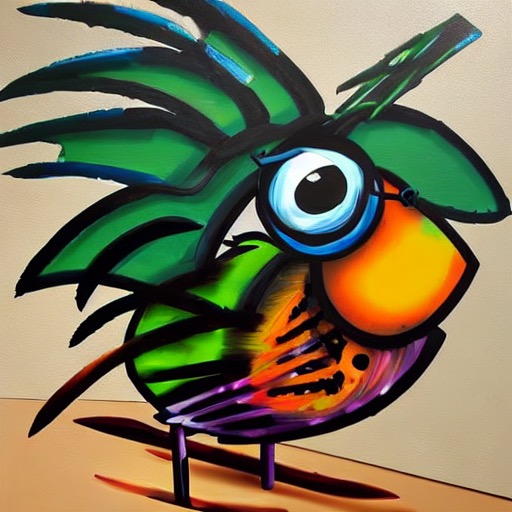} &
        \includegraphics[width=0.102\textwidth]{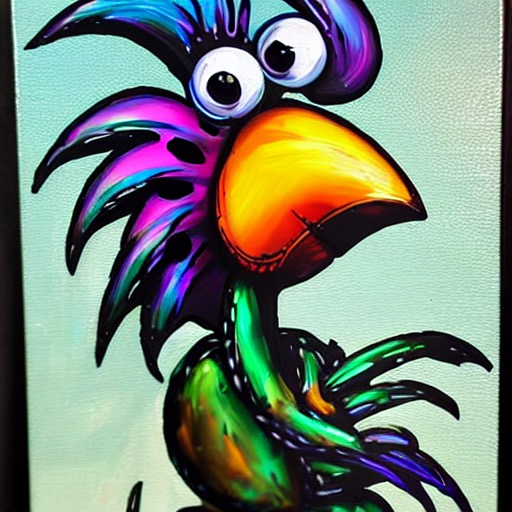} &
        \hspace{0.05cm}
        \includegraphics[width=0.102\textwidth]{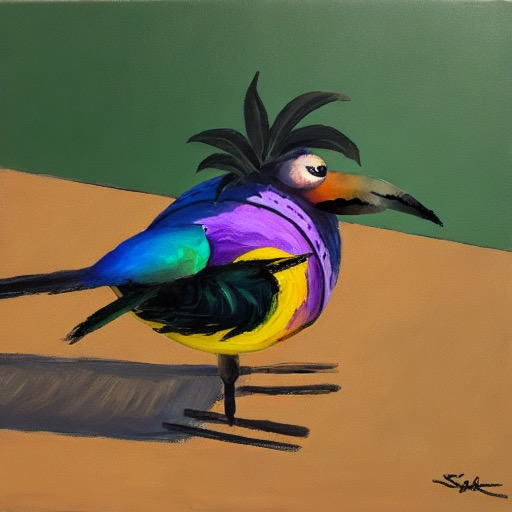} &
        \includegraphics[width=0.102\textwidth]{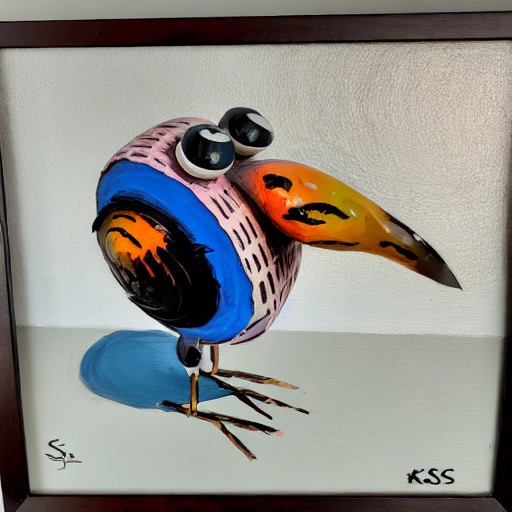} &
        \hspace{0.05cm}
        \includegraphics[width=0.102\textwidth]{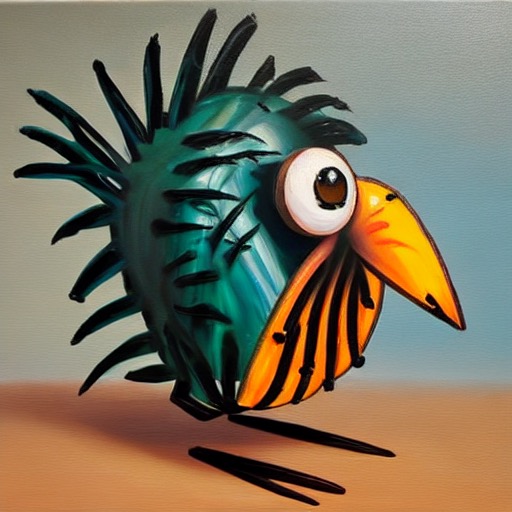} &
        \includegraphics[width=0.102\textwidth]{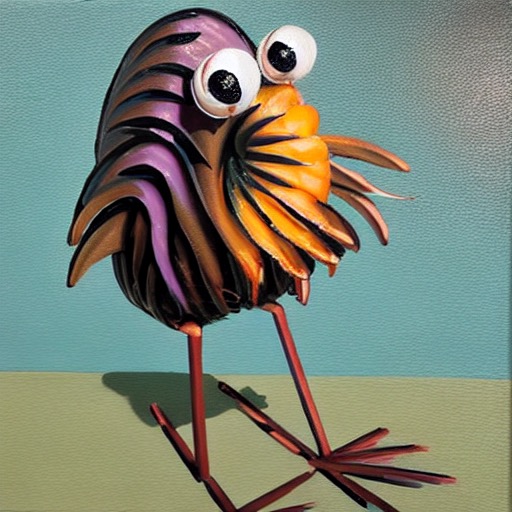} &
        \hspace{0.05cm}
        \includegraphics[width=0.102\textwidth]{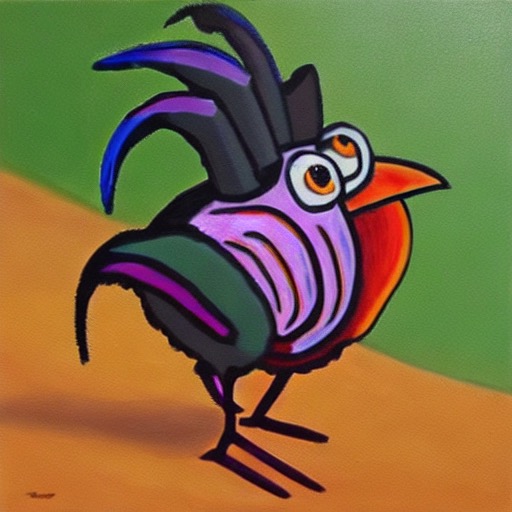} &
        \includegraphics[width=0.102\textwidth]{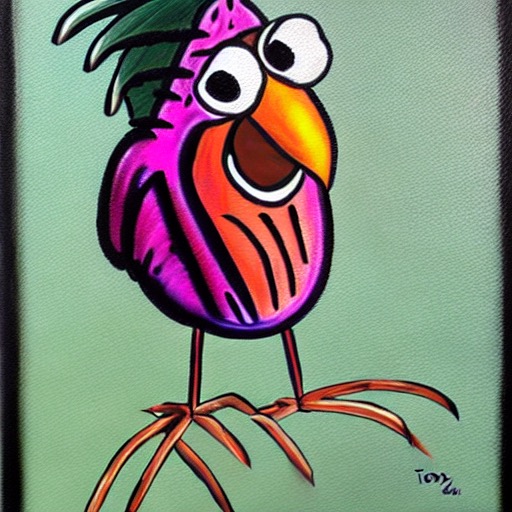} \\ \\

        \includegraphics[width=0.102\textwidth]{images/original/red_bowl.jpg} &
        \includegraphics[width=0.102\textwidth]{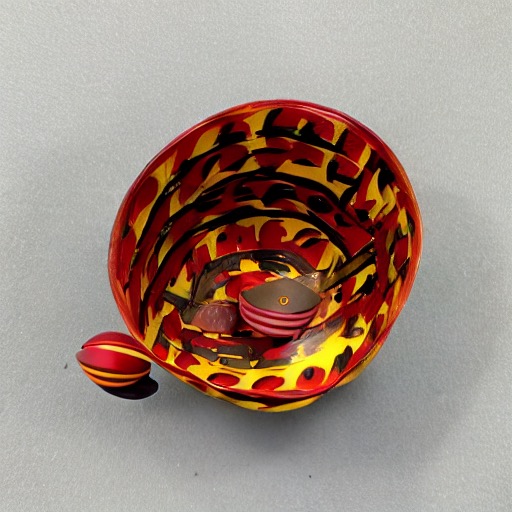} &
        \includegraphics[width=0.102\textwidth]{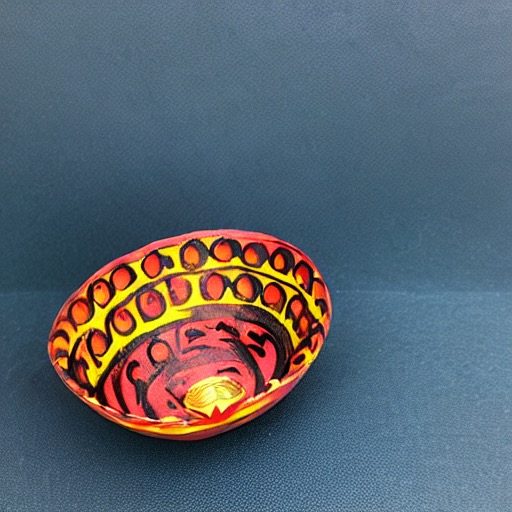} &
        \hspace{0.05cm}
        \includegraphics[width=0.102\textwidth]{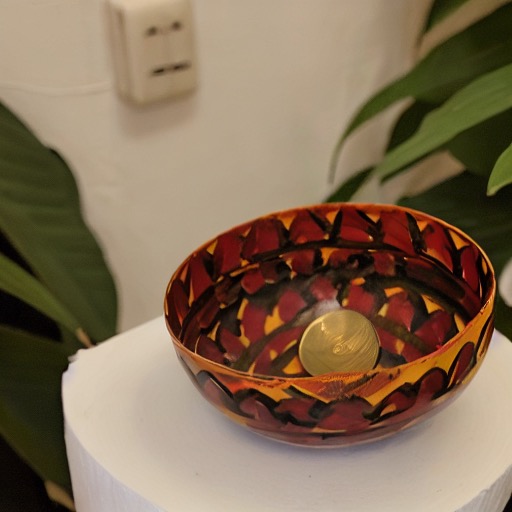} &
        \includegraphics[width=0.102\textwidth]{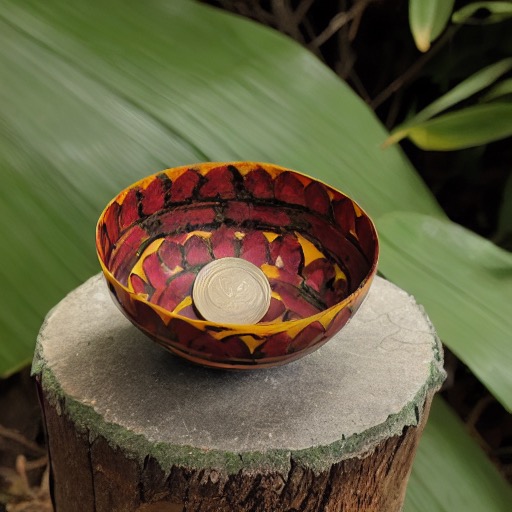} &
        \hspace{0.05cm}
        \includegraphics[width=0.102\textwidth]{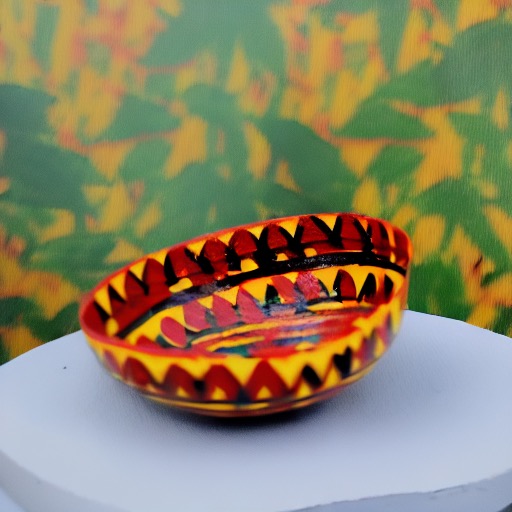} &
        \includegraphics[width=0.102\textwidth]{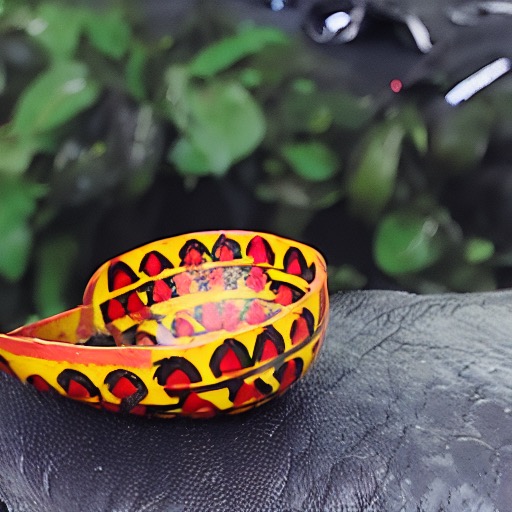} &
        \hspace{0.05cm}
        \includegraphics[width=0.102\textwidth]{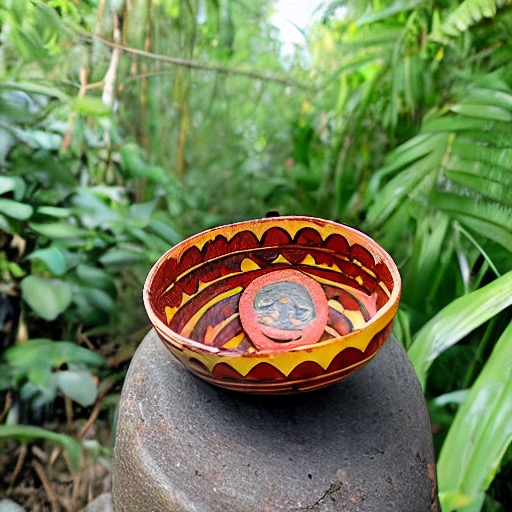} &
        \includegraphics[width=0.102\textwidth]{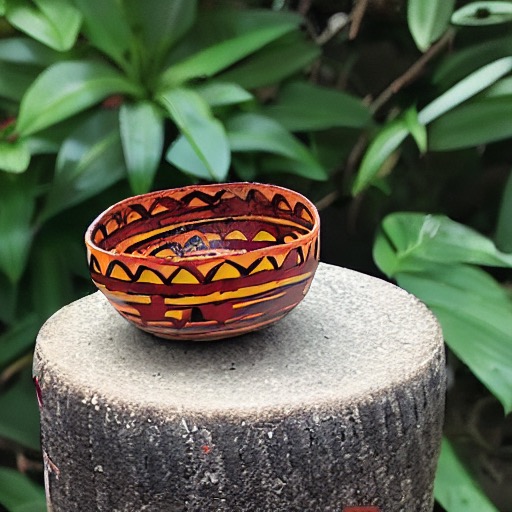} \\

        \raisebox{0.325in}{\begin{tabular}{c} ``A photo of $S_*$ \\ \\[-0.05cm] in the jungle''\end{tabular}} &
        \includegraphics[width=0.102\textwidth]{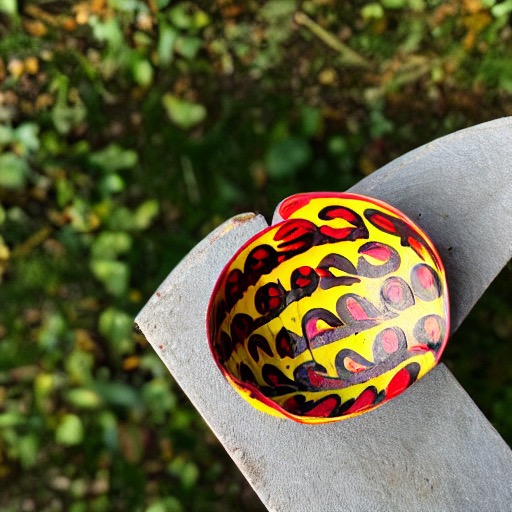} &
        \includegraphics[width=0.102\textwidth]{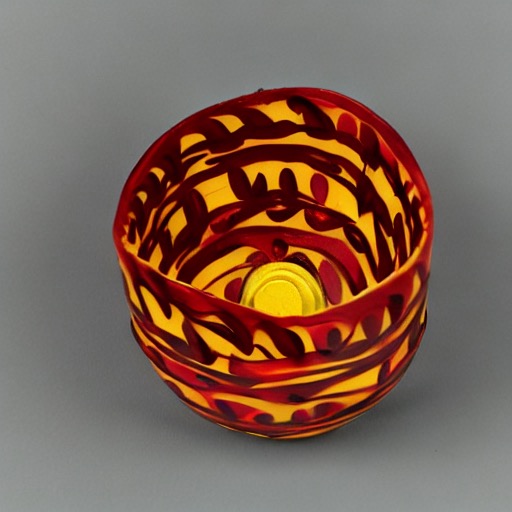} &
        \hspace{0.05cm}
        \includegraphics[width=0.102\textwidth]{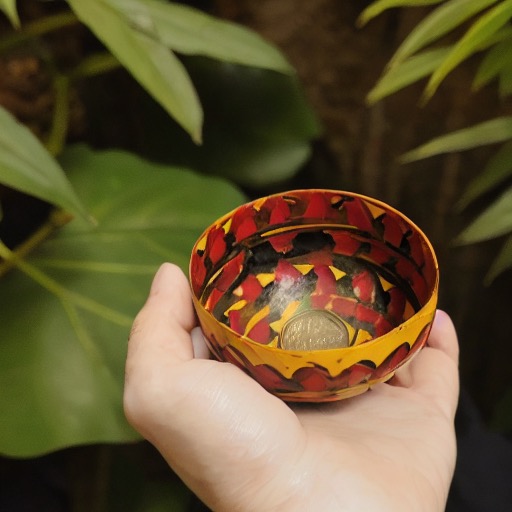} &
        \includegraphics[width=0.102\textwidth]{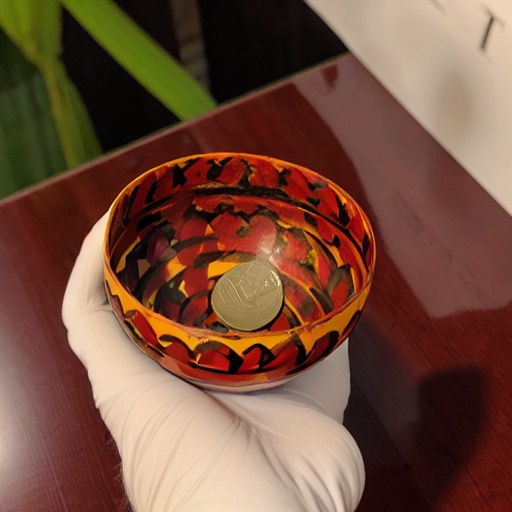} &
        \hspace{0.05cm}
        \includegraphics[width=0.102\textwidth]{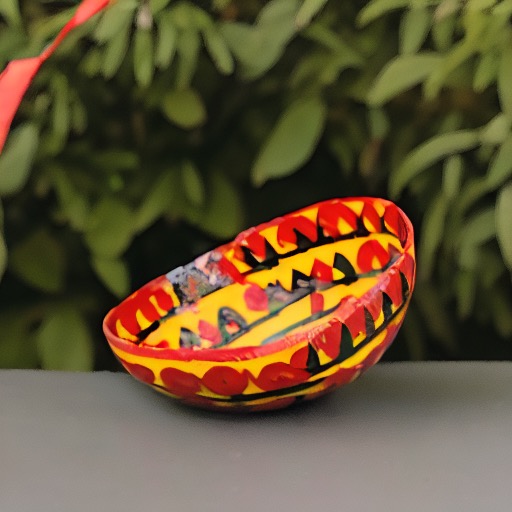} &
        \includegraphics[width=0.102\textwidth]{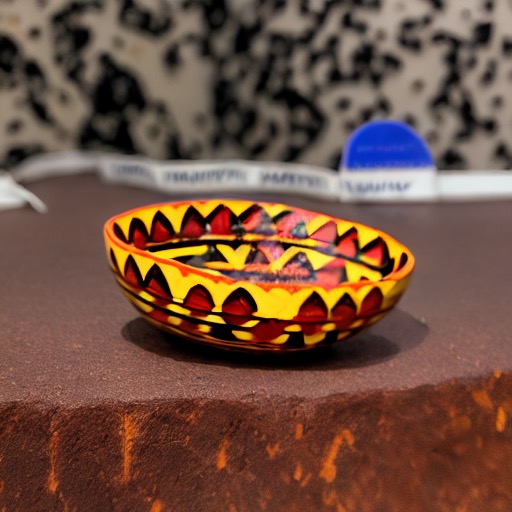} &
        \hspace{0.05cm}
        \includegraphics[width=0.102\textwidth]{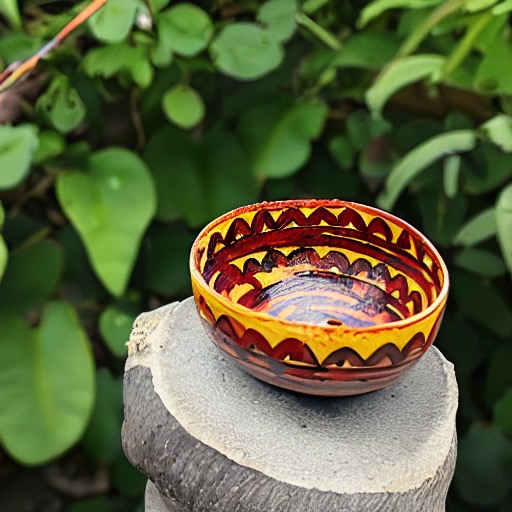} &
        \includegraphics[width=0.102\textwidth]{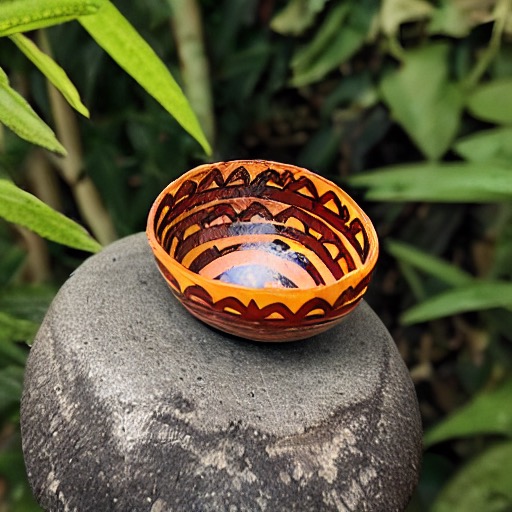} \\ \\

        \includegraphics[width=0.102\textwidth]{images/original/mugs_skulls.jpeg} &
        \includegraphics[width=0.102\textwidth]{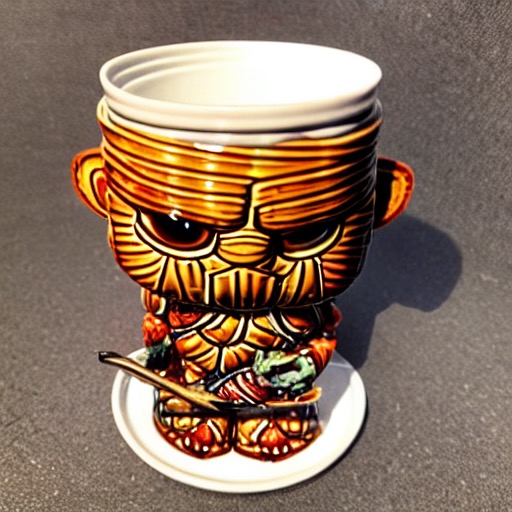} &
        \includegraphics[width=0.102\textwidth]{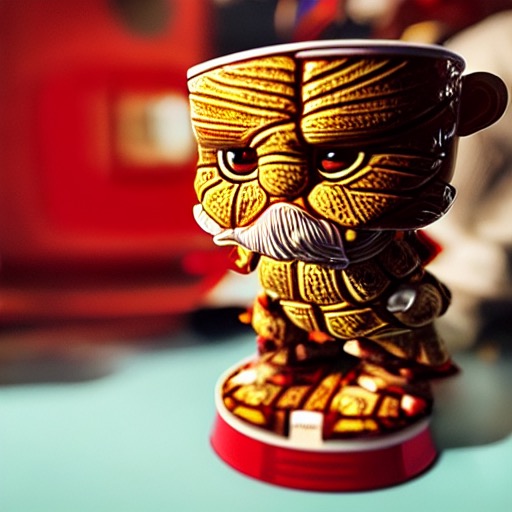} &
        \hspace{0.05cm}
        \includegraphics[width=0.102\textwidth]{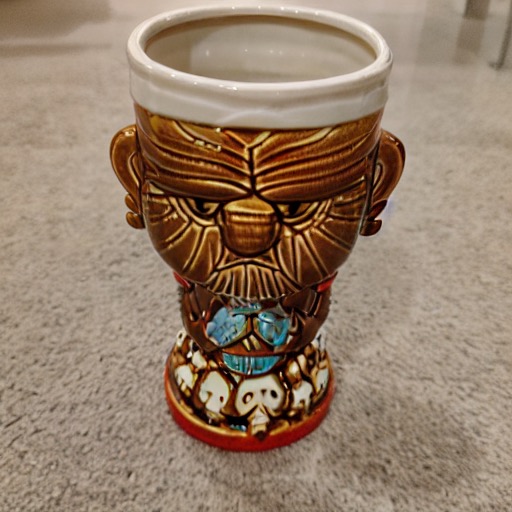} &
        \includegraphics[width=0.102\textwidth]{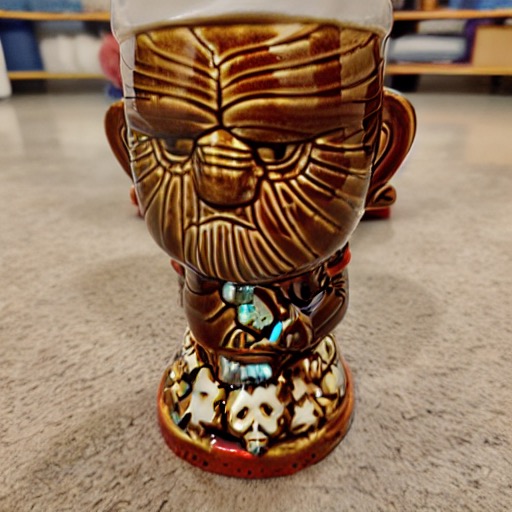} &
        \hspace{0.05cm}
        \includegraphics[width=0.102\textwidth]{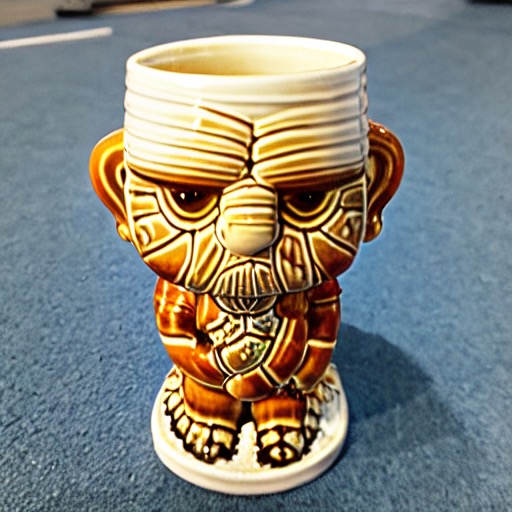} &
        \includegraphics[width=0.102\textwidth]{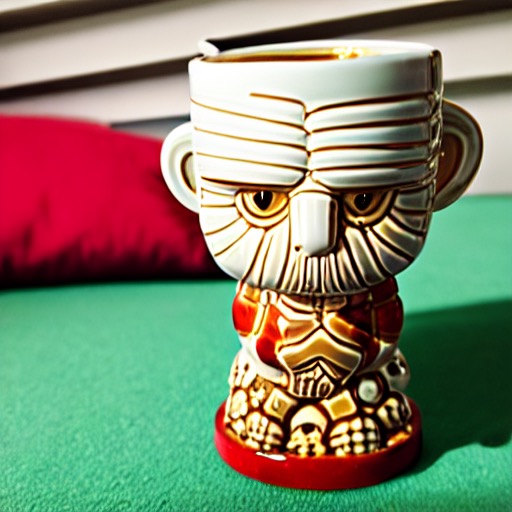} &
        \hspace{0.05cm}
        \includegraphics[width=0.102\textwidth]{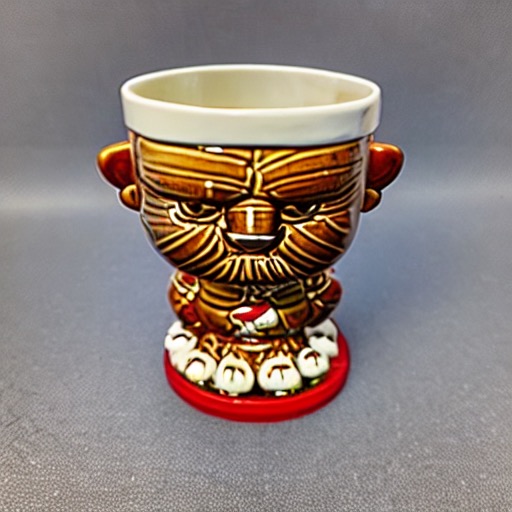} &
        \includegraphics[width=0.102\textwidth]{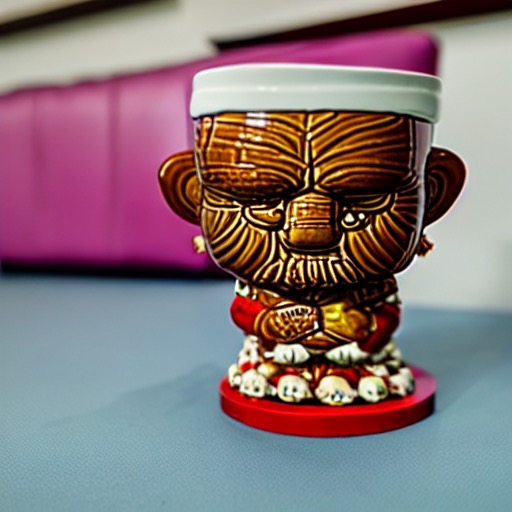} \\

        \raisebox{0.325in}{\begin{tabular}{c} ``A photo \\ \\[-0.05cm] of $S_*$''\end{tabular}} &
        \includegraphics[width=0.102\textwidth]{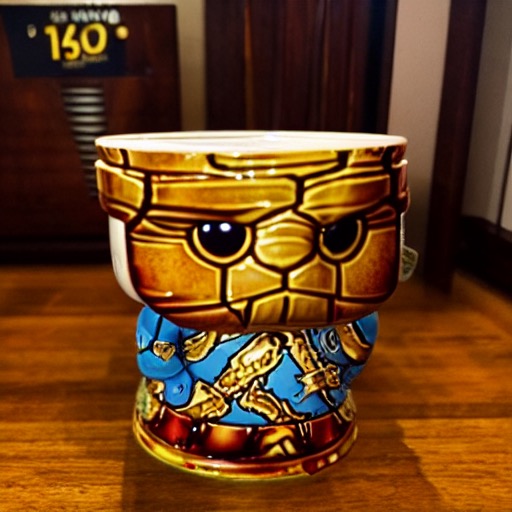} &
        \includegraphics[width=0.102\textwidth]{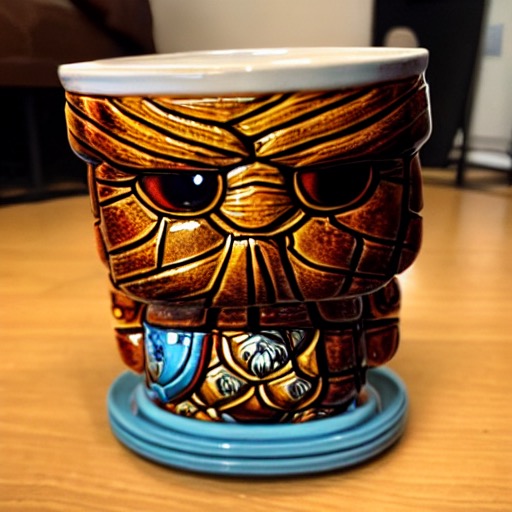} &
        \hspace{0.05cm}
        \includegraphics[width=0.102\textwidth]{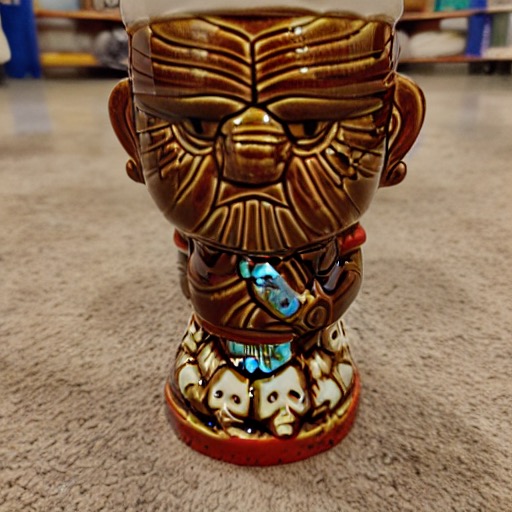} &
        \includegraphics[width=0.102\textwidth]{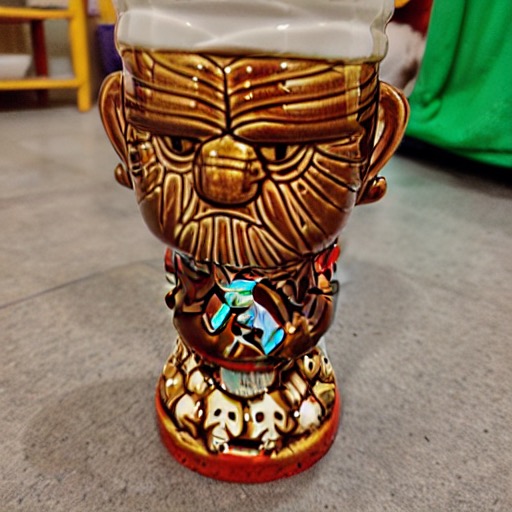} &
        \hspace{0.05cm}
        \includegraphics[width=0.102\textwidth]{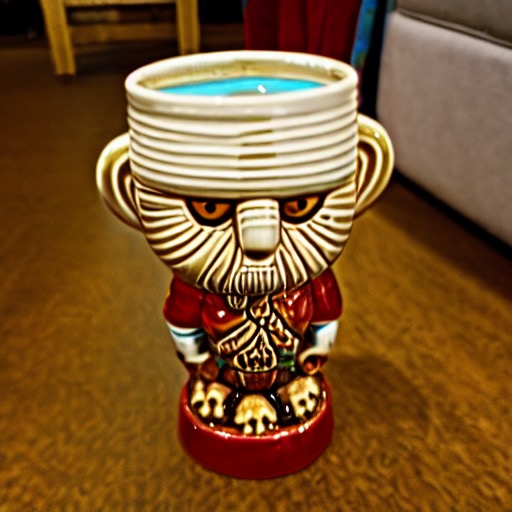} &
        \includegraphics[width=0.102\textwidth]{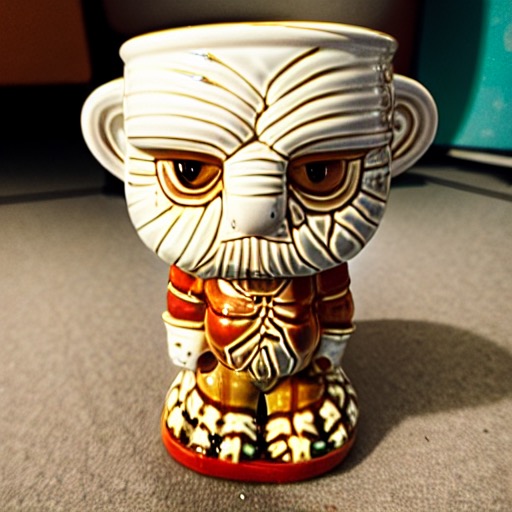} &
        \hspace{0.05cm}
        \includegraphics[width=0.102\textwidth]{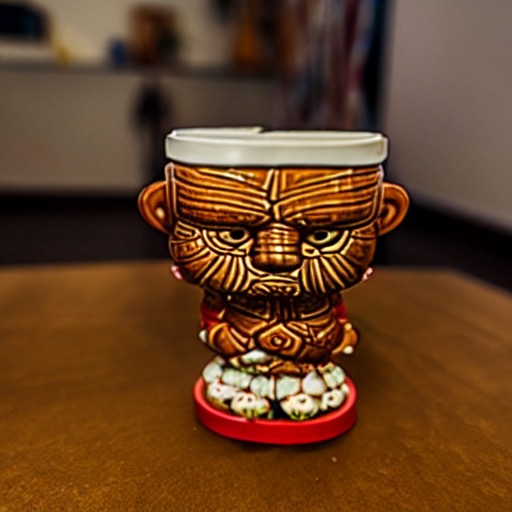} &
        \includegraphics[width=0.102\textwidth]{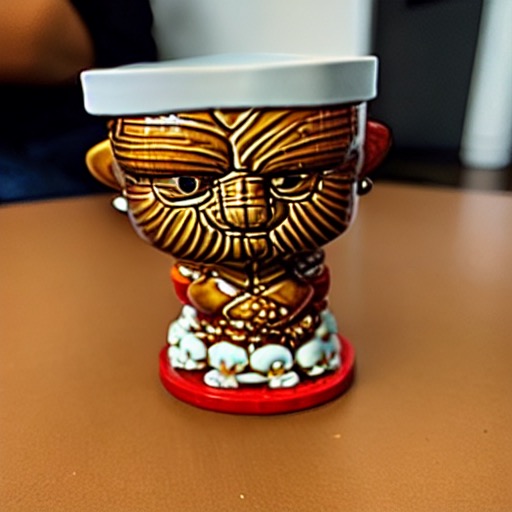} \\ \\

        \includegraphics[width=0.102\textwidth]{images/original/dangling_child.jpg} &
        \includegraphics[width=0.102\textwidth]{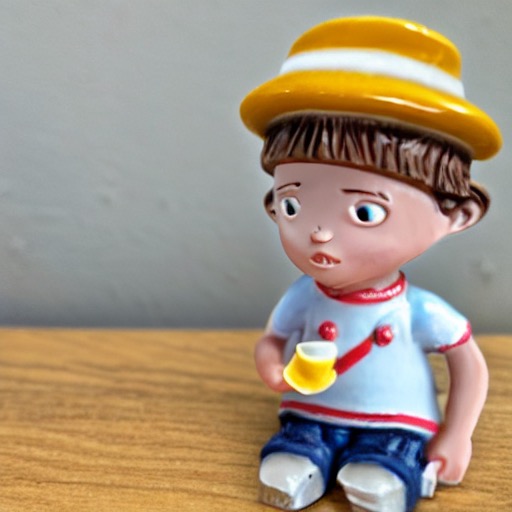} &
        \includegraphics[width=0.102\textwidth]{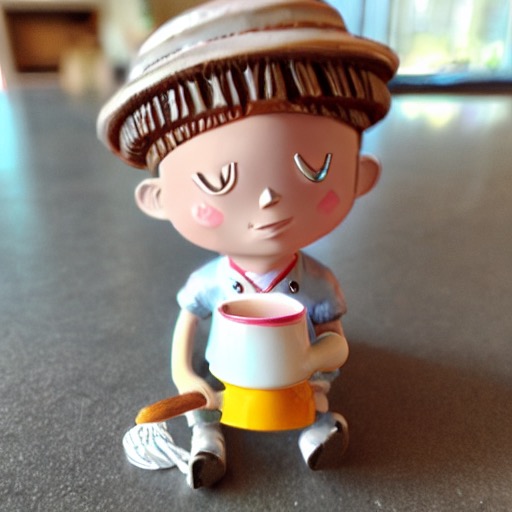} &
        \hspace{0.05cm}
        \includegraphics[width=0.102\textwidth]{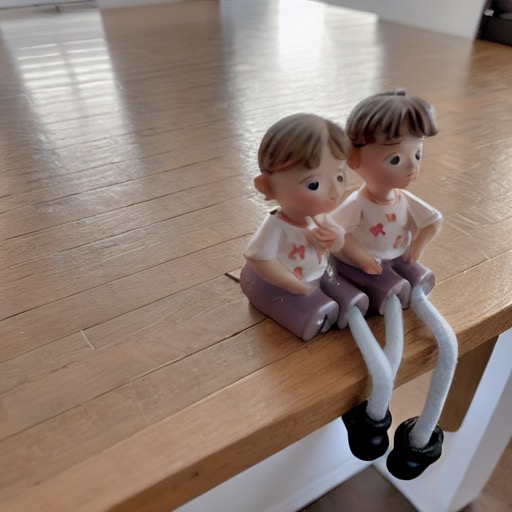} &
        \includegraphics[width=0.102\textwidth]{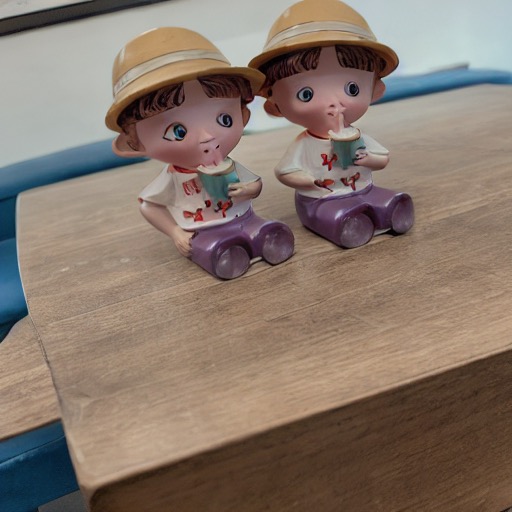} &
        \hspace{0.05cm}
        \includegraphics[width=0.102\textwidth]{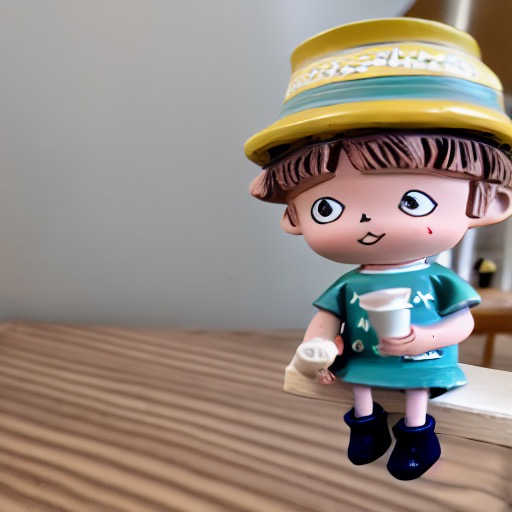} &
        \includegraphics[width=0.102\textwidth]{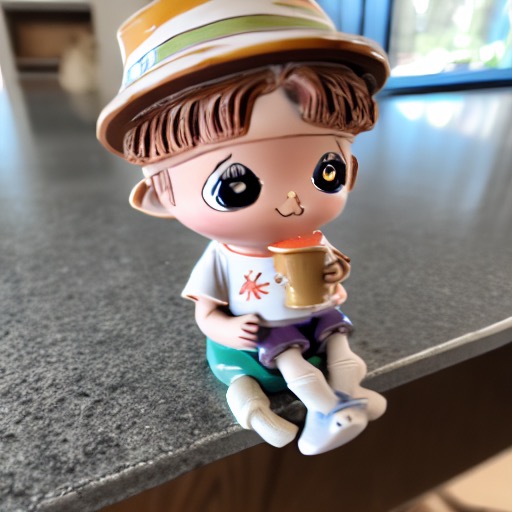} &
        \hspace{0.05cm}
        \includegraphics[width=0.102\textwidth]{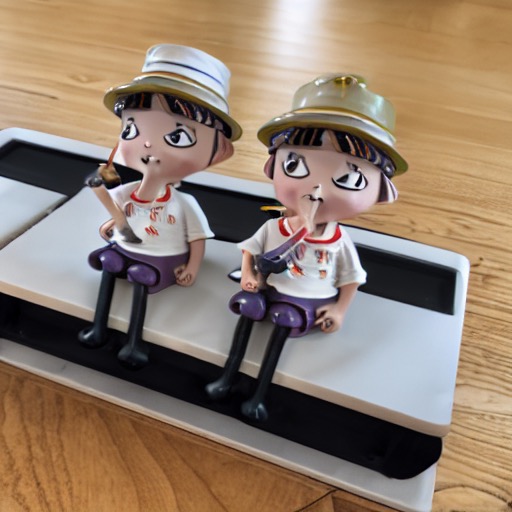} &
        \includegraphics[width=0.102\textwidth]{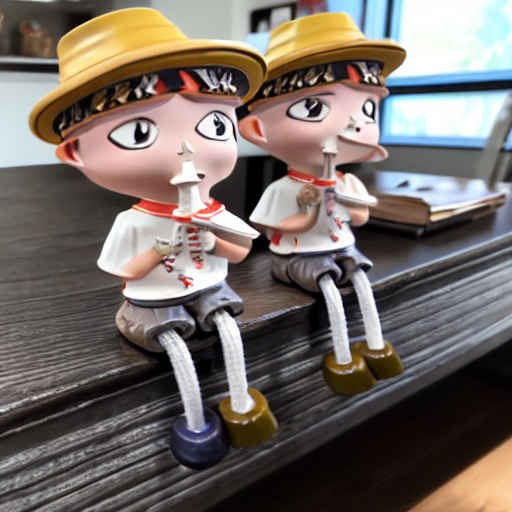} \\

        \raisebox{0.325in}{\begin{tabular}{c} ``A photograph \\ \\[-0.05cm] of two $S_*$ on \\ \\[-0.05cm] a table''\end{tabular}} &
        \includegraphics[width=0.102\textwidth]{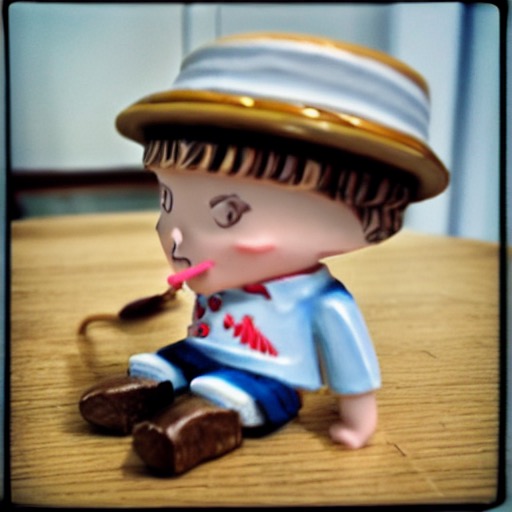} &
        \includegraphics[width=0.102\textwidth]{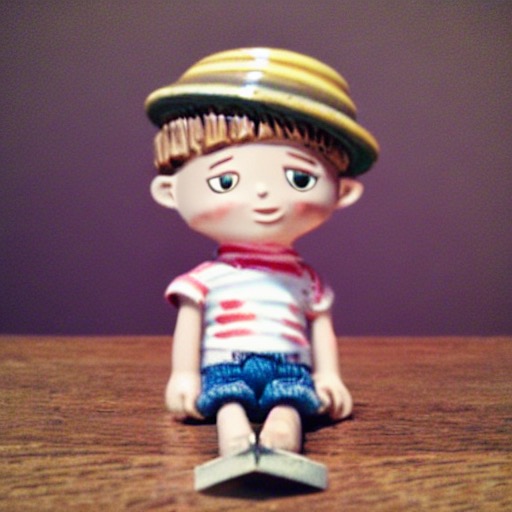} &
        \hspace{0.05cm}
        \includegraphics[width=0.102\textwidth]{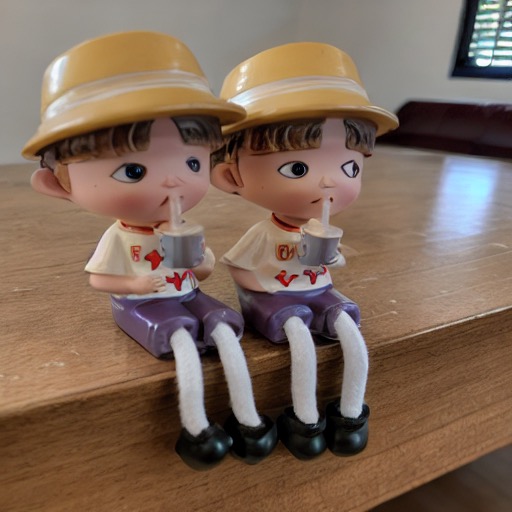} &
        \includegraphics[width=0.102\textwidth]{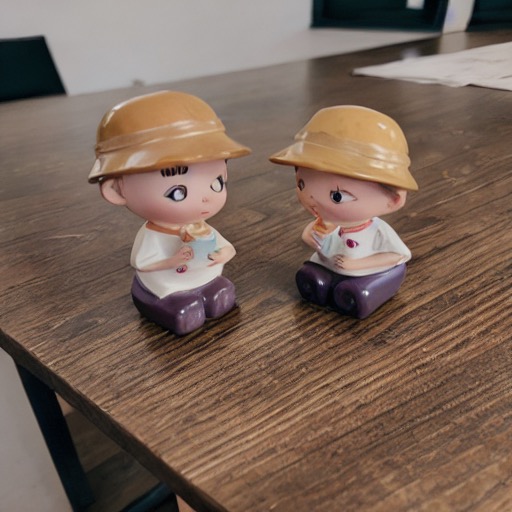} &
        \hspace{0.05cm}
        \includegraphics[width=0.102\textwidth]{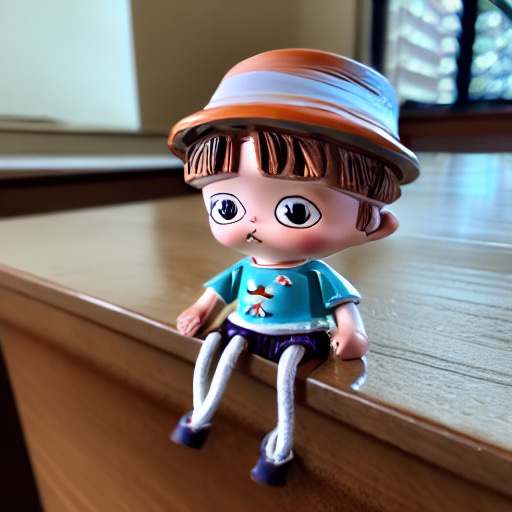} &
        \includegraphics[width=0.102\textwidth]{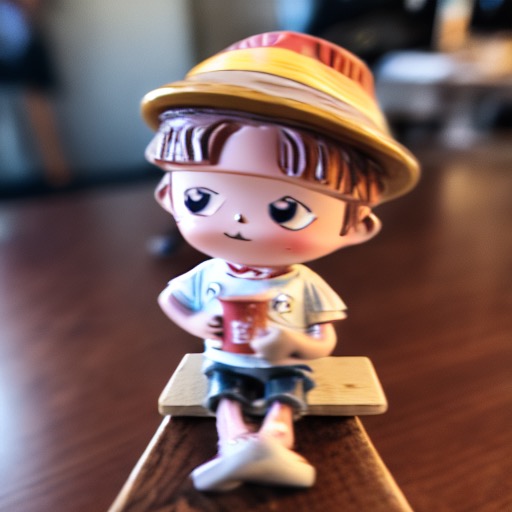} &
        \hspace{0.05cm}
        \includegraphics[width=0.102\textwidth]{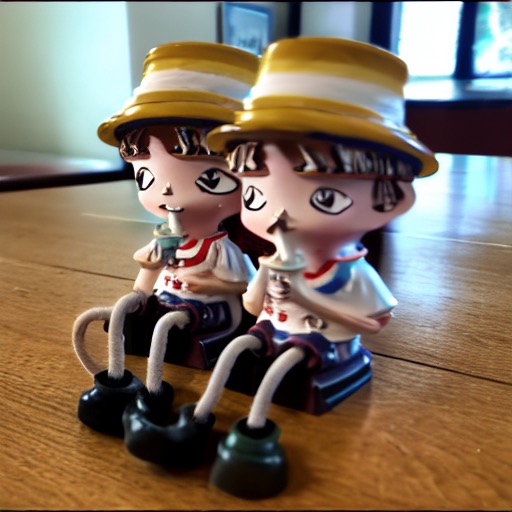} &
        \includegraphics[width=0.102\textwidth]{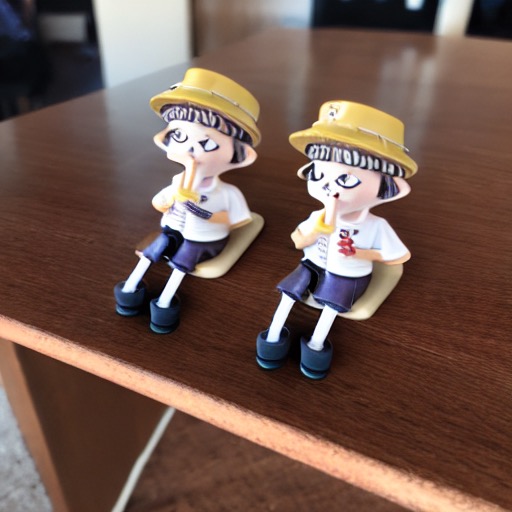} \\
        
    \end{tabular}
    \\[-0.1cm]
    }
    \caption{Additional qualitative comparisons. For each concept, we show four images generated by each method using the same set of random seeds. Results for TI are obtained after $5,000$ optimization steps while the remaining methods are all trained for $500$ steps. Results obtained with NeTI use our textual bypass.}
    \label{fig:additional_qualitative_comparison}
\end{figure*}

\begin{figure*}
    \centering
    \renewcommand{\arraystretch}{0.3}
    \setlength{\tabcolsep}{0.5pt}

    {\footnotesize

    \begin{tabular}{c@{\hspace{0.2cm}} c c @{\hspace{0.2cm}} c c @{\hspace{0.2cm}} c c @{\hspace{0.2cm}} c c }

        \begin{tabular}{c} Real Sample \\ \\[-0.05cm] \& Prompt \end{tabular} &
        \multicolumn{2}{c}{Textual Inversion (TI)} &
        \multicolumn{2}{c}{DreamBooth} &
        \multicolumn{2}{c}{Extended Textual Inversion} &
        \multicolumn{2}{c}{NeTI} \\

        \includegraphics[width=0.102\textwidth]{images/original/colorful_teapot.jpg} &
        \includegraphics[width=0.102\textwidth]{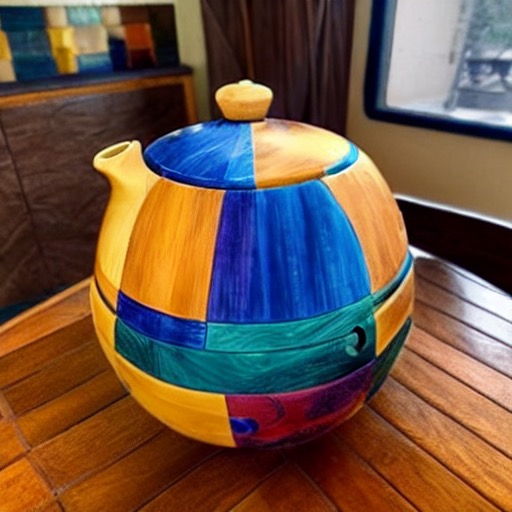} &
        \includegraphics[width=0.102\textwidth]{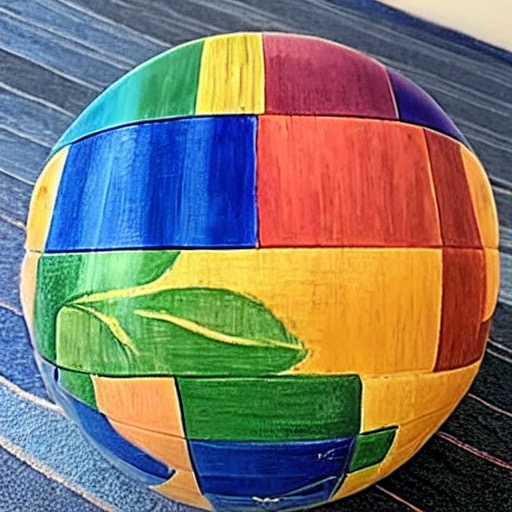} &
        \hspace{0.05cm}
        \includegraphics[width=0.102\textwidth]{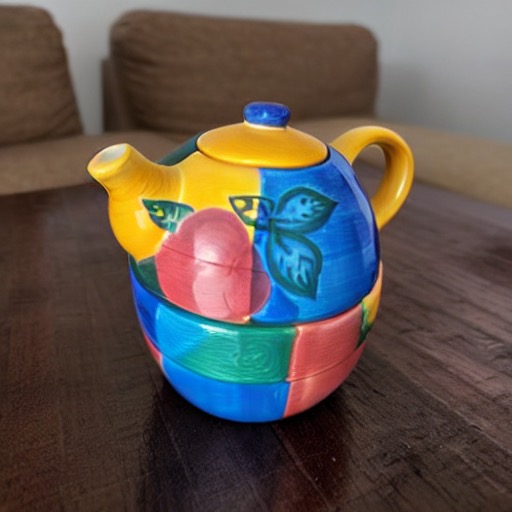} &
        \includegraphics[width=0.102\textwidth]{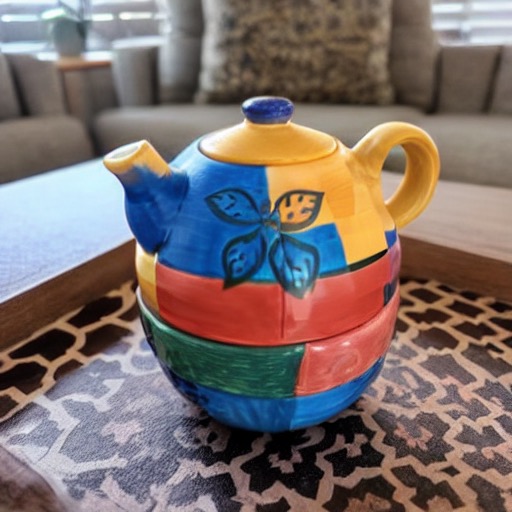} &
        \hspace{0.05cm}
        \includegraphics[width=0.102\textwidth]{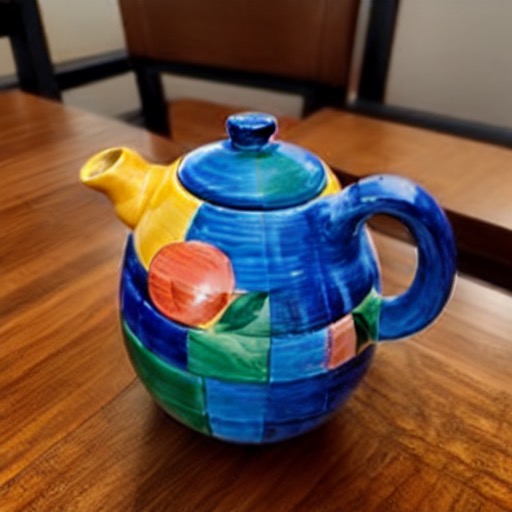} &
        \includegraphics[width=0.102\textwidth]{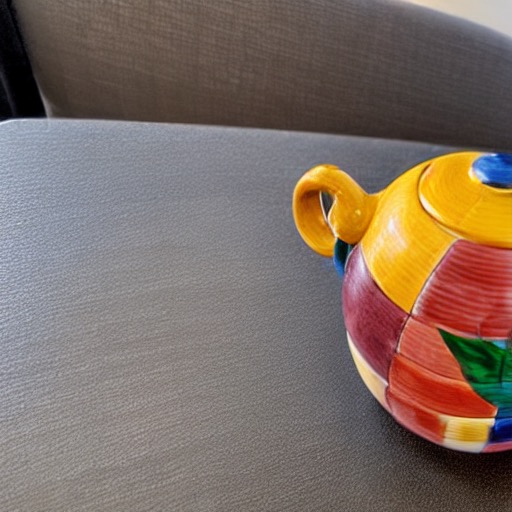} &
        \hspace{0.05cm}
        \includegraphics[width=0.102\textwidth]{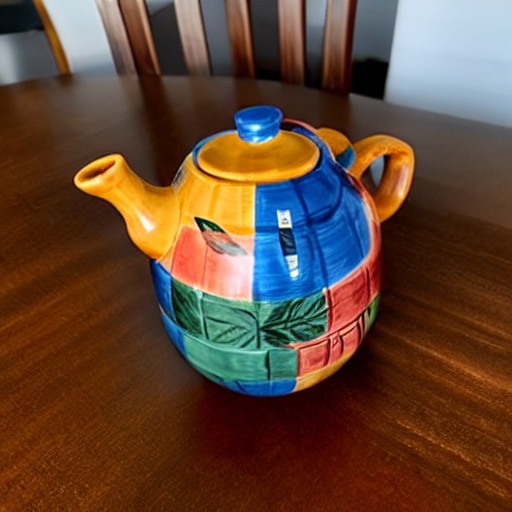} &
        \includegraphics[width=0.102\textwidth]{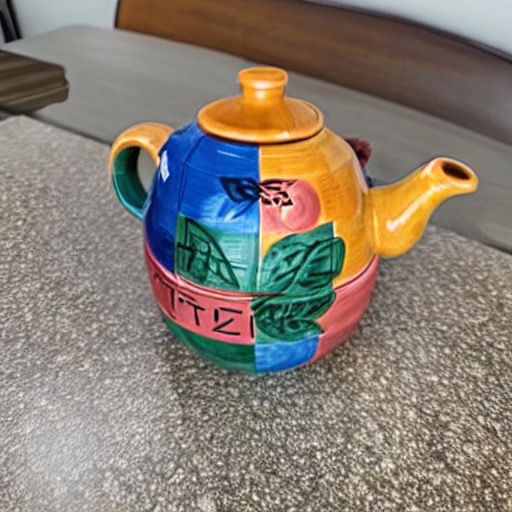} \\

        \raisebox{0.325in}{\begin{tabular}{c} ``A photo \\ \\[-0.05cm] of $S_*$''\end{tabular}} &
        \includegraphics[width=0.102\textwidth]{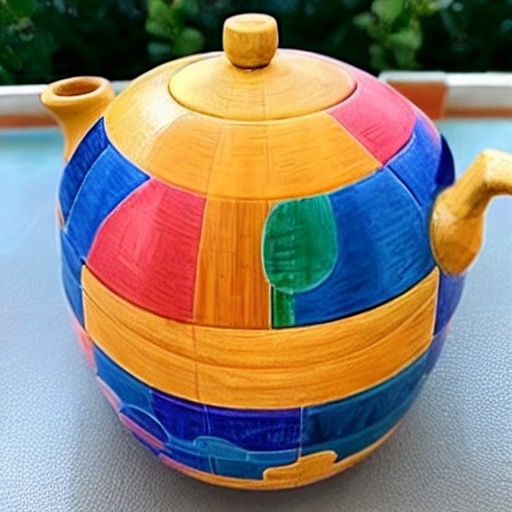} &
        \includegraphics[width=0.102\textwidth]{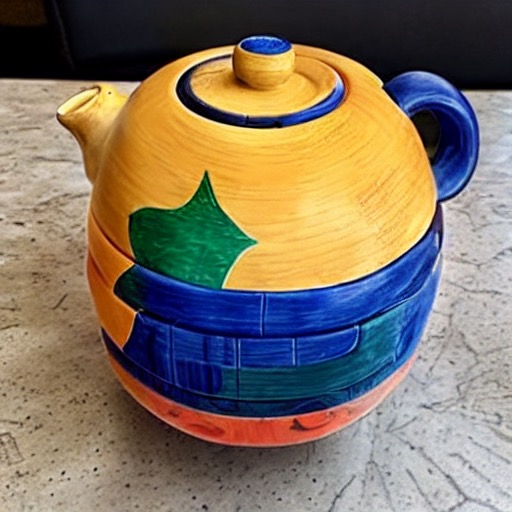} &
        \hspace{0.05cm}
        \includegraphics[width=0.102\textwidth]{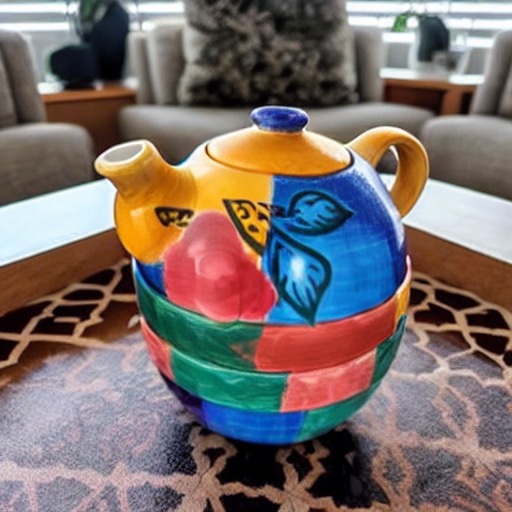} &
        \includegraphics[width=0.102\textwidth]{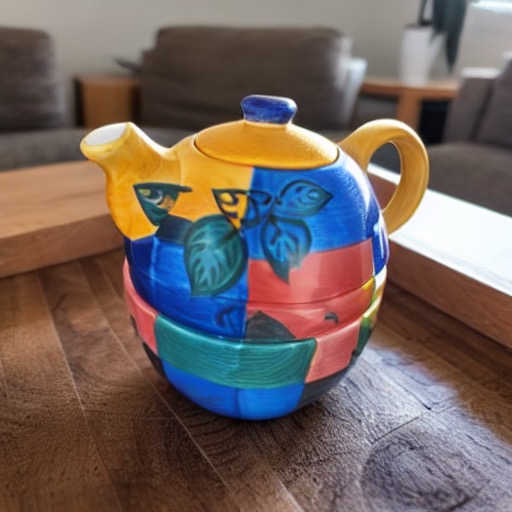} &
        \hspace{0.05cm}
        \includegraphics[width=0.102\textwidth]{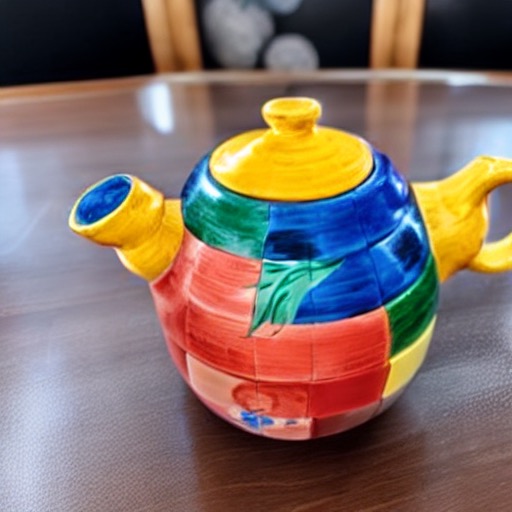} &
        \includegraphics[width=0.102\textwidth]{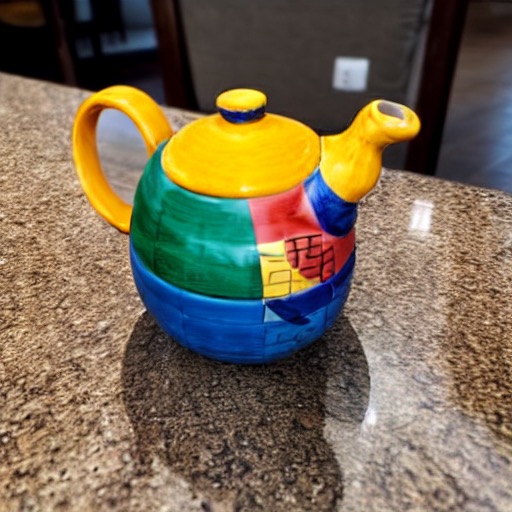} &
        \hspace{0.05cm}
        \includegraphics[width=0.102\textwidth]{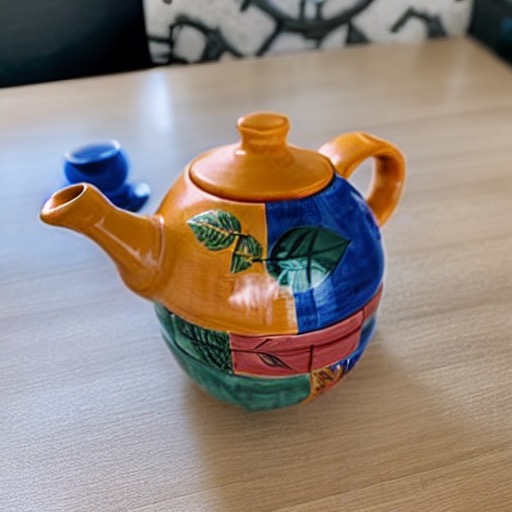} &
        \includegraphics[width=0.102\textwidth]{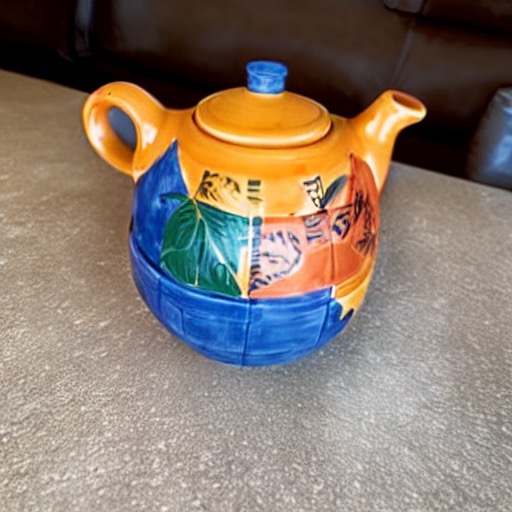} \\ \\
        
        \includegraphics[width=0.102\textwidth]{images/original/fat_stone_bird.jpg} &
        \includegraphics[width=0.102\textwidth]{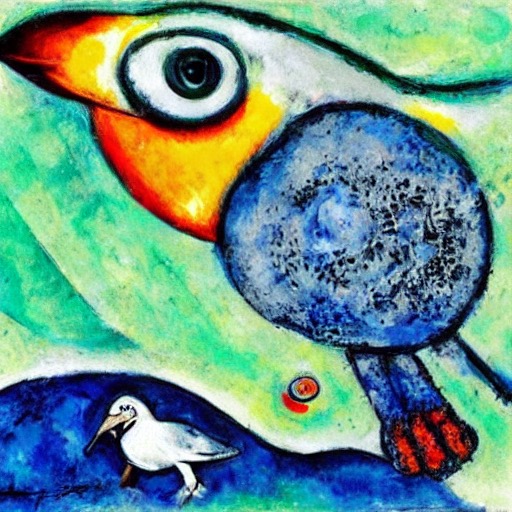} &
        \includegraphics[width=0.102\textwidth]{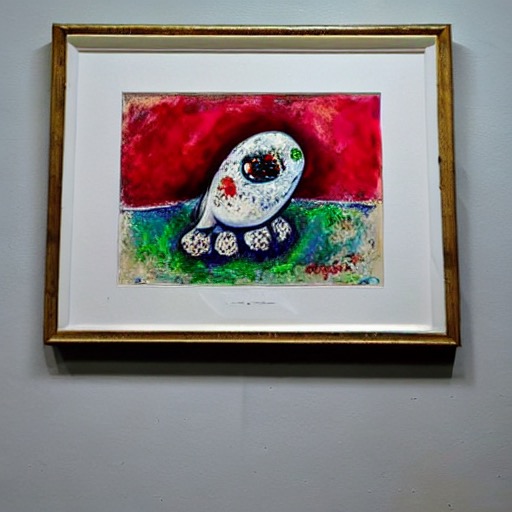} &
        \hspace{0.05cm}
        \includegraphics[width=0.102\textwidth]{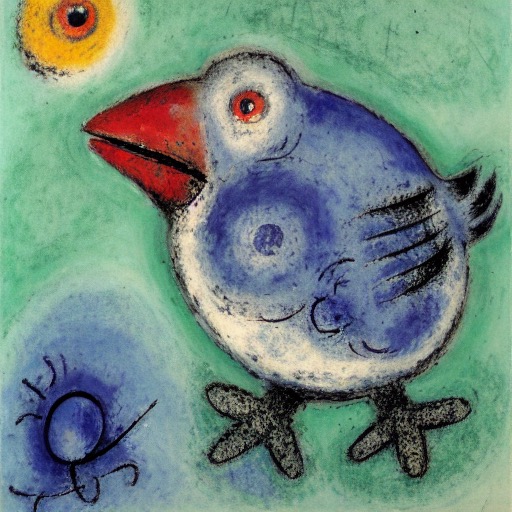} &
        \includegraphics[width=0.102\textwidth]{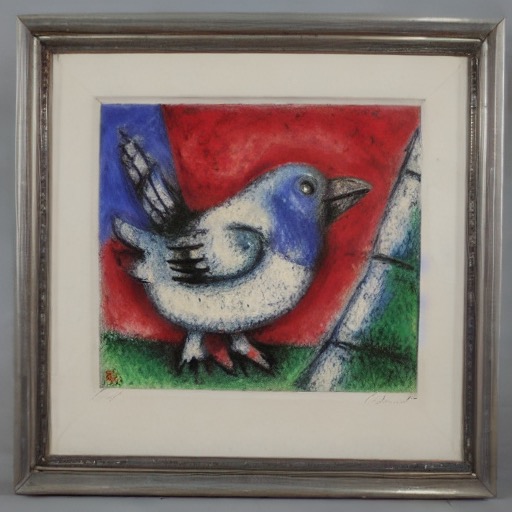} &
        \hspace{0.05cm}
        \includegraphics[width=0.102\textwidth]{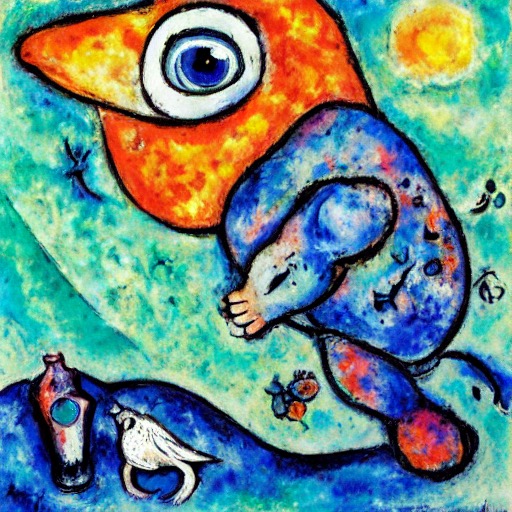} &
        \includegraphics[width=0.102\textwidth]{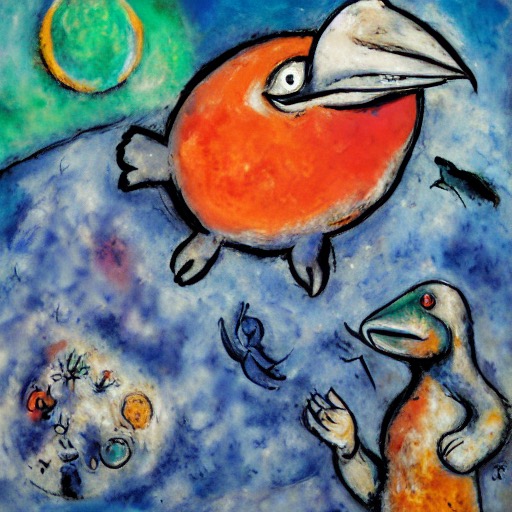} &
        \hspace{0.05cm}
        \includegraphics[width=0.102\textwidth]{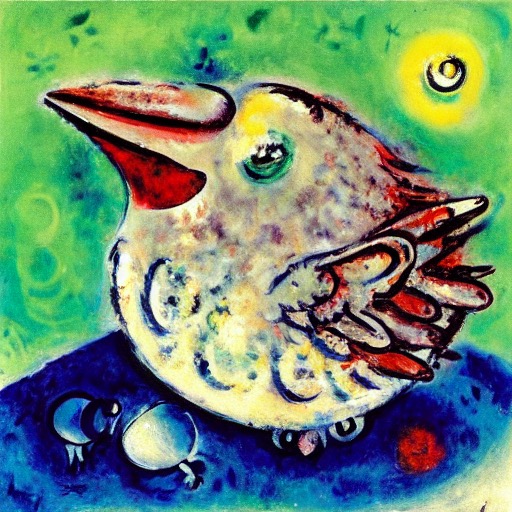} &
        \includegraphics[width=0.102\textwidth]{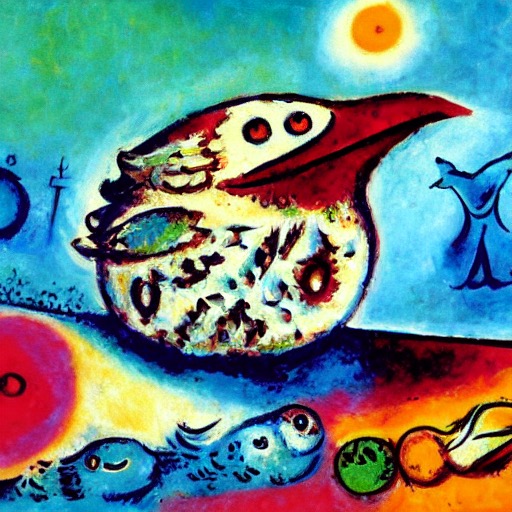} \\

        \raisebox{0.325in}{\begin{tabular}{c} ``A Marc \\ \\[-0.05cm] Chagall \\ \\[-0.05cm] painting of $S_*$''\end{tabular}} &
        \includegraphics[width=0.102\textwidth]{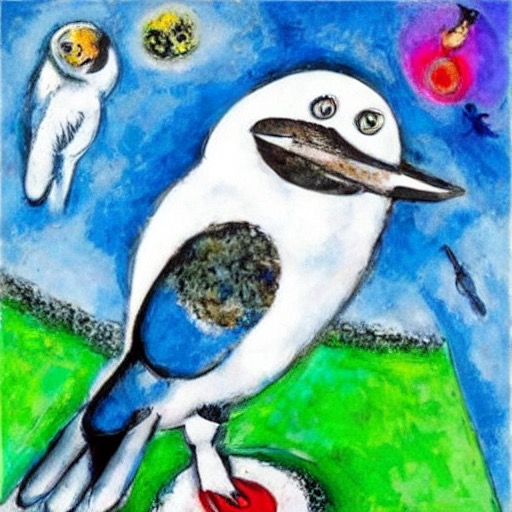} &
        \includegraphics[width=0.102\textwidth]{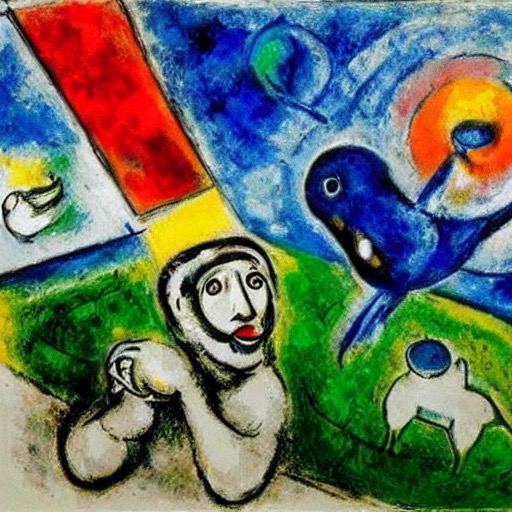} &
        \hspace{0.05cm}
        \includegraphics[width=0.102\textwidth]{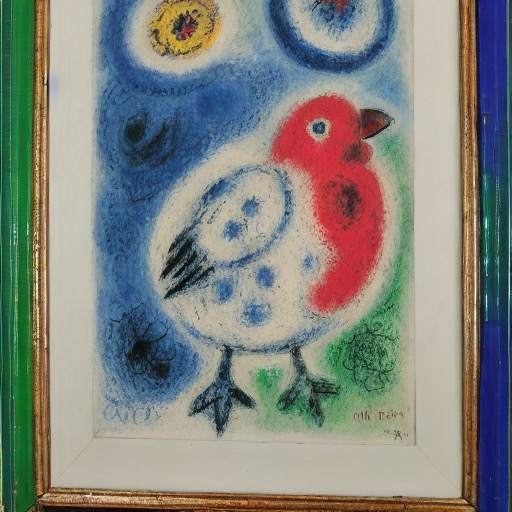} &
        \includegraphics[width=0.102\textwidth]{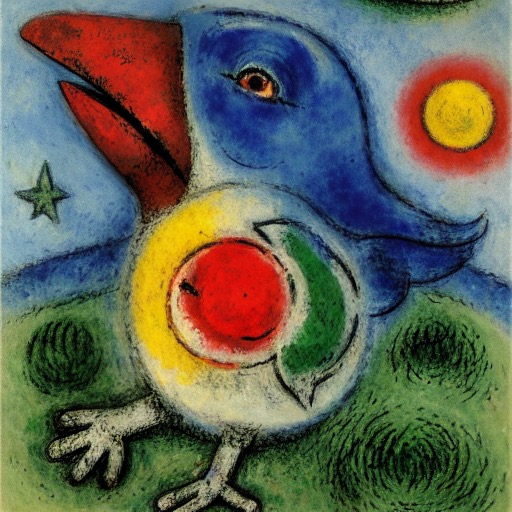} &
        \hspace{0.05cm}
        \includegraphics[width=0.102\textwidth]{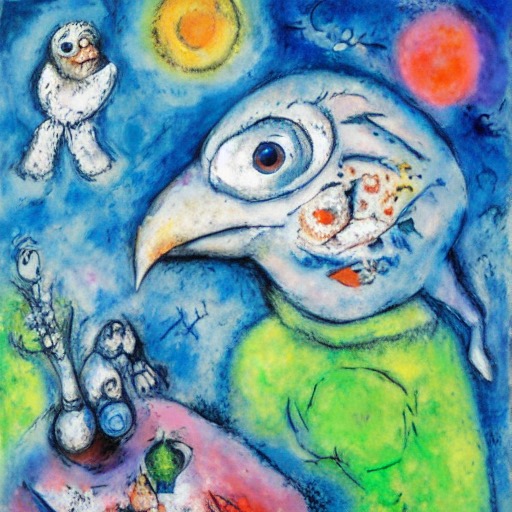} &
        \includegraphics[width=0.102\textwidth]{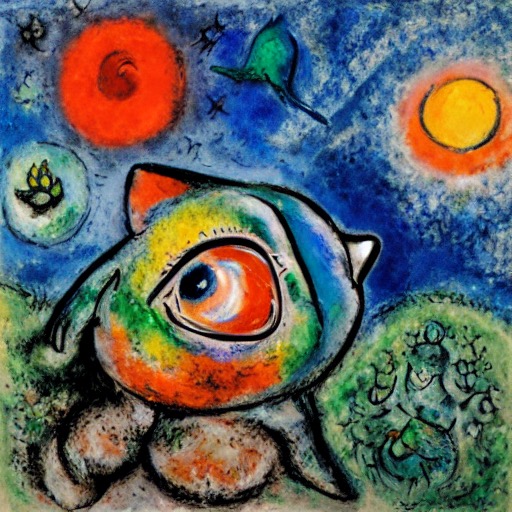} &
        \hspace{0.05cm}
        \includegraphics[width=0.102\textwidth]{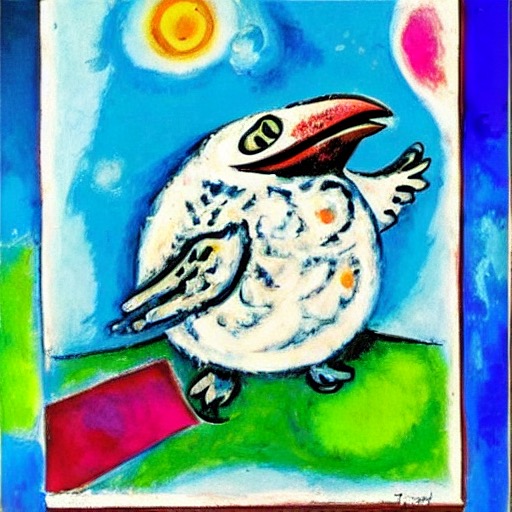} &
        \includegraphics[width=0.102\textwidth]{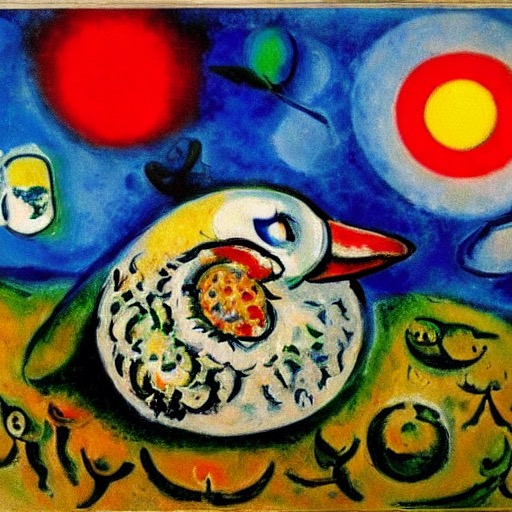} \\ \\

        \includegraphics[width=0.102\textwidth]{images/original/cat.jpg} &
        \includegraphics[width=0.102\textwidth]{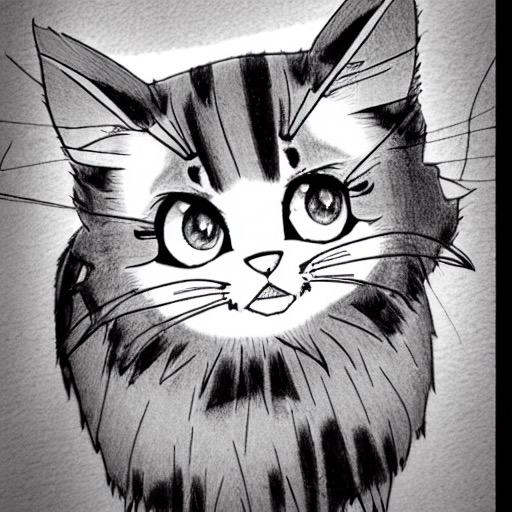} &
        \includegraphics[width=0.102\textwidth]{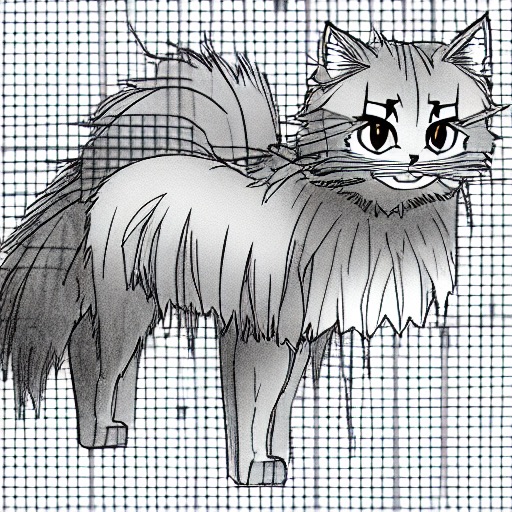} &
        \hspace{0.05cm}
        \includegraphics[width=0.102\textwidth]{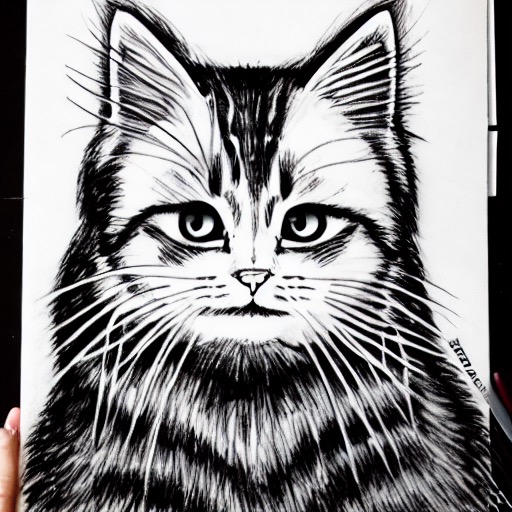} &
        \includegraphics[width=0.102\textwidth]{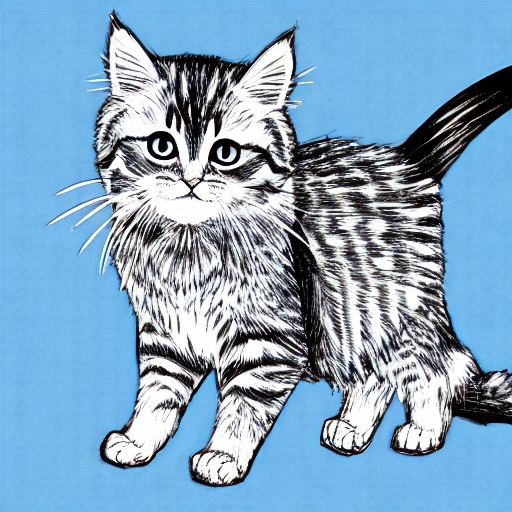} &
        \hspace{0.05cm}
        \includegraphics[width=0.102\textwidth]{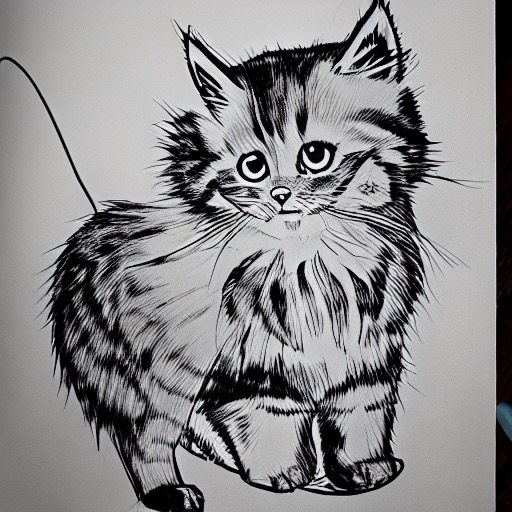} &
        \includegraphics[width=0.102\textwidth]{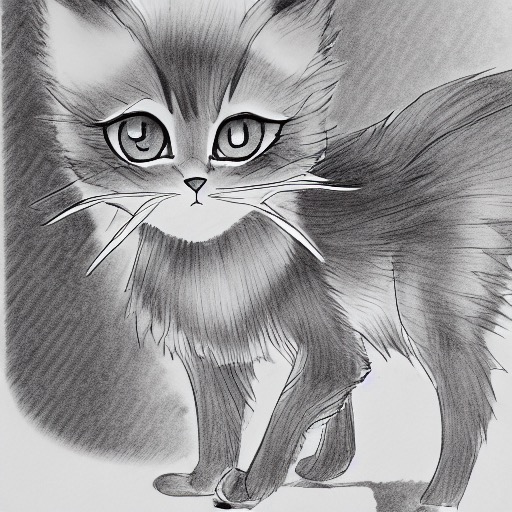} &
        \hspace{0.05cm}
        \includegraphics[width=0.102\textwidth]{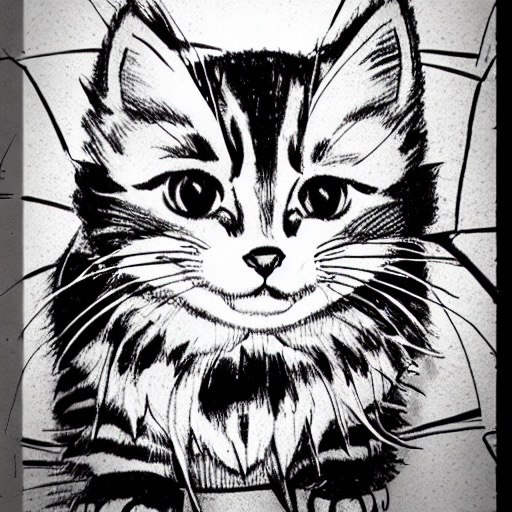} &
        \includegraphics[width=0.102\textwidth]{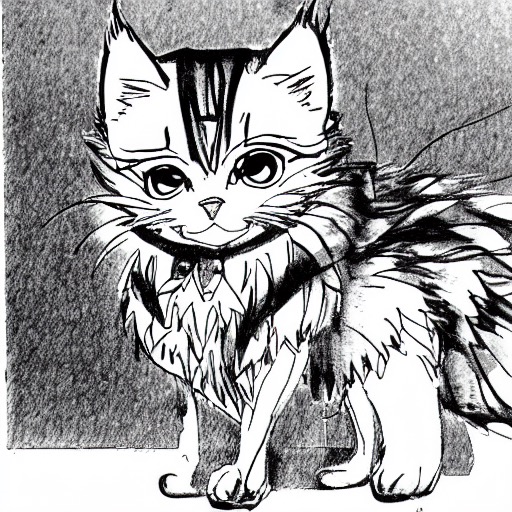} \\

        \raisebox{0.325in}{\begin{tabular}{c} ``A manga \\ \\[-0.05cm] drawing of $S_*$''\end{tabular}} &
        \includegraphics[width=0.102\textwidth]{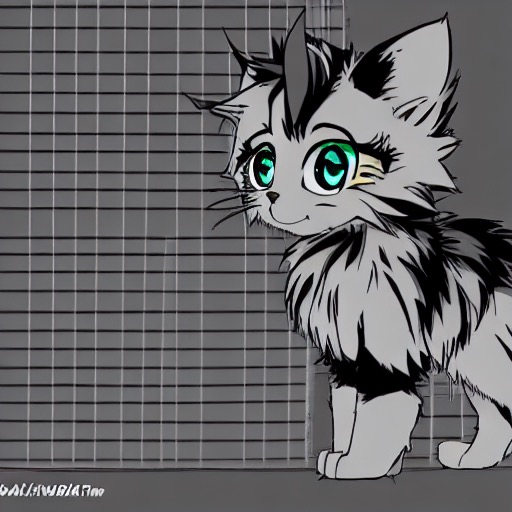} &
        \includegraphics[width=0.102\textwidth]{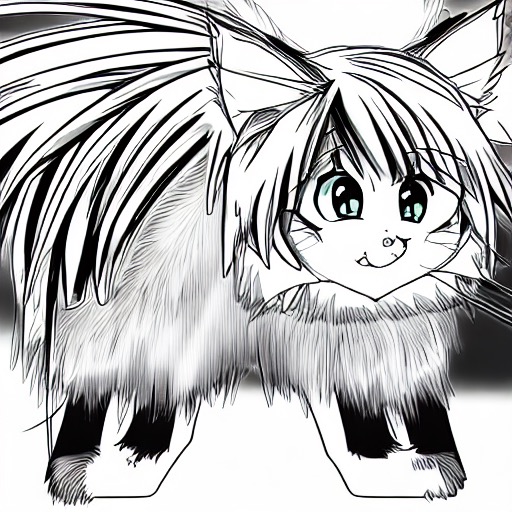} &
        \hspace{0.05cm}
        \includegraphics[width=0.102\textwidth]{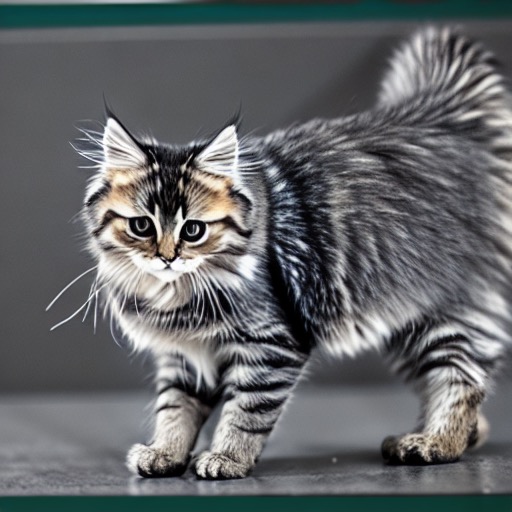} &
        \includegraphics[width=0.102\textwidth]{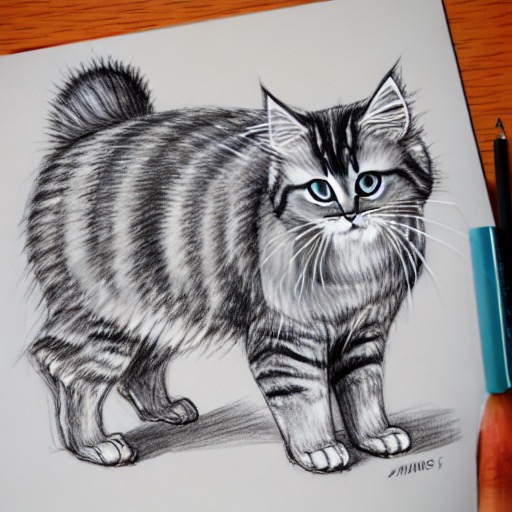} &
        \hspace{0.05cm}
        \includegraphics[width=0.102\textwidth]{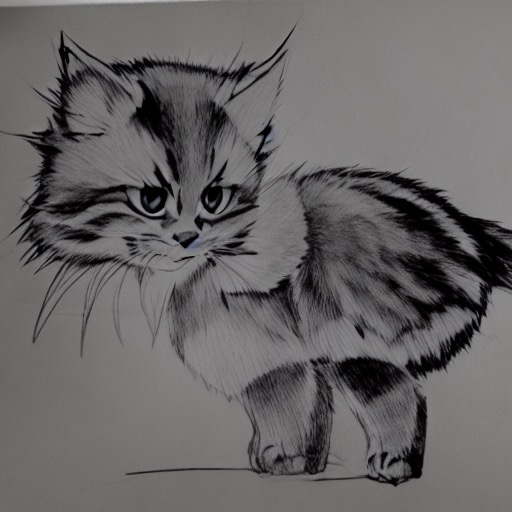} &
        \includegraphics[width=0.102\textwidth]{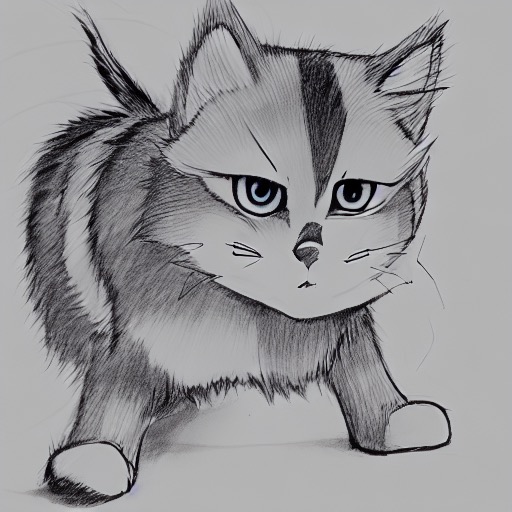} &
        \hspace{0.05cm}
        \includegraphics[width=0.102\textwidth]{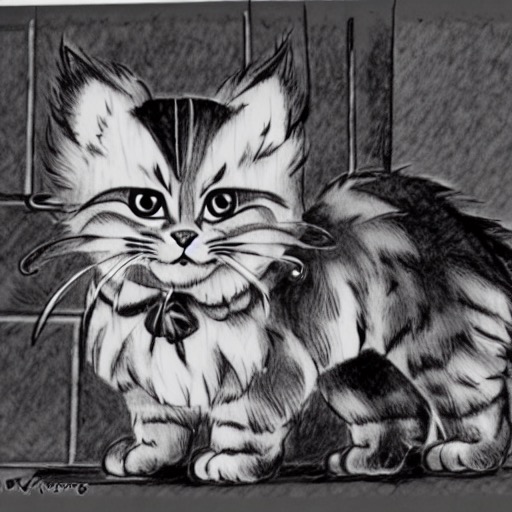} &
        \includegraphics[width=0.102\textwidth]{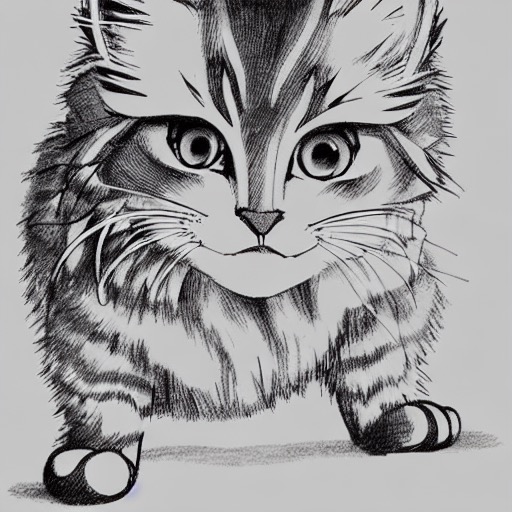} \\ \\

        \includegraphics[width=0.102\textwidth]{images/original/clock.jpeg} &
        \includegraphics[width=0.102\textwidth]{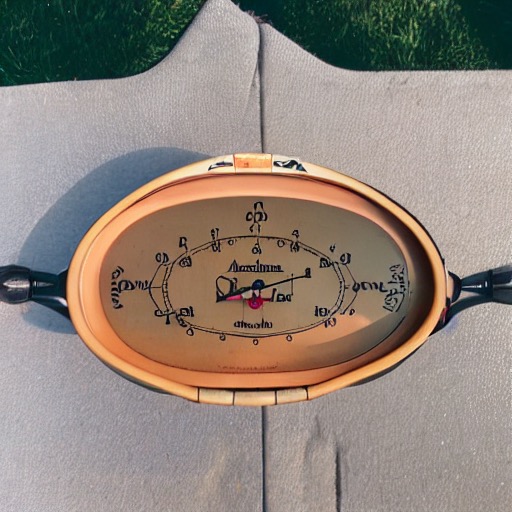} &
        \includegraphics[width=0.102\textwidth]{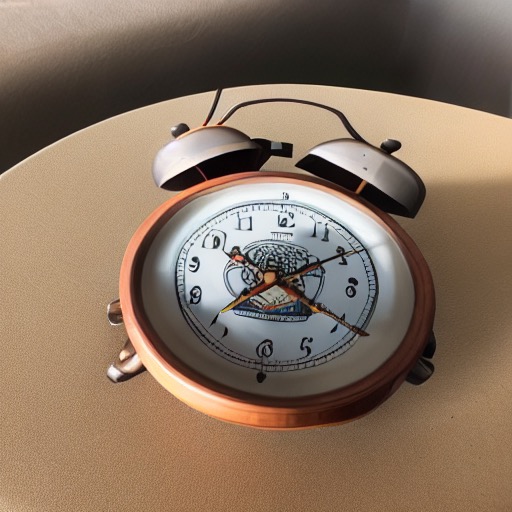} &
        \hspace{0.05cm}
        \includegraphics[width=0.102\textwidth]{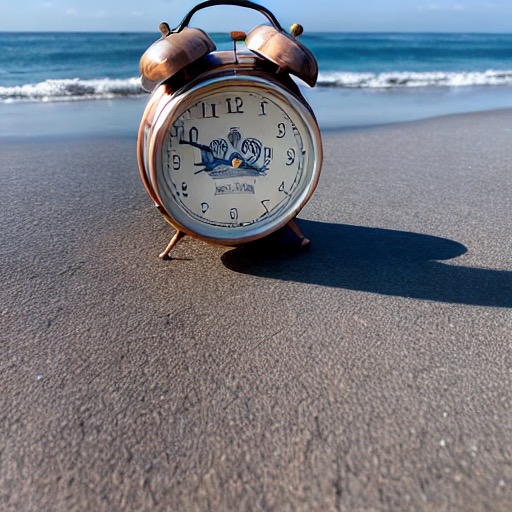} &
        \includegraphics[width=0.102\textwidth]{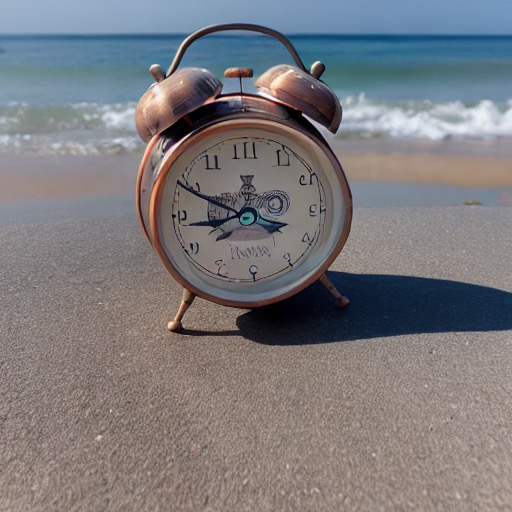} &
        \hspace{0.05cm}
        \includegraphics[width=0.102\textwidth]{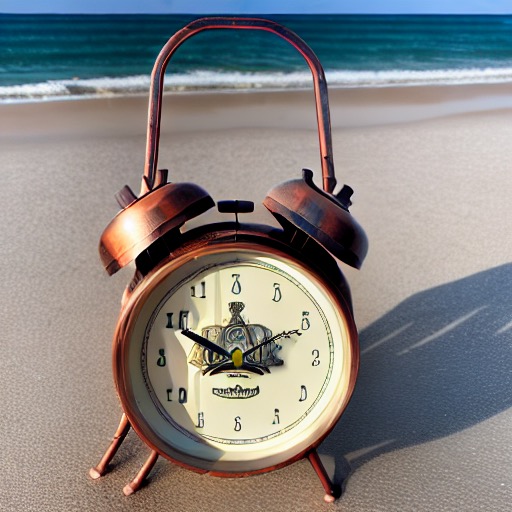} &
        \includegraphics[width=0.102\textwidth]{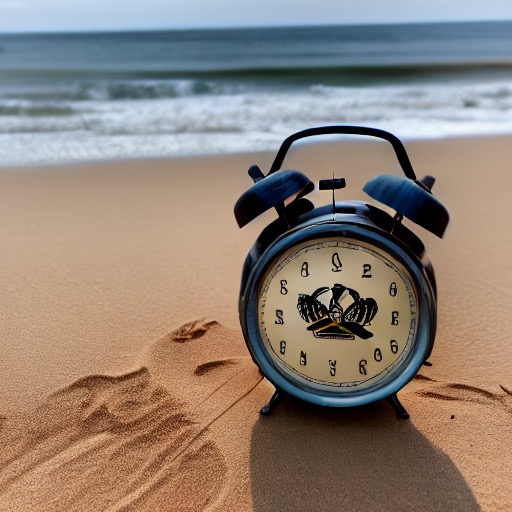} &
        \hspace{0.05cm}
        \includegraphics[width=0.102\textwidth]{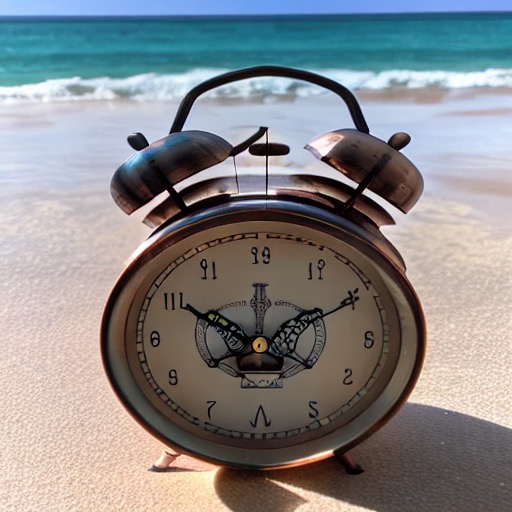} &
        \includegraphics[width=0.102\textwidth]{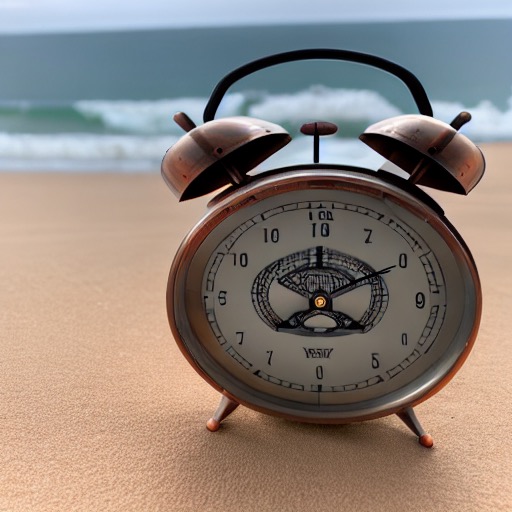} \\

        \raisebox{0.325in}{\begin{tabular}{c} ``A photo of \\ \\[-0.05cm] $S_*$ on a beach''\end{tabular}} &
        \includegraphics[width=0.102\textwidth]{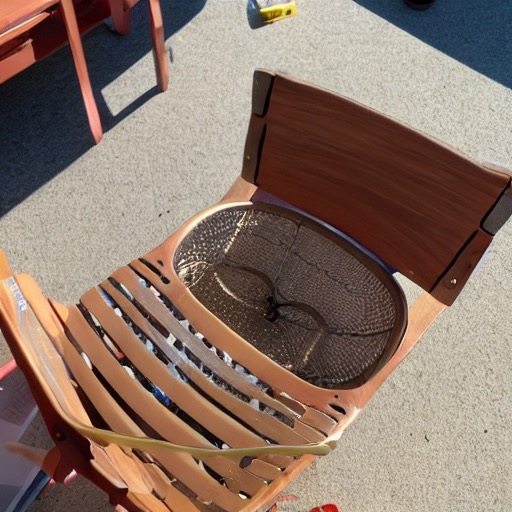} &
        \includegraphics[width=0.102\textwidth]{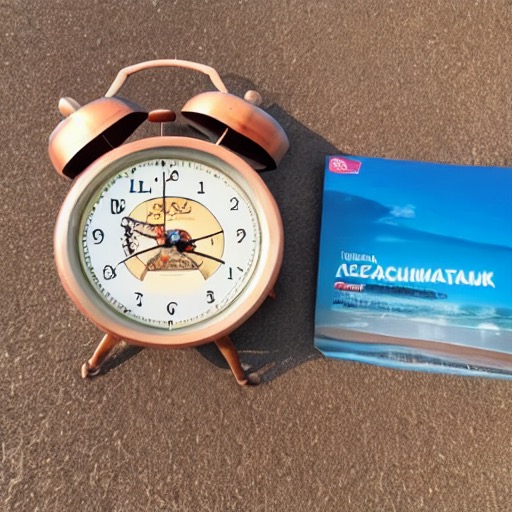} &
        \hspace{0.05cm}
        \includegraphics[width=0.102\textwidth]{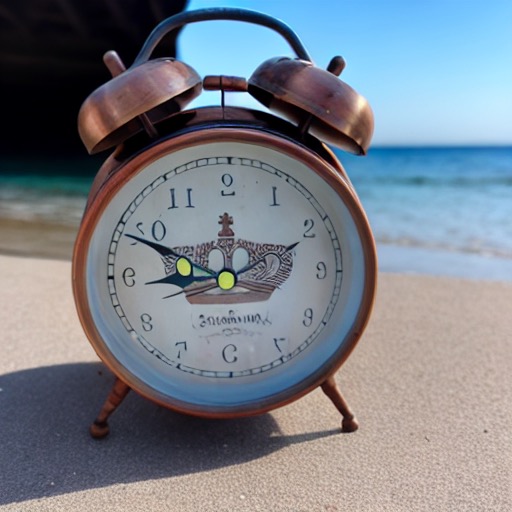} &
        \includegraphics[width=0.102\textwidth]{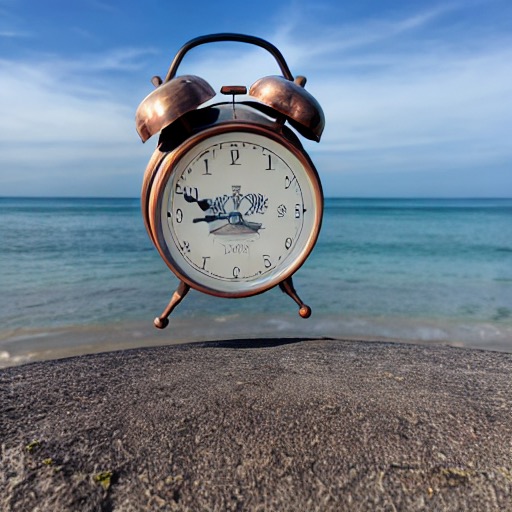} &
        \hspace{0.05cm}
        \includegraphics[width=0.102\textwidth]{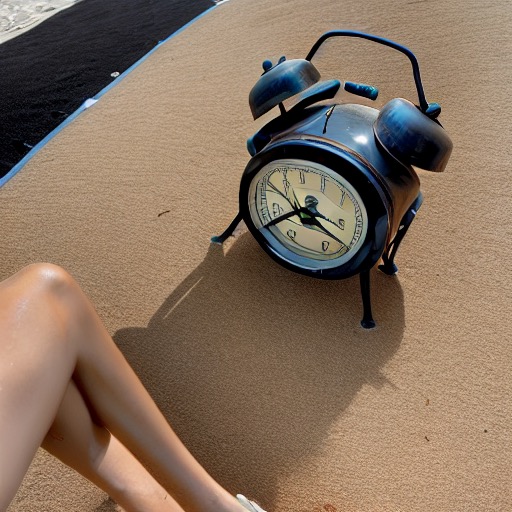} &
        \includegraphics[width=0.102\textwidth]{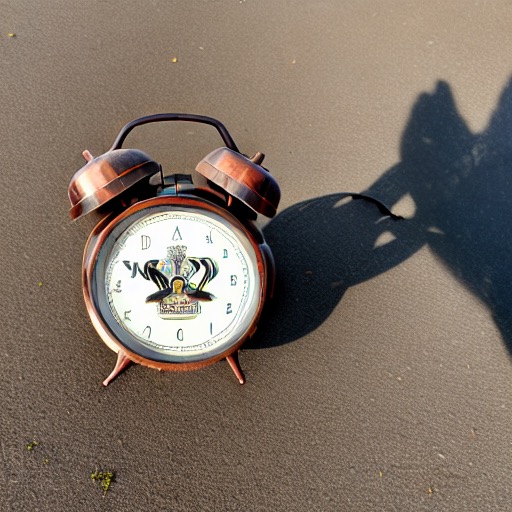} &
        \hspace{0.05cm}
        \includegraphics[width=0.102\textwidth]{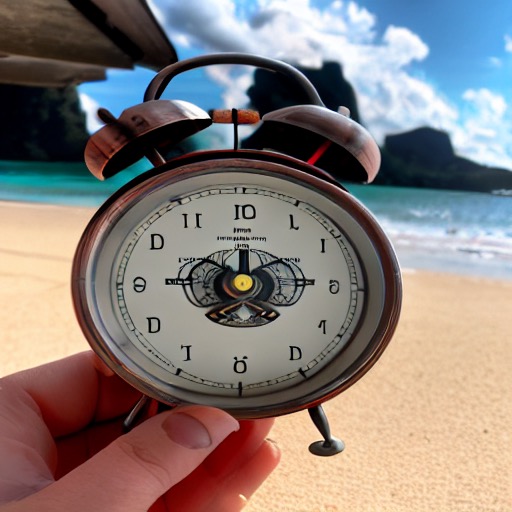} &
        \includegraphics[width=0.102\textwidth]{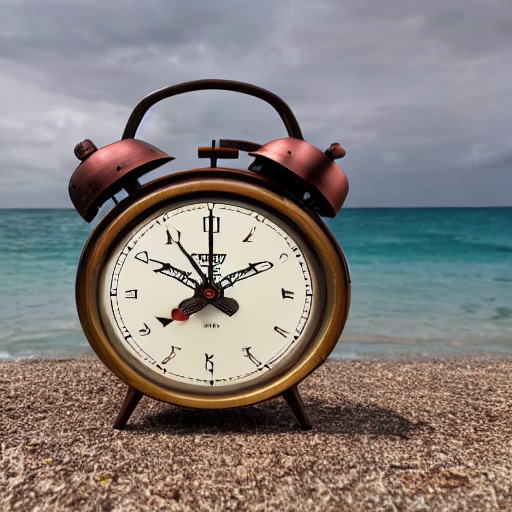} \\ \\

        \includegraphics[width=0.102\textwidth]{images/original/rainbow_cat.jpeg} &
        \includegraphics[width=0.102\textwidth]{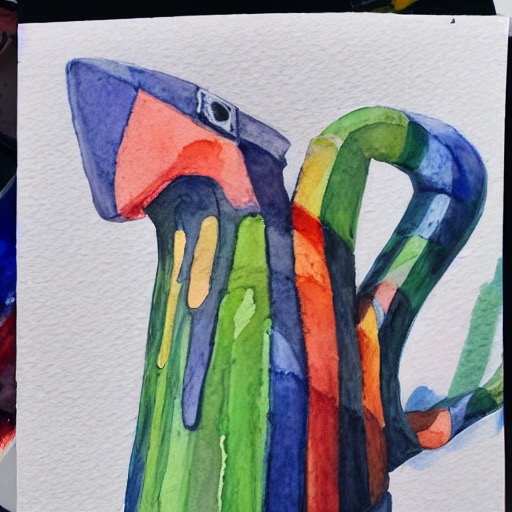} &
        \includegraphics[width=0.102\textwidth]{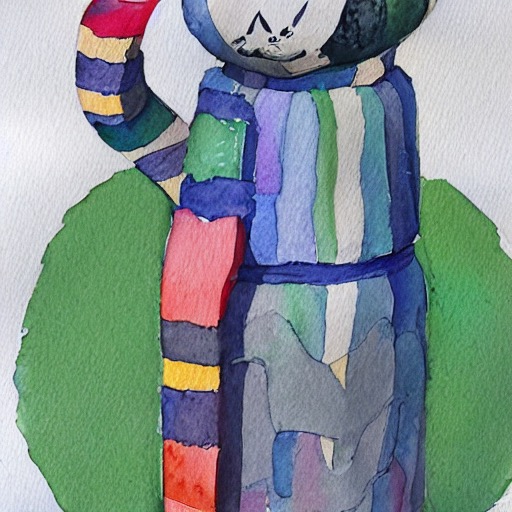} &
        \hspace{0.05cm}
        \includegraphics[width=0.102\textwidth]{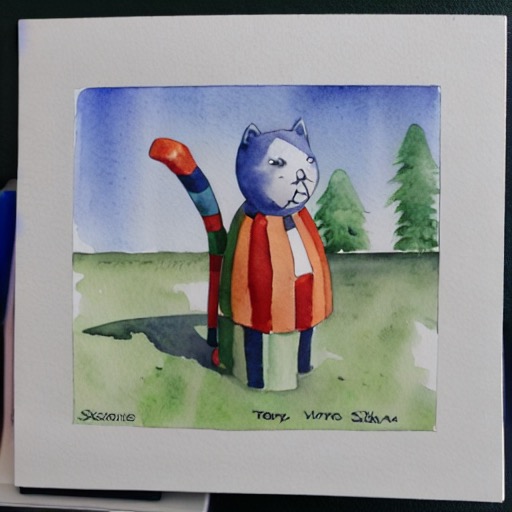} &
        \includegraphics[width=0.102\textwidth]{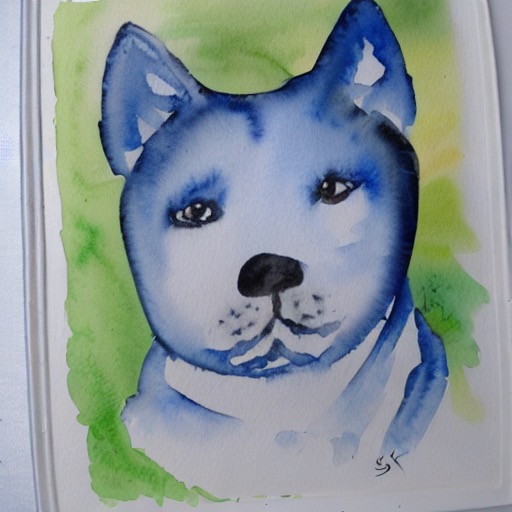} &
        \hspace{0.05cm}
        \includegraphics[width=0.102\textwidth]{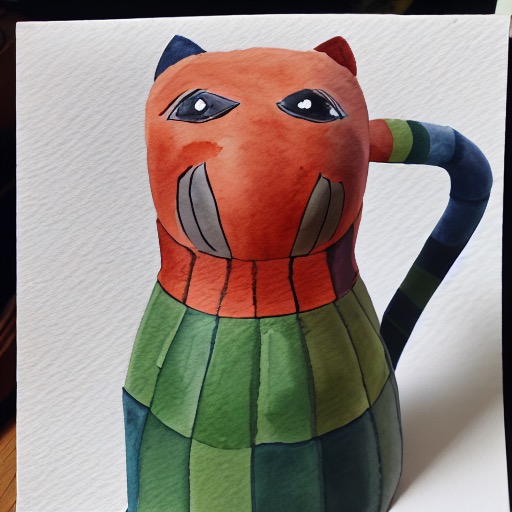} &
        \includegraphics[width=0.102\textwidth]{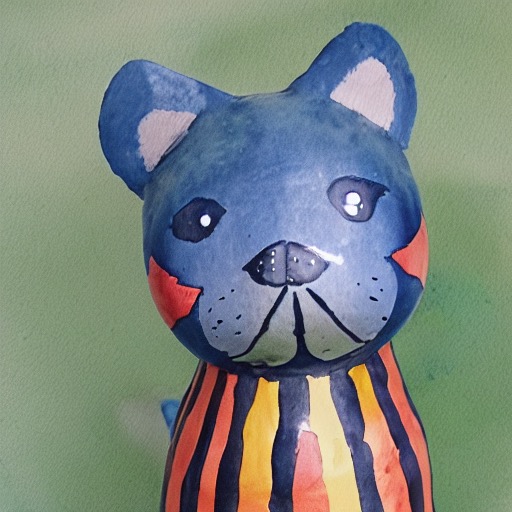} &
        
        \includegraphics[width=0.102\textwidth]{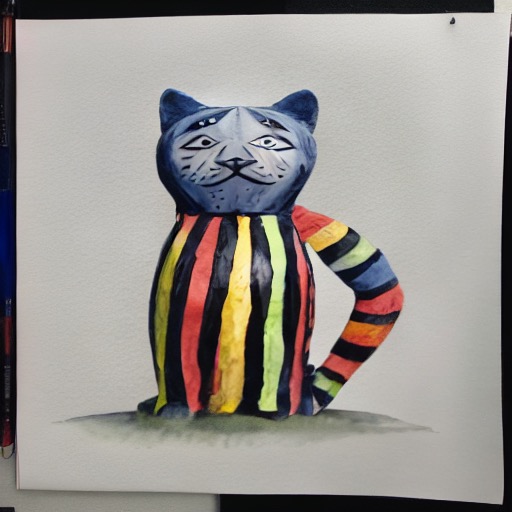} &
        \includegraphics[width=0.102\textwidth]{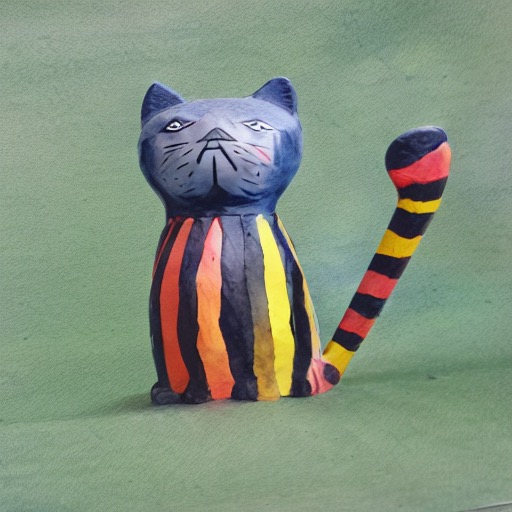} \\

        \raisebox{0.325in}{\begin{tabular}{c} ``A watercolor \\ \\[-0.05cm] painting of \\ \\[-0.05cm] $S_*$''\end{tabular}} &
        \includegraphics[width=0.102\textwidth]{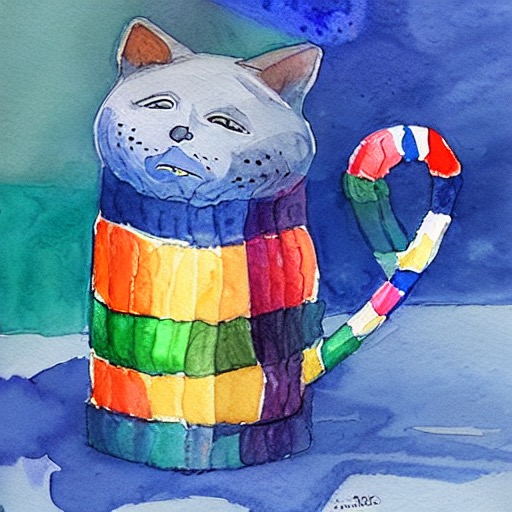} &
        \includegraphics[width=0.102\textwidth]{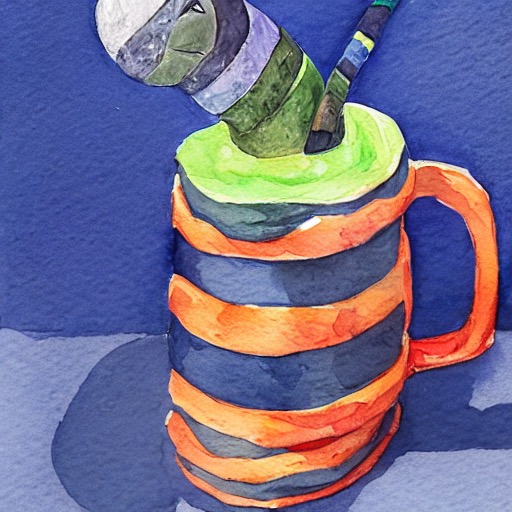} &
        \hspace{0.05cm}
        \includegraphics[width=0.102\textwidth]{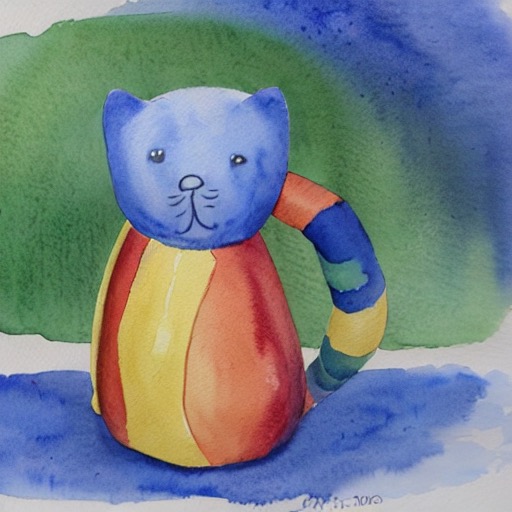} &
        \includegraphics[width=0.102\textwidth]{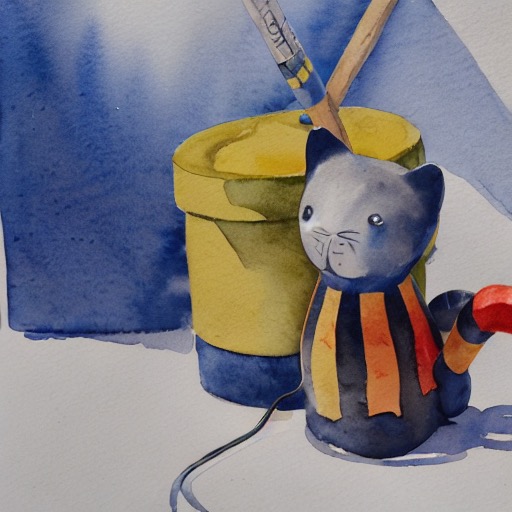} &
        \hspace{0.05cm}
        \includegraphics[width=0.102\textwidth]{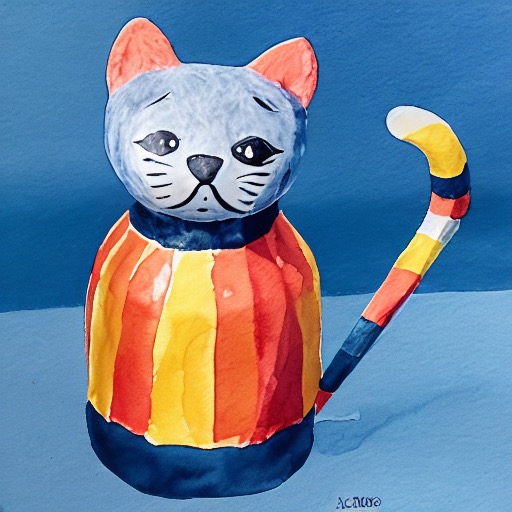} &
        \includegraphics[width=0.102\textwidth]{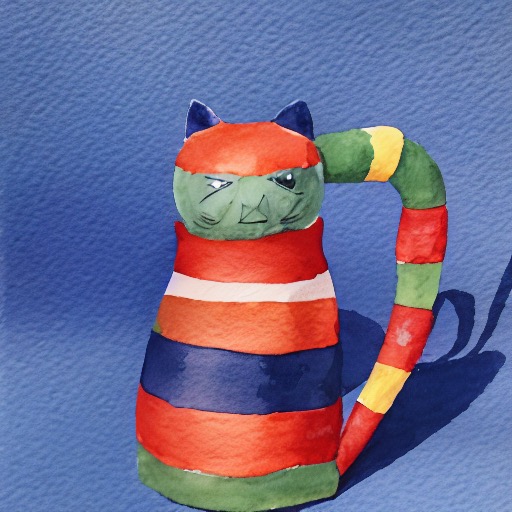} &
        \hspace{0.05cm}
        \includegraphics[width=0.102\textwidth]{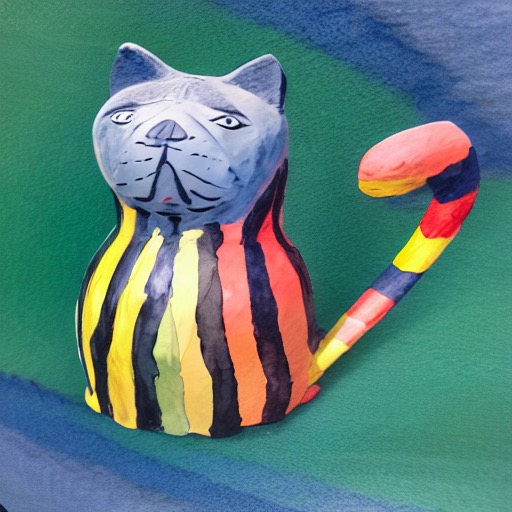} &
        \includegraphics[width=0.102\textwidth]{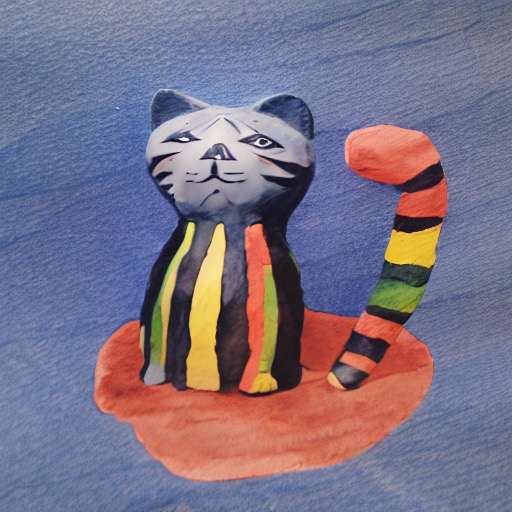} 
    \end{tabular}
    \\[-0.2cm]
    }
    \caption{Additional qualitative comparisons.  For each concept, we show four images generated by each method using the same set of random seeds. Results for TI are obtained after 5,000 optimization steps while the remaining methods are all trained for 500 steps. We show results obtained with NeTI using our textual bypass.}
    \label{fig:additional_qualitative_comparison_2}
\end{figure*}

\begin{figure*}
    \centering
    \setlength{\tabcolsep}{0.1pt}
    {\footnotesize
    \begin{tabular}{c@{\hspace{0.25cm}} c@{\hspace{0.25cm}} c@{\hspace{0.25cm}} c@{\hspace{0.25cm}} c@{\hspace{0.25cm}} c@{\hspace{0.25cm}} c@{\hspace{0.25cm}}}

        \includegraphics[width=0.13\textwidth]{images/original/rainbow_cat.jpeg} &
        \includegraphics[width=0.13\textwidth]{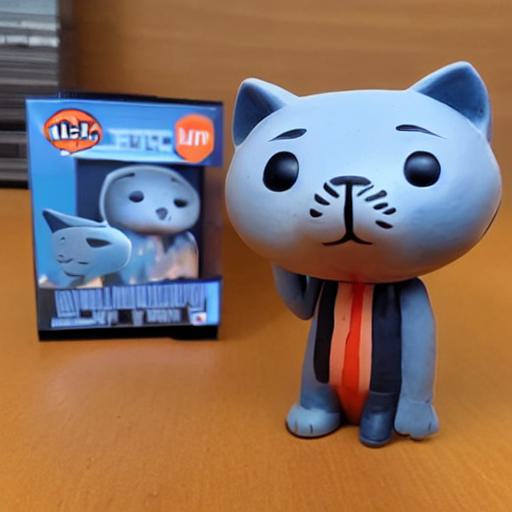} &
        \includegraphics[width=0.13\textwidth]{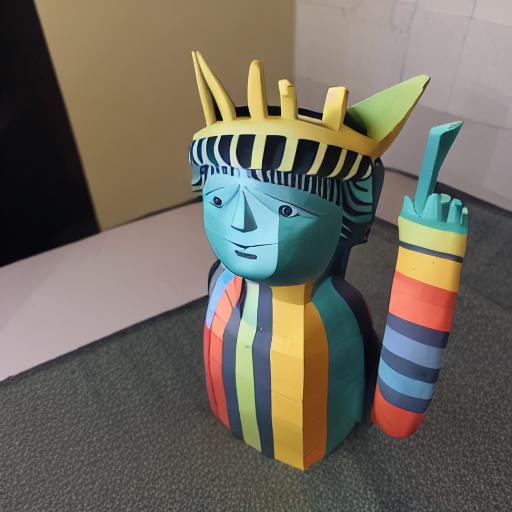} &
        \includegraphics[width=0.13\textwidth]{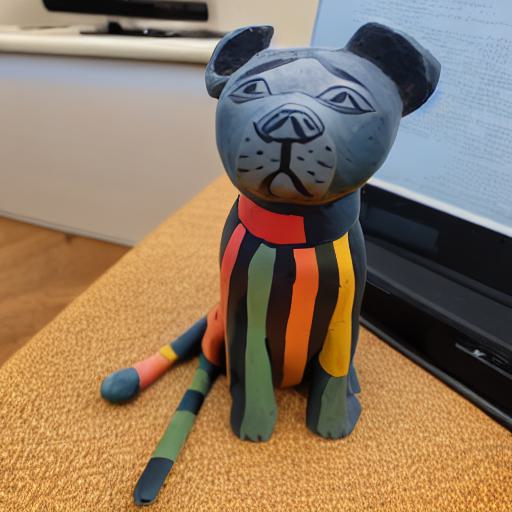} &
        \includegraphics[width=0.13\textwidth]{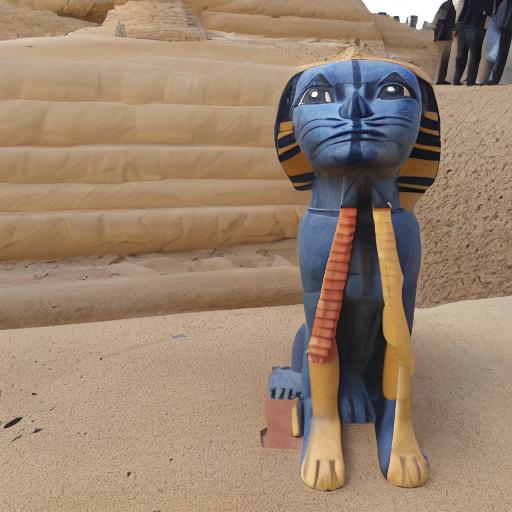} &
        \includegraphics[width=0.13\textwidth]{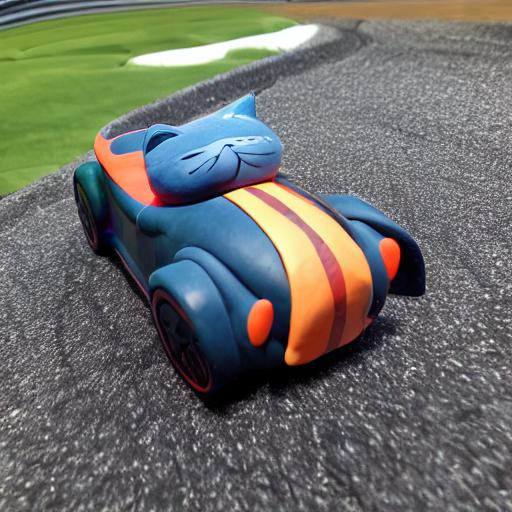} \\

        Real Sample &
        \begin{tabular}{c} ``A $S_*$ funko pop'' \end{tabular} &
        \begin{tabular}{c} ``A $S_*$ statue \\ of liberty'' \end{tabular} &
        \begin{tabular}{c} ``A dog statue of $S_*$'' \end{tabular} &
        \begin{tabular}{c} ``$S_*$ as Egypt's great \\ sphinx of Giza'' \end{tabular} &
        \begin{tabular}{c} ``A $S_*$ \\ sports car'' \end{tabular} \\ \\[-0.25cm]

        \includegraphics[width=0.13\textwidth]{images/original/metal_bird.jpg} &
        \includegraphics[width=0.13\textwidth]{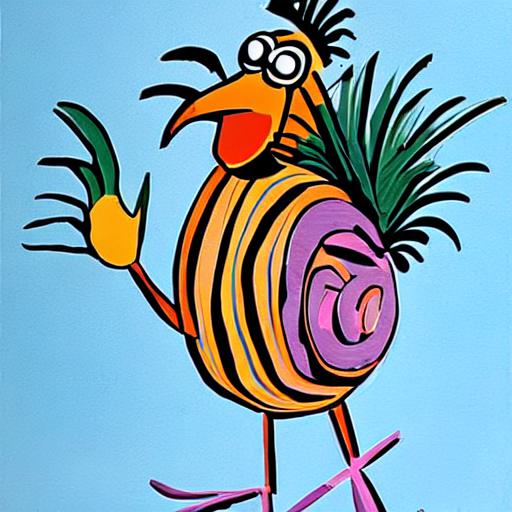} &
        \includegraphics[width=0.13\textwidth]{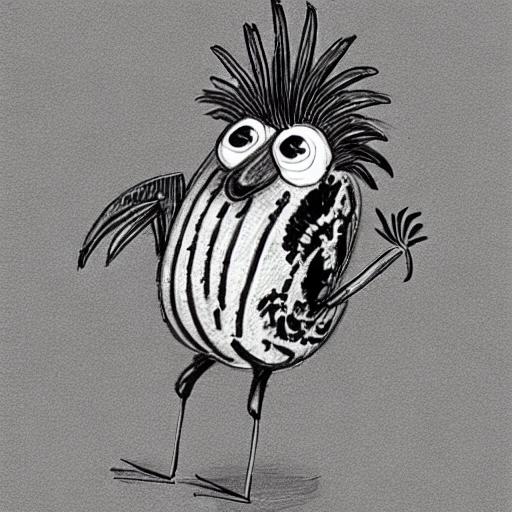} &
        \includegraphics[width=0.13\textwidth]{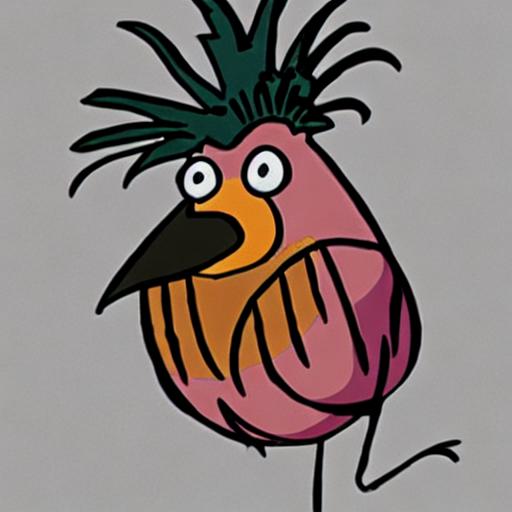} &
        \includegraphics[width=0.13\textwidth]{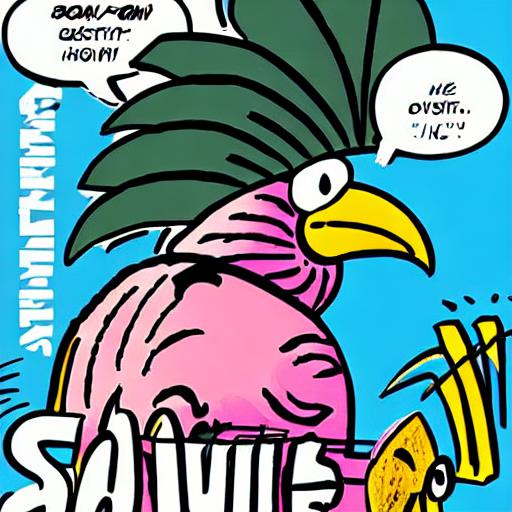} &
        \includegraphics[width=0.13\textwidth]{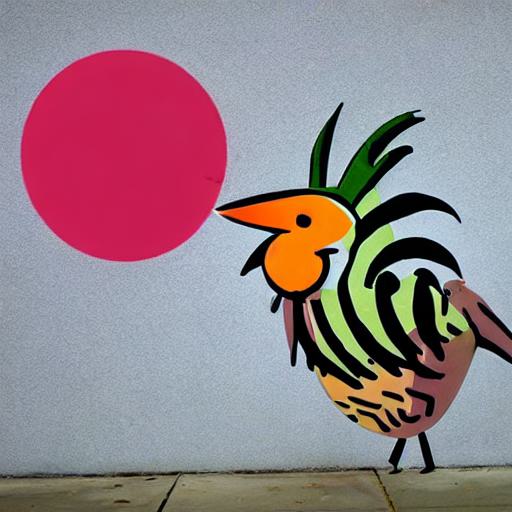} \\
        
        Real Sample &
        \begin{tabular}{c} ``A painting of $S_*$ \\ in the style of \\ Pablo Picasso'' \end{tabular} &
        \begin{tabular}{c} ``A black and white \\ pencil sketch of $S_*$'' \end{tabular} &
        \begin{tabular}{c} ``A cartoon \\ drawing of $S_*$'' \end{tabular} &
        \begin{tabular}{c} ``$S_*$ in a comic book'' \end{tabular} &
        \begin{tabular}{c} ``Banksy art of $S_*$'' \end{tabular} \\ \\[-0.25cm]

        \includegraphics[width=0.13\textwidth]{images/original/lecun.jpeg} &
        \includegraphics[width=0.13\textwidth]{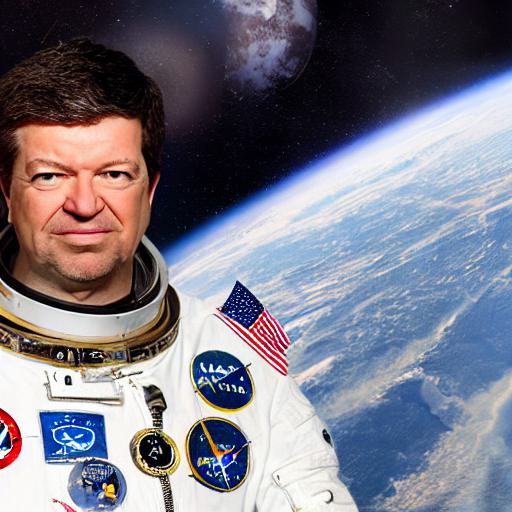} &
        \includegraphics[width=0.13\textwidth]{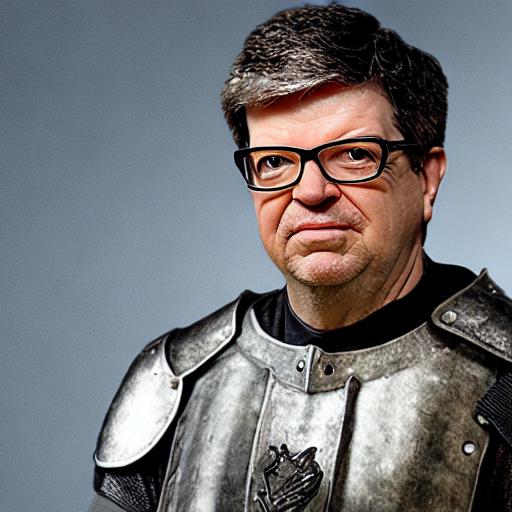} &
        \includegraphics[width=0.13\textwidth]{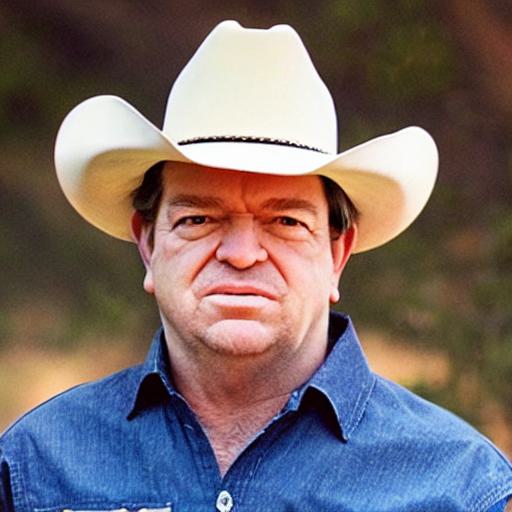} &
        \includegraphics[width=0.13\textwidth]{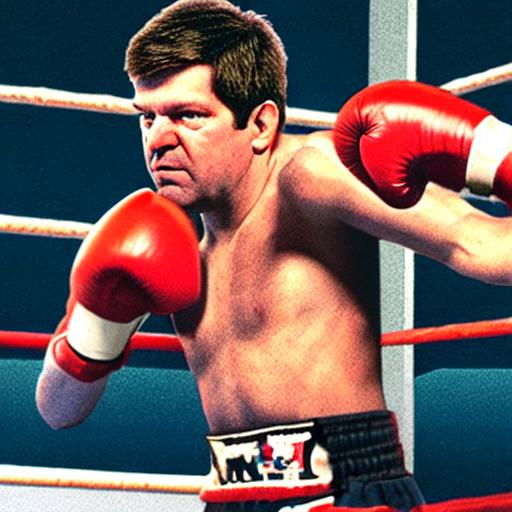} &
        \includegraphics[width=0.13\textwidth]{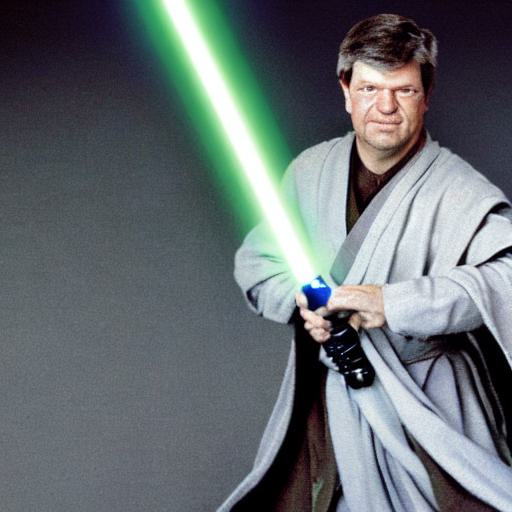} \\

        Real Sample &
        \begin{tabular}{c} ``A photo of $S_*$ \\ as an astronaut'' \end{tabular} &
        \begin{tabular}{c} ``A photo of $S_*$ as \\ an knight in armor'' \end{tabular} &
        \begin{tabular}{c} ``A photo of $S_*$ \\ as a cowboy'' \end{tabular} &
        \begin{tabular}{c} ``A photo of $S_*$ \\ as a boxer'' \end{tabular} &
        \begin{tabular}{c} ``A photo of $S_*$ \\ as a jedi'' \end{tabular} \\ \\[-0.25cm]
        
        \includegraphics[width=0.13\textwidth]{images/original/cat.jpg} &
        \hspace{0.05cm}
        \includegraphics[width=0.13\textwidth]{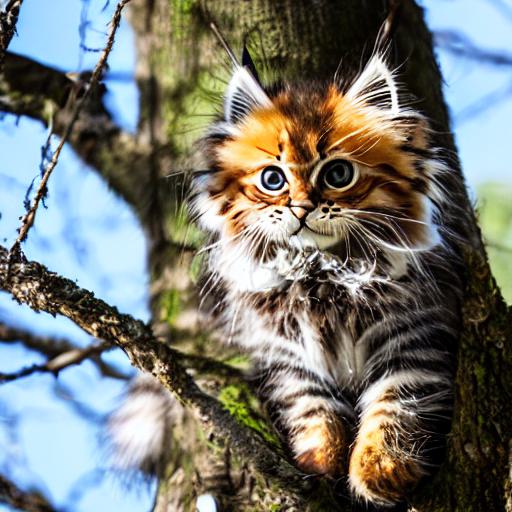} &
        \includegraphics[width=0.13\textwidth]{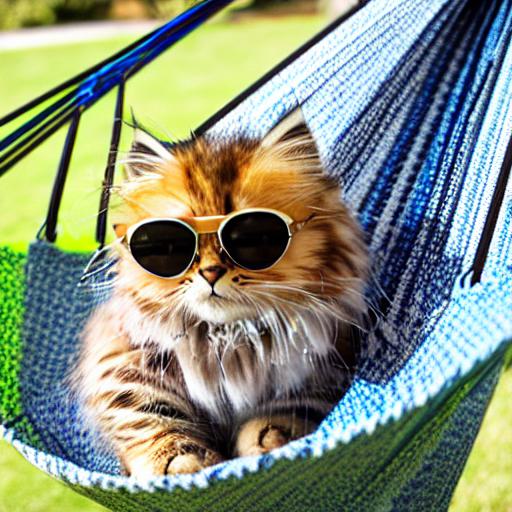} &
        \includegraphics[width=0.13\textwidth]{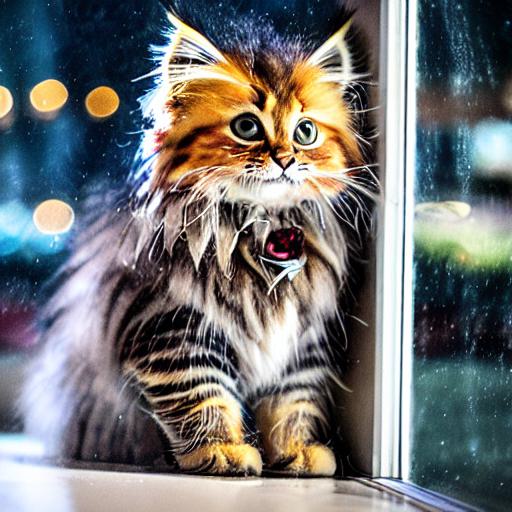} &
        \includegraphics[width=0.13\textwidth]{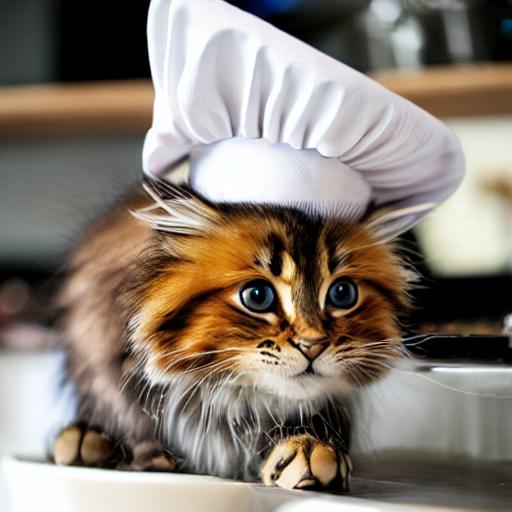} &
        \includegraphics[width=0.13\textwidth]{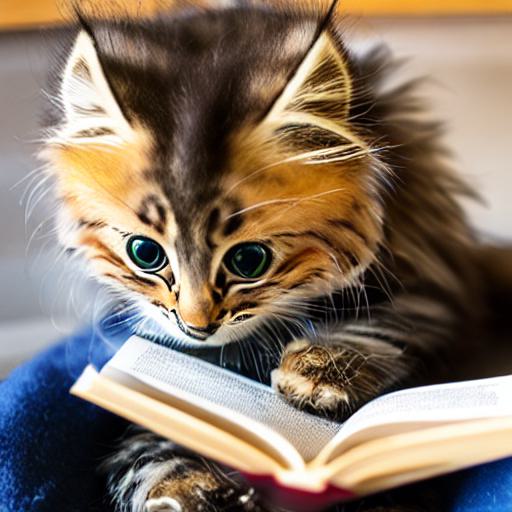} \\

        Real Sample &
        \hspace{0.05cm}
        \begin{tabular}{c} ``A photo of $S_*$ \\ sitting in a tree'' \end{tabular} &
        \begin{tabular}{c} ``$S_*$ sitting in \\ a hammock with \\ sunglasses on'' \end{tabular} &
        \begin{tabular}{c} ``$S_*$ looking out \\ of a window on \\ a rainy night'' \end{tabular} &
        \begin{tabular}{c} ``$S_*$ wearing a chefs \\ hat in the kitchen'' \end{tabular} &
        \begin{tabular}{c} ``A photo of $S_*$ \\ reading a book'' \end{tabular} \\ \\[-0.25cm]

        \includegraphics[width=0.13\textwidth]{images/original/fat_stone_bird.jpg} &
        \hspace{0.05cm}
        \includegraphics[width=0.13\textwidth]{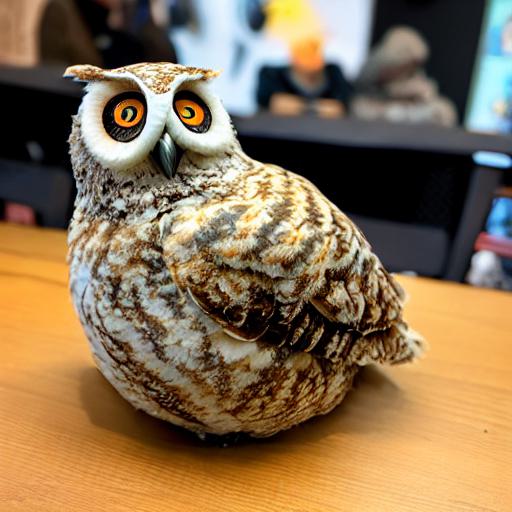} &
        \includegraphics[width=0.13\textwidth]{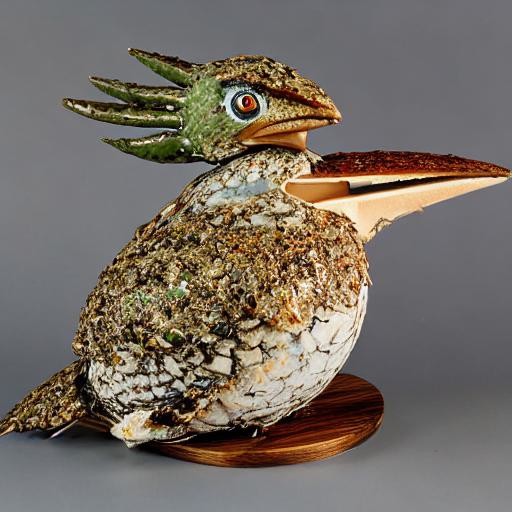} &
        \includegraphics[width=0.13\textwidth]{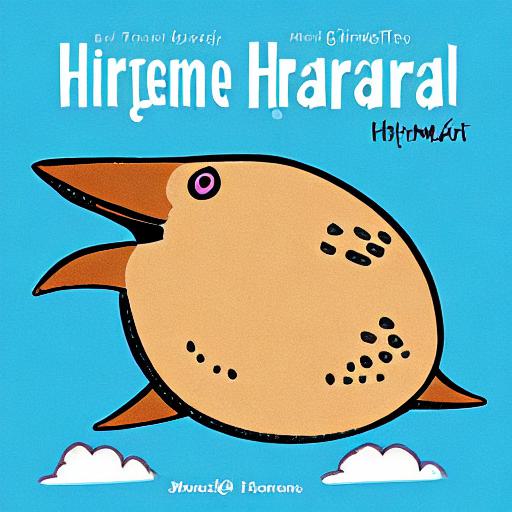} &
        \includegraphics[width=0.13\textwidth]{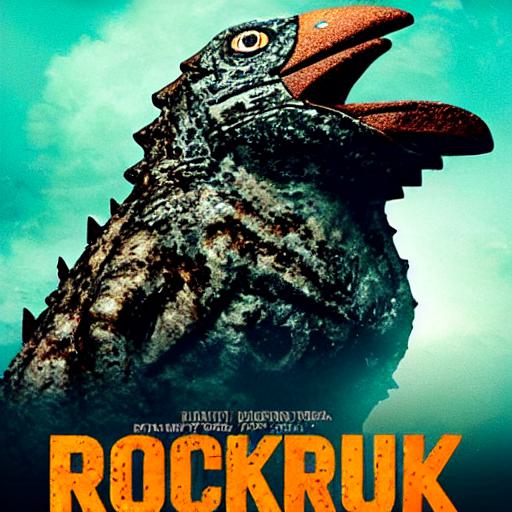} &
        \includegraphics[width=0.13\textwidth]{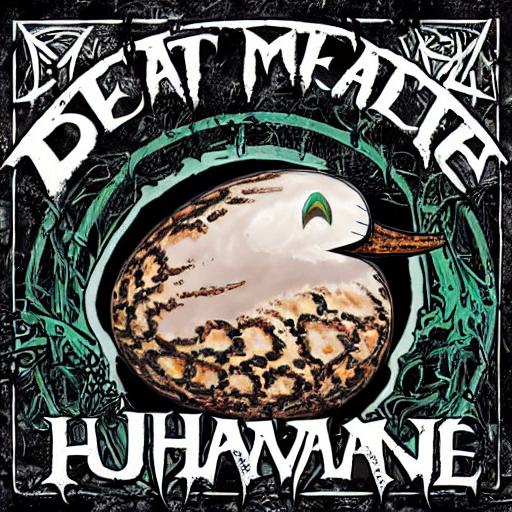} \\

        Real Sample &
        \hspace{0.05cm}
        \begin{tabular}{c} ``An owl that \\ looks like $S_*$'' \end{tabular} &
        \begin{tabular}{c} ``$S_*$ as a dragon'' \end{tabular} &
        \begin{tabular}{c} ``A children's book \\ cover about $S_*$'' \end{tabular} &
        \begin{tabular}{c} ``A movie poster of \\ The Rock, featuring \\ $S_*$ based on Godzilla'' \end{tabular} &
        \begin{tabular}{c} ``A death metal album \\ cover featuring $S_*$'' \end{tabular} \\ \\[-0.25cm]

        \includegraphics[width=0.13\textwidth]{images/original/red_bowl.jpg} &
        \hspace{0.05cm}
        \includegraphics[width=0.13\textwidth]{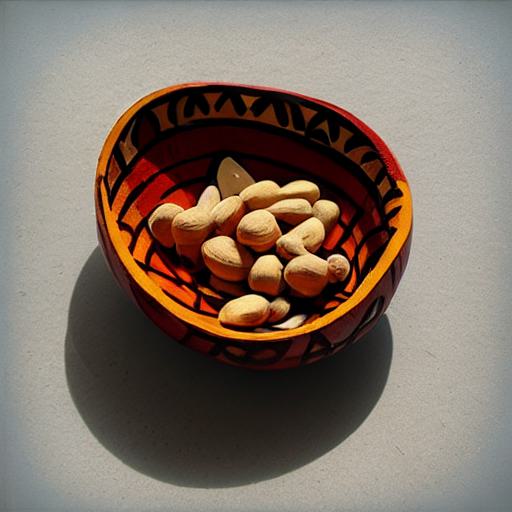} &
        \includegraphics[width=0.13\textwidth]{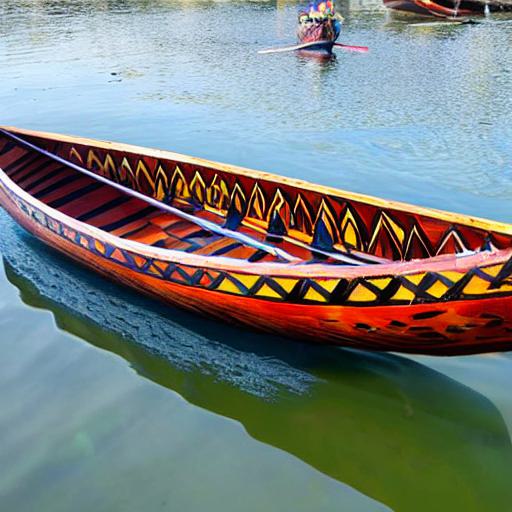} &
        \includegraphics[width=0.13\textwidth]{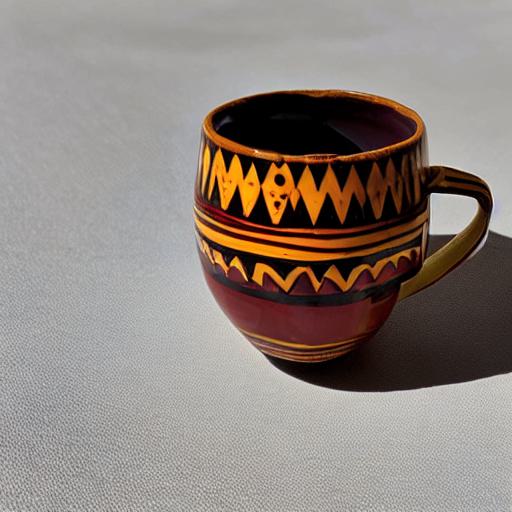} &
        \includegraphics[width=0.13\textwidth]{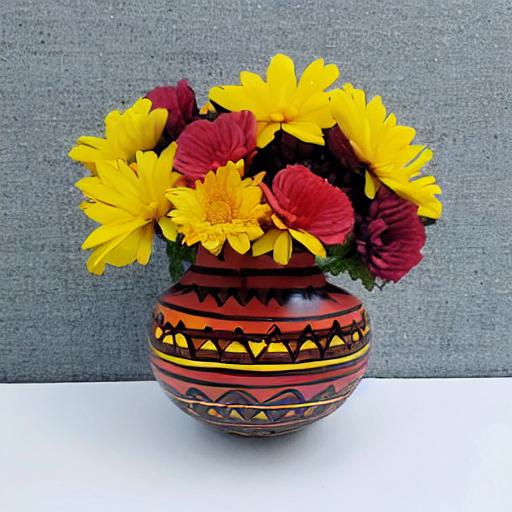} &
        \includegraphics[width=0.13\textwidth]{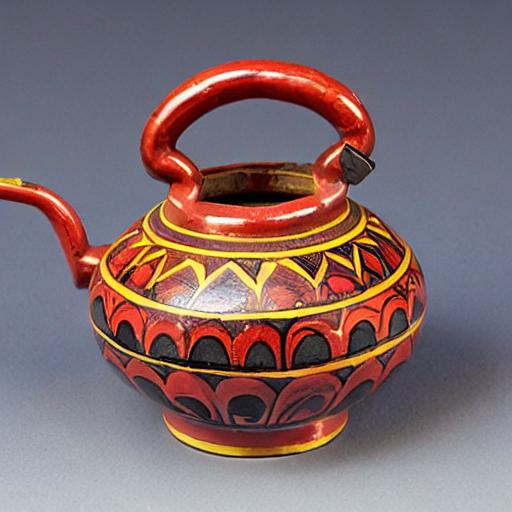} \\

        Real Sample &
        \hspace{0.05cm}
        \begin{tabular}{c} ``A photo of $S_*$ \\ with peanuts'' \end{tabular} &
        \begin{tabular}{c} ``A $S_*$ row boat'' \end{tabular} &
        \begin{tabular}{c} ``A $S_*$ coffee mug'' \end{tabular} &
        \begin{tabular}{c} ``A $S_*$ vase filled \\ with flowers'' \end{tabular} &
        \begin{tabular}{c} ``A $S_*$ Chinese \\ kettle'' \end{tabular} \\ \\[-0.25cm]

        \includegraphics[width=0.125\textwidth]{images/original/elephant.jpg} &
        \hspace{0.05cm}
        \includegraphics[width=0.125\textwidth]{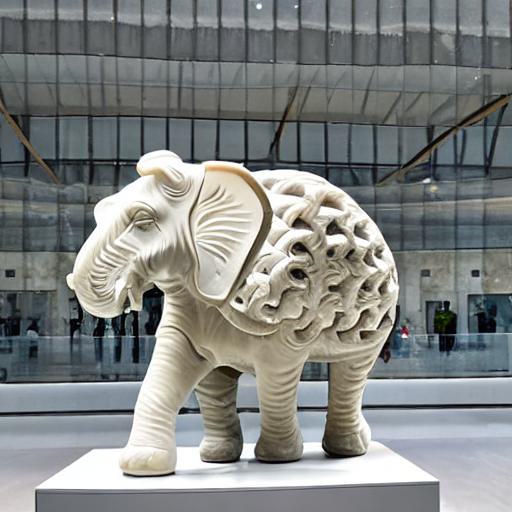} &
        \includegraphics[width=0.125\textwidth]{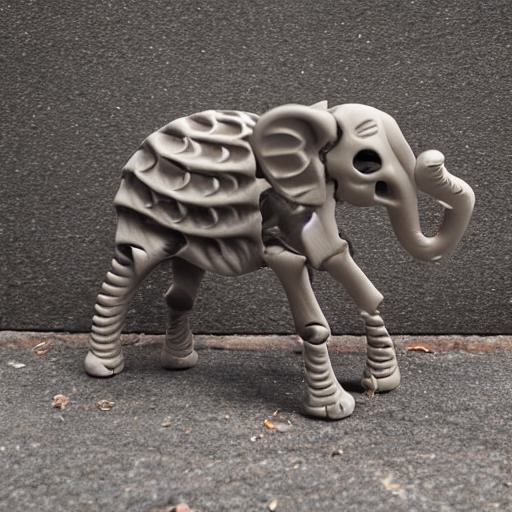} &
        \includegraphics[width=0.125\textwidth]{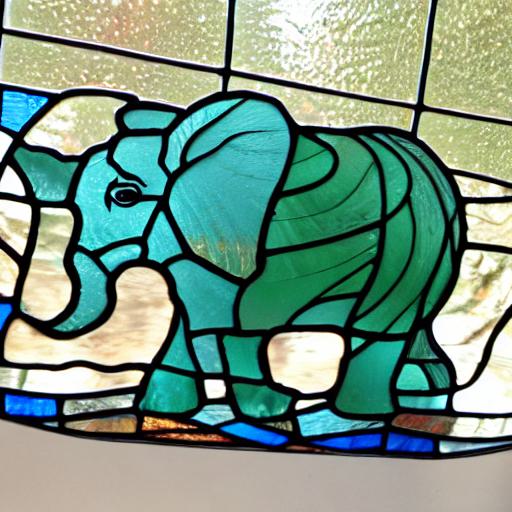} &
        \includegraphics[width=0.125\textwidth]{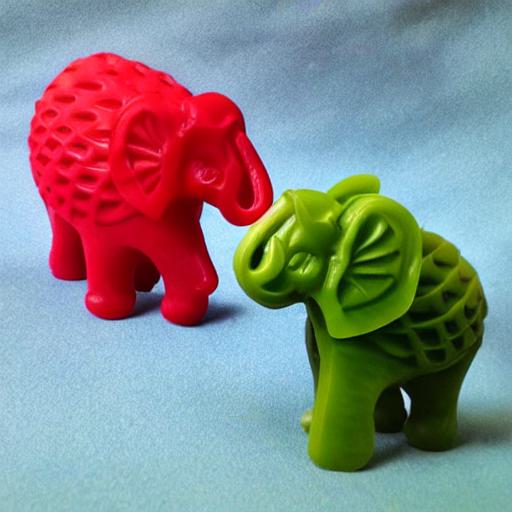} &
        \includegraphics[width=0.125\textwidth]{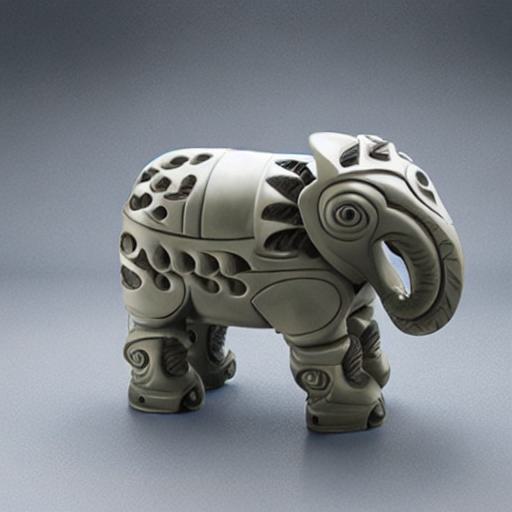} \\ 
        
        Real Sample &
        \hspace{0.05cm}
        \begin{tabular}{c} ``A marble statue \\ of $S_*$ in a museum'' \end{tabular} &
        \begin{tabular}{c} ``A skeleton of $S_*$'' \end{tabular} &
        \begin{tabular}{c} ``A $S_*$ stained \\ glass window'' \end{tabular} &
        \begin{tabular}{c} ``$S_*$ made out \\ of gummy candy'' \end{tabular} &
        \begin{tabular}{c} ``A futuristic \\ robot of $S_*$'' \end{tabular}

    \\[-0.25cm]        
    \end{tabular}
    }
    \caption{Sample text-guided personalized generation results obtained with NeTI.}
    \label{fig:ours_results_supplementary}
\end{figure*}

Below, we provide additional qualitative results, as follows: 
\begin{enumerate}
    \item In~\Cref{fig:timestep_analysis_sup}, we provide additional visualizations illustrating which concept-specific details are captured at different denoising timesteps using our neural mapper.
    \item In~\Cref{fig:nested_dropout_supplementary} we demonstrate how using our dropout technique at inference time allows for controlling the reconstruction and editability tradeoff over various concepts and prompts. 
    \item In~\Cref{fig:additional_qualitative_comparison,fig:additional_qualitative_comparison_2} we provide additional qualitative comparisons to the alternative personalization methods using concepts and prompts from our evaluation protocol.
    \item In~\Cref{fig:ours_results_supplementary} we provide additional results obtained using our NeTI scheme on a diverse set of prompts.
\end{enumerate}


\end{document}